\documentclass[sts]{imsart}
\usepackage{geometry}
\usepackage{subfiles}
\usepackage{lineno}
\usepackage{amsfonts}
\usepackage{amssymb}
\usepackage{mathrsfs}
\usepackage{textgreek}
\usepackage{mathtools}
\usepackage{amsthm,amsmath}
\usepackage[round]{natbib}
\usepackage{bbm}
\usepackage{bm}
\usepackage{tikz}
\usepackage[colorlinks=true,
            linkcolor=blue,
            urlcolor=blue,
            citecolor=blue]{hyperref}
\usepackage[nameinlink]{cleveref}
\usepackage{enumitem}
\usepackage{fancyhdr}
\usepackage{float}
\usepackage{soul}
\usepackage{xr}

\makeatletter
\newcounter{savesection}
\newcounter{apdxsection}
\renewcommand\appendix{\par
  \setcounter{savesection}{\value{section}}%
  \setcounter{section}{\value{apdxsection}}%
  \setcounter{subsection}{0}%
  \gdef\thesection{\@Alph\c@section}}
\newcommand\unappendix{\par
  \setcounter{apdxsection}{\value{section}}%
  \setcounter{section}{\value{savesection}}%
  \setcounter{subsection}{0}%
  \gdef\thesection{\@arabic\c@section}}
\makeatother

\startlocaldefs
\endlocaldefs

\RequirePackage[OT1]{fontenc}
\usetikzlibrary{positioning, shapes.geometric, arrows, shapes, snakes, automata, backgrounds, petri}
\usetikzlibrary{shapes.geometric}

\def\BE{\mathbb{E}}

\def\IIFF{\mathbb{IF}}
\def\NH{\mathsf{NH}}
\def\IF{\mathsf{IF}}
\def\log{\mathsf{log}}

\def\var{\mathsf{var}}

\def\cov{\mathsf{cov}}

\def\Bias{\mathsf{Bias}}
\def\H{\mathsf{H}}

\def\se{\mathsf{s.e.}}

\def\2{$\text{2}^{\text{nd}}$-order influence function U-statistics}
\def\3{$\text{3}^{\text{rd}}$-order influence function U-statistics}
\def\4{$\text{4}^{\th}$-order influence function U-statistics}
\def\5{$\text{5}^{\th}$-order influence function U-statistics}

\def\th{\mathsf{th}}
\def\TC{\mathsf{TC}}

\def\TB{\mathsf{TB}}
\def\CSBias{\mathsf{CSBias}}

\def\tr{\mathsf{tr}}
\def\est{\mathsf{est}}

\def\quasi{\mathsf{quasi}}
\def\debiased{\mathsf{debiased}}

\def\P{\mathsf{P}}

\def\oraclepsi{\hat{\psi}_{2, k} }
\def\genericpsi{\hat{\psi}_{2, k} (\widehat{\Omega}_{k}^{-1} )}

\def\oracleSOIF{\widehat{\IIFF}_{22, k} }

\def\zbar{\bar{\mathsf{z}}}
\def\z{\mathsf{z}}
\def\Zbar{\bar{\mathsf{Z}}}
\def \UCB{\mathsf{UCB}}

\def\Holder{\text{H\"{o}lder}}
\def\genericSOIF{\widehat{\IIFF}_{22, k} (\widehat{\Omega}_{k}^{-1} )}

\def\cf{\mathsf{cf}}

\def\shrink{\mathsf{shrink}}
\def\min{\mathsf{min}}
\def\B{\mathsf{B}}

\def\max{\mathsf{max}}

\def\NR{\mathsf{NR}}
\numberwithin{equation}{section}
\theoremstyle{definition}
\newtheorem{thm}{Theorem}[section]
\newtheorem{lem}[thm]{Lemma}

\newtheorem{definition}[thm]{Definition}

\newtheorem{proposition}[thm]{Proposition}

\newtheorem{cor}[thm]{Corollary}

\theoremstyle{remark}
\newtheorem{rem}[thm]{Remark}

\allowdisplaybreaks[1]
\newtheorem{innercustomthm}{Condition}
\newenvironment{customthm}[1]
  {\renewcommand\theinnercustomthm{#1}\innercustomthm}
  {\endinnercustomthm}
\newtheorem{innercustomgeneric}{\customgenericname}
\providecommand{\customgenericname}{}
\newcommand{\newcustomtheorem}[2]{  \newenvironment{#1}[1]
  {   \renewcommand\customgenericname{#2}   \renewcommand\theinnercustomgeneric{##1}   \innercustomgeneric
  }
  {\endinnercustomgeneric}
}
\newcustomtheorem{customthm1}{Faithfulness}
\newcustomtheorem{customthm3}{Faithfulness($\delta'$)}
\newcustomtheorem{customthm2}{CS}
\def\BL{\mathbb{L}}

\begin{document}
\begin{frontmatter}

\title{On nearly assumption-free tests of nominal confidence interval coverage for causal parameters estimated by machine learning}
\runtitle{Higher Order Testing}


\author{\fnms{Lin} \snm{Liu}\ead[label=e1]{linliu@alumni.tongji.edu.cn}\thanksref{t1}}
\thankstext{t1}{Department of Epidemiology \printead{e1}}
\author{\fnms{Rajarshi} \snm{Mukherjee}\ead[label=e2]{ram521@mail.harvard.edu}\thanksref{t2}}
\thankstext{t2}{Department of Biostatistics \printead{e2}}
\and
\author{\fnms{James M.} \snm{Robins}\ead[label=e3]{robins@hsph.harvard.edu}\thanksref{t3}}
\thankstext{t3}{Department of Epidemiology and Biostatistics \printead{e3}}
\affiliation{Harvard T. H. Chan School of Public Health}

\runauthor{LL, RM, JMR}

\begin{abstract}
For many causal effect parameters of interest, doubly robust machine learning (DRML) estimators $\hat{\psi}_{1}$ are the state-of-the-art, incorporating the good prediction performance of machine learning; the decreased bias of doubly robust estimators; and the analytic tractability and bias reduction of sample splitting with cross fitting. Nonetheless, even in the absence of confounding by unmeasured factors, the nominal $(1 - \alpha)$ Wald confidence interval $\hat{\psi}_{1} \pm z_{\alpha / 2} \widehat{\se} [\hat{\psi}_{1}]$ may still undercover even in large samples, because the bias of $\hat{\psi}_{1}$ may be of the same or even larger order than its standard error of order $n^{-1/2}$.

In this paper, we introduce essentially assumption-free tests that (i) can falsify the null hypothesis that the bias of $\hat{\psi}_{1}$ is of smaller order than its standard error, (ii) can provide an upper confidence bound on the true coverage of the Wald interval, and (iii) are valid under the null under no smoothness/sparsity assumptions on the nuisance parameters. The tests, which we refer to as \underline{A}ssumption \underline{F}ree \underline{E}mpirical \underline{C}overage \underline{T}ests (AFECTs), are based on a U-statistic that estimates part of the bias of $\hat{\psi}_{1}$.

Our claims need to be tempered in several important ways. First no test, including ours, of the null hypothesis that the ratio of the bias to its standard error is smaller than some threshold $\delta$ can be consistent [without additional assumptions (e.g. smoothness or sparsity) that may be incorrect]. Second the above claims only apply to certain parameters in a particular class. For most of the others, our results are unavoidably less sharp. In particular, for these parameters, we cannot directly test whether the nominal Wald interval $\hat{\psi}_{1} \pm z_{\alpha / 2} \widehat{\se} [\hat{\psi}_{1}]$ undercovers. However, we can often test the validity of the smoothness and/or sparsity assumptions used by an analyst to justify a claim that the reported Wald interval's actual coverage is no less than nominal. Third, in the main text, with the exception of the simulation study in \Cref{sec:background}, we assume we are in the semisupervised data setting (wherein there is a much larger dataset with information only on the covariates), allowing us to regard the covariance matrix of the covariates as known. In the simulation in \Cref{sec:background}, we consider the setting in which estimation of the covariance matrix is required. In the simulation we used a data adaptive estimator which performs very well in our simulations, but the estimator's theoretical sampling behavior remains unknown.
\end{abstract}



\end{frontmatter}











\section{Introduction and motivation}\label{sec:background} 
Valid inference (i.e. valid confidence intervals) for causal effects is of importance in many subject matter areas. For example, in medicine it is critical to evaluate whether a non-null treatment effect estimate could differ from zero simply because of sampling variability and, conversely, whether a null treatment effect estimate is compatible with a clinically important effect.

In observational studies, control of confounding is necessary for valid inference. Historically, and assuming no confounding by unmeasured covariates, two statistical approaches have been used to control confounding by potential measured confounders, both of which require the building of non-causal purely predictive algorithms:

\begin{itemize}
\item One approach builds an algorithm to predict the conditional mean $b(x)$ of the outcome of interest given data on potential confounders and (sometimes) treatment (referred to as the outcome regression);

\item The other approach builds an algorithm to predict the conditional probability $p(x)$ of treatment given data on potential confounders (referred to as the propensity score).
\end{itemize}

The validity of a nominal $(1-\alpha )$ Wald confidence interval (CI) $\hat{\psi}_{1} \pm z_{\alpha / 2} \widehat{\se} (\hat{\psi}_{1})$\footnote{In this paper, we use the standard notation $z_{\alpha}$ to denote the $1-\alpha$ standard normal quantile and $\Phi (x)$ to denote the standard normal CDF.} for a parameter $\psi$ of interest centered at a particular estimator $\hat{\psi}_{1}$ quite generally requires that the bias of $\hat{\psi}_{1}$ is much less than than its estimated standard error $\widehat{\se} (\hat{\psi}_{1})$. A nominal $(1 - \alpha)$ confidence interval is said to be valid if the actual coverage rate under repeated sampling is no smaller than $(1 - \alpha)$. Under either of the above approaches, obtaining estimators with small bias generally depends on good performance of the corresponding prediction algorithm. This has motivated the application of modern machine learning (ML) methods to these prediction problems for the following reason. When the vector of potential confounding factors is high-dimensional, as is now standard owing to the ``big data revolution'', it has become noted that, so-called machine learning algorithms (e.g. neural nets \citep{krizhevsky2012imagenet}, support vector machines \citep{cortes1995support}, boosting \citep{freund1997decision}, regression trees and random forests \citep{breiman2001random}, etc., especially when combined with cross-validation) can often do a much better job of prediction than traditional parametric or non-parametric approaches (e.g. kernel or series regression). However, even the best machine learning methods may fail to give predictions that are sufficiently accurate to provide nearly unbiased causal effect estimates and, thus, may fail to control bias due to confounding.

To partially guard against this possibility, so-called doubly robust machine learning (DRML) \citep{chernozhukov2018double} estimators have been developed that can be nearly unbiased for the causal effect $\psi$, even when both of the above approaches fail. DRML estimators employ ML estimators of both the outcome regression $b(x)$ and the propensity score $p(x)$. DRML estimators are the state-of-the-art for estimation of causal effects, combining the benefits of sample splitting, machine learning, and double robustness \citep{scharfstein1999adjusting, scharfstein1999rejoinder, robins2001comments, bang2005doubly}. By sample splitting we mean that the data is randomly divided into two (or more) samples - the estimation sample and the training sample. The ML estimators $\hat{b}(x)$ and $\hat{p}(x)$ of $b(x)$ and $p(x)$ are fit using the training sample data. The estimator $\hat{\psi}_{1}$ of our causal parameter $\psi$ is computed from the estimation sample treating the ML estimators as fixed functions. This approach is required because the ML estimates of the regression functions generally have unknown statistical properties and, in particular, may not lie in a so-called Donsker class - a condition often needed for valid inference when sample splitting is not employed. Under conditions given in \Cref{thm:drml}, the efficiency lost due to sample splitting can be recovered by cross-fitting. The cross-fitting estimator $\hat{\psi}_{\cf, 1}$ averages $\hat{\psi}_{1}$ with its `twin' obtained by exchanging the roles of the estimation and training sample. In the semiparametric statistics literature, the possibility of using sample-splitting with cross-fitting to avoid imposing Donsker conditions has a long history \citep[Page 391]{schick1986asymptotically, van1998asymptotic}, although the idea of explicitly combining cross-fitting with ML was not emphasized until recently. \citet{ayyagari2010applications} Ph.D. thesis (subsequently published as \citet{robins2013new}) and \citet{zheng2011cross} are early examples that emphasized the theoretical and finite sample advantages of DRML estimators.

However, even the use of DRML estimators is not guaranteed to provide valid inferences owing to the possibility that the two ML prediction algorithms are not sufficiently accurate for the bias to be small compared to the standard error. In particular, if the bias of the DRML estimator is of the same (or greater) order than its standard error, the actual coverage of nominal $(1 - \alpha)$ CIs for the causal effect will be smaller (and often much smaller) than the nominal level, thereby producing misleading inferences.

Suppose an author publishes a paper with a nominal $(1 - \alpha)$ Wald CI $\hat{\psi}_{\cf, 1} \pm z_{\alpha / 2} \widehat{\se} (\hat{\psi}_{\cf, 1})$ for a parameter $\psi$. The previous discussion leads to the following question. Can $\alpha^{\dag}$-level tests be developed that have the ability to falsify whether the bias of the DRML estimator $\hat{\psi}_{1}$ or $\hat{\psi}_{\cf, 1}$ is of the same or greater order than its standard error? In particular, can we provide an upper confidence bound on the actual coverage of a nominal $(1 - \alpha)$ CI $\hat{\psi}_{\cf, 1} \pm z_{\alpha / 2} \widehat{\se} (\hat{\psi}_{\cf, 1})$? If so, when such excess bias is detected, can we construct new estimators $\hat{\psi}_{2}$ that are less biased? Furthermore, is it possible to construct such tests and estimators without: i) refitting, modifying, or even having knowledge of the ML algorithms that have been employed and ii) without making any assumptions about the smoothness or sparsity of the true outcome regression $b(x)$ or propensity score function $p(x)$? 

Throughout we assume that we have been given access to the data set used to obtain both the estimate $\hat{\psi}_{1}$ and the estimated regression functions outputted by some ML prediction algorithms. We do not require any knowledge of or access to the ML algorithms used, other than the functions $\hat{b}(x)$ and $\hat{p}(x)$ that they outputted.

In this paper, we show that, perhaps surprisingly, for parameters in a certain class, the {\it monotone bias class} defined in \Cref{monotone} of \Cref{sec:intro}, the answer to these questions is ``yes'' by using higher-order influence function tests and estimators \citep{robins2008higher, robins2017minimax, mukherjee2017semiparametric}. We refer to such tests as \underline{A}ssumption-\underline{F}ree \underline{E}mpirical \underline{C}overage \underline{T}ests (AFECTs). For parameters not in the {\it monotone bias class}, we cannot test whether the bias of $\hat{\psi}_{1}$ is small compared to its standard error. The best we can do is to empirically test the author's \textit{justification} for the claim that his intervals are valid. In general a data analyst who reports the interval $\hat{\psi}_{\cf, 1} \pm z_{\alpha / 2} \widehat{\se} (\hat{\psi}_{\cf, 1})$ justifies its validity by (i) imposing restrictive assumptions on the complexities of $b$ and $p$ (in terms of smoothness or sparsity) and then (ii) appealing to theorems that guarantee the asymptotic validity of the Wald CI under these assumptions. However, these assumptions may be incorrect. We show that we can often construct AFECTs that can falsify the complexity reducing assumptions on $b$ and $p$.

To make the above more concrete, we describe our approach at a high level. Throughout, we let $A$ denote the treatment indicator, $Y$ a bounded outcome of interest, and $X$ the vector of potential confounders with compact support. Let $\hat{\psi}_{1}$ and $\hat{\psi}_{1} \pm z_{\alpha / 2} \widehat{\se} (\hat{\psi}_{1})$ denote a DRML estimator of and associated $(1 - \alpha)$ Wald CI for a particular parameter $\psi$. In this paper, for didactic purposes only, we will choose $\psi $ to be (components) of the so-called variance-weighted average treatment effect (ATE) of a binary treatment $A$ on $Y$ given a vector $X$ of confounding variables. Specifically these components are the expected conditional variance $\BE [\var (A | X)]$ of $A$ given $X$ and the expected conditional covariance $\BE [\cov (A, Y | X)]$ of $A$ and $Y$ given $X$, with the variance weighted ATE being $\BE[\cov(A, Y | X)] / \BE[\var(A | X)]$. We chose the variance weighted ATE precisely because $\BE [\var (A | X)]$ is in the {\it monotone bias class} but $\BE [\cov (A, Y | X)]$ is not, thereby allowing us to highlight the critical difference between these classes. The methods developed herein can be applied essentially unchanged to many other causal effect parameters (e.g. the average treatment effect and the effect of treatment on the treated) regardless of the state spaces of $A$ and $Y$, as well as to many non-causal parameters.  

Even for the parameter $\BE [\var (A | X)]$, as explained in \Cref{rem:consistency}, there is an unavoidable limitation to what can be achieved with our or any other method: No test, including ours, of the null hypothesis that the bias of a DRML estimator is negligible compared to its standard error can be consistent [without making additional, possibly incorrect, complexity reducing assumptions on $b(x)$ and $p(x)$]. Thus, when our $\alpha^{\dag}$-level test rejects the null for $\alpha^{\dag}$ small, we can have strong evidence that the estimators $\hat{\psi}_{1}$ and $\hat{\psi}_{\cf, 1}$ have bias at least the order of its standard error; nonetheless when the test does not reject, we cannot conclude that there is good evidence that the bias is less than the standard error, no matter how large the sample size. In fact, in the absence of complexity reducing assumptions, no consistent estimator of $\BE [\var (A | X)]$ exists; hence we can never empirically rule out that the bias of $\hat{\psi}_{1}$ and $\hat{\psi}_{\cf, 1}$ is as large as order 1 and thus $n^{1/2}$ times greater than $\widehat{\se} (\hat{\psi}_{1})$! Put another way, because we make essentially no assumptions, no methodology can (non-trivially) upper bound the bias of any estimator or lower bound the coverage of any confidence interval.

In this paper, we are adopting a skeptic's stance, which is illuminated by comparing two social norms. The first is the social norm most of our parents taught us and the second is the skeptics social norm.

\begin{itemize}
\item Parental Social Norm: If You Don't Have Anything Positive to Contribute, Don't Go Criticizing Others.

\item Skeptics's Social Norm: Not Having Anything Positive to Contribute Does Not Relieve You of Your Duty to Criticize What Others Say.
\end{itemize}

As we saw above, because we do not impose complexity reducing assumptions on $b$ and $p$, we have nothing to contribute if we follow parental social norms. However, in this paper, we adopt the \textit{skeptic's social norms} and criticize, where possible, an author who reports a state of the art $(1 - \alpha)$ Wald CI $\hat{\psi}_{\cf, 1} \pm z_{\alpha / 2} \widehat{\se} (\hat{\psi}_{\cf, 1})$ as valid. However, our critique will have to be stronger than simply informing the author that one can prove (when possible) that his interval will not be valid if his complexity-reducing assumptions are incorrect, as he will likely respond that he believes his assumptions to be reasonable and likely true under the law actually generating the data. Instead for parameters in the {\it monotone bias class}, we will employ AFECTs to prove to the author that his Wald CI is invalid. 

For other parameters such as the $\BE[\cov(A, Y | X)]$, we can only falsify the validity of the author's Wald interval under the so-called faithfulness assumption given in Section \ref{sec:null_cov}. Heuristically, faithfulness is the assumption that near perfect cancelling of the non-negligible bias of two separate components of the the bias of $\hat{\psi}_{\cf, 1}$ (one estimable and the other not) to give near zero total bias will essentially never occur. 

If we do not assume faithfulness, we must consider the less ambitious goal of demonstrating to the author, when possible, that his complexity reducing assumptions are incorrect [without being able to ever empirically prove the bias of his estimator is of the order of its standard error or greater]. If successfully achieved, the author would then have to admit that he can no longer justify his earlier claim of validity for his state-of-the-art confidence interval. The approach described here is one of being `in dialogue with current practices and practitioners'. This is not surprising, as it is the justifications of the practitioners that the skeptic is critiquing.

To be concrete, suppose, as is often the case, an author justifies the validity of $\hat{\psi}_{1} \pm z_{\alpha / 2} \widehat{\se} (\hat{\psi}_{1})$ and thus its cross-fit version $\hat{\psi}_{\cf, 1} \pm z_{\alpha / 2} \widehat{\se} (\hat{\psi}_{\cf, 1})$ by (i) first proving that, under his complexity reducing assumptions, the Cauchy Schwarz (CS) bias functional 
\begin{equation*}
\CSBias (\hat{\psi}_{1}) = \left\{ \BE \left[ \left( \hat{b}(X) - b(X) \right)^{2} \right] \right\}^{1/2} \left\{ \BE \left[ \left( \hat{p}(X) - p(X) \right)^{2} \right] \right\}^{1/2}
\end{equation*}
is $o(n^{-1/2})$\footnote{Here the asymptotic statement would be in probability had we not treat the training sample as fixed.} , conditional on the training sample\footnote{In this paper, essentially all expectations and probabilities are to be understood as being conditional on the training sample. Hence we can and do omit this conditioning event in our notation.} (and thus also on the functions $\hat{b}, \hat{p}$ computed from the training sample) and (ii) then noting the CS bias upper bounds the absolute conditional bias 
\begin{equation*}
\left\vert \BE \left[ \left( \hat{b}(X) - b(X) \right) \left( \hat{p}(X) - p(X) \right) \right] \right\vert 
\end{equation*}
of $\hat{\psi}_{1}$ for $\psi (\theta) = \BE_{\theta}[\cov_{\theta}(A, Y | X)]$. It then follows if we can empirically show that Cauchy Schwarz bias $\CSBias (\hat{\psi}_{1})$ exceeds some given multiple $\delta > 0$, e.g. $\delta = 0.75$, times $\hat{\psi}_{1}$'s conditional standard error of order $n^{-1/2}$, then we have falsified the analysts' justification of the claim that his nominal $(1 - \alpha)$ Wald CIs are valid.

To this end, we shall construct $\alpha^{\dag}$-level AFECTs of the null hypothesis $\CSBias (\hat{\psi}_{1}) < \se (\hat{\psi}_{1}) \delta$, which can be done because, as we shall see, the $\CSBias (\hat{\psi}_{1})$ parameter is in the {\it monotone bias class}.

We now describe our AFECT tests and related estimators at a high level. DRML estimators are based on the first order influence function of the parameter $\psi$ \citep{van1998asymptotic}. Our proposed approach begins by computing a \textit{second order influence function estimator} $\widehat{\IIFF}_{22, k}$ of the estimable part of the conditional bias $\BE \left[ \hat{\psi}_{1} - \psi \right]$ of $\hat{\psi}_{1}$ \textit{given the training sample data}. The bias corrected estimator is $\hat{\psi}_{2, k} \equiv \hat{\psi}_{1} - \widehat{\IIFF}_{22, k}$, where $\widehat{\IIFF}_{22, k}$ is a second-order U-statistic that depends on a choice of $k$ (with $k = o(n^{2})$ for reasons explained in \Cref{rem:k1}), a vector of basis functions $\Zbar_{k} \equiv \zbar_{k}(X) \equiv (\mathsf{z}_{1}(X), \ldots, \mathsf{z}_{k}(X))^{\top}$ of $X$ and an estimator $\widehat{\Omega}_{k}^{-1}$ of the inverse expected outer product $\Omega_{k}^{-1} \coloneqq \{\BE[\zbar_{k}(X) \zbar_{k}(X)^{\top}]\}^{-1}$. Both $\hat{\psi}_{2, k}$ and $\widehat{\IIFF}_{22, k}$ will be asymptotically normal when, as in our asymptotic set-up, $k \rightarrow \infty$ and $k = o(n^{2})$ as $n \rightarrow \infty$ (If $k$ did not increase with $n$, the asymptotic distribution of $\widehat{\IIFF}_{22, k}$ would be the so-called Gaussian chaos distribution \citep{rubin1980asymptotic}). 

The degree of the bias corrected by $\widehat{\IIFF}_{22, k}$ depends critically on (i) the choice of $k$, (ii) the accuracy of the estimator $\widehat{\Omega}_{k}^{-1}$ of $\Omega_{k}^{-1}$ when $\Omega_{k}^{-1}$ is unknown (see \Cref{sec:cov}), and (iii) the particular $k$-vector of (basis) functions $\Zbar_{k} \equiv \zbar_{k}(X)$ selected from a much larger, possibly countably infinite, dictionary of candidate functions.

One sometimes has $X$-semisupervised data available; that is, a data set in which the number $N$ of subjects with complete data on $\left( A, Y, X\right)$ is many fold less than the number of subjects on whom only data on the covariates $X$ are available. In that case, assuming the subjects with complete data are effectively a random sample of all subjects, we can estimate $\Omega_{k}$ by the empirical covariance matrix from subjects with incomplete data; and then treat $\Omega_{k}^{-1}$ as known in an analysis based on the $N$ subjects with complete data \citep{chapelle2010semi, chakrabortty2018efficient}. Since, for the most of the paper we assume access to semisupervised data, we will omit the notational dependence on $\Omega_{k}^{-1}$ and denote $\widehat{\IIFF}_{22, k} (\Omega_{k}^{-1})$ and $\hat{\psi}_{2, k} (\Omega_{k}^{-1})$ by $\widehat{\IIFF}_{22, k}$ and $\hat{\psi}_{2, k}$. However we write $\widehat{\IIFF}_{22, k} (\widehat{\Omega}_{k}^{-1})$ and $\hat{\psi}_{2, k} (\widehat{\Omega}_{k}^{-1})$ when an estimator $\widehat{\Omega}_{k}^{-1}$ is substituted for $\Omega_{k}^{-1}$. In the simulations below we use a particular data-adaptive estimator $\widehat{\Omega}_{k}^{-1}$, described in \Cref{sec:simulation}. Both $\widehat{\IIFF}_{22, k} (\widehat{\Omega}_{k}^{-1})$ and $\hat{\psi}_{2, k} (\widehat{\Omega}_{k}^{-1})$ performed very well in our simulations; nonetheless, in contrast to $\widehat{\IIFF}_{22, k}$ and $\hat{\psi}_{2, k}$, we, as yet, lack a theoretical understanding of their statistical behavior. Consequently, we have relegated the definition and discussion of the estimators $\widehat{\IIFF}_{22, k} (\widehat{\Omega}_{k}^{-1})$ and $\hat{\psi}_{2, k} (\widehat{\Omega}_{k}^{-1})$ to \Cref{sec:simulation} and the supplementary materials, as requested by a referee.

For further motivation, we now summarize the results from one of the simulation studies that are described in detail in \Cref{sec:simulations}. We simulated 100 estimation samples each with sample size $n = 5000$. The same training sample, also of size 5000, and thus the same estimates of the nuisance regression functions were used in each simulation. Thus the results are conditional on that training sample. The dimension $d$ of $X$ is chosen to be 2 in order to allow estimation of the nuisance functions by kernel regression (with bandwidth selected by cross validation) in a timely fashion. We let $\psi = \BE [\var (A | X)]$. We took $k$ to be less than $n$ for the following three reasons: $k < n$ is necessary i) for CIs centered at $\hat{\psi}_{2, k} \equiv \hat{\psi}_{1} - \widehat{\IIFF}_{22, k}$ to have length approximately equal to CIs centered at $\hat{\psi}_{1}$, ii) for $\se (\widehat{\IIFF}_{22, k})$ to be of order smaller than or equal to the order $n^{-1/2}$ of the standard error of $\hat{\psi}_{1}$, thereby creating the possibility of detecting that the ratio of the bias of $\hat{\psi}_{1}$ to its standard error exceeds a constant $\delta$, if $n$ is sufficiently large and iii) to be able to estimate $\Omega_{k}^{-1}$ accurately without imposing the additional (possibly incorrect) smoothness or sparsity assumptions on the marginal density $f_{X}$. Thus we were able to use quite nonsmooth densities $f_{X}$ in simulations. See \Cref{sec:simulations}.

In simulation studies we chose a data generating process for which the minimax rates of estimation were known, in order to be able to better evaluate the properties of our proposed procedures. Specifically, both the true propensity score and outcome regression functions in our simulation studies were chosen to lie in particular \Holder{} smoothness classes chosen to ensure that $\hat{\psi}_{1}$ had significant asymptotic bias. We estimated these regression functions using nonparametric kernel regression estimators that are known to obtain the minimax optimal rate of convergence for these smoothness classes \citep{tsybakov2009book}, thereby guaranteeing that $\hat{\psi}_{1}$ performed close to as well as any other DRML estimator. [Out of interest, in \Cref{sec:simulations}, we also report simulation results when the regression functions are estimated by neural networks.] The basis functions $\zbar_{k}(x)$ were chosen to be particular Cohen-Vial-Daubechies wavelets that \citet{robins2009semiparametric, robins2017minimax} showed to be minimax optimal for estimation of $\psi$ by 
$\hat{\psi}_{2, k}$ for the chosen smoothness classes. In summary, we used optimal versions of $\hat{\psi}_{1}$ and $\hat{\psi}_{2, k}$ to ensure a fair comparison.

\begin{table}[tbp]
\caption{}
\label{tab:intro}
\resizebox{\columnwidth}{!}{
\begin{tabular}{l | lclc}
\hline
$k$ & $\oracleSOIF$ & \shortstack{MC Coverage \\ ($\hat{\psi}_{2, k} $ 90\% Wald CI)} & $\mathsf{Bias} ( \oraclepsi )$ & $\widehat\chi_{k} ( \Omega_{k}^{-1}; z_{0.10}, \delta = 0.75 (1.5))$ \\
\hline
$0$ & 0 (0) & 0\% & 0.229 (0.0161) & 0\% (0\%) \\
$64$ & 0.0457 (0.00782) & 0\% & 0.183 (0.0144) & 99\% (44\%) \\
$128$ & 0.0484 (0.00831) & 0\% & 0.180 (0.0145) & 100\% (54\%) \\
$256$ & 0.125 (0.0144) & 0\% & 0.103 (0.0114) & 100\% (100\%) \\
$512$ & 0.127 (0.0147) & 0\% & 0.101 (0.0122) & 100\% (100\%) \\
$1024$ & 0.129 (0.0172) & 0\% & 0.100 (0.0147) & 100\% (100\%) \\
$2048$ & 0.161 (0.0238) & 4\% & 0.0672 (0.0191) & 100\% (100\%) \\
$4096$ & 0.180 (0.0271) & 46\% & 0.0483 (0.0259) & 100\% (100\%) \\
\hline
$k$ & $\genericSOIF$ & \shortstack{MC Coverage \\ ($\hat{\psi}_{2, k} (\widehat{\Omega}_{k}^{-1} )$ 90\% Wald CI)} & $\mathsf{Bias} \left( \genericpsi \right)$ & $\widehat\chi_{k} ( \widehat{\Omega}_{k}^{-1}; z_{0.10}, \delta = 0.75 (1.5))$ \\
\hline
$0$ & 0 (0) & 0\% & 0.229 (0.0252) & 0\% (0\%) \\
$64$ & 0.0465 (0.00785) & 0\% & 0.182 (0.0143) & 100\% (47\%) \\
$128$ & 0.0498 (0.00831) & 0\% & 0.180 (0.0143) & 100\% (64\%) \\
$256$ & 0.131 (0.0142) & 0\% & 0.0972 (0.0116) & 100\% (100\%) \\
$512$ & 0.136 (0.0150) & 0\% & 0.0922 (0.0125) & 100\% (100\%) \\
$1024$ & 0.142 (0.0173) & 0\% & 0.0868 (0.0143) & 100\% (100\%) \\
$2048$ & 0.165 (0.0222) & 4\% & 0.0636 (0.0185) & 100\% (100\%) \\
$4096$ & 0.225 (0.0374) & 90\% & 0.00314 (0.0296) & 100\% (100\%) \\
\hline
\end{tabular}
} \newline
Here the parameter of interest is $\psi(\theta) = \mathbb{E}_\theta [ \mathsf{var}_\theta (A | X) ]$. We reported the Monte Carlo averages (MCavs) of point estimates and Monte Carlo standard deviations (MCsds) in the parenthesis (first column in each panel) of $\widehat{\mathbb{IF}}_{22, k}$ and $\widehat{\mathbb{IF}}_{22, k} (\widehat{\Omega}_{k}^{-1})$, together with the coverage probability of 90\% CIs (second column in each panel) of $\hat{\psi}_{2, k}$ and $\hat{\psi}_{2, k} (\widehat{\Omega}_{k}^{-1})$, the MCavs of the bias and MCsds in the parenthesis (third column in each panel) of $\hat{\psi}_{2, k}$ and $\hat{\psi}_{2, k} (\widehat{\Omega}_{k}^{-1})$ and the empirical rejection rate based on the test statistic $%
\widehat\chi_{k} (\zeta_{k}, \delta = 0.75 \text{ or } 1.5)$ and $\widehat\chi_{k} (\widehat{\Omega}_{k}^{-1}; \zeta_{k}, \delta = 0.75 \text{ or } 1.5)$ (see \Cref{sec:intro}) with $\zeta_{k} = z_{0.10} = 1.28$ (fourth column in each panel). In the simulation, we choose $A \sim p(X) + N(0, 1)$. For more details on the simulation setup, see \Cref{sec:simulations}.
\end{table}

\Cref{tab:intro} reports results from this simulation study. We examined the empirical behavior of our data adaptive estimator as $k$ varies by comparing the estimators $\widehat{\IIFF}_{22, k} (\widehat{\Omega}_{k}^{-1})$ and $\hat{\psi}_{2, k}(\widehat{\Omega}_{k}^{-1})$ that use $\widehat{\Omega}_{k}^{-1}$ to the oracle estimators $\widehat{\IIFF}_{22, k}$ and $\hat{\psi}_{2, k}$ that use the true inverse covariance matrix $\Omega_{k}^{-1}$ (see \Cref{sec:simulation} and \Cref{sec:cov}). The target parameter $\psi$ of this simulation study is the expected conditional variance of $A$ given $X$. Simulation results for the expected conditional covariance were similar and are reported in \Cref{sec:simulations}.

Note the unmodified estimator $\hat{\psi}_{1}$ is included as the first row of \Cref{tab:intro} as, by definition, it equals $\hat{\psi}_{2, k}$ for $k = 0$. Also by definition, $\widehat{\mathbb{IF}}_{22, k = 0}$ and $\widehat{\mathbb{IF}}_{22, k = 0} (\widehat{\Omega}_{k}^{-1})$ are zero. As seen in row 1, column 2 of \Cref{tab:intro}, nominal 90\% Wald CIs centered at $\hat{\psi}_{1} = \hat{\psi}_{2, k = 0}$ had empirical coverage of 0\% in 100 simulations! However, as seen in column 2 of both the upper and lower panels of the last row, 90\% Wald CIs centered at $\hat{\psi}_{2, k}$ at $k = 4096$ had empirical coverage around 46\%\footnote{Our data generating process implied that $\hat{\psi}_{2, k}$ was $\sqrt{n}$-consistent but asymptotically biased, so the expected coverage of the Wald CI centered at $\hat{\psi}_{2, k}$ was less than 90\%.}. The standard error of $\hat{\psi}_{2, k}$ did not greatly exceed that of $\hat{\psi}_{1}$.

In more detail, the left panel of \Cref{tab:intro} displays the Monte Carlo averages (MCavs) of the point estimates and Monte Carlo standard deviations (MCsds) (in parentheses) of $\widehat{\mathbb{IF}}_{22, k}$ in the first column; the empirical probability that a nominal 90\% Wald CI centered at $\hat{\psi}_{2, k}$ covered the true parameter value in the second column; the MC bias (i.e. MCav of $\hat{\psi}_{2,k} - \psi$) and MCsd of $\hat{\psi}_{2, k} $ in the third column; and, in the fourth column, the empirical rejection rate of a one sided $\alpha^{\dag} = 0.10$ level test $\widehat{\chi}_{k}^{(1)}(z_{\alpha^{\dag} = 0.10}, \delta = 0.75\text{ or }1.5)$ (defined in \cref{oneside} of \Cref{sec:intro}) of the null hypothesis that the bias of $\hat{\psi}_{1}$ is smaller than $\delta = 0.75 \text{ or } 1.5$ of its standard error. The test rejects when the ratio $\widehat{\mathbb{IF}}_{22, k} / \widehat{\se} (\hat{\psi}_{1})$ is large. Similarly, the bottom panel displays these same summary statistics but with the data adaptive estimator $\widehat{\Omega}_{k}^{-1}$ in place of $\Omega _{k}^{-1}$. The difference between the MC bias of $\hat{\psi}_{2, k} (\widehat{\Omega}_{k}^{-1})$ and $\hat{\psi}_{2, k}$ is an estimate of the additional bias due to the estimation of $\Omega_{k}^{-1}$ by $\widehat{\Omega}_{k}^{-1}$. (The uncertainty in the estimate of the bias itself is not given in the table but it is negligible as it approximately equals $(1/100)^{1/2}$ times the standard error given in the table.)

Reading from the first row of \Cref{tab:intro}, we see that the MC bias of $\hat{\psi}_{1}$ was 0.229. The MC bias of $\hat{\psi}_{2, k}$ and $\hat{\psi}_{2, k} (\widehat{\Omega}_{k}^{-1})$ decreased with increasing $k$, becoming nearly zero at $k = 4096$. The observation that the bias decreases as $k$ increases is predicted by the theory developed in \Cref{sec:intro} and reflects the fact that  $\psi = \BE [\var (A | X)]$ is in the {\it monotone bias class}. The decrease in bias reflects the increase in the absolute value of $\widehat{\IIFF}_{22, k}$ with $k$. Note further that both the MCavs of $\widehat{\IIFF}_{22, k}$ and $\widehat{\IIFF}_{22, k} (\widehat{\Omega}_{k}^{-1})$ are relatively close, as are their MCsds, implying that our estimator $\widehat{\Omega}_{k}^{-1}$ performs similarly to the true $\Omega_{k}^{-1}$. The actual coverages of 90\% Wald CIs centered at $\hat{\psi}_{2, k}$ and $\hat{\psi}_{2, k}(\widehat{\Omega}_{k}^{-1})$ both increase from 0\% at $k = 0$ to more than 40\% at $k=4096$. Also, reading from the third column, we see that the MCsd (0.0259) of $\hat{\psi}_{2, k = 4096}$ is only 1.6 times the standard error (0.0161) of $\hat{\psi}_{1}$, confirming that the dramatic difference in coverage rates of their associated CIs is due to the bias of $\hat{\psi}_{1}$. Reading from the 4th column of each panel, we see that the rejection rates of both $\widehat{\chi}_{k}^{(1)} (z_{\alpha^{\dag} = 0.10}, \delta)$ and $\widehat{\chi}_{k}^{(1)} (\widehat{\Omega}_{k}^{-1}; z_{\alpha^{\dag} = 0.10}, \delta)$ for $\delta = 0.75$ (for $\delta = 1.5$) are already 100\% when $k$ is 64 (256), indicating that the bias of $\hat{\psi}_{1}$ is much greater than $0.75$ ($1.5$) of its standard error. Indeed, reading from row 1 of column 3, we see that the ratio of the MC bias of $\hat{\psi}_{1} = \hat{\psi}_{2, k = 0}$ (0.229) to its MCsd (0.0161) is nearly $14$! In \Cref{rem:sim}, we show that this ratio is close to that predicted by theory.

\Cref{fig:ucb} in \Cref{app:ucb} provides a histogram over the 100 estimation samples of $(1 - \alpha^{\dag})$ upper confidence bounds $\mathsf{UCB}^{(1)} (\Omega_{k = 2048}^{-1}; \alpha, \alpha^{\dag})$ (defined in \cref{ucbone} of \Cref{sec:intro}) and $\mathsf{UCB}^{(1)} (\widehat{\Omega}_{k = 2048}^{-1}; \alpha, \alpha^{\dag})$ (defined in \cref{ucbone-adapt} of \Cref{sec:simulation}) for the actual conditional asymptotic coverage of the nominal $(1 - \alpha)$ CI $\hat{\psi}_{1} \pm z_{\alpha / 2} \widehat{\se} (\hat{\psi}_{1})$. To clarify the meaning of $\mathsf{UCB}^{(1)}(\Omega_{k = 2048}^{-1}; \alpha, \alpha^{\dag})$, let $\mathsf{coverage} (\alpha) = P(\psi \in \{\hat{\psi}_{1} \pm z_{\alpha / 2} \widehat{\se} (\hat{\psi}_{1})\})$ be the conditional actual coverage of $\psi$, given the training sample. Then, by definition, a $(1 - \alpha^{\dag})$ conditional upper confidence bound $\mathsf{UCB}^{(1)}(\Omega_{k = 2048}^{-1}; \alpha, \alpha^{\dag})$ is a random variable satisfying\footnote{For example if $\mathsf{UCB}^{(1)}(\Omega_{k = 2048}^{-1}; \alpha = 0.10, \alpha^{\dag} = 0.10) = 0.14$, then the actual coverage of the nominal 90\% interval $\hat{\psi}_{1} \pm 1.64 \widehat{\se} (\hat{\psi}_{1})$ is no more than 14\% with confidence at least $1 - \alpha^{\dag} = 0.90$. More precisely, the random interval $[0, \mathsf{UCB}^{(1)}(\Omega_{k = 2048}^{-1}; \alpha = 0.10, \alpha^{\dag} = 0.10)]$ is guaranteed to include the actual coverage of $\hat{\psi}_{1} \pm 1.64 \widehat{\se} (\hat{\psi}_{1})$ at least 90\% of the time in repeated sampling of the estimation sample with the training sample fixed.} 
\begin{equation}  \label{def-ucb}
P \left\{ \mathsf{coverage}(\alpha) \leq \mathsf{UCB}^{(1)} (\Omega_{k=2048}^{-1}; \alpha, \alpha^{\dag}) \right\} \geq 1-\alpha^{\dag}
\end{equation}
Recall from row 1, column 2 of the right panel of \Cref{tab:intro}, that the actual Monte Carlo coverage of the nominal 90\% interval $\hat{\psi}_{1} \pm 1.64\widehat{\se} (\hat{\psi}_{1})$ was 0\%. As expected, our nominal 90\% upper confidence bounds $\mathsf{UCB}^{(1)}(\Omega_{k = 2048}^{-1}; \alpha, \alpha^{\dag})$ and $\mathsf{UCB}^{(1)}(\widehat{\Omega}_{k=2048}^{-1}; \alpha, \alpha^{\dag})$ were nearly 0\% in all the 100 simulated estimation samples.

\subsubsection*{Organization of the paper}

The remainder of the paper is organized as follows. In \Cref{sec:param} to \Cref{sec:drml2} we describe our data structure, our parameters of interest $\psi$, the state of the art DRML estimators, and the statistical properties of these estimators.

In \Cref{sec:intro}, we present a second order U-statistic $\widehat{\mathbb{IF}}_{22, k}$ that is an unbiased estimator of the `estimable' part of the bias of $\hat{\psi}_{1}$.

In \Cref{sec:var} and \Cref{sec:cov}, we develop $\alpha^{\dag}$ level tests that have the ability to detect whether the bias of $\hat{\psi}_{1}$ is of the same or greater order than its standard error, for the expected conditional variance; in the case of the expected conditional covariance we test whether the Cauchy-Schwarz bias is the same or greater than the standard error of $\hat{\psi}_{1}$.

In \Cref{sec:simulation} and Supplementary Materials \citep{hot_supp}, we propose an estimator $\widehat{\Omega}_{k}^{-1}$ of $\Omega_{k}^{-1}$ which performs well in simulations but lacks theoretical guarantees .

In \Cref{sec:hierarchy}, we consider a semisupervised setting with $k > n$, based on the following motivation. The estimator $\hat{\psi}_{2, k} = \hat{\psi}_{1} - \widehat{\mathbb{IF}}_{22, k}$ of $\psi = \BE[\cov(Y, A | X)]$ with $k$ less than but near $n$ has standard error not much larger than the standard error of $\hat{\psi}_{1}$, but has smaller bias. This suggests foregoing the estimation of an upper bound on the actual coverage of a nominal $(1 - \alpha)$ Wald CI centered at $\widehat{\psi }_{1}$; rather always report, with $\Omega_{k}^{-1}$ known, the nominal $(1 - \alpha)$ Wald CI $\hat{\psi}_{2, k} \pm z_{\alpha / 2} \widehat{\se} (\hat{\psi}_{2, k})$ with $k$ just less than $n$. However doing so naturally raises the question of whether the interval $\hat{\psi}_{2, k} \pm z_{\alpha / 2} \widehat{\se} (\hat{\psi}_{2, k})$ itself covers $\psi$ at its nominal $1 - \alpha$ rate. In \Cref{sec:hierarchy} we develop a test of the null hypothesis that the ratio of the conditional bias of $\hat{\psi}_{2, k}$ to its standard error is smaller than a fraction $\delta$ using an AFECT statistic based on $\widehat{\mathbb{IF}}_{22, k'}$ with $k' > n$.

In \Cref{sec:discussion}, we conclude by discussing several open problems.

The following common asymptotic notations are used throughout the paper: $x \lesssim y$ (equivalently $x = O(y)$) denotes that there exists some constant $C > 0$ such that $x \leq C y$, $x \asymp y$ means there exist some constants $c_{1} > c_{2} > 0$ such that $c_{2} |y| \leq |x| \leq c_{1} |y|$. $x = o(y)$ or $y \gg x$ is equivalent to $\lim_{x, y \rightarrow \infty}\frac{x}{y} = 0$. For a random variable $X_{n}$ with law $P$ possibly depending on the sample size $n$, $X_{n} = O_{P} (a_{n})$ denotes that $X_{n} / a_{n}$ is bounded in $P$-probability, and $X_{n} = o_{P} (a_{n})$ means that $\lim_{n \rightarrow \infty} P(|X_{n} / a_{n}| \geq \epsilon) = 0$ for every positive real number $\epsilon$.

\subsection{Parameter of interest}\label{sec:param} 
In this part we begin to make precise the issues discussed above. For didactic purposes, we will restrict our discussion to the variance-weighted average treatment effect (variance weighted ATE, defined below) for a binary treatment $A$ and binary outcome $Y$ given a vector $X$ of $d$-dimensional baseline covariates compactly supported in $[0, 1]^{d}$. We suppose we observe $N$ iid copies from the joint distribution of $(Y, A, X)$. 

We parametrize the joint distribution $P_{\theta}$ of $(Y, A, X)$ by the variation independent parameters $\theta \equiv \left(b, p, f_{X}, \mathit{OR}_{Y A | X = x} \right)$, where, 
\begin{eqnarray*}
b(X) &\equiv & \BE_{\theta} \left[ Y | X \right] \\
p(X) &\equiv & \BE_{\theta} \left[ A | X \right]
\end{eqnarray*}
are respectively the regression of $Y$ on $X$ and the regression of $A$ on $X$, $f_{X}$ is the marginal density of $X$, and $\mathit{OR}_{Y A | X = x}$ is the conditional odds ratio. We let $\hat{\theta} = (\hat{b},\hat{p},\theta \setminus \{b, p\})$. Throughout the paper, we use $\BE_{\theta}$, $\var_{\theta}$ and $\cov_{\theta}$ with subscript $\theta$ to indicate the conditional expectation, variance, and covariance, given the training sample, under the probability measure $P_{\theta}$ indexed by $\theta$. We assume a nonparametric infinite dimensional model $\mathcal{M} \left( \Theta \right) \coloneqq \left\{ P_{\theta}, \theta \in \Theta \right\} $ where $\Theta$ indexes all possible $\theta$ subject to weak regularity conditions given later in \Cref{cond:w}.

Under the assumption that the vector $X$ of measured covariates suffices to control confounding, the variance weighted ATE $\tau \left( \theta \right)$ is identified as $\tau \left( \theta \right) \coloneqq \frac{\BE_{\theta} [\gamma_{\theta} (X) \var_{\theta} \left( A | X\right) ]}{\BE_{\theta} \left[ \var_{\theta} \left( A  | X \right) \right]}$ where $\gamma_{\theta} \left( X \right) \equiv \BE_{\theta} \left[ Y | A = 1, X \right] - \BE_{\theta} \left[ Y | A = 0, X \right]$ is the conditional treatment effect given $X$ and $\var_{\theta} \left( A | X \right) = p \left( X \right) \left( 1 - p \left( X \right) \right)$. In applications, the variance weighted ATE arises when we want to down-weight the subjects whose propensity scores are extreme. Moreover, the parameter $\tau(\theta)$ can also be identified as the regression coefficient of $A$\footnote{$A$ does not need to be binary.} in the classical semiparametric partially linear model $Y = \tau A + b(X) + \mathsf{noise}$.

Some algebra shows that 
\begin{equation*}
\tau \left( \theta \right) = \frac{\BE_{\theta} \left[ \cov_{\theta} \left( Y, A | X \right) \right]}{\BE_{\theta} \left[ \var_{\theta} \left( A | X \right) \right]}.
\end{equation*}
Henceforth, we shall restrict attention to the estimation of the expected conditional covariance 
\begin{equation*}
\psi \left( \theta \right) \equiv \BE_{\theta} \left[ \cov_{\theta} \left( Y, A | X \right) \right] = \BE_{\theta} \left[ \left\{ Y - b(X) \right\} \left\{ A - p(X) \right\} \right].
\end{equation*}
and the expected conditional variance $\BE_{\theta} \left[ \var_{\theta} \left( A | X \right) \right]$, which is simply the special case of $\BE_{\theta} \left[ \cov_{\theta} \left( Y, A | X \right) \right]$ in which $A = Y$ w.p.1. If we can construct asymptotically unbiased and normal estimators of $\BE_{\theta} \left[ \cov_{\theta} \left( Y, A | X \right) \right]$ and $\BE_{\theta} \left[ \var_{\theta} \left( A | X \right) \right]$, we also can construct the same for $\tau (\theta)$ by the functional delta method.

\begin{rem}
We shall see that the statistical guarantees of our bias correction methodology differ depending on whether the parameter of interest is $\BE_{\theta} \left[ \cov_{\theta} \left( Y, A | X \right) \right]$ versus $\BE_{\theta} \left[ \var_{\theta} \left( A | X \right) \right]$. In fact, the insight into our methodology offered by this difference is the reason we chose the variance weighted average treatment effect rather than the average treatment effect as the causal effect of interest in this paper.
\end{rem}

In the next section, we describe the current state-of-the-art DRML estimators $\hat{\psi}_{1}$ and $\hat{\psi}_{\cf, 1}$. They will depend on estimators $\hat{b}(x)$ and $\hat{p}(x)$ of $b(x)$ and $p(x)$, which may have been outputted by machine learning algorithms for estimating conditional means, with completely unknown statistical properties.

\begin{rem}\label{rem:consistency} 
The methods in \citet{robins2009semiparametric} and \citet{ritov2014bayesian} can be straightforwardly combined to show that, without further unverifiable assumptions(such as smoothness or sparsity assumptions that may be incorrect), for some $\sigma > 0$, no consistent $\alpha$-level test of the null hypothesis $\BE_{\theta}[\cov_{\theta}(A, Y | X)] = \sigma$ for $\sigma > 0$ versus the alternative $\BE_{\theta}[\cov_{\theta}(A, Y | X)] = \sigma + c$ for some fixed constant $c > 0$ exists, whenever some components of $X$ have a continuous distribution. Furthermore, there is no consistent estimator of the expected conditional covariance without further unverifiable assumptions. The above negative result also applies to the expected conditional variance $\BE_{\theta}[\var_{\theta}(A | X)]$.
\end{rem}

\subsection{State-of-the-art estimators $\hat{\psi}_1$ and $\hat{\psi}_{\cf, 1}$ and their asymptotic properties}\label{sec:drml1} 
The state-of-the-art DRML estimator $\hat{\psi}_{1}$ uses sample splitting, because $\hat{b}(x)$ and $\hat{p}(x)$ have unknown statistical properties and, in particular, may not lie in a so-called Donsker class (see e.g. \citet[Chapter 2]{van1996weak}) - a condition often needed for valid inference when we do not split the sample. The cross-fitting estimator $\hat{\psi}_{\mathsf{cf}, 1}$ is a DRML estimator that can recover the information lost by $\hat{\psi}_{1}$ due to sample
splitting, provided that $\hat{\psi}_{1}$ is asymptotically unbiased given the training sample.

The following algorithm defines $\hat{\psi}_{1}$ and $\hat{\psi}_{\mathsf{cf}, 1}$ for $\psi (\theta) = \BE_{\theta}[\cov_{\theta}(A, Y | X)]$ and can be easily modified for $\psi (\theta) = \BE_{\theta}[\var_{\theta}(A | X)]$:

\begin{itemize}
\item[(i)] The $N$ study subjects are randomly split into 2 parts: an estimation sample of size $n$ and a training (nuisance) sample of size $n_\mathsf{tr} = N - n$ with $n / N \approx 1 / 2$. Without loss of generality we shall assume that $i=1,\ldots,n$ corresponds to the estimation sample.

\item[(ii)] Estimators $\hat{b} (x), \hat{p} (x)$ are constructed from the training sample data using ML methods.

\item[(iii)] Compute 
\begin{equation*}
\hat{\psi}_{1} = \frac{1}{n} \sum_{i = 1}^n \left[ \{ Y_i - \hat{b}(X_i) \} \{ A_i - \hat{p}(X_i) \} \right]
\end{equation*}
from $n$ subjects in the estimation sample and 
\begin{equation*}
\hat{\psi}_{\mathsf{cf}, 1} = \left( \hat{\psi}_{1} + \overline{\hat{\psi}}_1 \right) / 2
\end{equation*}
where $\overline{\hat{\psi}}_1$ is $\hat{\psi}_{1}$ but with the training and estimation samples reversed.
\end{itemize}

\subsection{Asymptotic properties of $\hat{\psi}_1$ and $\hat{\psi}_{\cf, 1}$}\label{sec:drml2} 
The following theorems (\Cref{thm:drml_cond} and \Cref{thm:drml}) give the asymptotic properties of the estimator $\hat{\psi}_{1}$ of the expected conditional covariance, conditional on the training sample.

\begin{thm}\label{thm:drml_cond} 
Conditional on the training sample, $\hat{\psi}_{1}$ is asymptotically normal with conditional bias 
\begin{equation}
\mathsf{Bias}_{\theta} (\hat{\psi}_{1}) \coloneqq \mathbb{E}_{\theta} \left[ \hat{\psi}_{1} - \psi (\theta) \right] = \mathbb{E}_{\theta} \left[ \left\{ b(X) - \hat{b}(X) \right\} \left\{ p(X) - \hat{p}(X) \right\} \right]. \label{cbias}
\end{equation}
\end{thm}

\begin{proof}
Since conditionally $\hat{b}(x)$ and $\hat{p}(x)$ are fixed functions, $\hat{\psi}_{1}$ is the sum of i.i.d. bounded random variables and thus is asymptotically normal. A straightforward calculation shows $\Bias_\theta ( \hat{\psi}_{1} )$ is the conditional bias.
\end{proof}

We note that $\hat{\psi}_{1}$ is, by definition, doubly robust because $\mathsf{Bias}_{\theta} (\hat{\psi}_{1}) = 0$ if either $b(X)=\hat{b}(X)$ or $p(X)=\hat{p}(X)$ with $P_{\theta}$-probability 1. Finally, before proceeding, we summarize the statistical properties of the DRML estimator in the following theorem, the proof of which is standard and can be found in \citet{chernozhukov2018double}. Recall that $\mathsf{Bias}_\theta ( \hat{\psi}_{1} )$ is random only through its dependence on the training sample data via $\hat{b}$ and $\hat{p}$.

\begin{thm}
\label{thm:drml} If a) $\mathsf{Bias}_\theta ( \hat{\psi}_{1} )$ is $o ( n^{-1/2} )$ and b) $\hat{b}(x)$ and $\hat{p}(x)$ converge to $b(x)$ and $p(x)$ in $L_2 \left( P_\theta \right)$, then

\begin{enumerate}
\item 
\begin{eqnarray*}
\hat{\psi}_{1} - \psi (\theta) & = & n^{-1} \sum_{i = 1}^n \mathsf{IF}_{1, i} (\theta) + o ( n^{-1/2} ) \\
\hat{\psi}_{\mathsf{cf}, 1} - \psi (\theta) & = & N^{-1} \sum_{i = 1}^N \mathsf{IF}_{1, i} (\theta) + o ( N^{-1/2} )
\end{eqnarray*}
where $\mathsf{IF}_1 (\theta) = \{ Y - b (X) \} \{ A - p (X) \} - \psi (\theta)$ is the first order influence function of $\psi (\theta)$ under $P_{\theta}$. Further $n^{1/2} ( \hat{\psi}_{1} - \psi (\theta) )$ converges conditionally and unconditionally to a normal distribution with mean zero; $\hat{\psi}_{\mathsf{cf}, 1}$ is a regular, asymptotically linear estimator; i.e. $N^{1/2} \left( \hat{\psi}_{\mathsf{cf}, 1} - \psi (\theta) \right)$ converges unconditionally to a normal distribution with mean zero and variance equal to the semiparametric variance bound $\mathsf{var}_\theta \left[ \mathsf{IF}_1 (\theta) \right]$.

\item The $(1 - \alpha)$ nominal Wald CIs (CIs) 
\begin{eqnarray*}
&& \hat{\psi}_{1} \pm z_{\alpha / 2} \widehat{\mathsf{s.e.}} [\hat{\psi}_{1}] \\
&& \hat{\psi}_{\mathsf{cf}, 1} \pm z_{\alpha / 2} \widehat{\mathsf{s.e.}} [ \hat{\psi}_{\mathsf{cf}, 1} ]
\end{eqnarray*}
are $(1 - \alpha)$ asymptotic CI for $\psi (\theta)$. Here $\widehat{\mathsf{s.e.}} [ \hat{\psi}_{1} ] = \left( \widehat{\mathsf{var}} \left[ \hat{\psi}_{1} \right] \right)^{1/2}$ with 
\begin{gather*}
\widehat{\mathsf{var}} \left[ \hat{\psi}_{1} \right] = \frac{1}{n^2} \sum_{i=1}^{n} \left[ \{ Y_i - \hat{b} (X_i) \} \{ A_i - \hat{p} (X_i) \} \right]^{2} \\
\widehat{\mathsf{var}} \left[ \hat{\psi}_{\mathsf{cf}, 1} \right] = \frac{1}{4} \left\{ \widehat{\mathsf{var}} \left[ \hat{\psi}_{1} \right] + \widehat{\mathsf{var}} \left[ \overline{\hat{\psi}}_1 \right] \right\}.
\end{gather*}
\end{enumerate}
\end{thm}

\begin{rem}
Had we chosen $\psi (\theta) = \mathbb{E}_\theta [ \mathbb{E}_\theta [ Y | A = 1, X ] ]$, the mean response of $Y$ under missing at random rather than the variance weighted ATE as our parameter of interest, the outcome regression function appearing in the first order influence function would be $\mathbb{E}_\theta [ Y | A = 1, X ]$ rather than $\mathbb{E}_\theta [ Y | X ]$ and $\hat{\psi}_{1} = \frac{1}{n} \sum_{i = 1}^{n} \frac{A_{i}}{\hat{p}(X_i)} (Y - \hat{b}(X_i)) + \hat{b}(X_i)$.
\end{rem}

\begin{rem}[Training sample squared error loss cross-validation]\label{rem:ml} 
How can we choose among the many (say, $J$) available ML algorithms if our goal is to minimize the conditional mean squared error $\mathbb{E}_\theta [(b (X) - \hat{b} (X))^2]$? One approach is to let the data decide by applying cross-validation restricted to the training sample. Specifically, we randomly split the training sample into $S$ subsamples of size $n_{\mathsf{tr}} / S$. For each subsample $s$, we fit the $j$-th ML algorithm to the other $S - 1$ subsamples to obtain outputs $\hat{b}_{s}^{(j)} (\cdot)$, for $j = 1, \dots, J$. Next we compute, for each $j$, the squared error loss $CV^{(j)} = \sum_{s = 1}^{S} CV_{s}^{(j)}$ with $CV_s^{(j)} = \sum_{i \in s} \{ Y_i - \hat{b}_s^{(j)} (X_i) \}^2$, and finally select the ML algorithm $j_{\ast} = \mathsf{arg \ min}_{j} CV^{(j)}$. Analogous results apply to the estimation of $p(X)$.
\end{rem}

\begin{rem}\label{rem:n} 
Although a standard result, \Cref{thm:drml} is of minor interest to us in this paper for several reasons. First, because of their asymptotic nature, there is no finite sample size $n$ at which any test could empirically falsify $\mathsf{Bias}_{\theta }(\hat{\psi}_{1})=o (n^{-1/2})$. Rather, as discussed in \Cref{sec:background}, our interest, instead, lies in testing and rejecting hypotheses such as, at the actual estimation sample size $n$, the actual coverage of the interval $\hat{\psi}_{1} \pm z_{\alpha /2}\widehat{\mathsf{s.e.}}[\hat{\psi}_{1}]$, conditional on the training sample, is less than a fraction $\varrho <1$ of its nominal coverage. 

Second, we make no assumptions concerning either the complexity of the unknown functions $b$ and $p$ or the statistical behavior of their ML estimators $\hat{b}$ and $\hat{p}$, our inferential statements will regard the training sample as fixed rather than random. In particular, the only randomness referred to in any theorem is that of the estimation sample. Our inferences rely on being in `asymptopia' to be able to posit that, at our estimation sample size of $n$, (1) the quantiles of the finite sample distribution of a conditionally asymptotically normal statistic (e.g. $\widehat{\mathbb{IF}}_{22,k}$ defined later in \cref{eq:if22_oracle}) are close to the quantiles of a normal and (2) the standard error estimators of $\hat{\psi}_{1}$ and $\widehat{\mathbb{IF}}_{22,k}$ are close to their true standard errors. (It will often be useful to consider the power functions of our proposed tests as a function of the sample size, which we do by taking $n \rightarrow \infty$.) 
\end{rem}

\begin{rem}\label{rem:non-asymptotic} 
Indeed, when the constants in the non-asymptotic concentration inequalities \citep{boucheron2013concentration, vershynin2018high} are explicit and can be estimated from data, then our reliance on asymptotics could be eliminated at the expense of decreased power and increased CI width. However, such finite sample bounds are beyond the scope of this paper.
\end{rem}

Before starting to explain our methodology in detail, we collect some frequently used notations.

\subsubsection*{Notations}
For a (random) vector $V$, $\Vert V \Vert_{\theta} \equiv \mathbb{E}_{\theta} [V^{\top} V]^{1/2}$ denotes its $L_{2} (P_{\theta})$ norm conditioning on the training sample, $\Vert V \Vert \equiv (V^{\top} V)^{1/2} $ denotes its $\ell_2$ norm and $\Vert V \Vert_{\infty}$ denotes its $L_{\infty}$ norm. For any matrix $A$, $\Vert A \Vert$ will be used for its operator norm. Given a $k$, the random vector $\bar{\mathsf{Z}}_{k} = \bar{\mathsf{z}}_{k}(X) = (\mathsf{z}_{1}(X), \ldots, \mathsf{z}_{k}(X))^{\top}$, $\Pi \left[ \cdot | \bar{\mathsf{Z}}_{k} \right]$ denotes the population linear projection operator onto the space spanned by $\bar{\mathsf{Z}}_{k}$ conditioning on the training sample: with $\Omega_{k} \coloneqq \mathbb{E}_{\theta} [ \bar{\mathsf{Z}}_{k} \bar{\mathsf{Z}}_{k}^{\top} ]$, $\Pi \left[ \cdot | \bar{\mathsf{Z}}_{k}^{\perp} \right] = I - \Pi \left[ \cdot | \bar{\mathsf{Z}}_{k}\right]$ is the projection onto the orthogonal complement of $\bar{\mathsf{Z}}_{k}$ in the Hilbert space $L_{2} \left( f_{X} \right)$. Hence, for a random variable $W$, 
\begin{equation}  \label{def:proj}
\Pi \left[ W | \bar{\mathsf{Z}}_{k} \right] = \bar{\mathsf{Z}}_{k}^{\top} \beta_{k, W}, \Pi \left[ W | \bar{\mathsf{Z}}_{k}^{\perp} \right] = W - \Pi \left[ W | \bar{\mathsf{Z}}_{k} \right]
\end{equation}
where $\beta_{k, W} = \Omega_{k}^{-1} \mathbb{E}_{\theta} \left[ \bar{\mathsf{Z}}_{k} W \right]$ is the vector of population regression coefficients. It should be noted that we allow selection of the vector $\bar{\mathsf{Z}}_{k}$ to depend on the training sample data (for further discussions, see \Cref{sec:discussion}). $\widehat{\Omega}_{k}^{-1}$ denotes a generic estimator of $\Omega_{k}^{-1}$. When referring to a particular estimator of $\Omega_{k}^{-1}$ (mostly in \Cref{sec:simulation}), an identifying superscript will often be attached.

We also denote the following commonly used residuals as
$$\widehat{\varepsilon }_{b,i}\coloneqq Y_{i}-\hat{b}(X_{i}), \widehat{\varepsilon }_{p,i}\coloneqq A_{i}-\hat{p}(X_{i}), \widehat{\xi }_{b,i} \coloneqq b(X_{i})-\hat{b}(X_{i}), \widehat{\xi }_{p,i}\coloneqq p(X_{i})-\hat{p}(X_{i})$$ 
for $i=1,2,\dots ,n$, where $\hat{b}$ and $\hat{p}$ are estimated from the training sample.

If $\bar{\mathsf{Z}}_{k_1}$ and $\bar{\mathsf{Z}}_{k_2}$ are vectors depending on different values of $k,$ we impose the following restriction:

\begin{customthm}{B}\label{cond:b} 
For any $k_1 < k_2 = o(n^2)$, the space spanned by $\bar{\mathsf{Z}}_{k_1}$ is a subspace of the space spanned by $\bar{\mathsf{Z}}_{k_2}$.
\end{customthm}

\begin{rem}
For example, when choosing the basis functions $\bar{\mathsf{Z}}_{k}$ from a dictionary $\mathcal{V}$ of (candidate) functions greedily, \Cref{cond:b} holds.
\end{rem}

\section{The projected conditional bias and two differences between $\mathbb{E}_{\theta}[\lowercase{\var}_{\theta}(A | X)]$ and $\mathbb{E}_{\theta}[\lowercase{\cov}_{\theta}(A, Y | X)]$}\label{sec:intro}

In the main text, following the recommendation by a referee, we only discuss an ``oracle'' procedures that assume $\Omega_{k}^{-1}$ is known, as with semisupervised data.

Let $\mathcal{V}$ be a set (i.e. dictionary) of (basis) functions of $X$ that is either countable or finite with cardinality $p > n$. Given the vector $X = \left( X_l; l = 1, \dots, d \right)$ of $d$ covariates, many choices for $\mathcal{V}$ are possible. For example, $\mathcal{V}$ could be tensor products of spline, wavelet, or local polynomial partition series (or the union of all three types) in defining $\mathcal{V}$.

We decompose $b(X) - \hat{b}(X) = \Pi [ b(X) - \hat{b}(X) | \bar{\mathsf{Z}}_{k} ] + \Pi [ b(X) - \hat{b}(X) | \bar{\mathsf{Z}}_{k}^\perp ]$ (and similarly for $p(X) - \hat{p}(X)$), where the first term is the $L_2 (P_\theta)$-orthogonal (population least squares) projection of $b(X) - \hat{b}(X)$ on the linear span of the vector $\bar{\mathsf{Z}}_{k}$ and the second term is the projection onto the orthocomplement $\bar{\mathsf{Z}}_{k}^\perp$. Specifically, following eq. \eqref{def:proj}, we have 
\begin{align}
\Pi [ b (X) - \hat{b} (X) | \bar{\mathsf{Z}}_{k} ] & = \bar{\mathsf{Z}}_{k}^{\top} \beta_{k, b - \hat{b}} = \bar{\mathsf{Z}}_{k}^{\top} \Omega_{k}^{-1} \mathbb{E}_{\theta} \left[ \bar{\mathsf{Z}}_{k} (b(X) - \hat{b}(X)) \right]  \label{eq:proj_b} \\
& = \bar{\mathsf{Z}}_{k}^{\top} \Omega_{k}^{-1} \mathbb{E}_{\theta} \left[ \bar{\mathsf{Z}}_{k} (Y - \hat{b}(X)) \right] = \bar{\mathsf{Z}}_{k}^{\top} \Omega_{k}^{-1} \mathbb{E}_{\theta} \left[ \bar{\mathsf{Z}}_{k} \hat{\varepsilon}_{b} \right],  \notag \\
\Pi [ p (X) - \hat{p} (X) | \bar{\mathsf{Z}}_{k} ] & = \bar{\mathsf{Z}}_{k}^{\top} \beta_{k, p - \hat{p}} = \bar{\mathsf{Z}}_{k}^{\top} \Omega_{k}^{-1} \mathbb{E}_{\theta} \left[ \bar{\mathsf{Z}}_{k} (p(X) - \hat{p}(X)) \right]  \label{eq:proj_p} \\
& = \bar{\mathsf{Z}}_{k}^{\top} \Omega_{k}^{-1} \mathbb{E}_{\theta} \left[ \bar{\mathsf{Z}}_{k} (A - \hat{p}(X)) \right] = \bar{\mathsf{Z}}_{k}^{\top} \Omega_{k}^{-1} \mathbb{E}_{\theta} \left[ \bar{\mathsf{Z}}_{k} \hat{\varepsilon}_{p} \right]  \notag
\end{align}
where in the second lines of the above two equations we use the definition of $b(X)$, $p(X)$, $\hat{\varepsilon}_{b}$ and $\hat{\varepsilon}_{p}$.

Then we have the following lemma that decomposes $\mathsf{Bias}_{\theta }(\hat{\psi}_{1})$ (see the LHS of \cref{cbias}).

\begin{lem}
$\mathsf{Bias}_{\theta} (\hat{\psi}_{1})$ can be decomposed into the sum of two terms $\mathsf{Bias}_{\theta, k} (\hat{\psi}_{1})$ and $\mathsf{TB}_{\theta, k} (\hat{\psi}_{1})$\footnote{The notation $\mathsf{TB}_{\theta, k} (\hat{\psi}_{1})$ was adopted because it is the so-called \textit{truncation bias} in \cite{robins2008higher}.}: 
\begin{equation}  \label{decompose}
\mathsf{Bias}_{\theta} (\hat{\psi}_{1}) \equiv \mathsf{Bias}_{\theta, k} (\hat{\psi}_{1}) + \mathsf{TB}_{\theta, k} (\hat{\psi}_{1})
\end{equation}
where we define $\mathsf{Bias}_{\theta, k} (\hat{\psi}_{1}) \coloneqq \mathbb{E}_{\theta} \left[ \left\{ \Pi [ b(X) - \hat{b}(X) | \bar{\mathsf{Z}}_{k} ] \right\} \left\{ \Pi [ p(X) - \hat{p}(X) | \bar{\mathsf{Z}}_{k} ] \right\} \right]$. Then 
\begin{equation}\label{cbiastb}
\begin{split}
\Bias_{\theta, k} (\hat{\psi}_{1}) & = \beta_{k, b - \hat{b}}^{\top} \Omega_{k} \beta_{k, p - \hat{p}} \equiv \mathbb{E}_{\theta} \left[ \hat{\varepsilon}_{b} \bar{\mathsf{Z}}_{k} \right]^{\top} \Omega_{k}^{-1} \BE_{\theta} \left[ \bar{\mathsf{Z}}_{k} \hat{\varepsilon}_{p} \right], \\
\mathsf{TB}_{\theta, k} (\hat{\psi}_{1}) & = \mathbb{E}_{\theta} \left[ \left\{ \Pi [ b(X) - \hat{b}(X) | \bar{\mathsf{Z}}_{k}^{\perp} ] \right\} \left\{ \Pi [ p(X) - \hat{p}(X) | \bar{\mathsf{Z}}_{k}^{\perp} ] \right\} \right].
\end{split}
\end{equation}
\end{lem}

\begin{proof}
By definition,
\begin{align}
\Bias_{\theta, k} (\hat{\psi}_{1}) & \coloneqq \BE_{\theta} \left[ \left\{ \Pi [ b(X) - \hat{b}(X) | \Zbar_{k} ] \right\} \left\{ \Pi [ p(X) - \hat{p}(X) | \Zbar_{k} ] \right\} \right] \nonumber \\
& = \BE_{\theta} \left[ \beta_{k, b - \hat{b}}^{\top} \Zbar_{k} \Zbar_{k}^{\top} \beta_{k, p - \hat{p}} \right] = \beta_{k, b - \hat{b}}^{\top} \Omega_{k} \beta_{k, p - \hat{p}} \nonumber \\
& = \BE_{\theta} \left[ (Y - \hat{b}(X)) \Zbar_{k} \right]^{\top} \Omega_{k}^{-1} \BE_{\theta} \left[ \Zbar_{k} (A - \hat{p}(X)) \right] \label{unbias} 
\end{align}
where the last equality follows from \cref{eq:proj_b}. The second part of \cref{cbiastb} directly follows from Pythagorean theorem.
\end{proof}

We now define the {\it monotone bias class} of parameters that we mentioned in \Cref{sec:background}:
\begin{definition}[Monotone bias class of parameters]\label{monotone}
For the parameter $\psi(\theta)$, given any DRML estimator $\hat{\psi}_{1}$, under \Cref{cond:b}, if $\vert \TB_{\theta, k} (\hat{\psi}_{1}) \vert$ is nonincreasing with $k$, or equivalently if $\vert \Bias_{\theta, k} (\hat{\psi}_{1}) \vert$ is nondecreasing with $k$, $\psi(\theta)$ is said to be in the {\it monotone bias class}.
\end{definition}

\subsection{Orderings between $\mathsf{Bias}_{\theta }(\hat{\psi}_{1})$ and $\mathsf{Bias}_{\theta ,k}(\hat{\psi}_{1})$: Difference between $\mathbb{E}_{\theta }[\mathsf{cov}_{\theta }[Y,A|X]]$ and $\mathbb{E}_{\theta }[\mathsf{var}_{\theta }[A|X]]$}\label{sec:diff}
We first compare certain properties of the parameters $\BE_{\theta} [\cov_{\theta} (Y, A | X)] = \BE_{\theta} [(Y - b(X)) (A - p(X))]$ and $\BE_{\theta} [\var_{\theta} (A | X)] = \BE_{\theta} [(A - p(X))^2]$, where we note that all the earlier results and definitions concerning $\BE_{\theta} [\cov_{\theta} (Y, A | X)]$ also apply to $\psi (\theta) = \BE_{\theta} [\var_{\theta} (A | X)]$ when we everywhere substitute $A, p, \hat{p}$ for $Y, b, \hat{b}$. However, we observe a first key difference between these two parameters, which are collected in the following lemma, whose proof is trivial once we note that for $\BE_{\theta} [\var_{\theta} (A | X)]$, unlike $\BE_{\theta} [\cov_{\theta} (Y, A | X)]$, $\mathsf{Bias}_\theta (\hat{\psi}_{1}) = \mathbb{E}_\theta [ ( p(X) - \hat{p}(X) )^2 ]$, $\mathsf{Bias}_{\theta, k} ( \hat{\psi}_{1} ) = \mathbb{E}_\theta [ \{ \Pi [ p(X) - \hat{p}(X) | \bar{\mathsf{Z}}_{k} ] \}^{2} ]$ and $\mathsf{TB}_{\theta, k} ( \hat{\psi}_{1} ) = \mathbb{E}_\theta [ \{ \Pi [ p(X) - \hat{p}(X) | \bar{\mathsf{Z}}_{k}^\perp ]\}^{2} ]$ are all non-negative. We thus have the following:

\begin{lem}
\label{lem:var} \leavevmode
The following statements are true for $\psi (\theta) = \mathbb{E}_\theta [ 
\mathsf{var}_\theta [ A | X ] ]$ but not always true for $\psi (\theta) = 
\mathbb{E}_\theta [ \mathsf{cov}_\theta [ Y, A | X ] ]$:

\begin{enumerate}
[label=(\roman*)]

\item $\mathsf{Bias}_{\theta, k} ( \hat{\psi}_{1} )$ is non-decreasing in $k$
(since, by \Cref{cond:b}{}, the space spanned by $\bar{\mathsf{Z}}_{k}$
increases with $k$) and, thus, $\mathsf{TB}_{\theta, k} ( \hat{\psi}_{1} )$
is non-increasing in $k$. That is, for $k_2 > k_1$ 
\begin{align*}
0 \leq \mathsf{Bias}_{\theta, k_{1}} ( \hat{\psi}_{1} ) & \leq \mathsf{Bias}%
_{\theta, k_{2}} ( \hat{\psi}_{1} ) \leq \mathsf{Bias}_\theta (\hat{\psi}_{1}),
\\
\mathsf{TB}_{\theta, k_{1}} ( \hat{\psi}_{1} ) & \geq \mathsf{TB}_{\theta, k_{2}} ( \hat{\psi}_{1} ) \geq 0.
\end{align*}

\item $\Bias_\theta (\hat{\psi}_{2, k}) \leq \mathsf{Bias}_\theta (\hat{\psi}_{1})$.

\item For any $\delta > 0$, consider the null hypotheses 
\begin{equation}  \label{h0}
\mathsf{H}_0 (\delta): \frac{ | \mathsf{Bias}_\theta ( \hat{\psi}_{1} ) |}{%
\mathsf{s.e.}_\theta [ \hat{\psi}_{1} ]} \equiv \frac{ | \mathsf{Bias}%
_{\theta, k} ( \hat{\psi}_{1} ) + \mathsf{TB}_{\theta, k} (\hat{\psi}_{1}) | 
}{\mathsf{s.e.}_\theta [ \hat{\psi}_{1} ]} < \delta
\end{equation}
and its surrogate hypothesis 
\begin{equation}  \label{h0_k}
\mathsf{H}_{0, k} (\delta): \frac{ | \mathsf{Bias}_{\theta, k} ( \hat{\psi}%
_{1} ) |}{\mathsf{s.e.}_\theta [ \hat{\psi}_{1} ]} < \delta.
\end{equation}

If $\mathsf{H}_0 (\delta)$ (\ref{h0}) is true then the surrogate null $%
\mathsf{H}_{0, k} (\delta)$ (\ref{h0_k}) is true. Hence rejection of the
surrogate $\mathsf{H}_{0, k} (\delta)$ (\ref{h0_k}) implies rejection of $%
\mathsf{H}_0 (\delta)$ (\ref{h0}).
\end{enumerate}
\end{lem}

Thus $\psi (\theta) = \mathbb{E}_\theta [ \mathsf{var}_\theta (A | X) ]$, unlike $\psi (\theta) = \mathbb{E}_\theta [ \mathsf{cov}_\theta (A, Y | X) ]$, belongs to the {\it monotone bias class}. The null hypothesis $\mathsf{H}_{0}(\delta )$ (\ref{h0}) states that $\mathsf{Bias}_{\theta }(\hat{\psi}_{1})$ is less than a fraction $\delta $
of its standard error. In \Cref{thm:test_var} and \Cref{thm:test_cov} below,
we construct valid $\alpha ^{\dag }$-level tests for the null hypothesis $%
\mathsf{H}_{0,k}(\delta )$ (\ref{h0_k}). In \Cref{sec:null_var} and %
\Cref{sec:null_cov}, we consider the role of these null hypotheses when our
goal is to either falsify (i) an analyst's claim that the Wald confidence
interval centered at $\hat{\psi}_{1}$ has at least nominal coverage or (ii),
less ambitiously, the analyst's justification for the claim.

\begin{rem}
\label{rem:sim} The simulation study reported in \Cref{tab:intro} was for
the parameter $\psi (\theta) = \mathbb{E}_{\theta}[\mathsf{var}_{\theta}(A |
X)]$. Were it not, our claim that the observation that the bias of $\hat{\psi%
}_{2, k}$ decreases as $k$ increases as predicted by the theory developed in %
\Cref{sec:intro} would have been false. Similarly, our claim that the test $%
\widehat{\chi}_{k}^{(1)}(z_{\alpha^{\dag}}, \delta)$ is an $\alpha^{\dag}$%
-level test of $\mathsf{H}_{0} (\delta)$ \eqref{h0} would also have been
false.


In our simulation studies for the parameter $\BE_{\theta}[\mathsf{cov}_{\theta}(Y,A|X)]$ reported in \Cref{tab:smooth_kern_cov} and \Cref{tab:nonsmooth_kern_cov} in \Cref{sec:simulations}, the results were qualitatively similar to those in \Cref{tab:intro} (e.g. the MCav of $\hat{\psi}_{2, k}$ increased with $k$). However this was due to the particular data generating process used and is not always true for $\psi (\theta) = \BE_{\theta}[\mathsf{cov}_{\theta} (Y, A | X)]$.

An additional point in regard to the study reported in \Cref{tab:intro}, the
ratio of the MC bias 0.229 of $\hat{\psi}_{1}$ for $\psi (\theta )=\mathbb{E}%
_{\theta }[\mathsf{var}_{\theta }(A|X)]$ to the MCav 0.0161 of its estimated
standard error was approximately $14$. The theoretical prediction based on
rates of convergence, ignoring constants, was reasonably close (given that we
ignore unknown constants), being equal to $4.1$, calculated as follows. In
the simulation, $p(x)$ had a \Holder{} exponent $s_{p}$ of 0.25
and therefore the conditional bias $\mathbb{E}_{\theta }[\{\hat{p}%
(X)-p(X)\}^{2}]$ was of order $n^{-2s_{p}/(2s_{p}+1)}=n^{-1/3}$, because we
used a rate minimax estimator $\hat{p}(x)$ (see \Cref{sec:simulations}%
). Hence the order of the bias over the standard error is $%
n^{-1/3}/n^{-1/2}=n^{1/6}$, which evaluated at the sample size $n=5000$
gives $4.1=5000^{1/6}$.
\end{rem}

It follows from \Cref{rem:consistency} above that in the absence of further
assumptions, $\mathsf{TB}_{\theta, k} ( \hat{\psi}_{1} )$ could be of order
1 and cannot be consistently estimated without further assumptions on $(b,
p, \hat{b}, \hat{p})$. However, it is immediate from \cref{cbiastb} that the
oracle second-order U-statistic estimator $\widehat{\mathbb{IF}}_{22,k}$%
\footnote{Following the definitions in \citet{robins2008higher}, $\widehat{\mathbb{IF}}_{22, k}$ is the unique second order influence function of $\mathsf{Bias}_{\theta, k} (\hat{\psi}_{1})$ under the law $P_{\hat{\theta}}$. But the definition of $\widehat{\mathbb{IF}}_{22, k}$ in \citet{robins2008higher} differs from that in the current paper in the sign; thus $\hat{\psi}_{2, k} \equiv \hat{\psi}_{1} - \widehat{\mathbb{IF}}_{22, k}$ would be $\hat{\psi}_{1} + \widehat{\mathbb{IF}}_{22, k}$ in \citet{robins2008higher}. We reversed the sign because it seems didactically useful to have $\widehat{\mathbb{IF}}_{22,k}$ be an unbiased estimator of $\mathsf{Bias}_{\theta, k}(\hat{\psi}_{1})$. \cite{robins2008higher} refer to $\psi (\theta) + \TB_{\theta, k} (\hat{\psi}_{1})$ as the truncated parameter} is an unbiased estimator of $\mathsf{Bias}_{\theta, k} (\hat{\psi}_{1})$ conditional on the training sample\footnote{Because it simply replaces the expectations of \cref{unbias} by U-statistics}, where 
\begin{equation}  \label{eq:if22_oracle}
\begin{split}
\widehat{\mathbb{IF}}_{22, k} \equiv \widehat{\mathbb{IF}}_{22, k} (\Omega_{k}^{-1}) \coloneqq 
\frac{1}{n(n-1)} \sum_{1 \leq i_{1} \neq i_{2} \leq n} \widehat{\mathsf{IF}}%
_{22, k, i_{1}, i_{2}} \left( \Omega_{k}^{-1} \right), \\
\widehat{\mathsf{IF}}_{22, k, i_{1}, i_{2}} \left( \Omega_{k}^{-1} \right) = %
\left[ \hat{\varepsilon}_{b} \bar{\mathsf{z}}_{k}(X) \right]_{i_{1}}^{\top}
\Omega_{k}^{-1} \left[ \bar{\mathsf{z}}_{k}(X) \hat{\varepsilon}_{p} \right]%
_{i_{2}}.
\end{split}%
\end{equation}

Thus the conditional bias of the bias corrected estimator\footnote{\label{ft:triple}We discuss in \Cref{rem:undersmoothing} the connection between $\hat{\psi}_{2, k}$ and a triple sample splitting estimator proposed in \cite{newey2018cross} for $\psi (\theta) = \mathbb{E}_{\theta}[\mathsf{var}_{\theta}(A | X)]$.} $\hat{\psi}_{2, k} \equiv \hat{\psi}_{2, k}(\Omega_{k}^{-1}) \coloneqq \hat{\psi}_{1} - \widehat{\mathbb{IF}}_{22, k}$
for $\psi (\theta)$ and conditional mean of $\hat{\psi}_{2, k}$ are 
\begin{equation}  \label{eq:psi2}
\begin{split}
\mathsf{Bias}_{\theta} (\hat{\psi}_{2, k}) & \equiv \mathbb{E}_{\theta} \left[ \hat{\psi}_{2, k} - \psi(\theta) \right] =\mathsf{TB}_{\theta, k} (\hat{\psi}_{1}) \\
\mathbb{E}_{\theta} \left[ \hat{\psi}_{2, k} \right] & = \psi(\theta) + \mathsf{TB}_{\theta, k} (\hat{\psi}_{1})
\end{split}
\end{equation}
since 
\begin{align*}
\mathbb{E}_{\theta} [\hat{\psi}_{2, k} - \psi (\theta)] & = \mathbb{E}%
_{\theta} [ \hat{\psi}_{1} - \psi(\theta) ] - \mathbb{E}_{\theta} [\widehat{%
\mathbb{IF}}_{22, k}] = \mathsf{Bias}_{\theta} (\hat{\psi}_{1}) - \mathsf{%
Bias}_{\theta, k} (\hat{\psi}_{1}) = \mathsf{TB}_{\theta, k} (\hat{\psi}%
_{1}).
\end{align*}
Thus for $\psi (\theta) = \mathbb{E}_{\theta}[\mathsf{var}_{\theta}(A | X)]$%
, we are certain that $\hat{\psi}_{2, k}$ has smaller bias than $\hat{\psi}%
_{1}$ and the bias of $\hat{\psi}_{2, k}$ decreases as we increase $k$,
following \Cref{lem:var}(i).

\subsection{Statistical properties of $\widehat{\mathbb{IF}}_{22,k}$: Another difference between $\mathbb{E}_{\theta}[\mathsf{cov}_{\theta} (Y,A|X)]$ and $\mathbb{E}_{\theta}[\mathsf{var}_{\theta} (A|X)]$}

Throughout the rest of this paper, our results require the following weak regularity conditions (\Cref{cond:w}) to hold:

\begin{customthm}{W}\label{cond:w} \leavevmode
\begin{enumerate}
\item All the eigenvalues of $\Omega_{k}$ are bounded away from 0 and $\infty$;

\item $b(X)$, $\hat{b}(X)$, $p(X)$ and $\hat{p}(X)$ are bounded with probability 1;

\item $\Vert \bar{\mathsf{Z}}_{k}^{\top} \bar{\mathsf{Z}}_{k} \Vert_\infty \leq B k$ for some constant $B > 0$, $\Vert \Pi [b - \hat{b} \vert \bar{\mathsf{Z}}_{k}] \Vert_{\infty} \leq C$ (where $\Vert \Pi [b - \hat{b} \vert \bar{\mathsf{Z}}_{k}] \equiv \bar{\mathsf{Z}}_{k}^{\top} \beta_{k, b - \hat{b}}$) and $\Vert \Pi [p - \hat{p} \vert \bar{\mathsf{Z}}_{k}] \Vert_{\infty} \leq C$ (where $\Vert \Pi [p - \hat{p} \vert \bar{\mathsf{Z}}_{k}] \equiv \bar{\mathsf{Z}}_{k}^{\top} \beta_{k, p - \hat{p}}$) for some constant $C > 0$; 
\end{enumerate}
\end{customthm}

\begin{rem}\label{rem:w} 
\Cref{cond:w}(2) was assumed to allow us to focus on important issues. We believe we should be able replace the boundedness assumption with an assumption of light tails \citep{vershynin2018high, kuchibhotla2018moving}. However, most of the existing results on U-statistics that we use, require the U-statistic kernel to be bounded.

\Cref{cond:w}(3) will only be needed in \Cref{sec:cov} when $\Omega_{k}^{-1}$ is unknown. Even though the main text only concerns the case with known $\Omega_{k}^{-1}$, we still keep this assumption to emphasize its importance in the setting where $\Omega_{k}^{-1}$ must be estimated. \Cref{cond:w}(3) holds for Cohen-Daubechies-Vial wavelet series, B-spline series, and local polynomial partition series following from \citet[Examples 3.8 - 3.10]{belloni2015some}.
\end{rem}

We have the following result regarding the statistical properties of the oracle estimator $\widehat{\IIFF}_{22,k}$ of the projected bias $\Bias_{\theta, k}(\hat{\psi}_{1})$. For notational convenience, we define the following $L_{2}(P_{\theta})$ norms: 
\begin{align*}
\mathbb{L}_{2, b, k} \coloneqq \left\{ \mathbb{E}_{\theta} \left[ \Pi [ b(X) - \hat{b}(X) | \bar{\mathsf{Z}}_{k} ]^{2} \right] \right\}^{1 / 2}, \mathbb{L}_{2, p, k} \coloneqq \left\{ \mathbb{E}_{\theta} \left[ \Pi [ p(X) - \hat{p}(X) | \bar{\mathsf{Z}}_{k} ]^{2} \right] \right\}^{1 / 2}.
\end{align*}
Note that $\BL_{2, p, k}$ is equal to $\Bias_{\theta, k} (\hat{\psi}_{1})$ when $\psi (\theta) = \BE_{\theta}[\var_{\theta}(A | X)]$.

\begin{thm}
\label{thm:soif} Under \Cref{cond:w}, with $k$ , $n\rightarrow \infty$, and $k=o(n^{2})$, conditional on the training sample, we have

(i) $\widehat{\IIFF}_{22,k}$ is unbiased for $\Bias_{\theta, k} (\hat{\psi}_{1})$ with variance of order
\begin{equation*}
\frac{1}{n} \max \left\{ \frac{k}{n}, \BL_{2, b, k}^{2}, \BL_{2, p, k}^{2} \right\},
\end{equation*}
where $\BL_{2, b, k}$ and $\BL_{2, p, k}$ are defined above.

(ii) $\frac{\widehat{\IIFF}_{22, k} - \Bias_{\theta, k} (\hat{\psi}_{1})}{\se_{\theta} [\widehat{\IIFF}_{22,k}]}$ converges in law to a standard normal $N(0,1)$. Further, $\se_{\theta} [\widehat{\IIFF}_{22, k}] \coloneqq \var_{\theta}^{1/2} [\widehat{\mathbb{IF}}_{22,k}]$ can be estimated by $\widehat{\se}[\widehat{\mathbb{IF}}_{22,k}] \coloneqq \widehat{\var}^{1/2}[\widehat{\IIFF}_{22,k}]$ defined in \Cref{sec:est.var} satisfying $\frac{\widehat{\se}[\widehat{\IIFF}_{22, k}]}{\mathsf{s.e.}_{\theta }[\widehat{\mathbb{IF}}_{22,k}]} = 1 + o_{P_{\theta}} (1)$.

(iii) $\widehat{\IIFF}_{22, k} \pm z_{\alpha^{\dag} / 2} \widehat{\se} [ \widehat{\IIFF}_{22, k} ]$ (resp. $[\widehat{\IIFF}_{22, k} - z_{\alpha^{\dag}} \widehat{\se} [ \widehat{\IIFF}_{22, k} ], \infty)$) is a $(1 - \alpha^{\dag})$ asymptotic two-sided (resp. one-sided) Wald CI for $\Bias_{\theta, k} (\hat{\psi}_{1})$ with length of order 
\begin{equation*}
\dfrac{1}{\sqrt{n}} \max \left\{ \sqrt{\dfrac{k}{n}}, \mathbb{L}_{2, b, k}, \mathbb{L}_{2, p, k} \right\}.
\end{equation*}
\end{thm}

\begin{proof}
The variance order of $\widehat{\IIFF}_{22, k}$ is proved in \Cref{sec:est.var}. When $k = o(n^2)$ and $k \rightarrow \infty$ as $n \rightarrow \infty$, the conditional asymptotic normality of $\frac{\widehat{\IIFF}_{22, k} - \Bias_{\theta, k} (\hat{\psi}_{1})}{\se_{\theta} [\widehat{\IIFF}_{22, k}]}$ follows directly from Hoeffding decomposition, with the conditional asymptotic normality of the degenerate second-order U-statistic part implied by \citet[Corollary 1.2]{bhattacharya1992class}.
\end{proof}

\begin{rem}
\label{rem:diff} Now we consider a second key difference between the
parameters $\mathbb{E}_{\theta }[\mathsf{var}_{\theta}(A|X)]$ and $\mathbb{E}_{\theta
}[\mathsf{cov}_{\theta}(A,Y|X)]$. It follows from \Cref{thm:soif}(i)that for $\mathbb{%
E}_{\theta }[\mathsf{cov}_{\theta}(A,Y|X)]$, 
\begin{equation*}
\mathsf{var}_{\theta }[\widehat{\mathbb{IF}}_{22,k}]=O\left( \frac{1}{n}%
\left\{ \frac{k}{n}+\mathbb{L}_{2,b,k}^{2}+\mathbb{L}_{2,p,k}^{2}\right\}
\right) ,
\end{equation*}%
whereas for $\mathbb{E}_{\theta }[\mathsf{var}_{\theta}(A|X)]$, 
\begin{equation*}
\mathsf{var}_{\theta }[\widehat{\mathbb{IF}}_{22,k}]=O\left( \frac{1}{n}%
\left\{ \frac{k}{n}+\mathbb{L}_{2,p,k}^{2}\right\} \right) .
\end{equation*}%
For $\BE_{\theta }[\var_{\theta}(A|X)]$, when $\mathbb{L}_{2,p,k}^{2}=O (n^{-1/2})$, with $k=o(n)$, we always have $\mathsf{var}_{\theta }[\widehat{\IIFF}_{22,k}]\ll n^{-1}$. However, for $\mathbb{E}_{\theta }[\cov_{\theta}(A,Y|X)]$, when $\Bias_{\theta, k} (\hat{\psi}_{1}) = O (n^{-1/2})$, $\mathbb{L}_{2,b,k}^{2}$ and $\mathbb{L}_{2,p,k}^{2}$ can still be $O (1)$, with $k = o(n)$, we then have $\var_{\theta }[\widehat{\IIFF}_{22,k}] \asymp n^{-1}$. We shall see below that the above implies the statistical behavior of tests of the hypothesis $\mathsf{H}_{0,k}(\delta )$ differ for $\BE_{\theta }[\var_{\theta}(A|X)]$and $\BE_{\theta }[\cov_{\theta}(A,Y|X)]$.
\end{rem}

\begin{rem}\label{rem:normal} 
The qqplots in the left panel of \Cref{fig:normal} (see \Cref{app:norm}) provide empirical evidence that, in our simulation experiments, in \Cref{sec:simulations}, the quantiles of $\widehat{\IIFF}_{22, k} / \se_{\theta} [\widehat{\IIFF}_{22, k}]$ are close to normal quantiles.
\end{rem}

\begin{rem}
\label{rem:k1} When $k$ is of order greater than or equal to $n^2$, the
conditional asymptotic normality of $\frac{\widehat{\mathbb{IF}}_{22, k} - 
\mathsf{Bias}_{\theta, k} (\hat{\psi}_{1})}{\mathsf{s.e.}_{\theta} [\widehat{%
\mathbb{IF}}_{22, k}]}$ does not hold. Moreover, when $k \gg n^2$, $\mathsf{%
var}_\theta [ \widehat{\mathbb{IF}}_{22, k} ] \asymp \frac{k}{n^2}$ is of
order greater than 1, and therefore $\widehat{\mathbb{IF}}_{22, k}$ is not
consistent for $\mathsf{Bias}_{\theta, k} (\hat{\psi}_{1})$ even if $\mathsf{%
Bias}_{\theta, k} (\hat{\psi}_{1})$ is of order 1. As mentioned in %
\Cref{sec:background}, when $k$ is bounded (not growing with $n$), after
standardization $\widehat{\mathbb{IF}}_{22, k}$ converges to a Gaussian
chaos distribution instead of a normal distribution, conditional on the
training sample.
\end{rem}


\section{The null hypothesis and an oracle test for $\mathbb{E}_{\theta} \left[ \lowercase{\var}_{\protect\theta}(A | X) \right]$}\label{sec:var}

\subsection{The null hypothesis}\label{sec:null_var} 
We next consider the implications of rejection of the null hypothesis $\mathsf{H}_{0,k}(\delta )$ in the case of $\psi (\theta) = \mathbb{E}_{\theta }[\mathsf{var}_{\theta }(A|X)]$. In \Cref{sec:null_cov}, we extend this discussion to $\psi (\theta) = \mathbb{E}_{\theta }[\mathsf{cov}_{\theta }(A,Y|X)]$. We shall require the following elementary lemma, which follows from the conditional asymptotic normality of $\hat{\psi}_{1}$ in \Cref{thm:drml}.

\begin{lem}
If $\frac{\Bias_{\theta} (\hat{\psi}_{1})}{\se_{\theta} (\hat{\psi}_{1})} = \delta$, the actual asymptotic coverage of a two-sided $(1 - \alpha)$ Wald CI $\hat{\psi}_{1} \pm z_{\alpha / 2} \widehat{\se}[\hat{\psi}_{1}]$ for $\psi (\theta)$ is
\begin{equation}  \label{eq:tc}
\TC_\alpha (\delta) \coloneqq \Phi ( z_{\alpha / 2} - \delta) - \Phi ( - z_{\alpha / 2} - \delta).
\end{equation}
\end{lem}
 
\begin{figure}[h]
\caption{$\mathsf{TC}_\alpha (\delta) \equiv \Phi ( z_{\alpha / 2} - \delta) - \Phi ( - z_{\alpha / 2} - 
\delta)$ as a function of $\delta$ over several different $\alpha$'s}
\label{fig:delta}\centering
\includegraphics[width=0.5\textwidth]{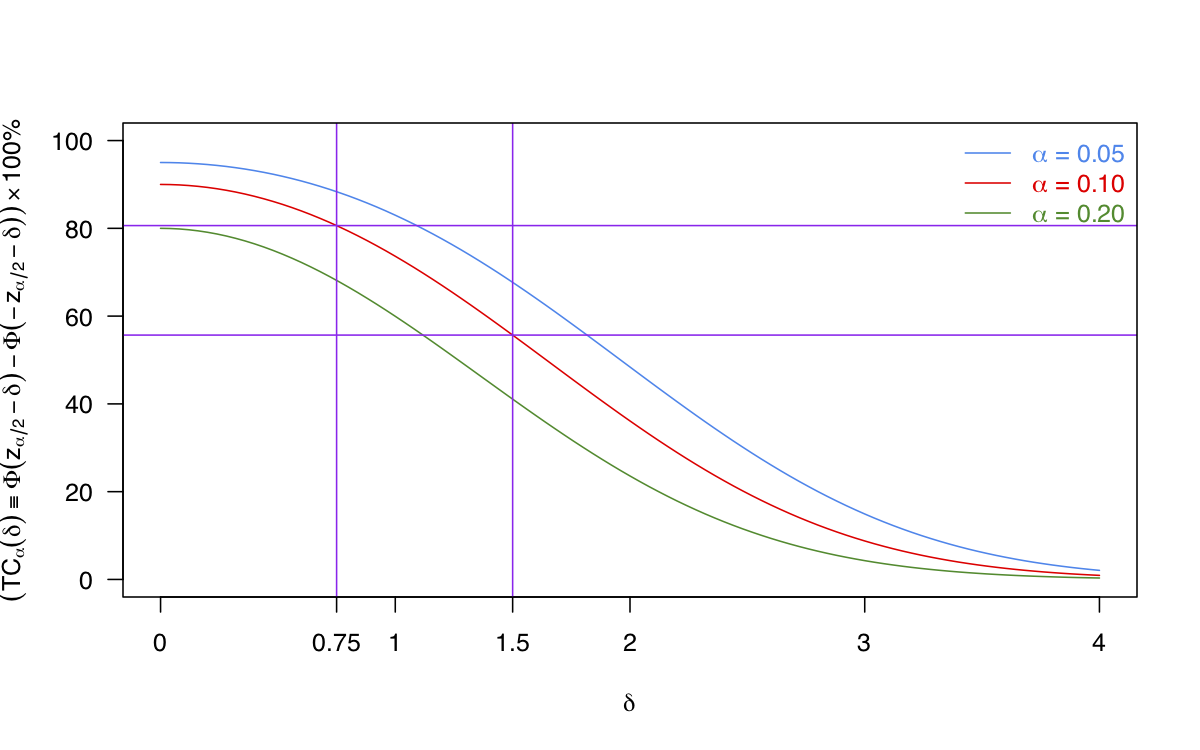}
\end{figure}

The dependence of $\TC_{\alpha} (\delta)$ on $\delta$ for several $\alpha$ is shown in \Cref{fig:delta}. It follows that if $\H_{0} (\delta)$ is false, the true coverage rate is no more than $\TC_{\alpha} (\delta)$. It follows that $\H_{0}(\delta)$ is equivalent to the null hypothesis that the actual asymptotic coverage (given the training sample) of $\hat{\psi}_{1} \pm z_{\alpha / 2} \widehat{\se}[\hat{\psi}_{1}]$ for $\psi (\theta)$ is greater than or equal to $\TC_{\alpha} (\delta)$. This result holds for both $\psi (\theta) = \mathbb{E}_{\theta }[\mathsf{cov}_{\theta}(A,Y|X)]$ and $\psi(\theta) = \mathbb{E}_{\theta }[\mathsf{var}_{\theta}(A|X)]$. For $\psi (\theta) = \mathbb{E}_{\theta }[\mathsf{var}_{\theta }(A|X)]$, but not for $\psi (\theta) = \mathbb{E}_{\theta}[\mathsf{cov}_{\theta }(A,Y|X)]$, if $\mathsf{H}_{0,k}(\delta)$ is false, and therefore $\mathsf{H}_{0}(\delta)$ is false, the true coverage rate is no more than $\TC_{\alpha} (\delta)$.

In \Cref{thm:test_var} below, we construct an asymptotically level $%
\alpha^{\dag}$ test for the surrogate null hypothesis $\mathsf{H}%
_{0,k}(\delta)$, which by \Cref{lem:var}(iii) is also an asymptotically
level $\alpha^{\dag}$ test of $\mathsf{H}_{0}(\delta)$ for $\psi (\theta) =%
\mathbb{E}_{\theta}[\mathsf{var}_{\theta}(A | X)]$ but not for $\psi
(\theta) = \mathbb{E}_{\theta}[\mathsf{cov}_{\theta}(Y,A|X)]$. Thus, one
might reasonably ask whether our methods are useful for inference concerning
the parameter $\mathbb{E}_{\theta}[\mathsf{cov}_{\theta}(Y,A|X)]$, a
question to which we return in \Cref{sec:cov}.

\subsection{An oracle test}
Based on the statistical properties of $\hat{\psi}_{1}$
and $\widehat{\IIFF}_{22, k}$ summarized in \Cref{thm:drml} and %
\Cref{thm:soif}, for $\psi(\theta) \coloneqq \BE_{\theta} [\var_{\theta} (A | X)]$,
we now consider the properties of the following one-sided test $\widehat{\chi%
}_{k}^{(1)} (\zeta_{k}, \delta)$ of the surrogate null $\mathsf{H}_{0,k}(\delta)$: 
\begin{equation}
\widehat{\chi}_{k}^{(1)} (\zeta_{k}, \delta) \equiv \widehat{\chi}_{k}^{(1)}
(\Omega_{k}^{-1}; \zeta_{k}, \delta) \coloneqq \mathbbm{1} \left\{ \frac{%
\widehat{\IIFF}_{22, k}}{\widehat{\mathsf{s.e.}}[\hat{\psi}_{1}]} -
\zeta_{k} \frac{\widehat{\mathsf{s.e.}}[\widehat{\IIFF}_{22, k}]}{%
\widehat{\mathsf{s.e.}}[\hat{\psi}_{1}]} > \delta \right\},  \label{oneside}
\end{equation}
for user-specified $\zeta_{k}, \delta > 0$. We use a one-sided test because the sign of $\Bias_{\theta, k}(\hat{\psi}_{1}) \geq 0$ is known \textit{a priori}.

The following theorem characterizes the asymptotic level and power of the oracle one-sided test $\widehat{\chi}_{k}^{(1)} (\zeta _{k}, \delta)$ of the surrogate null $\mathsf{H}_{0,k}(\delta)$ when $\psi (\theta) = \BE_{\theta}[\var_\theta(A | X)]$.

\begin{thm}\label{thm:test_var} 
For $\psi(\theta) = \BE_{\theta}[\mathsf{var}_\theta(A | X)]$, under \Cref{cond:w}, when $k \rightarrow \infty$ but $k =
o(n)$, for any given $\delta, \zeta_{k} > 0$, suppose that $\frac{\mathbb{L}_{2, p, k}^{2}}{\mathsf{s.e.}_\theta [\hat{\psi}_{1}]} = \frac{\mathsf{Bias}_{\theta, k} (\hat{\psi}_{1})}{\mathsf{s.e.}_\theta [\hat{\psi}_{1}]} = \gamma$ for some (sequence) $\gamma = \gamma(n)$ (where $\gamma(n)$ can diverge with $n$), then the rejection probability of $\widehat\chi_{k}^{(1)}
(\zeta_{k}, \delta)$ converges to 
\begin{align}
1 - \Phi \left( \zeta_{k} - \lim_{n \rightarrow \infty} (\gamma - \delta) \frac{\mathsf{s.e.}_\theta [\hat{\psi}_{1}]}{\mathsf{s.e.}_\theta [\widehat{\mathbb{IF}}_{22, k}]} \right)  \label{rejection_var}
\end{align}
as $n \rightarrow \infty$. In particular,

\begin{enumerate}
[label=(\arabic*)]

\item under $\mathsf{H}_{0, k} (\delta): \gamma \leq \delta$, $\widehat{\chi}_{k}^{(1)}(\zeta _{k}, \delta)$ rejects the null with probability less than or equal to $1 - \Phi(\zeta_{k})$, as $n \rightarrow \infty$;

\item under the following alternative to $\mathsf{H}_{0, k} (\delta)$: $\gamma = \delta + c$, for any fixed $c > 0$ or any diverging sequence $c = c(n) \rightarrow \infty$, $\widehat{\chi}_{k}^{(1)} (\zeta_{k}, \delta)$ rejects the null with probability converging to 1, as $n \rightarrow \infty$.
\end{enumerate}
\end{thm}

\begin{rem}\leavevmode\label{rem:test_var} 
In \Cref{app:oracletest}, we prove \cref{rejection_var}. We now prove that \cref{rejection_var} implies \Cref{thm:test_var}(1)-(2).
\begin{itemize}
\item Regarding (1), under $\mathsf{H}_{0, k} (\delta)$,
\begin{equation*}
- (\gamma - \delta) \frac{\mathsf{s.e.}_\theta [\hat{\psi}_{1}]}{\mathsf{s.e.}_\theta [\widehat{\IIFF}_{22, k}]} \geq 0,
\end{equation*}
which implies that the rejection probability is less than $1 - \Phi(\zeta_{k})$, as $n \rightarrow \infty$. Choose $\zeta_{k} = z_{\alpha^{\dag}}$, $1 - \Phi(\zeta _{k}) = 1 - \Phi(z_{\alpha^{\dag}}) = \alpha^{\dag}$ and conclude that the test is a valid level $\alpha^{\dag}$ test of the null.

\item Regarding (2), under the alternative $\gamma = \delta + c$ for some $c > 0$, it follows from \Cref{rem:diff} and \cref{rejection_var} that the rejection probability of $\widehat{\chi}_{k}^{(1)} (\zeta_{k}, \delta)$, as $n \rightarrow \infty$, is no smaller than 
\begin{align*}
1 - \Phi \left( \zeta_{k} - c \Theta(p, \hat{p}, f_{X}, \Zbar_{k}) \left\{ \dfrac{k}{n} + \BL_{2, p, k} \right\}^{-1} \right),
\end{align*}
where $\Theta(p, \hat{p}, f_{X}, \Zbar_{k})$ is some positive constant depending on the true regression function $p$, the estimated function $\hat{p}$ from the training sample, the density $f_{X}$ of $X$ and the chosen basis functions $\Zbar_{k}$. For fixed $c > 0$, $\Bias_{\theta, k} (\hat{\psi}_{1}) \equiv \BL_{2, p, k}^{2} = (\delta + c) \se_{\theta} (\hat{\psi}_{1}) = O (n^{-1/2}) = o (1)$, which implies that the power converges to $1 - \Phi(-\infty) = 1$.
\end{itemize}
\end{rem}

\Cref{thm:test_var} implies that $\widehat{\chi}_{k}^{(1)} (z_{\alpha^{\dag}}, \delta)$ is an asymptotically valid level $\alpha^{\dag}$ one-sided test of the surrogate null $\mathsf{H}_{0,k}(\delta)$. This allows us to define the following upper confidence bound that we briefly described in \Cref{sec:background}:
\begin{equation}\label{ucbone}
\UCB^{(1)} (\Omega_{k}^{-1}; \alpha, \alpha^{\dag}) \coloneqq \TC_\alpha \left( \left[ \frac{\widehat{\IIFF}_{22, k} - z_{\alpha^{\dag}} \widehat{\mathsf{s.e.}} [ \widehat{\IIFF}_{22, k} ]}{\widehat{\mathsf{s.e.}} [\hat{\psi}_{1}]} \right] \right).
\end{equation}
Given the mapping $\TC_{\alpha} (\delta)$ between $\delta$ and the minimal asymptotic coverage of a nominal $(1 - \alpha)$ two sided Wald CI centered at $\hat{\psi}_{1}$ under $\H_{0, k}(\delta)$, the following corollary is an immediate consequence of \Cref{thm:test_var}:
\begin{cor}\label{cor:ucb_var}
Under the conditions in \Cref{thm:test_var}, $\UCB^{(1)} (\Omega_{k}^{-1}; \alpha, \alpha^{\dag})$ is an asymptotically valid\footnote{Recall that the validity of a nominal $(1 - \alpha^{\dag})$ upper confidence bound is defined in \cref{def-ucb} with $P$ replaced by $P_{\theta}$. That is, $\UCB^{(1)} (\Omega_{k}^{-1}; \alpha, \alpha^{\dag})$ must be greater than the true asymptotic coverage probability of a $(1 - \alpha)$ two-sided Wald CI covering $\BE_{\theta} [\hat{\psi}_{2, k}]$ more than $(1 - \alpha^{\dag}) \times 100\%$ of the time over repeated sampling from the true data generating law $P_{\theta}$} nominal $(1 - \alpha^{\dag})$ upper confidence bound for the true coverage of a nominal $(1 - \alpha)$ two sided Wald CI centered at $\hat{\psi}_{1}$ for the parameter $\BE_{\theta} [\hat{\psi}_{2, k}] \equiv \psi (\theta) + \TB_{\theta, k} (\hat{\psi}_{1})$ when $\psi (\theta) = \BE_{\theta}[\var_{\theta}(A | X)]$.
\end{cor}

Finally, the following corollary, implied by \Cref{thm:test_var}, \Cref{cor:ucb_var} and \Cref{lem:var}, summarizes (1) the implication of $\widehat{\chi}_{k}^{(1)}(\zeta _{k}, \delta)$ on the actual null hypothesis of interest $\H_{0} (\delta)$ and (2) the implication of a nominal $(1 - \alpha^{\dag})$ upper confidence bound $\mathsf{UCB}^{(1)} (\Omega_{k}^{-1}; \alpha, \alpha^{\dag})$ on the true coverage of a nominal $(1 - \alpha)$ two-sided Wald CI centered at $\hat{\psi}_{1}$ for $\psi (\theta)$.

\begin{cor}\label{cor:test_var} 
Under the conditions in \Cref{thm:test_var}, 
\begin{itemize}
\item $\widehat{\chi}_{k}^{(1)}(\zeta _{k}, \delta)$ is an asymptotically level $1 - \Phi (\zeta_{k})$ one-sided test of $\H_{0} (\delta)$, as $n \rightarrow \infty$.
\item $\UCB^{(1)} (\Omega_{k}^{-1}; \alpha, \alpha^{\dag})$ is an asymptotically valid nominal $(1 - \alpha^{\dag})$ upper confidence bound for the true coverage of a nominal $(1 - \alpha)$ two sided Wald CI centered at $\hat{\psi}_{1}$ for $\psi (\theta) = \BE_{\theta}[\var_{\theta}(A | X)]$. That is, actual asymptotic coverage of a nominal $(1-\alpha)$ two-sided Wald CI centered at $\hat{\psi}_{1}$ is no greater than the random variable $\mathsf{UCB}^{(1)}(\Omega_{k}^{-1}; \alpha ,\alpha^{\dag})$ with probability at least $1 - \alpha ^{\dag}$.
\end{itemize}
\end{cor}

For $\psi (\theta) = \BE_{\theta} [\var_{\theta}(A | X)]$, when $\widehat{\chi}_{k}^{(1)}(\zeta _{k}, \delta)$ rejects $\mathsf{H}_{0, k} (\delta)$, we should also reject $\mathsf{H}_{0} (\delta)$. Nevertheless $\widehat{\chi}_{k}^{(1)}(\zeta _{k}, \delta)$ can be a powerless test under the alternative to $\mathsf{H}_{0} (\delta)$ for which $\mathsf{H}_{0,k}(\delta )$ holds. In fact, as
discussed earlier, $\Bias_{\theta, k} (\hat{\psi}_{1})$ may be zero and yet $\Bias_{\theta} (\hat{\psi}_{1}) = \TB_{\theta, k} (\hat{\psi}_{1})$ may be order 1, owing to the fact we are not controlling the magnitude of $\TB_{\theta, k} (\hat{\psi}_{1})$ by imposing sparsity or smoothness assumptions.

\section{The null hypothesis and an oracle test for $\mathbb{E}_{\protect%
\theta} \left[ \lowercase{\cov}_{\protect\theta}(A, Y | X) \right]$}

\label{sec:cov}

\subsection{The null hypothesis}\label{sec:null_cov} 
In this section, we turn our attention to the parameter $\psi (\theta )=\mathbb{E}_{\theta }[\mathsf{cov}_{\theta }(A,Y|X)]$. In fact the discussion in this section actually applies to any parameter $\psi (\theta)$ with a unique first order influence function depending on unknown regression functions or densities for which the absolute value $|\mathsf{TB}_{\theta ,k}(\hat{\psi}_{1})|$ of the truncation bias need not be a nonincreasing function of $k$, i.e. outside the {\it monotone bias class}. In particular it applies to the class of doubly robust functionals in \cite{robins2008higher}. Such parameters cover many causal parameters, including the average treatment effect and the effect of treatment on the treated, as well as many non-causal parameters. It is the class of parameters mentioned in the \Cref{sec:background} for which our results are unavoidably less sharp. For the {\it monotone bias class} we obtain much sharper results, as for $\psi (\theta )=\mathbb{E}_{\theta }[\mathsf{var}_{\theta }(A|X)]$ in \Cref{sec:var}.

In fact, for $\mathbb{E}_{\theta }[\mathsf{cov}_{\theta }(A,Y|X)]$ we shall have to settle for statements that are ``in dialogue'' with current practices and literature. To do so, we must return to the setting of \Cref{thm:drml} as, in current literature, authors often report a nominal $(1-\alpha )$ Wald CI $\hat{\psi}_{1}\pm z_{\alpha /2}\widehat{\mathsf{s.e.}}[\hat{\psi}_{1}]$, or, more commonly $\hat{\psi}_{\mathsf{cf},1}\pm z_{\alpha /2}\widehat{\mathsf{s.e.}}[\hat{\psi}_{\mathsf{cf},1}]$, and then appeal to \Cref{thm:drml} to support a claim that the true unconditional coverage is not less than nominal. Specifically \Cref{thm:drml} implies validity under the null hypothesis $\mathsf{Bias}_{\theta }(\hat{\psi}_{1})=o (n^{-1/2})$. The authors justification for the claim that $\mathsf{Bias}_{\theta }(\hat{\psi}_{1})=o (n^{-1/2})$ quite generally follows from making untestable complexity reducing assumptions (eg sparsity or smoothness) about the unknown nuisance regression functions appearing in the first order influence function. Even given such complexity reducing assumptions, their appeal to the asymptotic $o (n^{-1/2})$ is implicitly justified by the tacit assumption that, at their sample size of $N=2n=2n_{\mathsf{tr}}$, they are nearly in asymptopia both in regards to the estimation sample $n$ and in regards to the ratio $\mathsf{Bias}_{\theta }(\hat{\psi}_{1})/\mathsf{s.e.}_{\theta }[\hat{\psi}_{1}]$ being close to its asymptotic limit of $0$ (implied by their complexity reducing assumptions.)

However most authors fail to quantify or operationalize their claims. In line with the approach of this paper, whenever a null hypothesis is defined in terms of an asymptotic rate of convergence such as $o (n^{-1/2})$ in the training sample data, we will (1) ask the authors to specify a positive number $\delta =\delta(N)$ possibly depending on the actual sample size $N$ of their study and (2) then operationalize the asymptotic null hypothesis $\mathsf{Bias}_{\theta }(\hat{\psi}_{1})=o(n^{-1/2})$ as the null hypothesis $\mathsf{H}_{0}(\delta )$. That is, we have the operationalized pair 
\begin{eqnarray*}
\mathsf{NH}_{0} &:&\mathsf{Bias}_{\theta }\left( \hat{\psi}_{1}\right)
=o (n^{-1/2}) \\
\mathsf{H}_{0}(\delta ) &:&\frac{\left\vert \mathsf{Bias}_{\theta }(\hat{\psi%
}_{1})\right\vert }{\mathsf{s.e.}_{\theta }[\hat{\psi}_{1}]} < \delta 
\end{eqnarray*}%
by which we mean that if $\mathsf{H}_{0}(\delta )$ is (not) rejected, we, by convention, will declare $\mathsf{NH}_{0}$ (not) rejected. The authors' choice of $\delta$ depends on the degree of under coverage they are willing to tolerate. For example, if one allows the coverage of a 90\% two-sided Wald CI centered at $\hat{\psi}_{1}$ to be at least 80.6\% (or 55.6\%), then the authors choose $\delta = 0.75$ as $\TC_{\alpha = 0.1}(0.75) = 0.806$ (or choose $\delta = 1.5$ as $\TC_{\alpha = 0.1}(1.5) = 0.556$).

Similarly, we have the surrogate operationalized pair
\begin{eqnarray*}
\mathsf{NH}_{0,k} &:&\mathsf{Bias}_{\theta ,k}(\hat{\psi}_{1})=o (n^{-1/2}) \\
\mathsf{H}_{0,k}(\delta ) &:&\frac{\left\vert \mathsf{Bias}_{\theta ,k}(\hat{%
\psi}_{1})\right\vert }{\mathsf{s.e.}_{\theta }[\hat{\psi}_{1}]}<\delta .
\end{eqnarray*}

Suppose now the authors of a research paper agree that in reporting $\hat{%
\psi}_{1}\pm z_{\alpha /2}\widehat{\mathsf{s.e.}}[\hat{\psi}_{1}]$ as a $%
(1-\alpha )$ Wald CI for $\psi (\theta )=\mathbb{E}_{\theta }[\mathsf{cov}%
_{\theta }[Y,A|X]]$, their implicit or explicit null hypothesis is that $%
\mathsf{Bias}_{\theta }(\hat{\psi}_{1})$ is $o (n^{-1/2})$.
Further suppose the test $\widehat{\chi }_{k}^{(2)}(z_{\alpha ^{\dag
}},\delta )$ developed in \Cref{sec:test_cov} rejects the surrogate $\mathsf{%
H}_{0,k}(\delta )$, equivalently $\mathsf{NH}_{0,k}$. However, unlike for $%
\mathbb{E}_{\theta }[\mathsf{var}_{\theta }[A|X]]$, rejecting the surrogate $%
\mathsf{H}_{0,k}(\delta )$ does not logically imply rejecting $\mathsf{H}%
_{0}(\delta )$, equivalently $\mathsf{NH}_{0}$.

What, if anything, can be done? One approach is to adopt an additional
``faithfulness'' assumption under which rejection of the surrogate $\mathsf{%
NH}_{0, k}$ logically implies rejection of $\mathsf{NH}_0$.

\begin{customthm}{Faithfulness}
\label{cond:f} Given a fixed $k$, $\frac{\mathsf{Bias}_\theta( \hat{\psi}%
_{1} )}{\mathsf{Bias}_{\theta, k}( \hat{\psi}_{1} )} = 1 + \frac{\mathsf{TB}%
_{\theta, k} ( \hat{\psi}_{1} )}{\mathsf{Bias}_{\theta, k} ( \hat{\psi}_{1} )%
}$ is not $o (1)$.
\end{customthm}

One might find this assumption rather natural because it holds unless $%
\mathsf{TB}_{\theta ,k}(\hat{\psi}_{1})$ and $\mathsf{Bias}_{\theta ,k}(\hat{%
\psi}_{1})$ are of the same order and their leading constants sum to zero,
which seems highly unlikely to be the case. In finite samples, we can also
operationalize the above asymptotic faithfulness condition by choosing some $%
\delta ^{\prime }>0$ and imposing:

\begin{customthm}{Faithfulness$(\delta')$}
\label{cond:ff} For a given $k$, $\left\vert \frac{\mathsf{Bias}_{\theta }(%
\hat{\psi}_{1})}{\mathsf{Bias}_{\theta ,k}(\hat{\psi}_{1})}\right\vert
=\left\vert 1+\frac{\mathsf{TB}_{\theta ,k}(\hat{\psi}_{1})}{\mathsf{Bias}%
_{\theta ,k}(\hat{\psi}_{1})}\right\vert \geq \delta ^{\prime }$.
\end{customthm}

Under \Cref{cond:ff}, rejection of $\mathsf{H}_{0,k}(\delta )$ implies rejection of $\mathsf{H}_{0}(\delta \delta ^{\prime })$. If we choose $\delta' =0.15$, \Cref{cond:ff} holds unless $-1.15\leq \frac{\mathsf{TB}_{\theta ,k}(\hat{\psi}_{1})}{\mathsf{Bias}_{\theta ,k}(\hat{\psi}_{1})}\leq -0.85$. When we reject $\mathsf{H}_{0,k}(\delta )$ for some large $\delta$, say $\delta =10$, we will reject $\mathsf{H}_{0}(\delta \delta^{\prime }=1.5)$, suggesting that the true asymptotic coverage of a 90\% two-sided Wald CI should be lower than 55.6\%. To some extent, imposing \Cref{cond:f} or \Cref{cond:ff} may seem inconsistent with the goal of falsifying the validity of reported Wald CIs without unverifiable assumptions.

\subsubsection*{Cauchy Schwarz bias}
What else can be done if we are not willing to impose \Cref{cond:f} or \Cref{cond:ff}?

In what follows, we shall assume that the implicit or explicit goal in using a machine learning algorithm to learn the regression functions $b(x)$ and $p(x)$ is to construct $\hat{b}(x)$ and $\hat{p}(x)$ that (nearly) minimize the conditional mean square errors $\mathbb{E}_\theta [ \{ b(X) - \hat{b}(X) \}^2]$ and $\mathbb{E}_\theta [ \{ p(X) - \hat{p}(X) \}^2]$ over the set of functions computable by the algorithm. In fact, researchers who use the ``training sample squared-error loss cross-validation'' algorithm described in \Cref{rem:ml} are explicitly acknowledging this as their goal.

It follows that researchers who report a nominal $(1-\alpha )$ Wald CI $\hat{\psi}_{1} \pm z_{\alpha / 2} \widehat{\se} [\hat{\psi}_{1}]$ or $\hat{\psi}_{\mathsf{cf},1}\pm z_{\alpha / 2} \widehat{\se} (\hat{\psi}_{\mathsf{cf},1})$, based on a DRML estimator $\hat{\psi}_{1}$ for $\psi (\theta) = \BE_{\theta} [\cov_{\theta} (A, Y | X)]$ should naturally appeal to the following Cauchy-Schwarz (CS) null hypothesis $\mathsf{NH}_{0,CS}$ and its operationalization $\mathsf{H}_{0,CS} (\delta)$ 
\begin{equation}
\begin{split}
\mathsf{NH}_{0,CS} &: \ \mathsf{CSBias}_{\theta }(\hat{\psi}_{1}) \coloneqq \{ \mathbb{E}_{\theta }[\{b(X)-\hat{b}(X)\}^{2}]\mathbb{E}_{\theta }[\{p(X)-\hat{p}(X)\}^{2}]\}^{1/2}=o (n^{-1/2}), \\
\mathsf{H}_{0,CS}(\delta )& :\;\frac{\mathsf{CSBias}_{\theta }(\hat{\psi}_{1})}{\mathsf{s.e.}_{\theta }[\hat{\psi}_{1}]}<\delta 
\end{split}
\label{eq:csbias}
\end{equation}
as the \textit{justification} of a validity claim that the Wald CI's true coverage of $\psi (\theta)$ is (within the tolerance level set by $\delta$) nominal. The CS null hypothesis $\mathsf{NH}_{0,CS}$ is the hypothesis that the Cauchy-Schwarz (CS) bias, $\mathsf{CSBias}_{\theta }(\hat{\psi}_{1})$, is $o (n^{-1/2})$. We have the following logical orderings between the null hypotheses defined above:

\begin{lem}
\label{lem:cs_logic}\leavevmode

\begin{enumerate}
\item $\mathsf{NH}_{0, CS} \Rightarrow \mathsf{NH}_{0}$, and similarly $\mathsf{H}_{0, CS} (\delta) \Rightarrow \mathsf{H}_{0} (\delta)$;

\item $\mathsf{NH}_{0,CS}\Rightarrow \mathsf{NH}_{0,k}$ for all $k$, and similarly $\mathsf{H}_{0,CS}(\delta )\Rightarrow \mathsf{H}_{0,k}(\delta )$ for all $k$.
\end{enumerate}
\end{lem}

\begin{proof}
The first part simply follows from CS inequality. The second part follows from the derivation below:
\begin{equation}\label{eq:cs_bound}
\begin{split}
\vert \Bias_{\theta, k} (\hat{\psi}_{1}) \vert & = \left\vert \BE_{\theta} \left[ \Pi \left[ b(X) - \hat{b}(X) \vert \Zbar_{k} \right] \Pi \left[ p(X) - \hat{p}(X) \vert \Zbar_{k} \right] \right] \right\vert \\
& \leq \BL_{2, b, k} \BL_{2, p, k} \\ 
& \leq \left\{ \BE_{\theta} \left[ (b(X) - \hat{b}(X))^{2} \right] \right\}^{1 / 2} \left\{ \BE_{\theta} \left[ (p(X) - \hat{p}(X))^{2} \right] \right\}^{1 / 2} \equiv \CSBias_{\theta} (\hat{\psi}_{1})
\end{split}
\end{equation}
where the first inequality follows from CS inequality and the second inequality is a consequence of the fact that a projection contracts $L_2(P_\theta)$ norms.
\end{proof}

However the converse statements of \Cref{lem:cs_logic} are not always true: for example, $\mathsf{NH}_{0}$ may be true (and thus, by \Cref{thm:drml} the above the Wald CI centered at $\hat{\psi}_{1}$ is valid) even when the CS null hypothesis is false. Suppose we empirically falsify the \textit{justification} $\mathsf{NH}_{0,CS}$ ($\mathsf{H}_{0,CS}(\delta )$) for the null hypothesis of actual interest $\mathsf{NH}_{0}$ ($\mathsf{H}_{0}(\delta )$). Then, although logically $\mathsf{NH}_{0}$ may be true, there seems, to us, neither a substantive nor a philosophical reason to assume $\mathsf{NH}_{0}$ is true in the absence of $\mathsf{NH}_{0,CS}$. In Bayesian language, our (subjective) posterior probability that $\mathsf{NH}_{0}$ is true conditional on $\mathsf{NH}_{0,CS}$ being false is small; equivalently the rejection of $\mathsf{NH}_{0,CS}$ undermines our belief in $\mathsf{NH}_{0}$. Thus we will make the following

\begin{customthm}{CS}\label{cond:cs} 
If the CS null hypothesis $\NH_{0, CS}$ and $\H_{0, CS} (\delta)$ being true is used as the justification for the validity of the Wald interval $\hat{\psi}_{1} \pm z_{\alpha / 2} \widehat{\se} (\hat{\psi}_{1})$, but in fact are false, one should refuse to support claims whose validity rests on the truth of $\mathsf{NH}_0$ or $\mathsf{H}_{0} (\delta)$; in particular, the claims that the Wald CIs centered at $\hat{\psi}_{1}$ have true coverage greater than or equal to their nominal.
\end{customthm}

Clearly \Cref{cond:cs} will allow meaningful inferences regarding $\psi (\theta) = \mathbb{E}_{\theta }[\mathsf{cov}_{\theta }(A,Y|X)]$ only if it is possible to empirically reject the CS null hypothesis $\H_{0,CS}(\delta )$. Indeed, it follows from \Cref{lem:cs_logic}(2), that the rejection of the surrogate $\H_{0,k}(\delta )$ implies rejection of $\H_{0,CS}(\delta )$. In the next section, we will construct a test $\widehat{\chi }_{k}^{(2)}(\zeta _{k},\delta )$ that can empirically reject $\mathsf{H}_{0,k}(\delta )$, and hence reject $\mathsf{H}_{0,CS}(\delta )$ (and also reject $\mathsf{H}_{0}(\delta \delta ^{\prime })$ under \Cref{cond:ff}).

\subsection{An oracle test}\label{sec:test_cov} 
Based on the statistical properties of $\hat{\psi}_{1}$ and $\widehat{\IIFF}_{22, k}$ summarized in \Cref{thm:drml} and \Cref{thm:soif}, for $\psi(\theta) \coloneqq \BE_{\theta} [\cov_{\theta} (A, Y|X)]$, we now consider the properties of the following two-sided test $\widehat{\chi}_{k}^{(2)} (\zeta_{k}, \delta)$ for $\mathsf{H}_{0,k}(\delta)$ (\ref{h0_k}): 
\begin{equation}
\widehat{\chi}_{k}^{(2)} (\zeta_{k}, \delta) \equiv \widehat{\chi}_{k}^{(2)} (\Omega_{k}^{-1}; \zeta_{k}, \delta) \coloneqq \mathbbm{1} \left\{ \frac{\vert \widehat{\IIFF}_{22, k} \vert}{\widehat{\mathsf{s.e.}}[\hat{\psi}_{1}]} - \zeta_{k} \frac{\widehat{\mathsf{s.e.}}[\widehat{\IIFF}_{22, k}]}{\widehat{\mathsf{s.e.}}[\hat{\psi}_{1}]} > \delta \right\}, \label{twoside}
\end{equation}
for user-specified $\zeta_{k}, \delta > 0$. We use a two-sided test rather than a one-sided test because the sign of $\Bias_{\theta, k}(\hat{\psi}_{1})$ is unknown \textit{a priori}.

The following theorem characterizes the asymptotic level and power of the oracle two-sided test $\widehat{\chi}_{k}^{(2)} (\zeta _{k}, \delta)$ for $\mathsf{H}_{0, k}(\delta)$ (\ref{h0_k}) when $\psi (\theta) = \BE_{\theta}[\cov_\theta(A, Y | X)]$.

\begin{thm}\label{thm:test_cov} 
For $\psi (\theta) = \BE_{\theta}[\mathsf{cov}_\theta(A, Y | X)]$, under \Cref{cond:w}, when $k \rightarrow \infty$ but $k = o(n)$, for any given $\delta, \zeta_{k} > 0$, suppose that $\frac{\vert \Bias_{\theta, k} (\hat{\psi}_{1}) \vert}{\mathsf{s.e.}_\theta [\hat{\psi}_{1}]} = \gamma$ for some (sequence) $\gamma = \gamma(n)$ (where $\gamma(n)$ can diverge with $n$), then the rejection probability of $\widehat\chi_{k}^{(2)} (\zeta_{k}, \delta)$ converges to 
\begin{equation}\label{rejection:2}
2 - \Phi \left( \zeta_{k} - \lim_{n \rightarrow \infty} (\gamma - \delta) \frac{\mathsf{s.e.}_\theta [\hat{\psi}_{1}]}{\mathsf{s.e.}_\theta [\widehat{\IIFF}_{22, k}]} \right) - \Phi \left( \zeta_{k} + \lim_{n \rightarrow \infty} (\gamma + \delta) \frac{\mathsf{s.e.}_\theta [\hat{\psi}_{1}]}{\mathsf{s.e.}_\theta [\widehat{\IIFF}_{22, k}]} \right) 
\end{equation}
as $n \rightarrow \infty$. In particular,

\begin{enumerate}[label=(\arabic*)]
\item under $\mathsf{H}_{0, k} (\delta): \gamma \leq \delta$, $\widehat{\chi}_{k}^{(2)}(\zeta _{k}, \delta)$ rejects the null with probability less than or equal to $2 (1 - \Phi (\zeta _{k}))$, as $n \rightarrow \infty$;

\item under the following alternative to $\mathsf{H}_{0, k} (\delta)$: $\gamma = \delta + c$, for any diverging sequence $c = c(n) \rightarrow \infty$, $\widehat{\chi}_{k}^{(2)} (\zeta_{k}, \delta)$ rejects the null with probability converging to 1, as $n \rightarrow \infty$.
\end{enumerate}

\begin{enumerate}[label=(\arabic*')]
\setcounter{enumi}{1}
\item If $\hat{b}$ and $\hat{p}$ converge to $b$ and $p$ in $L_{2}(P_{\theta})$ norm, under the following alternative to $\mathsf{H}_{0, k} (\delta)$: $\gamma = \delta + c$, for any fixed $c > 0$ or any diverging sequence $c = c(n) \rightarrow \infty$, $\widehat{\chi}_{k}^{(2)} (\zeta_{k}, \delta)$ has rejection probability converging to 1, as $n \rightarrow \infty$.
\end{enumerate}
\end{thm}

\begin{rem}\leavevmode\label{rem:test_cov}
In \Cref{app:oracletest}, we prove \cref{rejection:2}. We now prove that \cref{rejection:2} implies \Cref{thm:test_cov}(1)-(2) and (2').
\begin{itemize}
\item Regarding (1), under $\mathsf{H}_{0, k} (\delta): \gamma \leq \delta$,
\begin{equation*}
- (\gamma - \delta) \frac{\mathsf{s.e.}_\theta [\hat{\psi}_{1}]}{\mathsf{s.e.}_\theta [\widehat{\IIFF}_{22, k}]} \geq 0 \text{ and } (\gamma + \delta) \frac{\mathsf{s.e.}_\theta [\hat{\psi}_{1}]}{\mathsf{s.e.}_\theta [\widehat{\mathbb{IF}}_{22, k}]} \geq 0,
\end{equation*}
which implies that the rejection probability is less than or equal to $2 - 2 \Phi(\zeta_{k})$. Choose $\zeta_{k} = z_{\alpha^{\dag} / 2}$, $2 (1 - \Phi(\zeta _{k})) = 2 \alpha^{\dag} / 2 = \alpha^{\dag}$ and conclude that the test is a valid level $\alpha^{\dag}$ test of the null.

\item \Cref{thm:test_cov}(2) and (2') are less sharp than \Cref{thm:test_var}(2) when $\psi (\theta) = \BE_{\theta}[\var_{\theta}(A | X)]$. Under the alternative to $\H_{0, k} (\delta)$ with $\gamma = \delta + c$ for some $c > 0$, it follows from \Cref{thm:soif} and \cref{rejection:2} that the rejection probability of $\widehat{\chi}_{k}^{(2)} (\zeta_{k}, \delta)$, as $n \rightarrow \infty$, is no smaller than
\begin{align*}
\begin{array}{c}
2 - \Phi \left( \zeta_{k} - c \Theta(b, p, \hat{b}, \hat{p}, f_{X}, \Zbar_{k}) \left\{ \dfrac{k}{n} + \BL_{2, p, k} + \BL_{2, b, k} \right\}^{-1} \right) - \Phi (\infty)
\end{array}
\end{align*}
where $\Theta(b, p, \hat{b}, \hat{p}, f_{X}, \Zbar_{k})$ is some positive constant depending on the true regression functions $b$ and $p$, the estimated functions $\hat{b}, \hat{p}$ from the training sample, the density $f_{X}$ of $X$ and the chosen basis functions $\Zbar_{k}$. To have power approaching 1 to reject $\mathsf{H}_{0, k} (\delta)$, we need one of the following:
\begin{itemize}
\item If one of $\mathbb{L}_{2, p, k}$ and $\mathbb{L}_{2, b, k}$ is $O(1)$, we need $c \rightarrow \infty$ to guarantee the rejection probability of $\widehat{\chi}_{k}^{(2)} (\zeta_{k}, \delta)$ to converge to $1 - \Phi(-\infty) = 1$. Hence we have \Cref{thm:test_cov}(2).
\item If $c$ is fixed, we need both $\mathbb{L}_{2, p, k}$ and $\mathbb{L}_{2, b, k}$ to be $o(1)$ to guarantee the rejection probability of $\widehat{\chi}_{k}^{(2)} (\zeta_{k}, \delta)$ to converge to $1 - \Phi(-\infty) = 1$. Note if $\hat{b}$ and $\hat{p}$ converge to $b$ and $p$ in $L_{2} (P_{\theta})$-norm, then both $\mathbb{L}_{2, p, k}$ and $\mathbb{L}_{2, b, k}$ are $o(1)$. Hence we have \Cref{thm:test_cov}(2').
\end{itemize}
\end{itemize}
\end{rem}

\Cref{thm:test_cov} implies that $\widehat{\chi}_{k}^{(2)} (z_{\alpha^{\dag} / 2}, \delta)$ is an asymptotically valid level $\alpha ^{\dag}$ two-sided test of the surrogate null $\H_{0, k} (\delta)$, and hence by \Cref{lem:cs_logic}(2) it is also an asymptotically $\alpha^{\dag}$ level test of $\H_{0, CS} (\delta)$. Thus when $\widehat{\chi}_{k}^{(2)} (z_{\alpha^{\dag} / 2}, \delta)$ rejects $\H_{0, k} (\delta)$, we also reject $\H_{0, CS} (\delta)$ and by \Cref{cond:cs}, we conclude that we have no justification for assuming the validity of the Wald CI centered at $\hat{\psi}_{1}$ (even though $\H_{0, k} (\delta)$ and $\H_{0, CS} (\delta)$ being false does not logically imply that $\H_{0} (\delta)$ is false and therefore does not logically imply a Wald CI centered at $\hat{\psi}_{1}$ is invalid).

On the other hand, $\widehat{\chi}_{k}^{(2)} (z_{\alpha^{\dag} / 2}, \delta)$ can be a powerless test for $\H_{0, CS} (\delta)$ under certain laws $P_{\theta}$: even when $\widehat{\chi}_{k}^{(2)} (z_{\alpha^{\dag} / 2}, \delta)$ fails to reject $\H_{0, k} (\delta)$ with (conditional) probability 1, $\H_{0, CS} (\delta)$ may still be false. 

Furthermore $\widehat{\chi }_{k}^{(2)} (z_{\alpha^{\dag} / 2}, \delta)$ is not an asymptotically valid level $\alpha^{\dag}$ test of $\H_{0} (\delta)$. However, if we assume \Cref{cond:ff}, then $\widehat{\chi}_{k}^{(2)} (z_{\alpha^{\dag} / 2}, \delta)$ is an asymptotically valid level $\alpha^{\dag}$ test of $\H_{0}(\delta \delta')$. But it can be a powerless test of $\mathsf{H}_{0}(\delta \delta')$: when $\widehat{\chi}_{k}^{(2)} (z_{\alpha^{\dag} / 2}, \delta)$ fails to reject $\H_{0, k} (\delta)$, $\H_{0} (\delta \delta')$ may still be false even under \Cref{cond:ff}.

Finally because $\vert \Bias_{\theta} (\hat{\psi}_{1}) \vert$ need not exceed $\vert \mathsf{Bias}_{\theta, k} (\hat{\psi}_{1}) \vert$, the concept of upper confidence bound is not particularly useful for $\psi (\theta) = \BE_{\theta} [\cov_{\theta} (A, Y | X)]$.

\begin{rem}
We have shown that it is indeed possible to empirically reject the CS null hypothesis $\mathsf{H}_{0, CS} (\delta)$ by testing $\mathsf{H}_{0, k} (\delta)$ using the two-sided test $\widehat\chi_{k}^{(2)} (\zeta_{k}, \delta)$. However, it is possible that $\mathsf{H}_{0, k} (\delta)$ is true whereas $\mathsf{H}_{0, CS} (\delta)$ is false, as we only have $\Bias_{\theta, k} (\hat{\psi}_{1}) \leq \mathsf{CSBias}_{\theta} (\hat{\psi}_{1})$ but do not have control over the gap between these two quantities without making further unverifiable assumptions on the true regression functions $b$ and $p$ and their estimators $\hat{b}$ and $\hat{p}$. This raises the question whether we can test $\mathsf{H}_{0, CS} (\delta): \frac{\mathsf{CSBias}_{\theta} (\hat{\psi}_{1})}{\se_{\theta}(\hat{\psi}_{1})} \leq \delta$ more directly by instead testing the following surrogate null hypothesis $\mathsf{H}_{0, CS, k} (\delta): \frac{\mathsf{CSBias}_{\theta, k} (\hat{\psi}_{1})}{\se_{\theta}(\hat{\psi}_{1})} \leq \delta$ where $\CSBias_{\theta, k} (\hat{\psi}_{1}) = \BL_{2, b, k} \BL_{2, p, k}$. We show in \Cref{sec:csbias} that it is still possible but we require multiple testing to increase the power to reject $\mathsf{H}_{0, CS, k} (\delta)$ when it is in fact false.
\end{rem}

\section{Testing the validity of Wald CI\lowercase{s} of $\hat{\psi}_{1}$ with $\lowercase{k} > \lowercase{n}$ for $\psi(\theta) = \BE_{\theta} [\lowercase{\var}_{\theta} (A | X)]$}\label{sec:hierarchy}
The tests developed in the previous sections restrict $k = o(n)$. In this section, we instead consider the case $k \gg n$ yet $k = o(n^2)$. We only consider the parameter $\psi(\theta) = \BE_{\theta} [\var_{\theta} (A | X)]$\footnote{The variance of $\widehat{\IIFF}_{22, k}$ is of order $k / n^{2}$ when $k \gg n$.}. Recall that $\Bias_{\theta, k} (\hat{\psi}_{1})$ is nondecreasing in $k$ (see \Cref{lem:var}) under \Cref{cond:b}. Further, when $k > n$, the variance of $\widehat{\IIFF}_{22, k}$ is always of order $k / n^{2}$ and thus increases with $k$ and exceeds the order of $\var_{\theta} (\hat{\psi}_{1})$. We exploit this bias-variance trade-off below. Although $\psi(\theta) = \BE_{\theta} [\cov_{\theta} (A, Y | X)]$ does not have a bias nondecreasing in $k$, results we obtained concerning $\psi(\theta) = \BE_{\theta} [\var_{\theta} (A | X)]$ can be extended to the parameter $\CSBias_{\theta} (\hat\psi_{1})$ discussed above and in \Cref{sec:csbias}, although we omit the details. We continue to assume that $\Omega_{k}^{-1}$ is known.

If $k_{0} = o(n),$ then, for $\psi (\theta )=\mathbb{E}_{\theta }[\mathsf{var}_{\theta }(A|X)]$, we may always prefer to report a Wald CI centered at $\hat{\psi}_{2,k_{0}}$\footnote{\label{ft:k0}Without loss of generality, we assume $\mathsf{var}_{\theta }[\widehat{\mathbb{IF}}_{22,k_{0}}] \asymp k_{0}/n^{2}$.} than one centered at $\hat{\psi}_{1}$ for the following reason: we know $\mathsf{Bias}_{\theta }(\hat{\psi}_{2,k_{0}})\leq \mathsf{Bias}_{\theta }(\hat{\psi}_{1})$ and yet the variances of $\hat{\psi}_{2,k_{0}}$ and $\hat{\psi}_{1}$ are close (i.e. of the same order). This choice naturally raises the question as to whether $\hat{\psi}_{2,k_{0}} \pm z_{\alpha /2}\widehat{\mathsf{s.e.}}[\hat{\psi}_{2,k_{0}}]$ covers $\psi (\theta )$ at its nominal level, which we operationalize as the null hypothesis $\mathsf{H}_{0,2,k_{0}}(\delta ):\frac{\mathsf{Bias}_{\theta }(\hat{\psi}_{2,k_{0}})}{\mathsf{s.e.}_{\theta }(\hat{\psi}_{2,k_{0}})}\leq \delta$. 

If $\mathsf{H}_{0,2,k_{0}}(\delta)$ is rejected, we may choose to report  $\hat{\psi}_{2,k}$ for some $k>n$ to further reduce bias at the the price of inflating the variance $\mathsf{var}_{\theta }(\hat{\psi}_{2,k}) \asymp k/n^{2}$ whose order then exceeds $\mathsf{var}_{\theta }(\hat{\psi}_{1}) \asymp 1/n$. Our goal is to find the values of $k$ for which we do not have empirical evidence that the Wald CI centered at $\hat{\psi}_{2,k}$ undercovers. We operationalize this goal as testing the null hypotheses in the following set, with cardinality $J$ bounded
\begin{align} \label{h0k}
\left\{ \H_{0, 2, k} (\delta): \frac{\Bias_{\theta} ( \hat{\psi}_{2, k} )}{\se_{\theta} (\hat{\psi}_{2, k})} = \frac{\TB_{\theta, k} (\hat{\psi}_{1})}{\se_{\theta} (\hat{\psi}_{2, k})} \leq \delta, k \in \mathcal{K}_{J} \right\}, \text{ where}
\end{align}
\begin{equation*}
\mathcal{K}_{J} \coloneqq \{ k_{0} < n < k_{1} < \ldots < k_{J - 1} = o(n^{2}): k_0 = o(n), k_{j - 1} = o(k_{j}), j = 1, \ldots, J - 1 \}.
\end{equation*}

Note that the hypotheses in the set \cref{h0k} are ordered: for any $k_{1} < k_{2} \in \mathcal{K}_{J}$, $\H_{0, 2, k_{1}} (\delta) \Rightarrow \H_{0, 2, k_{2}} (\delta)$ because $\Bias_{\theta} (\hat{\psi}_{2, k_{1}}) \geq \Bias_{\theta} (\hat{\psi}_{2, k_{2}})$ whereas $\se_{\theta} (\hat{\psi}_{2, k_{1}}) \ll \se_{\theta} (\hat{\psi}_{2, k_{2}})$. Hence if for each $k \in \mathcal{K}_{J}$ we have a level $\alpha^{\dag}_{k}$ test, the following sequential test protects the level for each hypothesis $\mathsf{H}_{0, 2, k}(\delta)$. See \citet[Proposition 1]{rosenbaum2008testing} for the proof.
\begin{definition}\label{def:seq}
Given a sequence of desired levels $\{ 0 < \alpha^{\dag}_{k} \le \frac{1}{2}, k \in \mathcal{K}_{J} \}$. For $j = 0, \cdots, J - 1$, at $k = k_j$:
\begin{itemize}
\item If the level $\alpha^{\dag}_{k}$ test of $\mathsf{H}_{0, 2, k}(\delta)$ rejects, set $k = k_{j + 1}$ and repeat.
\item Otherwise, we declare failure to reject $\mathsf{H}_{0, 2, k_{j'}}(\delta)$ for all $j' \geq j$ and stop.
\end{itemize}
\end{definition}

In particular, for any $j = 0, 1, \ldots, J - 2$, we define the following test of $\H_{0, 2, k_{j}} (\delta)$, given the desired level $\alpha^{\dag}_{k_{j}}$
\begin{align}\label{test_joint}
\widehat\chi_{2, k_j} (z_{\alpha^{\dag}_{k_{j}}}, \delta) \coloneqq \max \left\{ \widehat\chi_{2, k_j \rightarrow k'} ( z_{\alpha^{\dag}_{k_{j}} / (J - j - 1)}, \delta), k' \in \mathcal{K}^{-j}_{J} \coloneqq \mathcal{K}_{J} \setminus \{ k_0, \dots, k_j \} \right\},
\end{align}
where\footnote{We can choose $\widehat{\se} (\hat{\psi}_{2, k_{0}}) = \widehat{\se} (\hat{\psi}_{1})$ (as we have assumed $\se_{\theta}(\widehat{\IIFF}_{22, k_0}) \asymp \sqrt{k_{0}} / n \ll n^{-1/2}$ in \cref{ft:k0}) and $\widehat{\se} (\hat{\psi}_{2, k}) = \widehat{\se} (\widehat{\IIFF}_{22, k})$ for any $k \gg n$, where $\widehat{\se}(\widehat{\IIFF}_{22, k})$ is given in \Cref{thm:soif} (as $\se_{\theta}(\widehat{\IIFF}_{2, k}) \asymp \sqrt{k} / n \gg n^{-1/2}$).} 
\begin{align}\label{test_k}
\widehat\chi_{2, k_j \rightarrow k'} ( z_{\alpha^{\dag}_{k_{j}} / (J - j - 1)}, \delta) \coloneqq \mathbbm{1} \left\{ \frac{\widehat{\IIFF}_{22, k'} - \widehat{\IIFF}_{22, k_{j}}}{\widehat{\se} (\hat{\psi}_{2, k_{j}})} - z_{\alpha^{\dag}_{k_{j}} / (J - j - 1)} \frac{\widehat{\se} [\widehat{\IIFF}_{22, k'}]}{\widehat{\se} (\hat{\psi}_{2, k_{j}})} > \delta \right\}.
\end{align}

$\widehat\chi_{2, k_j} (z_{\alpha^{\dag}}, \delta)$ implicitly tests $J - j - 1$ surrogate hypotheses\footnote{We explain why we test multiple surrogate hypotheses instead of single hypothesis in \Cref{rem:multiple}} associated with the actual null hypothesis of interest $\H_{0, 2, k_{j}} (\delta)$. We choose the cutoff $z_{\alpha^{\dag} / (J - j - 1)}$ in $\widehat\chi_{2, k_j \rightarrow k'} ( z_{\alpha^{\dag} / (J - j - 1)}, \delta)$ to protect the level of $\widehat\chi_{2, k_j} (z_{\alpha^{\dag}}, \delta)$ by adjusting for multiple testing.

\begin{figure}[]
\caption{An illustration of the \textit{sequential test}.}
\label{fig:lepski2}\centering
\includegraphics[width=0.7\textwidth]{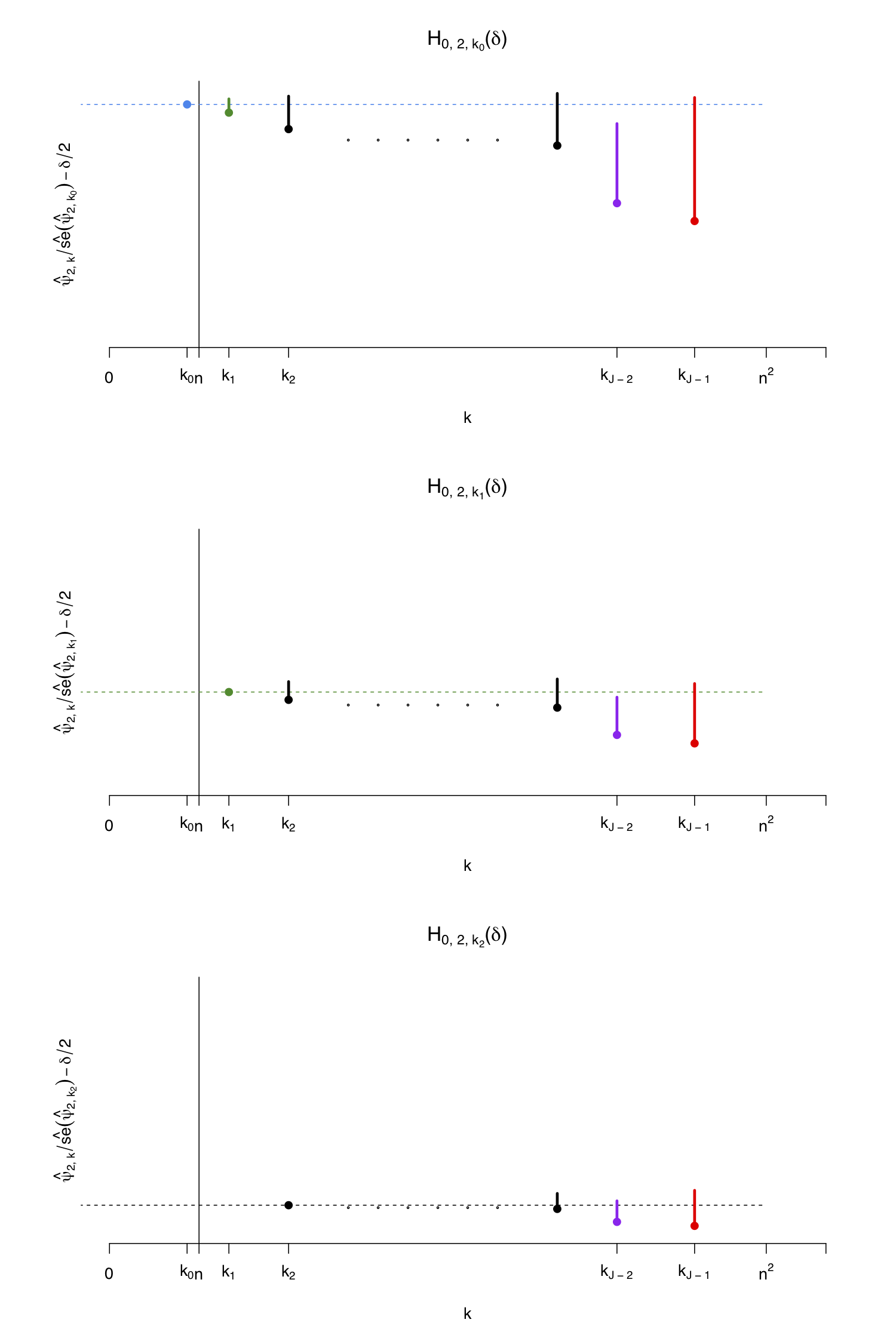} \newline
Depicted is a hypothetical data (one realization from the true data
generating law $P_{\theta}$ in which the \textit{sequential test} rejects
both $\mathsf{H}_{0, 2, k_0}(\delta)$ and $\mathsf{H}_{0, 2,
k_1}(\delta)$ but fails to reject $\mathsf{H}_{0, 2, k_2}(\delta)
$. The error bars and points are defined in \Cref{rem:heu}.
\end{figure}

\begin{rem}\label{rem:heu} 
We use \Cref{fig:lepski2} to visually illustrate the {\it sequential test} given in \Cref{def:seq} using $\widehat\chi_{2, k_j} (z_{\alpha^{\dag}}, \delta)$. We use the same level $\alpha^{\dag}$ for each $k_{j}$ in this example. \Cref{fig:lepski2} displays one hypothetical dataset drawn from $P_{\theta}$. Reading from the top ($j' = 0$) to the bottom panel ($j' = 2$):
\begin{enumerate}
\item The y-values of the points are $\frac{\hat{\psi}_{2,k_{j}}}{\widehat{\mathsf{s.e.}}[\hat{\psi}_{2,k_{j'}}]} - \frac{\delta}{2}$ for $j = j' + 1, \ldots, J - 1$. As shown in the plot, any given point moves closer to 0 from top ($j' = 0$) to bottom ($j' = 2$) because $\widehat{\se} (\hat{\psi}_{2, k_{0}}) \ll \widehat{\se} (\hat{\psi}_{2, k_{1}}) \ll \widehat{\se} (\hat{\psi}_{2, k_{2}})$ when $k_{0} \ll k_{1} \ll k_{2}$. 
\item The length of the error bar associated with $k_{j}$ is $z_{\alpha^{\dag} / (J - j' - 1)} \frac{\widehat{\mathsf{s.e.}}[\widehat{\mathbb{IF}}_{22, k_{j}}]}{\widehat{\mathsf{s.e.}}[\hat{\psi}_{2, k_{j'}}]}$, which decreases as we go from the top ($j' = 0$) to the bottom ($j' = 2$) panel. This reflects the fact that $\widehat{\se} (\hat{\psi}_{2, k_{0}}) \ll \widehat{\se} (\hat{\psi}_{2, k_{1}}) \ll \widehat{\se} (\hat{\psi}_{2, k_{2}})$ when $k_{0} \ll k_{1} \ll k_{2}$ while $z_{\alpha^{\dag} / (J - 1)} \asymp z_{\alpha^{\dag} / (J - 2)} \asymp z_{\alpha^{\dag} / (J - 3)}$.
\end{enumerate}
The {\it sequential test} for this example proceeds as follows:
\begin{itemize}
\item The upper panel of \Cref{fig:lepski2} corresponds to be the test of $\mathsf{H}_{0, 2, k_{0}}(\delta)$. The length of the error bar at each $k_{j}$ is $z_{\alpha^{\dag} / (J - 1)} \frac{\widehat{\mathsf{s.e.}}[\widehat{\mathbb{IF}}_{22, k_{j}}]}{\widehat{\mathsf{s.e.}}[\hat{\psi}_{2,k_{0}}]}$. The upper end of each error bar is $\frac{\hat{\psi}_{2,k_{j}}}{\widehat{\se}[\hat{\psi}_{2,k_{0}}]} - \frac{\delta}{2} + z_{\alpha^{\dag} / (J - 1)} \frac{\widehat{\se}[\widehat{\mathbb{IF}}_{22,k_{j}}]}{\widehat{\se}[\hat{\psi}_{2,k_{0}}]}$. If the point at $k_{0}$ (blue colored) lies outside at least one of the error bars to its right, we reject $\H_{0, 2, k_{0}}(\delta)$. This corresponds to the test $\widehat\chi_{2, k_0} (z_{\alpha^{\dag}}, \delta)$ (see \cref{test_joint}). We choose the cutoff $z_{\alpha^{\dag} / (J - 1)}$ to adjust for the $J - 1$ multiple comparisons. As shown in the plot, we reject $\mathsf{H}_{0,2,k_{0}}(\delta)$ because the blue point at $k_{0}$ is outside the error bar at $k_{J - 2}$ (purple).

\item As $\mathsf{H}_{0,2,k_{0}}(\delta)$ is rejected, we next test $\mathsf{H}_{0,2,k_{1}}(\delta)$, as shown in the middle panel of \Cref{fig:lepski2}. To test $\mathsf{H}_{0,2,k_{1}}(\delta)$, we follow the above procedure. In the middle panel, the upper end of the error bars for a given $k_{j}$ equals $\frac{\hat{\psi}_{2,k_{j}}}{\widehat{\mathsf{s.e.}}[\hat{\psi}_{2,k_{1}}]}-\frac{\delta}{2} + z_{\alpha^{\dag} / (J - 2)} \frac{\widehat{\mathsf{s.e.}}[\widehat{\mathbb{IF}}_{22,k_{j}}]}{\widehat{\mathsf{s.e.}}[\hat{\psi}_{2,k_{1}}]}$, $j = 2, \cdots, J - 1$. When $\frac{\hat{\psi}_{2,k_{1}}}{\widehat{\mathsf{s.e.}}[\hat{\psi}_{2,k_{1}}]} - \frac{\delta}{2}$ (the leftmost green point) lies outside at least one of the error bars to its right, we reject $\mathsf{H}_{0,2,k_{1}}(\delta)$. This corresponds to the test $\widehat\chi_{2, k_1} (z_{\alpha^{\dag}}, \delta)$ (see \cref{test_joint}). We reject $\mathsf{H}_{0,2,k_{1}}(\delta)$ because the green point $\frac{\hat{\psi}_{2,k_{1}}}{\widehat{\mathsf{s.e.}}[\hat{\psi}_{2,k_{1}}]}-\frac{\delta}{2}$ at $k_{1}$ is outside the error bar at $k_{J - 2}$ (purple).

\item We continue to test $\mathsf{H}_{0,2,k_{2}}(\delta)$, as shown in the lower panel of \Cref{fig:lepski2}. The upper end of the error bars for a given $k_{j}$ equals $\frac{\hat{\psi}_{2,k_{j}}}{\widehat{\mathsf{s.e.}}[\hat{\psi}_{2,k_{2}}]}-\frac{\delta}{2} + z_{\alpha^{\dag} / (J - 3)} \frac{\widehat{\mathsf{s.e.}}[\widehat{\mathbb{IF}}_{22,k_{j}}]}{\widehat{\mathsf{s.e.}}[\hat{\psi}_{2,k_{2}}]}$ for $j = 3, \cdots, J - 1$. We fail to reject $\mathsf{H}_{0,2,k_{2}}(\delta)$ because $\frac{\hat{\psi}_{2,k_{2}}}{\widehat{\mathsf{s.e.}}(\hat{\psi}_{2,k_{2}})} - \frac{\delta}{2}$ (the leftmost black point) is covered by all the error bars to its right.

\item We thus terminate the sequential test and declare failure to reject $\mathsf{H}_{0,2,k}(\delta)$ for all $k \geq k_{2}$.
\end{itemize}
\end{rem}

The result below shows that the sequential test given in \Cref{def:seq} using $\widehat\chi_{2, k_{j}} (z_{\alpha^{\dag}_{k_{j}}}, \delta)$ protects the desired level for each null hypothesis $\mathsf{H}_{0, 2, k_{j}}(\delta)$ in the set \cref{h0k}. It follows from \Cref{cor:test_{k}} below.
\begin{proposition}\label{prop:sbta} 
Under \Cref{cond:w}, for every $k_{j} \in \mathcal{K}_{J}$, $\widehat\chi_{2, k_{j}} (z_{\alpha^{\dag}_{k_{j}}}, \delta)$ is an asymptotic level $\alpha^{\dag}_{k_{j}}$ test of the null hypothesis $\mathsf{H}_{0, 2, k_{j}} (\delta)$. Consequently, the \textit{sequential test} defined in \Cref{def:seq} using $\widehat\chi_{2, k_{j}} (z_{\alpha^{\dag}_{k_{j}}}, \delta)$ is an asymptotically level $\alpha^{\dag}_{k_{j}}$ test for every individual null hypothesis $\mathsf{H}_{0, 2, k_{j}} (\delta)$ in $\mathcal{K}_{J}$.
\end{proposition}

\begin{rem}
We have assumed that $J$ is bounded for technical reasons: we need the joint conditional asymptotic normality of $\widehat{\IIFF}_{22, k}$ for $k \in \mathcal{J}$, which is not guaranteed if $J \rightarrow \infty$ as $n \rightarrow \infty$. It is possible to relax the boundedness assumption on $J$ using exponential inequalities for U-statistics rather than normality to set critical values. But to do so requires we estimate the constants in the exponential inequalities, which is left for future work.
\end{rem} 

The following result, which is a consequence of \Cref{prop:test_{k}}, summarizes the asymptotic power of the test $\widehat\chi_{2, k_j} (z_{\alpha^{\dag}_{k_{j}}}, \delta)$ when the null hypothesis $\mathsf{H}_{0, 2, k_j}(\delta)$ is false, for any given $k_{j} \in \mathcal{K}_{J}$.

\begin{proposition}\label{cor:test_{k}}\leavevmode
Under \Cref{cond:w}, for a given $j = 0, \ldots, J - 1$, let $k = k_{j}$. Given any $\delta > 0$, suppose that $\frac{\mathsf{Bias}_{\theta} (\hat{\psi}_{2, k})}{\se_{\theta} (\hat{\psi}_{2, k})} = \gamma$ for some (sequence) $\gamma \equiv \gamma(n)$ and $\frac{\mathsf{Bias}_{\theta, k'} (\hat{\psi}_{2, k})}{\mathsf{s.e.}_\theta (\hat{\psi}_{2, k})} = \gamma_{k'}$ for some (sequence) $\gamma_{k'} \equiv \gamma_{k'} (n)$\footnote{$\gamma \geq \gamma_{k'}$ for any $k' \in \mathcal{K}^{-j}_{J}$}, $\widehat\chi_{2, k} (z_{\alpha^{\dag}_{k}}, \delta)$ rejects $\mathsf{H}_{0, 2, k}(\delta)$ with probability that lies in the following interval
\begin{equation}\label{eq:rej}
\begin{split}
\left[ \begin{array}{c}
\max \left\{ 1 - \Phi \left( z_{\alpha^{\dag} / (J - j - 1)} - \lim_{n \rightarrow \infty} (\gamma_{k'} - \delta) \frac{\se_\theta (\hat{\psi}_{2, k})}{\se_\theta [\widehat{\mathbb{IF}}_{22, k'}]} \right), k' \in \mathcal{K}^{-j}_{J} \right\}, \\
\min \left\{ \sum\limits_{k' \in \mathcal{K}^{-j}_{J}} 1 - \Phi \left( z_{\alpha^{\dag} / (J - j - 1)} - \lim_{n \rightarrow \infty} (\gamma_{k'} - \delta) \frac{\se_\theta (\hat{\psi}_{2, k})}{\se_\theta [\widehat{\mathbb{IF}}_{22, k'}]} \right), 1 \right\}
\end{array} \right]
\end{split}
\end{equation} 
as $n \rightarrow \infty$. In particular, under the following alternative to $\mathsf{H}_{0, 2, k}(\delta)$: if there exists $k' \in \mathcal{K}^{-j}_{J}$ such that $\gamma_{k'} = \delta + c$ with $c \gg \sqrt{\frac{k'}{\max\{k, n\}}}$, then the test $\widehat\chi_{2, k} (z_{\alpha^{\dag}_{k}}, \delta)$ rejects the null with probability approaching 1, as $n \rightarrow \infty$.
\end{proposition}

\begin{rem}\label{rem:q1}\leavevmode
\Cref{cor:test_{k}} follows from \Cref{prop:test_{k}} (analogous to \Cref{thm:test_var}) and the definition of $\widehat\chi_{2, k} (z_{\alpha^{\dag}_{k}}, \delta)$ in \cref{test_joint}. 

In \Cref{prop:test_{k}}, we prove that for $k = k_{j}$, $\widehat\chi_{2, k \rightarrow k'} ( z_{\alpha^{\dag} / (J - j - 1)}, \delta)$ rejects the null hypothesis $\H_{0, 2, k \rightarrow k'} (\delta): \frac{\Bias_{\theta} (\hat{\psi}_{2, k_{j}}) - \Bias_{\theta} (\hat{\psi}_{2, k'})}{\se_{\theta} (\widehat{\psi}_{2, k_{j}})} \leq \delta$ with probability $$1 - \Phi \left( z_{\alpha^{\dag} / (J - j - 1)} - \lim_{n \rightarrow \infty} (\gamma_{k'} - \delta) \frac{\se_\theta (\hat{\psi}_{2, k})}{\se_\theta [\widehat{\mathbb{IF}}_{22, k'}]} \right).$$

Here $\Bias_{\theta} (\hat{\psi}_{2, k_{j}}) - \Bias_{\theta} (\hat{\psi}_{2, k'}) = \BE_{\theta} [\widehat{\IIFF}_{22, k'} - \widehat{\IIFF}_{22, k_{j}}] \geq 0$. $\H_{0, 2, k_{j} \rightarrow k'} (\delta)$ is the surrogate null hypothesis associated with $\mathsf{H}_{0, 2, k_j}(\delta)$ in the following sense (see also \Cref{lem:var_{k}}): $\mathsf{H}_{0, 2, k_j}(\delta) \Rightarrow \H_{0, 2, k_{j} \rightarrow k'} (\delta)$ for all $k' \in \mathcal{K}^{-j}_{J}$, therefore if one of $\H_{0, 2, k_{j} \rightarrow k'} (\delta)$ is false, $\mathsf{H}_{0, 2, k_j}(\delta)$ is false.

Under $\H_{0, 2, k_{j} \rightarrow k'} (\delta)$, $\widehat\chi_{2, k \rightarrow k'} ( z_{\alpha^{\dag} / (J - j - 1)}, \delta)$ rejects $\H_{0, 2, k \rightarrow k'} (\delta)$ no more than $\alpha^{\dag} / (J - j - 1)$. Under the following alternative $\gamma_{k'} - \delta \gg \sqrt{\frac{k'}{\max\{k_{j}, n\}}}$\footnote{The need for a diverging alternative is a consequence of the variance of the statistic $\frac{\widehat{\mathbb{IF}}_{22, k'}}{\widehat{\se} (\hat{\psi}_{2, k_{j}})}$ being of order $k' / \max\{k_{j}, n\}$}, $\widehat\chi_{2, k_j \rightarrow k'} ( z_{\alpha^{\dag} / (J - j - 1)}, \delta)$ rejects $\H_{0, 2, k_{j} \rightarrow k'} (\delta)$ with probability approaching 1.
\end{rem}

\section{Concluding remarks}\label{sec:discussion} 
We conclude by mentioning some open problems:

\begin{itemize}
\item We did not consider how to optimally select the basis functions $\bar{\mathsf{Z}}_{k}$ from a dictionary of $K > k$ basis functions. Data driven basis selection in the training sample has the potential of markedly increased power.

\item As mentioned in \Cref{sec:background} (also see \Cref{sec:cov} in \citet{hot_supp}), for unknown $\Omega_{k}^{-1}$, we lack theoretical guarantees as to the statistical properties of the estimators/tests that performed the best in our simulation studies.
\end{itemize}

Once these open problems are solved, we would suggest that testing the undercoverage of Wald confidence intervals centered at DRML estimators would become routine.





\section*{Acknowledgements}

We would like to thank the editor Cun-Hui Zhang, the associate editor and
the anonymous referee for their constructive comments which significantly
improve our paper. We would also like to thank Thomas M. Kolokotrones
(Harvard University), Weiming Li (Shanghai University of Finance and
Economics), Thomas S. Richardson (University of Washington), Linbo Wang (University of Toronto), Michael Wolf (ETH Zurich)
for valuable discussions.

\section*{Supplementary Materials}

In Supplementary Materials \citep{hot_supp}, we discuss estimators/tests
when $\Omega_{k}^{-1}$ is unknown, other technical details, the details of
the simulation reported in \Cref{tab:intro} and other simulation studies.

\bibliographystyle{imsart-nameyear}
\bibliography{soif}


\appendix

\section{Estimators for $\mathsf{B\lowercase{ias}}_{\lowercase{k}, \theta} (\hat{\psi}_{1})$ when $\Omega_{\lowercase{k}}^{-1}$ is unknown}\label{sec:simulation} 
In this appendix, we describe the data-adaptive test and the upper confidence bound used in the simulation studies of \Cref{sec:background} when $\Omega_{k}^{-1}$ is unknown:
\begin{align}
\widehat\chi_{k}^{(1)} (\widehat{\Omega}_{k}^{-1}; \zeta_{k}, \delta) = \mathbbm{1} \left\{ \frac{\widehat{\mathbb{IF}}_{22, k} (\widehat{\Omega}_{k}^{-1})}{\widehat{\mathsf{s.e.}} (\hat{\psi}_{1})} - \zeta_{k} \frac{\widehat{\mathsf{s.e.}} (\widehat{\mathbb{IF}}_{22, k} (\widehat{\Omega}_{k}^{-1}))}{\widehat{\mathsf{s.e.}} (\hat{\psi}_{1})} > \delta \right\}, \label{oneside-adapt} \text{ (See \Cref{tab:intro})} \\
\mathsf{UCB}^{(1)} (\widehat{\Omega}_{k}^{-1}; \alpha, \alpha^{\dag}) \coloneqq \mathsf{TC}_\alpha \left( \left[ \frac{\widehat{\mathbb{IF}}_{22, k} (\widehat{\Omega}_{k}^{-1}) - z_{\alpha^{\dag}} \widehat{\mathsf{s.e.}} [ \widehat{\mathbb{IF}}_{22, k} (\widehat{\Omega}_{k}^{-1}) ]}{\widehat{\mathsf{s.e.}} [\hat{\psi}_{1}]} \right] \right). \text{ (See \Cref{fig:ucb})}  \label{ucbone-adapt}
\end{align}
Both statistics depend on a data-adaptive estimator $\widehat{\IIFF}_{22, k} (\widehat{\Omega}_{k}^{-1})$, which we next define. At a given $k$, $\widehat{\IIFF}_{22, k} (\widehat{\Omega}_{k}^{-1})$ is equal to either $\widehat{\mathbb{IF}}_{22, k} ([\widehat{\Omega}_{k}^{\mathsf{shrink}}]^{-1})$ or $\widehat{\mathbb{IF}}_{22, k}^{\mathsf{quasi}} ([\widehat{\Omega}_{k}^{\mathsf{est}}]^{-1})$, defined as follows:
\begin{align}
\widehat{\mathbb{IF}}_{22, k} ([\widehat{\Omega}_{k}^{\mathsf{shrink}}]^{-1}) & \coloneqq \frac{(n - 2)!}{n!} \sum_{1 \leq i_{1} \neq i_{2} \leq n} \left[ \hat{\varepsilon}_{b} \bar{\mathsf{Z}}_{k} \right]_{i_{1}}^{\top} [\widehat{\Omega}_{k}^{\mathsf{shrink}}]^{-1} \left[ \bar{ \mathsf{Z}}_{k} \hat{\varepsilon}_{p} \right]_{i_{2}}  \label{eq:if22shrink} \\
\widehat{\mathbb{IF}}_{22, k}^{\mathsf{quasi}} ([\widehat{\Omega}_{k}^{\mathsf{est}}]^{-1}) & \coloneqq \frac{(n - 2)!}{n!} \sum_{1 \leq i_{1} \neq i_{2} \leq n} \left[ \hat{\varepsilon}_{b} \bar{\mathsf{Z}}_{k} \right]_{i_{1}}^{\top} Q \left( [\widehat{\Omega}_{k}^{\mathsf{est}}]^{-1}, \bar{\mathsf{Z}}_{k, i_{1}}, \bar{\mathsf{Z}}_{k, i_{2}} \right) \left[ \bar{\mathsf{Z}}_{k} \hat{\varepsilon}_{p} \right]_{i_{2}}  \label{eq:if22quasi}
\end{align}
where 
\begin{align*}
Q \left( [\widehat{\Omega}_{k}^{\mathsf{est}}]^{-1}, \bar{\mathsf{Z}}_{k, 1}, \bar{\mathsf{Z}}_{k, 2} \right) & \coloneqq [\widehat{\Omega}_{k}^{\est}]^{-1} + \frac{1}{n} [\widehat{\Omega}_{k}^{\mathsf{est}}]^{-1} \left( \bar{\mathsf{Z}}_{k, 1} \bar{\mathsf{Z}}_{k, 1}^{\top} + \bar{\mathsf{Z}}_{k, 2} \bar{\mathsf{Z}}_{k, 2}^{\top} \right) [\widehat{\Omega}_{k}^{\mathsf{est}} ]^{-1}, \\
\widehat{\Omega}_{k}^{\mathsf{est}} & \coloneqq \frac{1}{n} \sum_{i \in \mathsf{est}} \bar{\mathsf{Z}}_{k, i} \bar{\mathsf{Z}}_{k, i}^{\top}
\end{align*}
and $\widehat{\Omega}_{k}^{\mathsf{shrink}}$ is the nonlinear shrinkage covariance matrix estimator developed in \cite{ledoit2012nonlinear}, computed from the training sample. We briefly describe below how we choose between $\widehat{\mathbb{IF}}_{22, k}^{\mathsf{quasi}} ([\widehat{\Omega}_{k}^{\mathsf{est}}]^{-1})$ and $\widehat{\mathbb{IF}}_{22, k} ([\widehat{\Omega}_{k}^{\mathsf{shrink}}]^{-1})$. More details can be found in \Cref{sec:cov}, \Cref{sec:adaptive} and \Cref{sec:simulations}. Their variance estimators are described in \Cref{sec:var_quasi} and \Cref{rem:var_shrink} respectively.

\begin{itemize}
\item In simulations, for every $k$, $\widehat{\mathbb{IF}}_{22, k}^{\mathsf{quasi}} ([\widehat{\Omega}_{k}^{\mathsf{est}}]^{-1})$ is always numerically stable. We know that $\mathsf{Bias}_{\theta, k}(\hat{\psi}_{1})$ increases with $k$. In contrast, although $\widehat{\IIFF}_{22, k}^{\mathsf{quasi}} ([\widehat{\Omega}_{k}^{\mathsf{est}}]^{-1})$ initially increases with $k$, we observe that after some $k^{\ast}$, it begins to decrease. Our adaptive estimator switches to $\widehat{\mathbb{IF}}_{22, k^{\ast}} ([\widehat{\Omega}_{k^{\ast}}^{\mathsf{shrink}}]^{-1}))$ at this $k^{\ast}$, if the variance estimator of $\widehat{\mathbb{IF}}_{22, k^{\ast}} ([\widehat{\Omega}_{k}^{\mathsf{shrink}}]^{-1}))$ does not blow up. Empirically, $\widehat{\mathbb{IF}}_{22, k} ([\widehat{\Omega}_{k}^{\mathsf{shrink}}]^{-1}))$ performs well as an estimator of $\mathsf{Bias}_{\theta, k}(\hat{\psi}_{1})$ when its variance estimator does not blow up.

\item In our simulation study, at each $k$, the empirical probability of either choosing $\widehat{\mathbb{IF}}_{22, k}^{\mathsf{quasi}} ([\widehat{\Omega}_{k}^{\mathsf{est}}]^{-1})$ or $\widehat{\mathbb{IF}}_{22, k} ([\widehat{\Omega}_{k}^{\mathsf{shrink}}]^{-1}))$ is 1. Thus we do not need to take into account the above data-driven selection step in estimating the variance of the data-adaptive estimator $\widehat{\IIFF}_{22, k} (\widehat{\Omega}_{k}^{-1})$.
\end{itemize}

We leave the problem of unknown $\Omega_{k}^{-1}$ with $k > n$ to future work, because estimation of $\Omega_{k}^{-1}$ with $k > n$ requires additional assumptions on the distribution of $X$ outside those in \Cref{cond:w} that may not hold.

\begin{frontmatter}

\title{Supplementary Materials for ``On nearly assumption-free tests of nominal confidence interval coverage for causal parameters estimated by machine learning''}

\end{frontmatter}
\maketitle
\allowdisplaybreaks
\setcounter{section}{0}
\section{Connections to related literatures}\label{sec:connection}
\subsection{Undersmoothing, three-way sample splitting, and $\hat{\psi}_{2,k}$}\label{rem:undersmoothing}
This section concerns \cref{ft:triple} in the main text. We aim to provide a heuristic understanding of the relationship between $\hat{\psi}_{1}$ and $\hat{\psi}_{2,k}$ by considering the relationship of each to the undersmoothed, triple sample splitting estimator $\hat{\psi}_{1, \NR}$ of \citet{newey2018cross}. We begin by comparing $\hat{\psi}_{1}$ with $\hat{\psi}_{1,NR}$. We focus on $\psi(\theta) = \mathbb{E}_{\theta}[\mathsf{var}_{\theta}(A | X)]$.

Recall that the bias of $\hat{\psi}_{1} = \frac{1}{n} \sum_{i=1}^{n} (A_{i}-\hat{p}(X_{i}))^{2}$ is $\Bias_{\theta} (\hat{\psi}_{1}) = \BE_{\theta} [ \{ p(X)-\hat{p}(X) \}^{2}]$. If $p(x)$ lies in a \Holder{} ball with exponent $s_{p}$, then if the density of the $d$-dimensional vector $X$ is known, arguments analogous to those in \citet{stone1980optimal, stone1982optimal} shows that for any estimator $\hat{p}(x)$ of $p\left( x\right)$, $\BE_{\theta} [ (p(X)-\hat{p}(X))^{2} ]$ is at best $O (n^{-\frac{2s_{p}}{2s_{p}+d}})$, which can be achieved by the estimator $\hat{p}(x) = \tilde{\beta}_{k, p}^{T} \zbar_{k} (x)$ with $\tilde{\beta}_{k, p} = \Omega_{k}^{-1} \frac{1}{n} \sum\nolimits_{i} A_{i} \zbar_{k}(X_{i})$ with $k \asymp n^{\frac{d}{2s_{p}+d}}$ and $\zbar_{k} (x)$ the first $k$ bases of a suitably chosen spline or wavelet orthonormal basis for $L_{2} (P_{\theta})$. This implies that $s_{p} > d / 2$ is needed for $\Bias_{\theta} (\hat{\psi}_{1})$ to be $o (n^{-1/2})$. Furthermore, the variance of $\hat{\psi}_{1}$ increases if we use an undersmoothed estimator $\hat{p}(x)$ obtained by choosing $k \gg n^{\frac{d}{2s_{p}+d}}$. (Note choosing $k \asymp n^{\frac{d}{2s_{p}+d}}$ equalizes the order $k / n$ of the variance of $\hat{p}(x)$ and the order $k^{-2 s_{p} / d}$ of the square of the approximation bias $\BE_{\theta} [ \Pi [ p(X) | \Zbar_{k}^{\perp} ] ]$). 

However, suppose as in \citet{newey2018cross}, we replaced $\hat{\psi}_{1}$ defined in \Cref{sec:drml1} by 
\begin{eqnarray*}
\hat{\psi}_{1, \NR} &=&\frac{1}{n}\sum_{i = 1}^n \{A_{i}-\hat{p}_{1}(X_{i})\}\left\{ A_{i}-\hat{p}_{2}(X_{i})\right\} \\
&=&\frac{1}{n}\sum_{i = 1}^n A_{i}^{2}-\left( A_{i}-\hat{p}_{2}(X_{i})\right) \hat{p}_{1}(X_{i})-\left( A_{i}-\hat{p}_{1}(X_{i})\right) \hat{p}_{2}(X_{i})-\hat{p}_{1}(X_{i})\hat{p}_{2}(X_{i})
\end{eqnarray*}
where the training sample is itself randomly split into two subsamples $I_1$ and $I_2$ of equal size and the regression coefficients in $\hat{p}_{1}(x) = \tilde{\beta}_{1, k, p}^{\top}\zbar_{k}(x)$ and $\hat{p}_{2}(x)=\tilde{\beta}_{2, k, p}^{\top} \zbar_{k}(x)$ are computed from subjects in subsamples $I_1$ and $I_2$ respectively. Hence $\hat{\psi}_{1, \NR}$ is computed from three independent samples and uses the true $\Omega_{k}$ and its inverse. \citet{newey2018cross} show that unconditionally the (random) bias $\BE_\theta [\left\{ p(X)-\hat{p}_{1}(X)\right\} \left\{ p(X) - \hat{p}_{2}(X)\right\}]$ of $\hat{\psi}_{1, \NR}$ is of order $\left(k / n^{2} + k^{-4s_{p}/d}\right)^{1/2}$ in probability which is minimized by choosing $k = n^{2d / ( d+ 4s_{p})}$, for which the bias is of order $n^{-\frac{4s_{p}}{4s_{p}+d}}$. Then the bias is $o (n^{-1/2})$ if $s_{p}>d/4$. Note also that $n^{\frac{2d}{4s_{p}+d}} > n^{\frac{d}{2s_{p}+d}}$, so, unlike with $\hat{\psi}_{1}$, it is optimal to undersmooth the estimators $\hat{p}_{1}(x)$ and $\hat{p}_{2}(x)$ to minimize the bias of $\hat{\psi}_{1, \NR}$. Hence the secret sauce behind the much better performance of $\hat{\psi}_{1, \NR}$ compared to $\hat{\psi}_{1}$ is a combination of three-way (rather than two-way) sample splitting combined with undersmoothing in the estimation of $\hat{p}_{1}(x)$ and $\hat{p}_{2}(x)$. 

We now turn to a comparison of $\hat{\psi}_{2, k}$ and $\hat{\psi}_{1, \NR}$. For didactic purposes, it will be useful to first consider the case in which $\hat{p}(X)$ is artificially chosen to be identically zero; then $\hat{\psi}_{1} = \frac{1}{n} \sum_{i=1}^{n} A_{i}^{2}$,
\begin{eqnarray*}
\widehat{\mathbb{IF}}_{22, k} &=& \frac{1}{n(n-1)} \sum_{1 \leq i_{1} \neq i_{2} \leq n}\left\{ \left[ A_{i_{1}} \zbar_{k}(X_{i_{1}})^{\top} \right] \Omega_{k}^{-1} \right\} \Omega_{k} \left\{ \Omega_{k}^{-1} \left[ \zbar_{k}(X_{i_{2}}) A_{i_{2}} \right] \right\} , \\
\text{and } \hat{\psi}_{2, k}  & = & \frac{1}{n} \sum_{i=1}^{n} A_{i}^{2} - \widehat{\mathbb{IF}}_{22, k}.
\end{eqnarray*}

We now show that $\hat{\psi}_{1, \NR}$ and $\hat{\psi}_{2, k}$ have the same mean and the same order of variance $1 / n$. Since both $\hat{\psi}_{1,\NR}$ and $\hat{\psi}_{2,k}$ have the common term $\frac{1}{n}\sum_{i=1}^{n}A_{i}^{2}$, our goal becomes to compare $\hat{\psi}_{1, \NR} - \frac{1}{n} \sum_{i=1}^{n}A_{i}^{2}$ with $\widehat{\mathbb{IF}}_{22, k}$. Rather than doing so directly, let us consider the following third-order U-statistic that substitutes the unbiased estimator $\zbar_{k} (X_{i_3}) \zbar_{k}(X_{i_3})^{\top}$ for $\Omega_{k}$ to give
\begin{align*}
\widehat{\mathbb{IF}}_{22,k}^{\ast} = \frac{(n - 2)!}{n!} \sum_{1\leq i_{1}\neq i_{2}\neq i_{3}\leq n}\left\{ \left[ A_{i_{1}}\zbar_{k}\left( X_{i_{1}}\right) ^{\top }\right] \Omega_{k}^{-1}\right\} \zbar_{k} (X_{i_3}) \zbar_{k}(X_{i_3})^{\top} \left\{ \Omega_{k}^{-1}\left[ \zbar_{k}\left( X_{i_{2}}\right) A_{i_{2}}\right] \right\}
\end{align*}
so
\begin{align*}
& \ \widehat{\mathbb{IF}}_{22,k}^{\ast} - \widehat{\mathbb{IF}}_{22, k} \\
= & \ \frac{(n-3)!}{n!} \sum_{1\leq i_{1}\neq i_{2}\neq i_{3}\leq n}\left\{ \left[ A_{i_{1}}\zbar_{k}\left( X_{i_{1}}\right) ^{\top }\right] \Omega_{k}^{-1}\right\} \left( \zbar_{k} (X_{i_3}) \zbar_{k}(X_{i_3})^{\top} - \Omega_{k}\right) \left\{ \Omega_{k}^{-1}\left[ \zbar_{k}\left( X_{i_{2}}\right) A_{i_{2}} \right] \right\}.
\end{align*}
It thus follows that $\widehat{\mathbb{IF}}_{22,k}^{\ast }$ and $\widehat{\mathbb{IF}}_{22, k}$ have the same mean and the same order of variance $1 / n$. 
Denote the estimation sample of size $n$ as $I_3$, and the sample sizes of $I_1$ and $I_2$ as $n_1$ and $n_2$ respectively. Then note that $\widehat{\mathbb{IF}}_{22,k}^{\ast }$ and 
\begin{align*}
& \frac{1}{n}\sum_{i = 1}^n \hat{p}_{1}(X_{i})\hat{p}_{2}(X_{i}) \\
& = \frac{1}{n n_1 n_2} \sum_{i_{1} \in I_{1}, i_{2} \in I_{2}, i_{3} \in I_{3}} \left\{ \left[ A_{i_{1}}\zbar_{k}\left( X_{i_{1}}\right) ^{\top }\right] \Omega_{k}^{-1}\right\} \zbar_{k} (X_{i_3}) \zbar_{k}(X_{i_3})^{\top} \left\{ \Omega_{k}^{-1}\left[ \zbar_{k}\left( X_{i_{2}}\right) A_{i_{2}}\right] \right\}
\end{align*} 
have identical kernels $$\left\{ \left[ A_{i_{1}}\zbar_{k}\left( X_{i_{1}}\right)^{\top }\right] \Omega_{k}^{-1}\right\} \zbar_{k} (X_{i_3}) \zbar_{k}(X_{i_3})^{\top} \left\{ \Omega_{k}^{-1}\left[ \zbar_{k}\left( X_{i_{2}}\right) A_{i_{2}} \right] \right\}$$ and thus identical expectations. They differ only in that $\widehat{\mathbb{IF}}_{22,k}^{\ast }$ is a third-order U-statistic while $\frac{1}{n}\sum_{i \in I_3} \hat{p}_{1}(X_{i})\hat{p}_{2}(X_{i})$ splits the sample into three subsets $I_{1}, I_{2}$ and $I_{3}$. Then both $n^{-1}\sum_{i=1}^{n}\hat{p}_{1}(X_{i})\hat{p}_{2}(X_{i})$ and $\widehat{\mathbb{IF}}_{22,k}^{\ast}$ have variance of order $1 / n$ although the constants will differ. Furthermore $\hat{\psi}_{1, \NR} - \frac{1}{n} \sum_{i=1}^{n} A_{i}^{2}$ has the two additional mean zero terms $\frac{1}{n} \sum_{i = 1}^{n} - (A_{i}-\hat{p}_{2}(X_{i})) \hat{p}_{1}(X_{i}) - ( A_{i}-\hat{p}_{1}(X_{i})) \hat{p}_{2}(X_{i})$ which both have variance of order $1 / n$.

In summary, $\hat{\psi}_{2,k}^{\ast} \coloneqq \frac{1}{n} \sum_{i = 1}^n A_i^2 - \widehat{\mathbb{IF}}_{22,k}^{\ast}$ and $\hat{\psi}_{1, \NR}$ have the same kernel and thus the same mean as $\hat{\psi}_{2, k}$. It also follows from the above arguments that $\hat{\psi}_{1, \NR}$, $\hat{\psi}_{2,k}^{\ast}$ and $\hat{\psi}_{2,k}$ have the same order of variance.

Lastly consider the case where, as earlier, we have a preliminary machine learning estimator $\hat{p}\left( x\right)$ computed from a second independent (training) sample in the case of $\hat{\psi}_{2,k}$ and from a fourth sample $I_4$ in the case of $\hat{\psi}_{1, \NR}$. If, in defining $\hat{\psi}_{1, \NR}$, we redefine $\tilde{\beta}_{j, k, p}$ as $ \Omega_{k}^{-1} \frac{1}{n_j}\left\{ \sum\nolimits_{i \in I_j}\left\{ A_{i}- \hat{p}\left( X_{i}\right) \right\} \zbar_{k}\left( X_{i}\right) \right\}$ with $j = 1, 2$, and use $\hat{p}\left( x\right) $ in $\hat{\psi}_{2,k}$ as earlier, the relationships between $\hat{\psi}_{1, \NR}$ and $\hat{\psi}_{2,k}$ remain as above.

\section{Rejection probability calculations}\label{app:oracletest}
Because of the similarity of the rejection probability calculations for \Cref{thm:test_var}, \Cref{thm:test_cov}, \Cref{thm:test_cov_cs} and \Cref{prop:test_{k}}, we only show the calculation for \Cref{thm:test_cov}.
\begin{proof}
We denote the oracle test as $\widehat\chi_{k} (\zeta_{k}, \delta)$. Then we have
\begin{align*}
& \; \lim_{n \rightarrow \infty} P_\theta \left( \frac{\vert \widehat{\mathbb{IF}}_{22, k} \vert}{\widehat{\se} [\hat{\psi}_{1}]} - \zeta_{k} \frac{\widehat{\se} [\widehat{\mathbb{IF}}_{22, k}]}{\widehat{\se} [\hat{\psi}_{1}]} > \delta \right) \\
= & \; \lim_{n \rightarrow \infty} \left\{ P_\theta \left( \frac{\widehat{\mathbb{IF}}_{22, k}}{\widehat{\se} [\widehat{\mathbb{IF}}_{22, k}]} > \zeta_{k} + \delta \frac{\widehat{\se} [\hat{\psi}_{1}]}{\widehat{\se} [\widehat{\mathbb{IF}}_{22, k}]} \right) + P_\theta \left( \frac{\widehat{\mathbb{IF}}_{22, k}}{\widehat{\se} [ \widehat{\mathbb{IF}}_{22, k} ]} < - \zeta_{k} - \delta \frac{\widehat{\se} [\hat{\psi}_{1}]}{\widehat{\se} [ \widehat{\mathbb{IF}}_{22, k} ]} \right) \right\} \\
= & \; \lim_{n \rightarrow \infty} P_\theta \left( \frac{\widehat{\mathbb{IF}}_{22, k} - \Bias_{\theta, k} ( \hat{\psi}_{1} )}{\widehat{\se} [ \widehat{\mathbb{IF}}_{22, k} ]} > \zeta_{k} - \frac{\Bias_{\theta, k} ( \hat{\psi}_{1} )}{\widehat{\se} [ \widehat{\mathbb{IF}}_{22, k} ]} + \delta \frac{\widehat{\se} [\hat{\psi}_{1}]}{\widehat{\se} [ \widehat{\mathbb{IF}}_{22, k} ]} \right) \\
& \; + \lim_{n \rightarrow \infty} P_\theta \left( \frac{\widehat{\mathbb{IF}}_{22, k} - \Bias_{\theta, k} ( \hat{\psi}_{1} )}{%
\widehat{\se} [ \widehat{\mathbb{IF}}_{22, k} ]} < - \zeta_{k} - \frac{\Bias_{\theta, k} ( \hat{\psi}_{1} )}{\widehat{\se} [ \widehat{\mathbb{IF}}_{22, k} ]} - \delta \frac{\widehat{\se} [\hat{\psi}_{1}]}{\widehat{\se} [ \widehat{\mathbb{IF}}_{22, k} ]} \right) \\
= & \; \lim_{n \rightarrow \infty} P_\theta \left( \frac{\widehat{\mathbb{IF}}_{22, k} - \Bias_{\theta, k} ( \hat{\psi}_{1} )}{\se_\theta [ \widehat{\mathbb{IF}}_{22, k} ]} (1 + o_{P_\theta}(1)) > \zeta_{k} - (\gamma - \delta) \frac{\se_\theta [ \hat{\psi}_{1} ]}{\se_\theta [ \widehat{\mathbb{IF}}_{22, k} ]} (1 + o_{P_\theta}(1)) \right) \\
& \; + \lim_{n \rightarrow \infty} P_\theta \left( \frac{\widehat{\mathbb{IF}}_{22, k} - \Bias_{\theta, k} ( \hat{\psi}_{1} )}{\se_\theta [ \widehat{\mathbb{IF}}_{22, k} ]} (1 + o_{P_\theta}(1)) < - \zeta_{k} - (\gamma + \delta) \frac{\se_\theta [ \hat{\psi}_{1} ]}{\se_\theta [ \widehat{\mathbb{IF}}_{22, k} ]} (1 + o_{P_\theta}(1)) \right) \\
= & \; 1 - \Phi \left( \zeta_{k} - \lim_{n \rightarrow \infty} (\gamma - \delta) \frac{\se_\theta [ \hat{\psi}_{1} ]}{\se_\theta [ \widehat{\mathbb{IF}}_{22, k} ]} \right) + \Phi \left( - \zeta_{k} - \lim_{n \rightarrow \infty} (\gamma + \delta) \frac{\se_\theta [ \hat{\psi}_{1} ]}{\se_\theta [ \widehat{\mathbb{IF}}_{22, k} ]} \right) \\
= & \; 2 - \Phi \left( \zeta_{k} - \lim_{n \rightarrow \infty} (\gamma - \delta) \frac{\se_\theta [ \hat{\psi}_{1} ]}{\se_\theta [ \widehat{\mathbb{IF}}_{22, k} ]} \right) - \Phi \left( \zeta_{k} + \lim_{n \rightarrow \infty} (\gamma + \delta) \frac{\se_\theta [ \hat{\psi}_{1} ]}{\se_\theta [ \widehat{\mathbb{IF}}_{22, k} ]} \right).
\end{align*}
\end{proof}

\section{On the choice of estimators when $\Omega_{k}^{-1}$ is unknown}\label{sec:cov} 
In the main text we assume that $\Omega_{k}^{-1}$ is known. Outside the $X$-semisupervised case, this assumption is usually untenable and $\Omega_{k}^{-1}$ must be estimated from data. To resolve this issue, one approach is to construct an estimator of the density $f_{X}$ of $X$ \citep{robins2008higher, robins2017minimax}. But when the dimension of the covariates $X$ is large, accurate density estimation is problematic. More recently, in the regime $k=o(n)$, \citet{mukherjee2017semiparametric} proposed to replace $\Omega_{k}$ by $\widehat{\Omega}_{k}^{\mathsf{tr}}$ in $\widehat{\mathbb{IF}}_{22, k}$, where $\widehat{\Omega}_{k}^{\mathsf{tr}}=n^{-1}\sum_{i\in \mathsf{tr}}\zbar_{k}(X_{i})\zbar_{k}(X_{i})^{\top }$ is the sample covariance matrix estimator from the training sample. They show that $\widehat{\mathbb{IF}}_{22, k} ([\widehat{\Omega}_{k}^{\mathsf{tr}}]^{-1})$ is a biased estimator of $\Bias_{\theta, k }(\hat{\psi}_{1})$ with estimation bias $\mathit{EB}_{\theta, 2, k} ([\widehat{\Omega}_{k}^{\mathsf{tr}}]^{-1})\equiv \mathbb{E}_{\theta}[\widehat{\mathbb{IF}}_{22, k} ([\widehat{\Omega}_{k}^{\mathsf{tr}}]^{-1})-\Bias_{\theta, k }(\hat{\psi}_{1})]$ of order $O (\BL_{2, b, k} \BL_{2, p, k} \sqrt{\frac{k\mathsf{log}(k)}{n}})$ under \Cref{cond:w}. [Note that $\mathit{EB}_{\theta, 2, k} ([\widehat{\Omega}_{k}^{\mathsf{tr}}]^{-1})$ is also the bias of $\hat{\psi}_{2, k} ([\widehat{\Omega}_{k}^{\mathsf{tr}}]^{-1})$ as an estimator of the truncated parameter $\psi (\theta) + \TB_{\theta, k} (\hat{\psi}_{1})$.] It follows that the bias in estimating $\Bias_{\theta, k }(\hat{\psi}_{1})$ converges to zero if we choose $k = o(n / \log(n))$ when $\BL_{2, b, k}$ and $\BL_{2, p, k}$ are bounded (as implied by \Cref{cond:w}).

We will complement the above discussion with the same simulation study reported in \Cref{tab:intro} when the goal is to estimate $\psi (\theta) = \BE_{\theta} [\var_{\theta}(A | X)]$. We found in simulation that $\widehat{\mathbb{IF}}_{22, k} ([\widehat{\Omega}_{k}^{\mathsf{tr}}]^{-1})$ has very unstable finite sample performance when $k$ is relatively large. For example, as shown in the second column of \Cref{tab:sample}, $\widehat{\mathbb{IF}}_{22, k} ([\widehat{\Omega}_{k}^{\mathsf{tr}}]^{-1})$ starts to break down when $k = 2048$, reflected by its MCsd being almost 20 times that of $\widehat{\mathbb{IF}}_{22, k}$. When $k = 4096$, $\widehat{\mathbb{IF}}_{22, k} ([\widehat{\Omega}_{k}^{\mathsf{tr}}]^{-1})$ and its standard error are more than $1000$ times those of $\widehat{\mathbb{IF}}_{22, k}$. This motivates us to find estimators that work better than $\widehat{\mathbb{IF}}_{22, k} ([\widehat{\Omega}_{k}^{\tr}]^{-1})$ in practice.

\subsection{An empirically stable estimator $\widehat{\mathbb{IF}}_{22, k} ( [\widehat{\Omega}_{k}^\mathsf{est}]^{-1} )$}\label{sec:est} 
A natural alternative estimator $\widehat{\mathbb{IF}}_{22, k} ( [\widehat{\Omega}_{k}^\mathsf{est}]^{-1} )$ is simply to replace $\widehat{\Omega}_{k}^\mathsf{tr}$ by $\widehat{\Omega}_{k}^\mathsf{est} \coloneqq n^{-1} \sum_{i \in \mathsf{est}} \Zbar_{k, i} \Zbar_{k, i}^{\top}$ from the estimation sample. Reading from \Cref{tab:sample}, we see that $\widehat{\mathbb{IF}}_{22, k} ( [\widehat{\Omega}_{k}^\mathsf{est}]^{-1} )$ and its standard error never blows up even for $k = 4096$ ($k / n \approx 0.8$). However, though numerically stable, \Cref{tab:sample} shows that the MCav of $\widehat{\mathbb{IF}}_{22, k} ([\widehat{\Omega}_{k}^\mathsf{est}]^{-1})$, in contrast to that of the oracle $\widehat{\mathbb{IF}}_{22, k}$, eventually decreases as $k$ increases and thus fails to correct nearly as much of the bias of $\hat{\psi}_{1}$ as does $\widehat{\mathbb{IF}}_{22, k}$. This can be seen from the third column of \Cref{tab:sample}: the MCav of $\widehat{\mathbb{IF}}_{22, k} ([\widehat{\Omega}_{k}^\mathsf{est}]^{-1})$ at $k = 256$ is 0.116 and close to that of $\widehat{\mathbb{IF}}_{22, k}$ (0.126). However the MCav of $\widehat{\mathbb{IF}}_{22, k} ([\widehat{\Omega}_{k}^\mathsf{est}]^{-1})$ decreases to 0.102 at $k = 512$, while the MCav of $\widehat{\mathbb{IF}}_{22, k}$ (0.127) continues to increase.

These numerical results raise the question whether we can find a stable estimator with MCav closer to that of the oracle $\widehat{\mathbb{IF}}_{22, k}$ than is the MCav of $\widehat{\mathbb{IF}}_{22, k} ([\widehat{\Omega}_{k}^\mathsf{est}]^{-1})$.

We proceed based on a theoretical analysis of the estimation bias of $\widehat{\mathbb{IF}}_{22, k} ([\widehat{\Omega}_{k}^{\mathsf{est}}]^{-1})$ as an estimator of $\Bias_{\theta, k }(\hat{\psi}_{1})$ conditional on the training sample, which we refer to as $\mathit{EB}_{\theta, 2, k} ([\widehat{\Omega }_{k}^{\mathsf{est}}]^{-1})\coloneqq\mathbb{E}_\theta [\widehat{\mathbb{IF}}_{22, k} ([\widehat{\Omega}_{k}^{\mathsf{est}}]^{-1})-\Bias_{\theta, k }(\hat{\psi}_{1}) ]=\mathbb{E}_\theta [\widehat{\mathbb{IF}}_{22, k} ([\widehat{\Omega}_{k}^{\mathsf{est}}]^{-1})-\widehat{\mathbb{IF}}_{22, k} (\Omega_{k}^{-1}) ]$.

As we describe now, this analysis led us to derive a de-biased version, $\widehat{\mathbb{IF}}_{22, k}^\mathsf{debiased} ( [\widehat{\Omega}_{k}^\mathsf{est}]^{-1} )$, defined below in \cref{eq:debiasedsoif}, of $\widehat{\mathbb{IF}}_{22, k} ( [\widehat{\Omega}_{k}^\mathsf{est}]^{-1} )$, with estimation bias 
\begin{equation*}
\mathit{EB}_{\theta, 2, k}^{\mathsf{debiased}}([\widehat{\Omega}_{k}^{\mathsf{est}}]^{-1})\coloneqq\mathbb{E}_\theta [\widehat{\mathbb{IF}}_{22,k}^{\mathsf{debiased}}([\widehat{\Omega}_{k}^{\mathsf{est}}]^{-1})-\Bias_{\theta, k }(\hat{\psi}_{1}) ]
\end{equation*}
under \Cref{cond:w} of order $O (\BL_{2, b, k} \BL_{2, p, k} \frac{k\mathsf{log}(k)}{n})$, which is of smaller order than $\mathit{EB}_{\theta, 2, k} ([\widehat{\Omega}_{k}^{\mathsf{tr}}]^{-1})$. Specifically our derivation used the following identity (see \Cref{app:id}): 
\begin{align}
& \;\mathit{EB}_{\theta, 2, k} ([\widehat{\Omega}_{k}^{\mathsf{est}}]^{-1})\equiv \mathbb{E}_\theta \left[ \widehat{\mathbb{IF}}_{22, k} ([\widehat{\Omega}_{k}^{\mathsf{est}}]^{-1})-\widehat{\mathbb{IF}}_{22, k}|\hat{\theta} \right]  \notag  \label{eq:heuristic} \\
=& \;\mathbb{E}_\theta \left[ \widehat{\xi }_{b,1}\zbar_{k}(X_{1})^{\top }\cdot \left( [\widehat{\Omega}_{k,-1,-2}^{\mathsf{est}}]^{-1}-\Omega_{k}^{-1}\right) \zbar_{k}(X_{2})\widehat{\xi }_{p,2} \right]  \notag \\
& \;-\frac{1}{n}\mathbb{E}_\theta \left[ \widehat{\xi }_{b,1}\zbar_{k}(X_{1})^{\top }\cdot [\widehat{\Omega}_{k,-1,-2}^{\mathsf{est}}]^{-1}\cdot \underset{i=1,2}{\sum }\zbar_{k}(X_{i})\zbar_{k}(X_{i})^{\top} [\widehat{\Omega}_{k}^{\mathsf{est}}]^{-1}\cdot \zbar_{k}(X_{2})\widehat{\xi }_{p_{2}} \right]  \notag \\
\coloneqq& \;\text{(I)}+\text{(II)}.
\end{align}
where for any $(i_{1},i_{2}:1\leq i_{1}\neq i_{2}\leq n)$, we define $\widehat{\Omega}_{k,-i_{1},-i_{2}}^{\mathsf{est}}\coloneqq\frac{1}{n}\sum_{i\in \mathsf{est}:i\neq i_{1},i_{2}}\zbar_{k}(X_{i})\zbar_{k}(X_{i})^{\top }$.

Consider the first term $\text{(I)}$ in the last line of the RHS of \cref{eq:heuristic}. Due to the independence between the three product terms, we show in \Cref{app:I} that we can upper bound $\text{(I)}$, up to constant, by $\BL_{2, b, k} \BL_{2, p, k} \mathbb{E}_\theta [ \Vert \widehat{\Omega}_{k,-1,-2}^{\mathsf{est}}-\Omega \Vert^{2} ]$, under \Cref{cond:w}.

This upper bound is similar to the upper bound established for $\widehat{\mathbb{IF}}_{22, k} ([\widehat{\Omega}_{k}^\mathsf{tr}]^{-1})$ in \citet{mukherjee2017semiparametric}, with $\Vert \widehat{\Omega}_{k}^{\tr} - \Omega_{k} \Vert$ replaced by $\Vert \widehat{\Omega}_{k,-1,-2}^{\mathsf{est}} - \Omega_{k} \Vert^{2}$. In particular, under \Cref{cond:w}{}, $\text{(I)}$ is of order $O (\BL_{2, b, k} \BL_{2, p, k} \frac{k\mathsf{log}(k)}{n})$ \citep{rudelson1999random}.

Next consider the second term $\text{(II)}$ in \cref{eq:heuristic}. It may be the dominating term in \cref{eq:heuristic}. However, we can remove the contribution of $\text{(II)}$ to $\mathit{EB}_{\theta, 2, k} ([\widehat{\Omega}_{k}^{\mathsf{est}}]^{-1})$ by subtracting its unbiased estimator from $\widehat{\mathbb{IF}}_{22, k} ([\widehat{\Omega}_{k}^{\mathsf{est}}]^{-1})$, leading to $\widehat{\mathbb{IF}}_{22,k}^{\mathsf{debiased}}([\widehat{\Omega}_{k}^{\mathsf{est}}]^{-1})$ defined as\footnote{We cannot write $\widehat{\mathbb{IF}}_{22,k}^{\mathsf{debiased}}([\widehat{\Omega}_{k}^{\mathsf{est}}]^{-1})$ in the form of $\widehat{\mathbb{IF}}_{22, k} (\widehat{\Omega}_{k}^{-1})$ because the bias correction on $[\widehat{\Omega}_{k}^{\mathsf{est}}]^{-1}$ is not common for every pair of subjects in the summation $(i_{1},i_{2}:1\leq i_{1}\neq i_{2}\leq n)$ and this is reflected in the notation by attaching a superscript ``$\mathsf{debiased}$'' on $\widehat{\mathbb{IF}}_{22,k}$.}: 
\begin{align}
& \widehat{\mathbb{IF}}_{22,k}^{\mathsf{debiased}}([\widehat{\Omega}_{k}^{\mathsf{est}}]^{-1})\coloneqq\widehat{\mathbb{IF}}_{22,k}\left( [\widehat{\Omega}_{k}^{\mathsf{est}}]^{-1}\right)  \label{eq:debiasedsoif} \\
& + \ \frac{1}{n^{2}(n-1)}\sum_{1\leq i_{1}\neq i_{2}\leq n}\hat{\varepsilon}_{b,i_{1}}\zbar_{k}(X_{i_{1}})^{\top}[\widehat{\Omega}_{k, -i_{1}, -i_{2}}^{\mathsf{est}}]^{-1} \left( \sum_{i = i_{1},i_{2}} \zbar_{k}(X_{i}) \zbar_{k}(X_{i})^{\top} \right) [\widehat{\Omega}_{k}^{\mathsf{est}}]^{-1} \zbar_{k}(X_{i_{2}}) \hat{\varepsilon}_{p, i_{2}}. \notag
\end{align}
By the above calculation, $\mathit{EB}_{\theta, 2, k}^{\mathsf{debiased}}([\widehat{\Omega}_{k}^{\mathsf{est}}]^{-1})$ is of order $O (\BL_{2, b, k} \BL_{2, p, k} \frac{k \mathsf{log}(k)}{n})$.

\begin{rem}\label{rem:simpler}\leavevmode
We define $\widehat{\mathbb{IF}}_{22 \rightarrow 33, k} ([\widehat{\Omega}_{k}^{\mathsf{tr}}]^{-1})$ as
\begin{align}
\widehat{\mathbb{IF}}_{22 \rightarrow 33, k} ([\widehat{\Omega}_{k}^{\mathsf{tr}}]^{-1}) & \coloneqq \widehat{\mathbb{IF}}_{22, k} ([\widehat{\Omega}_{k}^{\mathsf{tr}}]^{-1}) + \widehat{\mathbb{IF}}_{33, k} ([\widehat{\Omega}_{k}^{\mathsf{tr}}]^{-1})  \label{eq:if22tr}
\end{align}
where, 
\begin{align*}
\widehat{\mathbb{IF}}_{22, k} ([\widehat{\Omega}_{k}^{\mathsf{tr}}]^{-1}) & \coloneqq \frac{(n - 2)!}{n!} \sum_{1 \leq i_{1} \neq i_{2} \neq i_{3} \leq n} \left[ \hat{\varepsilon}_{b} \bar{\mathsf{Z}}_{k} \right]_{i_{1}}^{\top} [\widehat{\Omega}_{k}^{\mathsf{tr}}]^{-1} \left[ \bar{\mathsf{Z}}_{k} \hat{\varepsilon}_{p} \right]_{i_{2}}, \\
\widehat{\mathbb{IF}}_{33, k} ([\widehat{\Omega}_{k}^{\mathsf{tr}}]^{-1}) & \coloneqq - \frac{(n - 3)!}{n!} \sum_{1 \leq i_{1} \neq i_{2} \neq i_{3} \leq n} \left[\hat{\varepsilon}_{b} \bar{\mathsf{Z}}_{k} \right]_{i_{1}}^{\top} [\widehat{\Omega}_{k}^{\mathsf{tr}}]^{-1} \left[ \bar{\mathsf{Z}}_{k} \bar{\mathsf{Z}}_{k}^{\top} - \widehat{\Omega}_{k}^{\mathsf{tr}} \right]_{i_{3}} [\widehat{\Omega}_{k}^{\mathsf{tr}}]^{-1} \left[ \bar{\mathsf{Z}}_{k} \hat{\varepsilon}_{p} \right]_{i_{2}},
\end{align*}
$\widehat{\mathbb{IF}}_{22 \rightarrow 33, k} ([\widehat{\Omega}_{k}^{\mathsf{tr}}]^{-1})$, given in \cref{eq:if22tr}, has the same order of estimation bias as $\widehat{\mathbb{IF}}_{22,k}^{\mathsf{debiased}}([\widehat{\Omega}_{k}^{\mathsf{est}}]^{-1})$ \citep[Theorem 4]{mukherjee2017semiparametric}: 
\begin{equation}
\mathit{EB}_{\theta, 3, k}([\widehat{\Omega}_{k}^{\mathsf{tr}}]^{-1})\coloneqq\mathbb{E}_{\theta}[\widehat{\mathbb{IF}}_{22 \rightarrow 33, k} ([\widehat{\Omega}_{k}^{\mathsf{tr}}]^{-1})-\Bias_{\theta, k }(\hat{\psi}_{1})] = O \left( \BL_{2, b, k} \BL_{2, p, k} \frac{k\mathsf{log}(k)}{n} \right). \label{eb:soiftoif}
\end{equation}
It adds a third order U-statistic $\widehat{\mathbb{IF}}_{33, k} ([\widehat{\Omega}_{k}^{\mathsf{tr}}]^{-1})$ to reduce the estimation bias $\mathit{EB}_{\theta, 2, k} ([\widehat{\Omega}_{k}^{\mathsf{tr}}]^{-1})$. Unfortunately, the instability of $\widehat{\mathbb{IF}}_{22, k} ([\widehat{\Omega}_{k}^{\mathsf{tr}}]^{-1})$ in finite sample cannot be resolved by correcting its estimation bias by adding $\widehat{\mathbb{IF}}_{33, k}([\widehat{\Omega}_{k}^{\mathsf{tr}}]^{-1})$: when $\widehat{\mathbb{IF}}_{22, k} ([\widehat{\Omega}_{k}^{\mathsf{tr}}]^{-1})$ starts to break down at $k = 2048$ (see the second column of \Cref{tab:sample}), $\widehat{\mathbb{IF}}_{33,k}([\widehat{\Omega}_{k}^{\mathsf{tr}}]^{-1})$ also starts to break down (see the second column of \Cref{tab:toif}).

\begin{table}[tbp]
\caption{Simulation setup I: $\psi(\theta) = \mathbb{E}_\theta [ \mathsf{var}_\theta (A | X) ]$, regression functions estimated by nonparametric kernel regression with cross validation} \label{tab:toif}\centering
\begin{tabular}{l|ll}
\hline
$k$ & $\widehat{\mathbb{IF}}_{33, k}$ & $\widehat{\mathbb{IF}}_{33, k} ( [\widehat{\Omega}_{k}^\mathsf{tr}]^{-1} )$ \\ 
\hline 
$64$ & $-8.330 \times 10^{-5}$ (0.00125) & $0.00200$ (0.00156) \\ 
$128$ & $-0.000122$ (0.00163) & 0.00388 (0.00257) \\ 
$256$ & $0.000355$ (0.00421) & 0.00222 (0.00874) \\ 
$512$ & $0.00139$ (0.00602) & 0.0450 (0.0166) \\ 
$1024$ & $0.00148$ (0.00912) & 0.143 (0.0699) \\
$2048$ & $0.00241$ (0.0148) & 43.775 (370.125) \\
$4096$ & $0.00343$ (0.0270) & $4.264 \times 10^{19}$ ($2.242 \times 10^{20}$) \\ 
\hline
\end{tabular}
\newline
A comparison between $\widehat{\mathbb{IF}}_{33, k}$ and $\widehat{\mathbb{IF}}_{33, k} ( [\widehat{\Omega}_{k}^\mathsf{tr}]^{-1} )$. Unlike the other tables, the numbers in the parentheses are Monte Carlo (MC) standard deviations of the corresponding estimators. For more details on the data generating mechanism, see \Cref{sec:simulations}.
\end{table}
\end{rem}

\begin{rem}\label{rem:est} 
Even though $\widehat{\mathbb{IF}}_{22,k}^{\mathsf{debiased}}([\widehat{\Omega}_{k}^{\mathsf{est}}]^{-1})$ has better estimation bias, it is extremely difficult to compute in practice. To compute $\widehat{\mathbb{IF}}_{22,k}^{\mathsf{debiased}}([\widehat{\Omega}_{k}^{\mathsf{est}}]^{-1})$, in the summation over all $1\leq i_{1}\neq i_{2}\leq n$, we have to evaluate $\widehat{\Omega}_{k,-i_{1},-i_{2}}^{\mathsf{est}}$ for all $(i_{1},i_{2}:1\leq i_{1}<i_{2}\leq n)$. Thus one needs to compute ${\binom{n}{2}}$ different inverse sample covariance matrices when computing $\widehat{\mathbb{IF}}_{22,k}^{\mathsf{debiased}}([\widehat{\Omega}_{k}^{\mathsf{est}}]^{-1})$. Moreover, the kernel of $\widehat{\mathbb{IF}}_{22,k}^{\mathsf{debiased}}([\widehat{\Omega}_{k}^{\mathsf{est}}]^{-1})$ is no longer separable, this is because for all $(i_{1},i_{2}:1\leq i_{1}<i_{2}\leq n)$, the kernel also depends on all the other subjects $\{i\neq i_{1},i_{2}:1\leq i\leq n\}$. We will introduce a computationally-feasible estimator $\widehat{\mathbb{IF}}_{22,k}^{\quasi}([\widehat{\Omega}_{k}^{\mathsf{est}}]^{-1})$ in \Cref{sec:quasi}, which enjoys the stability of $\widehat{\mathbb{IF}}_{22, k} ([\widehat{\Omega}_{k}^{\mathsf{est}}]^{-1})$ and has no greater estimation bias at least in simulations than $\widehat{\mathbb{IF}}_{22, k} ([\widehat{\Omega}_{k}^{\mathsf{est}}]^{-1}) $ as shown in the fourth column of \Cref{tab:sample}.
\end{rem}

\subsection{An easy-to-compute quasi de-biased estimator}\label{sec:quasi}
The estimator $\widehat{\mathbb{IF}}_{22,k}^{\mathsf{quasi}}([\widehat{\Omega}_{k}^{\mathsf{est}}]^{-1})$ differs from $\widehat{\mathbb{IF}}_{22,k}^{\mathsf{debiased}}([\widehat{\Omega}_{k}^{\mathsf{est}}]^{-1})$, only in that $\widehat{\Omega}_{k,-i_{1},-i_{2}}^{\mathsf{est}}$ in $\widehat{\mathbb{IF}}_{22,k}^{\mathsf{debiased}}([\widehat{\Omega}_{k}^{\mathsf{est}}]^{-1})$ is replaced by $\widehat{\Omega}_{k}^{\mathsf{est}}$: 
\begin{equation}\label{equasi}
\widehat{\IIFF}_{22,k}^{\quasi} ([\widehat{\Omega}_{k}^{\est}]^{-1}) \coloneqq \frac{1}{n (n - 1)} \sum_{1 \leq i_{1} \neq i_{2} \leq n} \hat{\varepsilon}_{b, i_{1}} \zbar_{k} (X_{i_{1}})^{\top} Q \left( [\widehat{\Omega}_{k}^{\est}]^{-1}, \zbar_{k}(X_{i_{1}}),\zbar_{k}(X_{i_{2}}) \right) \zbar_{k}(X_{i_{2}}) \hat{\varepsilon}_{p, i_{2}}
\end{equation}
where\footnote{Similar to $\widehat{\IIFF}_{22,k}^{\debiased} ([\widehat{\Omega}_{k}^{\est}]^{-1})$, we cannot write $\widehat{\IIFF}_{22, k}^{\quasi} ([\widehat{\Omega}_{k}^{\est}]^{-1})$ in the form of $\widehat{\IIFF}_{22, k} (\widehat{\Omega}_{k}^{-1})$ and this is reflected in the notation by attaching a superscript ``$\mathsf{quasi}$'' on $\widehat{\IIFF}_{22,k}$.} 
\begin{equation*}
Q \left( [\widehat{\Omega}_{k}^{\mathsf{est}}]^{-1},\zbar_{k}(X_{i_{1}}),\zbar_{k}(X_{i_{2}})\right) \coloneqq [\widehat{\Omega}_{k}^{\est}]^{-1} + \frac{1}{n} [\widehat{\Omega}_{k}^{\est}]^{-1} \left( \begin{array}{c}
\zbar_{k}(X_{i_{1}}) \zbar_{k}(X_{i_{1}})^{\top} \\ 
+ \ \zbar_{k}(X_{i_{2}}) \zbar_{k}(X_{i_{2}})^{\top}
\end{array} \right) [\widehat{\Omega}_{k}^{\est}]^{-1}.
\end{equation*}

In terms of finite sample performance, as shown in column 5 of \Cref{tab:sample}, $\widehat{\mathbb{IF}}_{22,k}^{\mathsf{quasi}}([\widehat{\Omega}_{k}^{\mathsf{est}}]^{-1}) $ did not blow up numerically even when $k = 4096 \approx n = 5000$. For small $k$ (e.g. $k \leq 256$), the four estimators for unknown $\Omega_{k}^{-1}$ are all very close to the oracle $\widehat{\IIFF}_{22, k}$. For $k$ large compared to $n$ (e.g. $256 < k \leq 2048$), the MCavs of $\widehat{\mathbb{IF}}_{22,k}^{\mathsf{quasi}}([\widehat{\Omega}_{k}^{\mathsf{est}}]^{-1})$ are closer to that of $\widehat{\mathbb{IF}}_{22, k}$ than are the MCavs of $\widehat{\mathbb{IF}}_{22, k} ([\widehat{\Omega}_{k}^{\mathsf{est}}]^{-1})$ or $\widehat{\mathbb{IF}}_{22, k} ([\widehat{\Omega}_{k}^{\mathsf{tr}}]^{-1})$ or $\widehat{\mathbb{IF}}_{22 \rightarrow 33, k} ([\widehat{\Omega}_{k}^{\mathsf{tr}}]^{-1})$. For example, when $k = 2048$, the MCav of $\widehat{\mathbb{IF}}_{22,k}^{\mathsf{quasi}}([\widehat{\Omega }_{k}^{\mathsf{est}}]^{-1})$ (0.165) is closer to that of $\widehat{\mathbb{IF}}_{22, k}$ (0.161), compared to the MCav of $\widehat{\mathbb{IF}}_{22, k} ([\widehat{\Omega}_{k}^{\mathsf{est}}]^{-1})$ (0.0733). Furthermore, for $k \leq 2048$, the MCsds of $\widehat{\mathbb{IF}}_{22,k}^{\mathsf{quasi}}([\widehat{\Omega }_{k}^{\mathsf{est}}]^{-1})$ are also very close to the MCsds of $\widehat{\IIFF}_{22, k}$ (see the numbers in the parentheses in \Cref{tab:sample}). Unfortunately, we have not been able to derive a satisfactory upper bound on the estimation bias and variance of $\widehat{\mathbb{IF}}_{22,k}^{\mathsf{quasi}}([\widehat{\Omega}_{k}^{\mathsf{est}}]^{-1})$.

For $k$ comparable to $n$, i.e. when $k = 4096$, $\widehat{\mathbb{IF}}_{22,k}^{\mathsf{quasi}}([\widehat{\Omega}_{k}^{\mathsf{est}}]^{-1})$ performs poorly: the MCav of $\widehat{\mathbb{IF}}_{22,k}^{\mathsf{quasi}}([\widehat{\Omega}_{k}^{\mathsf{est}}]^{-1})$ at $k = 4096$ is even lower than that at $k = 2048$. But $\Bias_{\theta, k} (\hat{\psi}_{1})$ should monotonically increase with $k$ for $\psi(\theta) = \mathbb{E}_\theta [ \mathsf{var}_\theta (A | X) ]$. Specifically, as shown in \Cref{tab:sample}, the MCav of $\widehat{\mathbb{IF}}_{22,k}^{\mathsf{quasi}}([\widehat{\Omega}_{k}^{\mathsf{est}}]^{-1})$ decreased from 0.165 at $k = 2048$ to 0.0733 at $k = 4096$, while the MCav of $\widehat{\mathbb{IF}}_{22, k}$ increased from 0.161 at $k = 2048$ to 0.180 at $k = 4096$, as expected for $\psi(\theta) = \mathbb{E}_\theta [\var_{\theta}(A | X)]$.

Finally, we want to remark that although $\widehat{\mathbb{IF}}_{22,k}^{\mathsf{quasi}}([\widehat{\Omega}_{k}^{\mathsf{est}}]^{-1})$ has better finite sample performance than $\widehat{\mathbb{IF}}_{22 \rightarrow 33,k} ([\widehat{\Omega}_{k}^{\tr}]^{-1})$ based on our simulation studies, we do not yet have theoretical understanding on (1) the orders of the bias and variance of $\widehat{\mathbb{IF}}_{22,k}^{\mathsf{quasi}}([\widehat{\Omega}_{k}^{\mathsf{est}}]^{-1})$ as an estimator of $\Bias_{\theta, k} (\hat{\psi}_{1})$ and (2) whether the variance estimator proposed in \Cref{sec:var_quasi} is close to $\var_{\theta} [\widehat{\mathbb{IF}}_{22,k}^{\mathsf{quasi}}([\widehat{\Omega}_{k}^{\mathsf{est}}]^{-1})]$ is true. In \Cref{sec:cancel}, we provide an explanation of its numerical stability.

\begin{table}[h]
\caption{Simulation setup I: $\psi(\theta) = \mathbb{E}_\theta [ \mathsf{var}_\theta (A | X) ]$, regression functions estimated by nonparametric kernel regression with cross validation}
\label{tab:sample}\centering
\begin{tabular}{c|c|c|c|c|c}
\hline
$k$ & $\widehat{\mathbb{IF}}_{22, k}$ & $\widehat{\mathbb{IF}}_{22, k}\left( [\widehat{\Omega}_{k}^\mathsf{tr}]^{-1} \right)$ & $\widehat{\mathbb{IF}}_{22 \rightarrow 33, k}\left( [\widehat{\Omega}_{k}^\mathsf{tr}]^{-1} \right)$ & $\widehat{\mathbb{IF}}_{22, k}\left( [\widehat{\Omega}_{k}^\mathsf{est}]^{-1}\right)$ & $\widehat{\mathbb{IF}}_{22, k}^{\mathsf{quasi}}\left([\widehat{\Omega}_{k}^\mathsf{est}]^{-1} \right)$ \\ 
\hline
$64$ & 0.0457 (0.00782) & 0.0473 (0.00801) & 0.0453 (0.00762) & 0.0452 (0.00763) & 0.0465 (0.00785) \\ 
$128$ & 0.0484 (0.00831) & 0.0509 (0.00855) & 0.0470 (0.00777) & 0.0471 (0.00787) & 0.0498 (0.00831) \\ 
$256$ & 0.126 (0.0144) & 0.138 (0.0155) & 0.116 (0.0133) & 0.118 (0.0128) & 0.131 (0.0142) \\ 
$512$ & 0.127 (0.0147) & 0.147 (0.0175) & 0.102 (0.0149) & 0.113 (0.0124) & 0.136 (0.0150) \\ 
$1024$ & 0.129 (0.0172) & 0.171 (0.0264) & 0.0284 (0.0551) & 0.101 (0.0120) & 0.142 (0.0173) \\ 
$2048$ & 0.161 (0.0238) & 0.445 (0.434) & 43.330 (369.787) & 0.0935 (0.0124) & 0.165 (0.0222) \\ 
$4096$ & 0.180 (0.0322) & $6.220 \times 10^{7}$ ($3.215 \times 10^{8}$) & $-4.264 \times 10^{19}$ ($2.242 \times 10^{20}$) & 0.0310 (0.00830) & 0.0733 (0.0198) \\
\hline
\end{tabular}
\newline
A comparison among the MCavs and MCsds (in the parentheses) of $\widehat{\mathbb{IF}}_{22, k}$, $\widehat{\mathbb{IF}}_{22, k}\left( [\widehat{\Omega}_{k}^\mathsf{tr}]^{-1} \right)$, $\widehat{\mathbb{IF}}_{22 \rightarrow 33, k}\left( [\widehat{\Omega}_{k}^\mathsf{tr}]^{-1} \right)$, $\widehat{\mathbb{IF}}_{22, k} \left( [\widehat{\Omega}_{k}^\mathsf{est}]^{-1} \right)$ and $\widehat{\mathbb{IF}}_{22, k}^{\mathsf{quasi}}\left( [\widehat{\Omega}_{k}^\mathsf{est}]^{-1} \right)$. For more details on the data generating mechanism, see \Cref{sec:simulations}.
\end{table}

\subsection{Shrinkage covariance matrix estimator}\label{sec:shrink} 
\begin{table}[h]
\caption{Simulation setup I ($s_{f_X} = 0.1$): $\psi(\theta) = \mathbb{E}_\theta [ \mathsf{var}_\theta (A | X) ]$, regression functions estimated by nonparametric kernel regression with cross validation}
\label{tab:smooth_shrink}\centering
\begin{tabular}{l|lll}
\hline
$k$ & $\widehat{\IIFF}_{22, k}$ & $\widehat{\IIFF}_{22, k}^{\quasi} \left( [\widehat{\Omega}_{k}^{\est}]^{-1} \right)$ & $\widehat{\IIFF}_{22, k} \left( [\widehat{\Omega}_{k}^{\shrink}]^{-1} \right)$ \\ 
\hline
$512$ & 0.127 (0.0147) & 0.136 (0.0150) & 0.0157 (0.0152) \\ 
$1024$ & 0.129 (0.0172) & 0.142 (0.0173) & 161776 (61666.49) \\ 
$2048$ & 0.161 (0.0238) & 0.165 (0.0222) & 0.185 (0.0262) \\ 
$4096$ & 0.180 (0.0322) & 0.0733 (0.0198) & 0.225 (0.0374) \\ 
\hline
\end{tabular}
\newline
A comparison between $\widehat{\IIFF}_{22, k}$, $\widehat{\IIFF}_{22, k}^{\quasi} \left( [\widehat{\Omega}_{k}^{\est}]^{-1} \right)$, and $\widehat{\IIFF}_{22, k}\left( [\widehat{\Omega}_{k}^{\shrink}]^{-1} \right)$. The numbers in the parentheses are MCsds of the corresponding estimators. For more details on the data generating mechanism, see \Cref{sec:simulations}.
\end{table}

In this section, we explore whether it is possible to find an estimator $\widehat{\Omega}_{k}^{-1}$ for which the estimation bias of $\widehat{\mathbb{IF}}_{22, k} (\widehat{\Omega}_{k}^{-1})$ remains small when $k$ is near $n$. As $k$ gets close to $n$, the relevant asymptotic regime is no longer $k=o\left( n\right)$ but rather $k/n\rightarrow c$ for some $c\in (0,1)$ as $n\rightarrow \infty$. This motivated us to try a non-linear shrinkage covariance matrix estimator proposed in \citet{ledoit2012nonlinear, ledoit2017numerical, ledoit2018optimal} for this latter asymptotic regime. In our simulations, we implemented the estimator $\widehat{\mathbb{IF}}_{22, k} ([\widehat{\Omega}_{k}^{\mathsf{shrink}}]^{-1}),$ where $\widehat{\Omega }_{k}^{\mathsf{shrink}}$ is the nonlinear shrinkage covariance matrix estimator $\widehat{\Omega}_{k}^{\mathsf{shrink}}$ \citep{ledoit2012nonlinear, ledoit2017numerical} computed from the training sample data. $\widehat{\mathbb{IF}}_{22, k} ([\widehat{\Omega}_{k}^{\mathsf{shrink}}]^{-1})$ had very small estimation bias when $k$ is near $n$. In \Cref{tab:smooth_shrink}, even when $k = 4096$ and $n = 5000$ ($k / n \approx 0.8$), the MCav of $\widehat{\mathbb{IF}}_{22, k} ([\widehat{\Omega}_{k}^{\mathsf{shrink}}]^{-1})$ (0.225) is still quite close to that of $\widehat{\mathbb{IF}}_{22, k} (\Omega_{k}^{-1})$ (0.180) whereas all the other estimators including $\widehat{\mathbb{IF}}_{22,k}^{\mathsf{quasi}}([\widehat{\Omega}_{k}^{\mathsf{est}}]^{-1})$ do not perform well.

However, $\widehat{\mathbb{IF}}_{22, k} ([\widehat{\Omega}_{k}^{\mathsf{shrink}}]^{-1})$ does not always work, as evidenced by the MCav of $\widehat{\mathbb{IF}}_{22, k} ([\widehat{\Omega}_{k}^{\mathsf{shrink}}]^{-1})$ being more than $10^{5}$ times that of $\widehat{\mathbb{IF}}_{22, k}$ when $k = 1024$ in \Cref{tab:smooth_shrink} (where $\widehat{\mathbb{IF}}_{22,k}^{\mathsf{quasi}}([\widehat{\Omega}_{k}^{\text{est}}]^{-1})$ still performs well). It is an open problem to theoretically explain when $\widehat{\mathbb{IF}}_{22, k} ([\widehat{\Omega}_{k}^{\mathsf{shrink}}]^{-1})$ can be used to estimate $\Bias_{\theta, k} (\hat{\psi}_{1})$.

Because of the limitations of $\widehat{\mathbb{IF}}_{22, k} ([\widehat{\Omega}_{k}^{\mathsf{shrink}}]^{-1})$ and $\widehat{\mathbb{IF}}_{22,k}^{\mathsf{quasi}}([\widehat{\Omega}_{k}^{\mathsf{est}}]^{-1})$ discussed here and in \Cref{sec:quasi}, we develop a data-adaptive estimator $\widehat{\mathbb{IF}}_{22,k} (\widehat{\Omega}_{k}^{-1})$ in \Cref{sec:adaptive} to choose for each $k$, which of $\widehat{\mathbb{IF}}_{22, k} ([\widehat{\Omega}_{k}^{\mathsf{shrink}}]^{-1})$ or $\widehat{\mathbb{IF}}_{22,k}^{\mathsf{quasi}}([\widehat{\Omega}_{k}^{\mathsf{est}}]^{-1})$ should be used. 
These procedures are motivated by our simulation studies, as theoretical justifications are not available. It is also interesting to further investigate if other types of penalized or shrinkage methods work for estimating $\Bias_{\theta, k} (\hat{\psi}_{1})$. For example, see \citet{ledoit2004well, wang2015shrinkage, bodnar2016direct, wei2017estimation, donoho2018optimal, ke2019user}.

\begin{rem}[Asymptotic normality when $\Omega_{k}^{-1}$ needs to be estimated]\label{rem:anormal} 
The conditional asymptotic normalities of $$\frac{\widehat{\mathbb{IF}}_{22, k} ([\widehat{\Omega}_{k}^{\mathsf{tr}}]^{-1}) - \mathbb{E}_\theta [\widehat{\mathbb{IF}}_{22, k} ([\widehat{\Omega}_{k}^{\mathsf{tr}}]^{-1}) ]}{\se_{\theta} [\widehat{\mathbb{IF}}_{22, k} ([\widehat{\Omega}_{k}^{\mathsf{tr}}]^{-1})]}$$ and
$$\frac{\widehat{\mathbb{IF}}_{22 \rightarrow 33, k} ([\widehat{\Omega}_{k}^{\mathsf{tr}}]^{-1}) - \mathbb{E}_\theta [\widehat{\mathbb{IF}}_{22 \rightarrow 33, k} ([\widehat{\Omega}_{k}^{\mathsf{tr}}]^{-1}) ]}{\se_{\theta} [\widehat{\mathbb{IF}}_{22 \rightarrow 33, k} ([\widehat{\Omega}_{k}^{\mathsf{tr}}]^{-1})]}$$ 
follow from the same argument as the conditional asymptotic normality of the oracle $\widehat{\mathbb{IF}}_{22, k}$ because we can treat $\widehat{\Omega}_{k}^{\mathsf{tr}}$ as fixed by conditioning on the training sample or a third independent covariance matrix sample other than training sample/estimation sample. This argument would also imply the conditional asymptotic normality of $$\frac{\widehat{\mathbb{IF}}_{22, k} ([\widehat{\Omega}_{k}^{\mathsf{shrink}}]^{-1}) - \mathbb{E}_\theta [\widehat{\mathbb{IF}}_{22, k} ([\widehat{\Omega}_{k}^{\mathsf{shrink}}]^{-1}) ]}{\se_{\theta} [\widehat{\mathbb{IF}}_{22, k} ([\widehat{\Omega}_{k}^{\mathsf{shrink}}]^{-1})]}$$ if the eigenvalues of $\widehat{\Omega}_{k}^{\mathsf{shrink}}$ are bounded with high probability. This is still an open problem.

We have yet to prove the conditional asymptotic normality of $$\frac{\widehat{\mathbb{IF}}_{22, k}^{\quasi} ([\widehat{\Omega}_{k}^{\mathsf{est}}]^{-1}) - \mathbb{E}_\theta [\widehat{\mathbb{IF}}_{22, k}^{\quasi} ([\widehat{\Omega}_{k}^{\mathsf{est}}]^{-1}) ]}{\se_{\theta} [\widehat{\mathbb{IF}}_{22, k}^{\quasi} ([\widehat{\Omega}_{k}^{\mathsf{est}}]^{-1})]}$$ as $k, n \rightarrow \infty$. However, based on the qqplots \Cref{fig:normal} in \Cref{app:norm}, we conjecture that all these aforementioned ``estimators'' of $\Bias_{\theta, k} (\hat{\psi}_{1})$ for unknown $\Omega_{k}^{-1}$ are both conditionally asymptotic normal as $k, n \rightarrow \infty$ when $k = o(n^2)$ when they are numerically stable (i.e. not blowing up). For an example of non-normal distribution when numerically blowing up, see the 2nd row of the right panel of \Cref{fig:normal}, $\widehat{\IIFF}_{22, k} ([\widehat{\Omega}_{k}^{\shrink}]^{-1})$ obviously deviates from normality at $k = 1024$. At $k = 1024$, reading from \Cref{tab:smooth_shrink}, we do observe that $\widehat{\IIFF}_{22, k} ([\widehat{\Omega}_{k}^{\shrink}]^{-1})$ blows up numerically.
\end{rem}

\subsection{Asymptotic properties of the tests for $\H_{0,k}(\delta)$ \cref{h0_k} based on $[\widehat{\Omega}_{k}^{\tr}]^{-1}$}\label{sec:test} 
In previous sections we considered the estimation of $\Bias_{\theta, k} (\hat{\psi}_{1})$ when $\Omega_{k}^{-1}$ is unknown. In this section we consider if estimating $\Omega_{k}^{-1}$ by $[\widehat{\Omega}_{k}^{\tr}]^{-1}$ has an effect on the statistical properties of the test of $\H_{0,k}(\delta)$. For the estimators $\widehat{\mathbb{IF}}_{22,k}^{\mathsf{quasi}}([\widehat{\Omega }_{k}^{\mathsf{est}}]^{-1})$ and $\widehat{\mathbb{IF}}_{22, k} ([\widehat{\Omega}_{k}^{\mathsf{shrink}}]^{-1})$ contributing to the data-adaptive estimator $\widehat{\mathbb{IF}}_{22,k} (\widehat{\Omega}_{k}^{-1})$ reported in the right panel of \Cref{tab:intro} (see also later \Cref{sec:adaptive}), we have not as yet obtained satisfactory theoretical results on the estimation biases, variances or asymptotic normalities.

We first define the following one-sided test statistics corresponding to $\widehat{\chi}_{k}^{(1)} (\zeta_{k}, \delta)$ for $\psi (\theta) = \BE_{\theta} [\var_{\theta} (A | X)]$ when $\Omega_{k}^{-1}$ is estimated by $[\widehat{\Omega}_{k}^{\tr}]^{-1}$:
\begin{align}
\widehat{\chi}_{k}^{(1)} ([\widehat{\Omega}_{k}^{\tr}]^{-1}; \zeta_{k}, \delta) & = \mathbbm{1} \left\{ \frac{\widehat{\IIFF}_{22, k} ([\widehat{\Omega}_{k}^{\tr}]^{-1})}{\widehat{\se} \left[ \hat{\psi}_{1} \right]} - \zeta_{k} \frac{\widehat{\se} [\widehat{\IIFF}_{22, k} ([\widehat{\Omega}_{k}^{\tr}]^{-1})]}{\widehat{\se} \left[ \hat{\psi}_{1} \right]} > \delta \right\} \label{oneside_tr_2} \\
\widehat{\chi}_{33, k}^{(1)} ([\widehat{\Omega}_{k}^{\tr}]^{-1}; \zeta_{k}, \delta) & = \mathbbm{1} \left\{ \frac{\widehat{\IIFF}_{22 \rightarrow 33, k} ([\widehat{\Omega}_{k}^{\tr}]^{-1})}{\widehat{\se} \left[ \hat{\psi}_{1} \right]} - \zeta_{k} \frac{\widehat{\se} [\widehat{\IIFF}_{22 \rightarrow 33, k} ([\widehat{\Omega}_{k}^{\tr}]^{-1})]}{\widehat{\se} \left[ \hat{\psi}_{1} \right]} > \delta \right\} \label{oneside_tr_3}
\end{align}
where $\widehat{\se} [\widehat{\IIFF}_{22, k} ([\widehat{\Omega}_{k}^{\tr}]^{-1})] = \left\{ \widehat{\var} [\widehat{\IIFF}_{22, k} ([\widehat{\Omega}_{k}^{\tr}]^{-1})] \right\}^{1/2}$ and $\widehat{\se} [\widehat{\IIFF}_{22 \rightarrow 33, k} ([\widehat{\Omega}_{k}^{\tr}]^{-1})] = \left\{ \widehat{\var} [\widehat{\IIFF}_{22 \rightarrow 33, k} ([\widehat{\Omega}_{k}^{\tr}]^{-1})] \right\}^{1/2}$, with $\widehat{\var} [\widehat{\IIFF}_{22, k} ([\widehat{\Omega}_{k}^{\tr}]^{-1})]$ and $\widehat{\var} [\widehat{\IIFF}_{22 \rightarrow 33, k} ([\widehat{\Omega}_{k}^{\tr}]^{-1})]$ defined later in \cref{eq:est.var} and \cref{eq:est.var_if2233} respectively. 

Similarly, we define the following two-sided test statistics corresponding to $\widehat{\chi}_{k}^{(2)} (\zeta_{k}, \delta)$ for $\psi (\theta) = \BE_{\theta} [\cov_{\theta} (A, Y | X)]$ when $\Omega_{k}^{-1}$ is estimated by $[\widehat{\Omega}_{k}^{\tr}]^{-1}$:
\begin{align}
\widehat{\chi}_{k}^{(2)} ([\widehat{\Omega}_{k}^{\tr}]^{-1}; \zeta_{k}, \delta) & = \mathbbm{1} \left\{ \frac{\vert \widehat{\IIFF}_{22, k} ([\widehat{\Omega}_{k}^{\tr}]^{-1}) \vert}{\widehat{\se} \left[ \hat{\psi}_{1} \right]} - \zeta_{k} \frac{\widehat{\se} [\widehat{\IIFF}_{22, k} ([\widehat{\Omega}_{k}^{\tr}]^{-1})]}{\widehat{\se} \left[ \hat{\psi}_{1} \right]} > \delta \right\} \label{twoside_tr_2} \\
\widehat{\chi}_{33, k}^{(2)} ([\widehat{\Omega}_{k}^{\tr}]^{-1}; \zeta_{k}, \delta) & = \mathbbm{1} \left\{ \frac{\vert \widehat{\IIFF}_{22 \rightarrow 33, k} ([\widehat{\Omega}_{k}^{\tr}]^{-1}) \vert}{\widehat{\se} \left[ \hat{\psi}_{1} \right]} - \zeta_{k} \frac{\widehat{\se} [\widehat{\IIFF}_{22 \rightarrow 33, k} ([\widehat{\Omega}_{k}^{\tr}]^{-1})]}{\widehat{\se} \left[ \hat{\psi}_{1} \right]} > \delta \right\} \label{twoside_tr_3}
\end{align}

The following proposition, which is a consequence of the variance bound of $\widehat{\IIFF}_{22 \rightarrow 33, k} ([\widehat{\Omega}_{k}^{\tr}]^{-1})$ given in \Cref{prop:var_if2233} and the estimation bias bound of $\widehat{\IIFF}_{22 \rightarrow 33, k} ([\widehat{\Omega}_{k}^{\tr}]^{-1})$ obtained in \Cref{rem:simpler}, shows that $\widehat{\chi}_{33, k}^{(1)} ([\widehat{\Omega}_{k}^{\tr}]^{-1}; \zeta_{k}, \delta)$ (or $\widehat{\chi}_{33, k}^{(2)} ([\widehat{\Omega}_{k}^{\tr}]^{-1}; \zeta_{k}, \delta)$) is asymptotically the same as the oracle test $\widehat{\chi}_{k}^{(1)} (\zeta_{k}, \delta)$ (or $\widehat{\chi}_{k}^{(2)} (\zeta_{k}, \delta)$) with known $\Omega_{k}^{-1}$.
\begin{proposition}\label{prop:test}
Under \Cref{cond:w} and the additional restriction $\Bias_{\theta, k} (\hat{\psi}_{1}) \neq o (\BL_{2, b, k} \BL_{2, p, k})$, in the event that $\widehat{\Omega}_{k}^{\tr}$ is invertible, if $k \rightarrow \infty$ as $n \rightarrow \infty$ and $k = o(n / \log^{2}(n))$, for any given $\delta, \zeta_{k} > 0$, suppose that $\frac{\vert \Bias_{\theta, k} (\hat{\psi}_{1}) \vert}{\se_\theta [\hat{\psi}_{1}]} = \gamma$ for some (sequence) $\gamma = \gamma(n)$ (where $\gamma(n)$ can diverge with $n$), we have
\begin{enumerate}[label=(\arabic*)]
\item when $\psi (\theta) = \BE_{\theta}[\var_\theta(A | X)]$ (or $\psi (\theta) = \BE_{\theta}[\cov_\theta(A, Y | X)]$) and under $\H_{0, k} (\delta): \gamma \leq \delta$, $\widehat{\chi}_{33, k}^{(1)} ([\widehat{\Omega}_{k}^{\tr}]^{-1}; \zeta_{k}, \delta)$ (or $\widehat{\chi}_{33, k}^{(2)} ([\widehat{\Omega}_{k}^{\tr}]^{-1}; \zeta_{k}, \delta)$) rejects the null with probability less than or equal to $1 - \Phi (\zeta _{k})$ (or less than or equal to $2 - 2 \Phi (\zeta _{k})$), as $n \rightarrow \infty$;

\item when $\psi (\theta) = \BE_{\theta}[\var_\theta(A | X)]$ and under the following alternative to $\H_{0, k} (\delta)$: $\gamma = \delta + c$, for any fixed $c > 0$ or any diverging sequence $c = c(n) \rightarrow \infty$, $\widehat{\chi}_{33, k}^{(1)} ([\widehat{\Omega}_{k}^{\tr}]^{-1}; \zeta_{k}, \delta)$ rejects the null probability converging to 1, as $n \rightarrow \infty$;

\item when $\psi (\theta) = \BE_{\theta}[\cov_\theta(A, Y | X)]$ and under the following alternative to $\H_{0, k} (\delta)$: $\gamma = \delta + c$, for any diverging sequence $c = c(n) \rightarrow \infty$, $\widehat{\chi}_{33, k}^{(2)} ([\widehat{\Omega}_{k}^{\tr}]^{-1}; \zeta_{k}, \delta)$ rejects the null probability converging to 1, as $n \rightarrow \infty$;
\end{enumerate}
\begin{enumerate}[label=(\arabic*')]
\setcounter{enumi}{2}
\item when $\psi (\theta) = \BE_{\theta}[\cov_\theta(A, Y | X)]$ and under the following alternative to $\H_{0, k} (\delta)$: $\gamma = \delta + c$, if both $\BL_{2, b, k}$ and $\BL_{2, p, k}$ are $o(1)$, then for any fixed $c > 0$ or any diverging sequence $c = c(n) \rightarrow \infty$, $\widehat{\chi}_{33, k}^{(2)} ([\widehat{\Omega}_{k}^{\tr}]^{-1}; \zeta_{k}, \delta)$ has rejection probability converging to 1, as $n \rightarrow \infty$.
\end{enumerate}
\end{proposition}

\begin{rem}\leavevmode
\begin{itemize}
\item We prove \Cref{prop:test} in \Cref{app:test}.

\item $\widehat{\chi}_{33, k}^{(1)} ([\widehat{\Omega}_{k}^{\tr}]^{-1}; \zeta_{k}, \delta)$ (or $\widehat{\chi}_{33, k}^{(2)} ([\widehat{\Omega}_{k}^{\tr}]^{-1}; \zeta_{k}, \delta)$) instead of $\widehat{\chi}_{k}^{(1)} ([\widehat{\Omega}_{k}^{\tr}]^{-1}; \zeta_{k}, \delta)$ (or $\widehat{\chi}_{k}^{(2)} ([\widehat{\Omega}_{k}^{\tr}]^{-1}; \zeta_{k}, \delta)$) is used due to its smaller estimation bias $\mathit{EB}_{\theta, 3, k} ([\widehat{\Omega}_{k}^{\mathsf{tr}}]^{-1}) = o \left\{\BL_{2, b, k} \BL_{2, p, k} \frac{k \log(k)}{n} \right\}$, compared to $\mathit{EB}_{\theta, 2, k} ([\widehat{\Omega}_{k}^{\mathsf{tr}}]^{-1}) = o \left\{\BL_{2, b, k} \BL_{2, p, k} \sqrt{\frac{k \log(k)}{n}} \right\}$. Under $\H_{0, k}(\delta)$ and the additional restriction $\Bias_{\theta, k }(\hat{\psi}_{1}) \neq o \left\{\BL_{2, b, k} \BL_{2, p, k}\right\}$, $\frac{\mathit{EB}_{\theta, 2, k} ([\widehat{\Omega}_{k}^{\mathsf{tr}}]^{-1})}{\se_{\theta} (\hat{\psi}_{1})} = \sqrt{\frac{k \log(k)}{n}}$ might be greater in order than $\frac{\se_{\theta} (\widehat{\IIFF}_{22, k} ([\widehat{\Omega}_{k}^{\mathsf{tr}}]^{-1}))}{\se_{\theta} (\hat{\psi}_{1})} = O \left( \max \left\{ \sqrt{\frac{k}{n}}, \BL_{2, b, k}, \BL_{2, p, k} \right\} \right)$, and this is not sufficient to control the level of the test to be smaller than the desired level determined by $\zeta_{k}$. However, $\frac{\mathit{EB}_{\theta, 3, k} ([\widehat{\Omega}_{k}^{\mathsf{tr}}]^{-1})}{\se_{\theta} (\hat{\psi}_{1})} = \frac{k \log(k)}{n}$ is guaranteed to be smaller in order than $\frac{\se_{\theta} (\widehat{\IIFF}_{22, k} ([\widehat{\Omega}_{k}^{\mathsf{tr}}]^{-1}))}{\se_{\theta} (\hat{\psi}_{1})}$. More detailed explanation can be found in \Cref{app:test}.

\item We need the additional restriction that $\Bias_{\theta, k }(\hat{\psi}_{1}) \neq o \left\{\BL_{2, b, k} \BL_{2, p, k}\right\}$ for the following reason: The known upper bound on the estimation bias $\mathit{EB}_{\theta, 3, k} ([\widehat{\Omega}_{k}^{\mathsf{tr}}]^{-1})$ of $\widehat{\IIFF}_{22 \rightarrow 33, k} ([\widehat{\Omega}_{k}^{\tr}]^{-1})$ is controlled in terms of $\BL_{2, b, k} \BL_{2, p, k}$, which equals $\Bias_{\theta, k}(\hat{\psi}_{1})$ for $\psi(\theta) = \mathbb{E}_\theta [ \mathsf{var}_\theta (A | X) ]$ but only upper bounds the absolute value of $\Bias_{\theta, k}(\hat{\psi}_{1})$ for $\psi(\theta) = \mathbb{E}_\theta [ \cov_\theta (A, Y | X) ]$. Without this additional restriction, $\Bias_{\theta, k}(\hat{\psi}_{1})$ can be $o(n^{-1/2})$ even when $\BL_{2, b, k} \BL_{2, p, k}$ is large. As a result, we are not able to show $\frac{\mathit{EB}_{\theta, 3, k} ([\widehat{\Omega}_{k}^{\mathsf{tr}}]^{-1})}{\se_{\theta} (\hat{\psi}_{1})} \ll \frac{\se_{\theta} (\widehat{\IIFF}_{22, k} ([\widehat{\Omega}_{k}^{\mathsf{tr}}]^{-1}))}{\se_{\theta} (\hat{\psi}_{1})}$ under $\H_{0, k} (\delta)$ as in the previous paragraph. We are now working on proving that it is possible to use higher order influence functions with order increasing with sample size to further relax this assumption and we plan to study this problem in a future paper. 
\end{itemize}
\end{rem}

Similar to \Cref{sec:var}, we define the following one-sided upper confidence bound when $\psi (\theta) = \BE_{\theta}[\var_{\theta}(A | X)]$:
\begin{align}
\UCB^{(1)} ([\widehat{\Omega}_{k}^{\tr}]^{-1}; \alpha, \alpha^{\dag}) \coloneqq \TC_\alpha \left( \left[ \frac{\widehat{\IIFF}_{22 \rightarrow 33, k} ([\widehat{\Omega}_{k}^{\tr}]^{-1}) - z_{\alpha^{\dag}} \widehat{\se} [\widehat{\IIFF}_{22 \rightarrow 33, k} ([\widehat{\Omega}_{k}^{\tr}]^{-1})]}{\widehat{\se} [\hat{\psi}_{1}]} \right] \right) \label{ucbone_tr_3}.
\end{align}

We have shown in \Cref{prop:test} that $\widehat{\chi}_{3, k}^{(1)} ([\widehat{\Omega}_{k}^{\tr}]^{-1}; \zeta_{k}, \delta)$ and $\widehat{\chi}_{3, k}^{(2)} ([\widehat{\Omega}_{k}^{\tr}]^{-1}; \zeta_{k}, \delta)$ are asymptotically valid $(1 - \Phi(\zeta_{k}))$ level one-sided and $(2 - 2 \Phi(\zeta_{k}))$ level two-sided tests of the surrogate null hypothesis $\H_{0, k} (\delta)$ for $\psi (\theta) = \BE_{\theta} [\var_{\theta}(A | X)]$ and $\psi (\theta) = \BE_{\theta} [\cov_{\theta}(A, Y | X)]$ respectively. It further implies the following result on the upper confidence bound when $\psi (\theta) = \BE_{\theta} [\var_{\theta}(A | X)]$:
\begin{cor}
Under the conditions in \Cref{prop:test}, $\UCB^{(1)} ([\widehat{\Omega}_{k}^{\tr}]^{-1}; \alpha, \alpha^{\dag})$ is an asymptotically valid nominal $(1 - \alpha^{\dag})$ one-sided confidence bound for the true coverage of a two-sided Wald CI centered at $\hat{\psi}_{1}$ for $\BE_{\theta} [\hat{\psi}_{2, k}] = \psi (\theta) + \TB_{\theta, k} (\hat{\psi}_{1})$.
\end{cor}

Finally, the following corollary of \Cref{prop:test} summarizes the implication of the test results and upper confidence bounds on the actual null hypothesis of interest $\H_{0} (\delta)$ and parameter $\psi (\theta)$.
\begin{cor}
Under the conditions in \Cref{prop:test}:
\begin{enumerate}
\item for $\psi (\theta) = \BE_{\theta} [\var_{\theta}(A | X)]$,
\begin{itemize}
\item $\widehat{\chi}_{33, k}^{(1)} ([\widehat{\Omega}_{k}^{\tr}]^{-1}; \zeta_{k}, \delta)$ is an asymptotically valid level $(1 - \Phi(\zeta_{k}))$ one-sided test of $\H_{0} (\delta)$, and
\item $\UCB^{(1)} ([\widehat{\Omega}_{k}^{\tr}]^{-1}; \alpha, \alpha^{\dag})$ is an asymptotically valid nominal $(1 - \alpha^{\dag})$ one-sided upper confidence bound for the true coverage of a two-sided Wald CI centered at $\hat{\psi}_{1}$ for $\psi (\theta)$;
\end{itemize}
\item for $\psi (\theta) = \BE_{\theta} [\cov_{\theta}(A, Y | X)]$, under \Cref{cond:ff},
\begin{itemize}
\item $\widehat{\chi}_{33, k}^{(2)} ([\widehat{\Omega}_{k}^{\tr}]^{-1}; \zeta_{k}, \delta)$ is an asymptotically valid level $(2 - 2 \Phi(\zeta_{k}))$ two-sided test of $\H_{0} (\delta \delta')$.
\end{itemize}
\end{enumerate}
\end{cor}

\section{Technical details for calculations related to $\widehat{\mathbb{IF}}_{22, k}^\mathsf{quasi} ( [\widehat{\Omega}_{k}^\mathsf{est}]^{-1} )$}\label{sec:cov_proof}
\subsection{Derivation of \cref{eq:heuristic}}\label{app:id} 
\begin{align}
& \; \mathit{EB}_{\theta, 2, k} ( [\widehat{\Omega}_{k}^\mathsf{est}]^{-1} ) \equiv \mathbb{E}_\theta \left[ \widehat{\mathbb{IF}}_{22, k} ( [\widehat{\Omega}_{k}^\mathsf{est}]^{-1} ) - \widehat{\mathbb{IF}}_{22, k} \right]  \notag \\
= & \; \mathbb{E}_\theta \left[ \widehat\xi_{b,1}\zbar_{k}(X_{1})^{\top} \cdot \left\{ \left( \frac{1}{n}\sum_{i=1}^{n}\zbar_{k}(X_{i}) \zbar_{k}(X_{i})^{\top}\right) ^{-1} - \Omega_{k}^{-1} \right\} \cdot \zbar_{k}(X_{2}) \widehat\xi_{p,2} \right]  \notag \\
= & \; \mathbb{E}_\theta \left[ \widehat\xi_{b,1}\zbar_{k}(X_{1})^{\top} \cdot \left\{ \left( \widehat{\Omega}_{k,-1,-2}^{\mathsf{est}} + \frac{\underset{i=1,2} \sum \zbar_{k}(X_{i})\zbar_{k}(X_{i})^{\top}}{n}\right) ^{-1} - \Omega_{k}^{-1} \right\} \cdot \zbar_{k}(X_{2})\widehat\xi_{p,2}  \right]  \notag \\
= & \; \mathbb{E}_\theta \left[ \widehat\xi_{b,1} \zbar_{k}(X_{1})^{\top} \cdot \left( [\widehat{\Omega}_{k,-1,-2}^{\mathsf{est}}]^{-1} - \Omega_{k}^{-1}\right) \cdot \zbar_{k}(X_{2})\widehat\xi_{p,2} \right]  \notag \\
& \; - \frac{1}{n} \mathbb{E}_\theta \left[ \widehat\xi_{b,1}\zbar_{k}(X_{1})^{\top} \cdot [\widehat{\Omega}_{k,-1,-2}^{\mathsf{est}}]^{-1}\cdot \underset{i=1,2} \sum \zbar_{k}(X_{i})\zbar_{k}(X_{i})^{\top} \cdot [\widehat{\Omega}_{k}^{\mathsf{est}}]^{-1}\cdot \zbar_{k}(X_{2})\widehat\xi_{p_{2}}\right]  \notag \\
\coloneqq & \; \text{(I)}+\text{(II)}.
\end{align}
where the second equality follows from the definition 
\begin{equation*}
\widehat{\Omega}_{k,-i_{1},-i_{2}}^{\mathsf{est}}\coloneqq\frac{1}{n} \sum_{i \in \mathsf{est}:i\neq i_{1},i_{2}}\zbar_{k}(X_{i})\zbar_{k}(X_{i})^{\top}
\end{equation*}
for any $1\leq i_{1}\neq i_{2}\leq n$, and the third equality is due to the exact expansion of the matrix inverse 
\begin{align*}
& \; [\widehat{\Omega}_{k}^{\mathsf{est}}]^{-1} = \left( \widehat{\Omega}_{k,-1,-2}^{\mathsf{est}}+\frac{\underset{i=1,2}{\sum }\zbar_{k}(X_{i}) \zbar_{k}(X_{i})^{\top}}{n}\right)^{-1} \\
= & \; [\widehat{\Omega}_{k,-1,-2}^{\mathsf{est}}]^{-1}-\frac{1}{n}[\widehat{\Omega }_{k,-1,-2}^{\mathsf{est}}]^{-1}\cdot \underset{i=1,2}{\sum }\zbar_{k}(X_{i})\zbar_{k}(X_{i})^{\top}\cdot [ \widehat{\Omega}_{k}^{\mathsf{est}}]^{-1}.
\end{align*}

\subsection{Upper bound on (I) of \cref{eq:heuristic}}\label{app:I} 
\begin{align*}
|\text{(I)}| = & \; \left\vert \mathbb{E}_\theta\left[ \widehat\xi_{b}\zbar_{k}(X)^{\top}\right] \cdot \mathbb{E}_\theta\left[ [\widehat{\Omega}_{k,-1,-2}^{\mathsf{est}}]^{-1}-\Omega_{k}^{-1}\right] \cdot \mathbb{E}_\theta\left[ \zbar_{k}(X)\hat{\xi}_{p} \right] \right\vert \\
\leq & \; \left\vert \mathbb{E}_\theta\left[ \widehat\xi_{b}\zbar_{k}(X)^{\top}\right] \cdot \mathbb{E}_\theta\left[ \Omega_{k}^{-1}\left( \widehat{\Omega}_{k,-1,-2}^{\mathsf{est}}-\Omega_{k}\right) \Omega_{k}^{-1}\right] \cdot \mathbb{E}_\theta\left[ \zbar_{k}(X)\widehat\xi_{p}\right] \right\vert \\
& + \left\vert \mathbb{E}_\theta\left[ \widehat\xi_{b}\zbar_{k}(X)^{\top}\right] \cdot \mathbb{E}_\theta\left[ \left\{ \Omega_{k}^{-1}\left( \widehat{\Omega}_{k,-1,-2}^{\mathsf{est}}-\Omega_{k}\right) [\widehat{\Omega}_{k,-1,-2}^{\mathsf{est}}]^{-1/2}\right\} ^{2}\right] \cdot \mathbb{E}_\theta\left[ \zbar_{k}(X)\widehat\xi_{p} \right] \right\vert \\
=& \; \frac{2}{n-2}\left\vert \mathbb{E}_\theta\left[ \widehat\xi_{b} \zbar_{k}(X)^{\top}\right] \cdot \Omega_{k}^{-1} \cdot \mathbb{E}_\theta\left[ \zbar_{k}(X)\widehat\xi_{p} \right] \right\vert \\
& + \left\vert \mathbb{E}_\theta\left[ \widehat\xi_{b}\zbar_{k}(X)^{\top}\right] \cdot \mathbb{E}_\theta\left[ \left\{ \Omega_{k}^{-1}\left( \widehat{\Omega}_{k,-1,-2}^{\mathsf{est}}-\Omega_{k}\right) [\widehat{\Omega}_{k,-1,-2}^{\mathsf{est}}]^{-1/2}\right\} ^{2}\right] \cdot \mathbb{E}_\theta\left[ \zbar_{k}(X)\widehat\xi_{p}\right] \right\vert \\
\leq & \; \BL_{2,b,k} \BL_{2,p,k} \left( \frac{2}{n-2}+\mathbb{E}_\theta \left[ \left\Vert \widehat{\Omega}_{k,-1,-2}^{\mathsf{est}}-\Omega \right\Vert^{2} \right] \cdot \Vert \Omega_{k}^{-1} \Vert \cdot \Vert [ \widehat{\Omega}_{k,-1,-2}^{\mathsf{est}}]^{-1}\Vert \right) \\
\lesssim & \; \BL_{2,b,k} \BL_{2,p,k} \left\{ \frac{2}{n} + \mathbb{E}_\theta\left[ \left\Vert \widehat{\Omega}_{k,-1,-2}^{\mathsf{est}}-\Omega \right\Vert^{2} \right] \right\} \\
\lesssim & \; \BL_{2,b,k} \BL_{2,p,k} \mathbb{E}_\theta\left[ \left\Vert  \widehat{\Omega}_{k,-1,-2}^{\mathsf{est}}-\Omega \right\Vert^{2} \right],
\end{align*}
where the second line inequality follows from the following exact expansion of matrix inverse 
\begin{equation*}
[ \widehat{\Omega}_{k,-1,-2}^{\mathsf{est}}]^{-1} \equiv \Omega_{k}^{-1} - \Omega_{k}^{-1} \left( \widehat{\Omega}_{k,-1,-2}^{\mathsf{est}} - \Omega_{k} \right) \Omega_{k}^{-1} + \left[ \Omega_{k}^{-1} \left( \widehat{\Omega}_{k,-1,-2}^{\mathsf{est}}-\Omega_{k} \right) \right]^{2} [\widehat{\Omega}_{k,-1,-2}^{\mathsf{est}}]^{-1}
\end{equation*}
and triangle inequality, the third line equality follows from the fact that $\widehat{\Omega}_{k,-1,-2}^{\mathsf{est}}$ unbiasedly estimates $\frac{n}{n-2} \Omega_{k}$, the fourth line inequality applies Cauchy-Schwarz inequality, the definition of operator norm and the last line inequality follows from the contraction norm property of linear projections and the assumption that $\Omega_{k}$ and $\widehat{\Omega }_{k,-1,-2}^{\mathsf{est}}$ both have bounded eigenvalues.

\subsection{A possible explanation of stability of $\widehat{\IIFF}_{22, k} ( [\widehat{\Omega}_{k}^\mathsf{est}]^{-1} )$ and $\widehat{\mathbb{IF}}_{22, k}^\mathsf{quasi} ( [\widehat{\Omega}_{k}^\mathsf{est}]^{-1} )$}\label{sec:cancel} 
In this section, as promised, we discuss why $\widehat{\mathbb{IF}}_{22, k} ( [\widehat{\Omega}_{k}^\mathsf{est}]^{-1} )$ and $\widehat{\mathbb{IF}}_{22, k}^\mathsf{quasi} ( [\widehat{\Omega}_{k}^\mathsf{est}]^{-1} )$ are more stable than $\widehat{\mathbb{IF}}_{22, k} ( [\widehat{\Omega}_{k}^\mathsf{tr}]^{-1} )$ in finite sample. We consider the matrix form of $\widehat{\mathbb{IF}}_{22, k} ( [\widehat{\Omega}_{k}^\mathsf{est}]^{-1} )$: 
\begin{align*}
& \; \widehat{\mathbb{IF}}_{22, k} ( [\widehat{\Omega}_{k}^\mathsf{est}]^{-1} ) \\
= & \;\frac{1}{n - 1} \widehat{\bm\varepsilon}_b^{\top} \left\{ \mathbf{\Zbar}_{k}^\mathsf{est} \cdot \left( \mathbf{\Zbar}_{k}^{\mathsf{est} \top} \mathbf{\Zbar}_{k}^\mathsf{est} \right)^{-1} \mathbf{\Zbar}_{k}^{\mathsf{est} \top} - \mathsf{Diag} \left( \mathbf{\Zbar}_{k}^\mathsf{est} \left( \mathbf{\Zbar}_{k}^{\mathsf{est} \top} \mathbf{\Zbar}_{k}^\mathsf{est} \right)^{-1} \cdot \mathbf{\Zbar}_{k}^{\mathsf{est} \top} \right) \right\} \widehat{\bm\varepsilon}_p
\end{align*}
where $\widehat{\bm\varepsilon}_b = \left( \hat{\varepsilon}_{b, 1}, \dots, \hat{\varepsilon}_{b, n} \right)^{\top}$, $\widehat{\bm\varepsilon}_p = \left( \hat{\varepsilon}_{p, 1}, \dots, \hat{\varepsilon}_{p, n} \right)^{\top}$, 
\begin{equation*}
\mathbf{\Zbar}_{k}^{\est} = \left( \begin{array}{c}
\zbar_{k} (X_1)^{\top} \\ 
\vdots \\ 
\zbar_{k} (X_n)^{\top}
\end{array} \right) = \left( \begin{array}{ccc}
\z_1 (X_1) & \dots & \z_{k} (X_1) \\ 
\vdots & \ddots & \vdots \\ 
\z_1 (X_n) & \dots & \z_{k} (X_n)
\end{array} \right)
\end{equation*}
and $\mathsf{Diag} \left( \mathbf{M} \right)$ denotes the diagonal matrix with the diagonal elements of matrix $\mathbf{M}$. Consider the singular value decomposition (SVD) of $\mathbf{\Zbar}_{k}^\mathsf{est} = \widehat{\mathbf{U}}_{k} \widehat{\mathbf{D}}_{k} \widehat{\mathbf{V}}_{k}^{\top}$ from the estimation sample, where $\widehat{\mathbf{D}}_{k}^{2}$ is the eigenvalues of the sample covariance matrix estimator $\widehat{\Omega}_{k}^\mathsf{est}$ up to constant. Then it is easy to see that 
\begin{align*}
& \widehat{\mathbb{IF}}_{22, k} ( [\widehat{\Omega}_{k}^\mathsf{est}]^{-1} ) \\
= & \frac{1}{n - 1} \widehat{\bm\varepsilon}_b^{\top} \left\{ \begin{array}{c}
\widehat{\mathbf{U}}_{k} \widehat{\mathbf{D}}_{k} \widehat{\mathbf{V}}_{k}^{\top} \left( \widehat{\mathbf{V}}_{k} \widehat{\mathbf{D}}_{k}^{2} \widehat{\mathbf{V}}_{k}^{\top}\right)^{-1} \widehat{\mathbf{V}}_{k} \widehat{\mathbf{D}}_{k} \widehat{\mathbf{U}}_{k}^{\top} \\ 
- \mathsf{Diag} \left( \widehat{\mathbf{U}}_{k} \widehat{\mathbf{D}}_{k} \widehat{\mathbf{V}}_{k}^{\top} \left( \widehat{\mathbf{V}}_{k} \widehat{\mathbf{D}}_{k}^2 \widehat{\mathbf{V}}_{k}^{\top} \right)^{-1} \widehat{\mathbf{V}}_{k} \widehat{\mathbf{D}}_{k} \widehat{\mathbf{U}}_{k}^{\top} \right)
\end{array} \right\} \cdot \widehat{\bm\varepsilon}_p \\
= & \frac{1}{n - 1} \widehat{\bm\varepsilon}_b^{\top} \left\{ \widehat{\mathbf{U}}_{k} \widehat{\mathbf{D}}_{k} \widehat{\mathbf{D}}_{k}^{-2} \widehat{\mathbf{D}}_{k} \widehat{\mathbf{U}}_{k}^{\top} - \mathsf{Diag} \left( \widehat{\mathbf{U}}_{k} \widehat{\mathbf{D}}_{k} \widehat{\mathbf{D}}_{k}^{-2} \widehat{\mathbf{D}}_{k} \widehat{\mathbf{U}}_{k}^{\top} \right) \right\} \cdot \widehat{\bm\varepsilon}_p
\\
= & \frac{1}{n - 1} \widehat{\bm\varepsilon}_b^{\top} \left\{ \widehat{\mathbf{U}}_{k} \widehat{\mathbf{U}}_{k}^{\top} - \mathsf{Diag} \left( \widehat{\mathbf{U}}_{k} \widehat{\mathbf{U}}_{k}^{\top} \right) \right\} \cdot \widehat{\bm\varepsilon}_p
\end{align*}
by which we can explicitly see how the eigenvalues $\widehat{\mathbf{D}}_{k}$ got cancelled from the second equality to the third equality.

Similarly, for $\widehat{\mathbb{IF}}_{22, k}^\mathsf{quasi} ( [\widehat{\Omega}_{k}^\mathsf{est}]^{-1} )$, we have 
\begin{align*}
& \; \widehat{\mathbb{IF}}_{22, k}^\mathsf{quasi} ( [\widehat{\Omega}_{k}^\mathsf{est}]^{-1} ) \\
= & \; \widehat{\mathbb{IF}}_{22, k} ( [\widehat{\Omega}_{k}^\mathsf{est}]^{-1} ) + \frac{1}{n - 1} \widehat{\bm\varepsilon}_b^{\top} \cdot \left[ \begin{array}{c}
\mathsf{Diag} \left( \widehat{\mathbf{U}}_{k} \cdot \widehat{\mathbf{U}}_{k}^{\top} \right) \left\{ \widehat{\mathbf{U}}_{k} \cdot \widehat{\mathbf{U}}_{k}^{\top} - \mathsf{Diag} \left( \widehat{\mathbf{U}}_{k} \cdot \widehat{\mathbf{U}}_{k}^{\top} \right) \right\} \\ 
+ \left\{ \widehat{\mathbf{U}}_{k} \cdot \widehat{\mathbf{U}}_{k}^{\top} - \mathsf{Diag} \left( \widehat{\mathbf{U}}_{k} \cdot \widehat{\mathbf{U}}_{k}^{\top} \right) \right\} \mathsf{Diag} \left( \widehat{\mathbf{U}}_{k} \cdot \widehat{\mathbf{U}}_{k}^{\top} \right) 
\end{array} \right] \cdot \widehat{\bm\varepsilon}_p,
\end{align*}
again without involving the eigenvalues of $\widehat{\Omega}_{k}^\mathsf{est}$. Moreover, with the SVD formulation, one can interpret $\widehat{\mathbb{IF}}_{22, k}^\mathsf{quasi} ( [\widehat{\Omega}_{k}^\mathsf{est}]^{-1} )$ as follows: given the basis matrix $\mathbf{\Zbar}_{k}$, one first obtains its left singular vector $\widehat{\mathbf{U}}_{k}$, then replaces $\mathbf{\Zbar}_{k}$ by $\widehat{\mathbf{U}}_{k}$ and replaces $\Omega_{k}$ by the identity matrix in $\widehat{\mathbb{IF}}_{22, k}$ to get $\widehat{\mathbb{IF}}_{22, k} ( [\widehat{\Omega}_{k}^\mathsf{est}]^{-1} )$, and finally adds the correction terms to get $\widehat{\mathbb{IF}}_{22, k}^\mathsf{quasi} ( [\widehat{\Omega}_{k}^\mathsf{est}]^{-1} )$.

Such ``cancellation of eigenvalues'' does not happen in $\widehat{\mathbb{IF}}_{22, k} ( [\widehat{\Omega}_{k}^\mathsf{tr}]^{-1} )$. Consider the SVD of $\mathbf{\Zbar}_{k}^\mathsf{tr} = \widetilde{\mathbf{U}}_{k} \widetilde{\mathbf{D}}_{k} \widetilde{\mathbf{V}}_{k}^{\top}$ from the training sample. Similar to $\widehat{\mathbb{IF}}_{22, k} ( [\widehat{\Omega}_{k}^\mathsf{est}]^{-1} )$, the matrix form of $\widehat{\mathbb{IF}}_{22, k} ( [\widehat{\Omega}_{k}^\mathsf{tr}]^{-1} )$ is 
\begin{align*}
& \widehat{\mathbb{IF}}_{22, k}\left( [\widehat{\Omega}_{k}^\mathsf{tr}]^{-1} \right) \\
= & \frac{1}{n - 1} \widehat{\bm\varepsilon}_b^{\top} \cdot \left\{ 
\begin{array}{c}
\widehat{\mathbf{U}}_{k} \widehat{\mathbf{D}}_{k} \widehat{\mathbf{V}}_{k}^{\top} \cdot \left( \widetilde{\mathbf{V}}_{k} \widetilde{\mathbf{D}}_{k}^2 \widetilde{\mathbf{V}}_{k}^{\top} \right)^{-1} \cdot \widehat{\mathbf{V}}_{k} \widehat{\mathbf{D}}_{k} \widehat{\mathbf{U}}_{k}^{\top} \\ 
- \mathsf{Diag} \left( \widehat{\mathbf{U}}_{k} \widehat{\mathbf{D}}_{k} \widehat{\mathbf{V}}_{k}^{\top} \cdot \left( \widetilde{\mathbf{V}}_{k} \widetilde{\mathbf{D}}_{k}^2 \widetilde{\mathbf{V}}_{k}^{\top} \right)^{-1} \cdot \widehat{\mathbf{V}}_{k} \widehat{\mathbf{D}}_{k} \widehat{\mathbf{U}}_{k}^{\top} \right)%
\end{array} \right\} \cdot \widehat{\bm\varepsilon}_p,
\end{align*}
in which case $\widehat{\mathbf{V}}_{k}^{\top} \widetilde{\mathbf{V}}_{k} \ne \text{Id}$, where $\text{Id}$ is the identity matrix and hence there is no cancellation in the eigenvalues as in $\widehat{\mathbb{IF}}_{22, k} ([\widehat{\Omega}_{k}^\mathsf{est}]^{-1} )$ or $\widehat{\mathbb{IF}}_{22, k}^\mathsf{quasi} ( [\widehat{\Omega}_{k}^\mathsf{est}]^{-1} )$.

\subsection{Proof of \Cref{prop:test}}\label{app:test}
As discussed in \Cref{sec:test}, since $\widehat{\IIFF}_{22, k}$ is generally unknown, it needs to be replaced by an estimator such as $\widehat{\mathbb{IF}}_{22 \rightarrow 33, k} ( [\widehat{\Omega}_{k}^\mathsf{tr}]^{-1} )$. For notational convenience, in this section we use $\widehat{\IIFF}_{22, k} (\widehat{\Omega}_{k}^{-1})$ to denote an estimator of $\Bias_{\theta, k} ( \hat{\psi}_{1} )$ with $\Omega_{k}^{-1}$ replaced by some generic covariance matrix estimator computed from the training sample. For example, If we use $\widehat{\mathbb{IF}}_{22 \rightarrow 33, k} ( [\widehat{\Omega}_{k}^\mathsf{tr}]^{-1} )$, then $\widehat{\mathbb{IF}}_{22, k} (\widehat{\Omega}_{k}^{-1}) = \widehat{\mathbb{IF}}_{22 \rightarrow 33, k} ( [\widehat{\Omega}_{k}^\mathsf{tr}]^{-1} )$. We denote $\BE_{\theta} [\widehat{\mathbb{IF}}_{22, k} (\widehat{\Omega}_{k}^{-1}) - \widehat{\mathbb{IF}}_{22, k}]$ as $\mathit{EB}_{\theta, k} (\widehat{\Omega}_{k}^{-1})$. Thus for $\widehat{\mathbb{IF}}_{22 \rightarrow 33, k} ( [\widehat{\Omega}_{k}^\mathsf{tr}]^{-1} )$, $\mathit{EB}_{\theta, k} (\widehat{\Omega}_{k}^{-1})$ is equivalent to $\mathit{EB}_{\theta, 3, k} ([\widehat{\Omega}_{k}^{\tr}]^{-1})$. We consider the following statistic used in the test $\widehat{\chi}_{k}^{(1)} (\widehat{\Omega}_{k}^{-1}; \zeta_{k}, \delta)$ (or $\widehat{\chi}_{k}^{(2)} (\widehat{\Omega}_{k}^{-1}; \zeta_{k}, \delta)$) after standardization: 
\begin{equation}  \label{eq:effect}
\begin{split}
& \; \frac{\widehat{\se} [\hat{\psi}_{1}]}{\widehat{\se} [ \widehat{\mathbb{IF}}_{22, k} (\widehat{\Omega}_{k}^{-1} ) ]} \left( \frac{\widehat{\mathbb{IF}}_{22, k} ( \widehat{\Omega}_{k}^{-1} )}{\widehat{\se} \left[ \hat{\psi}_{1} \right]} - \delta \right) \\
= & \; \left( \frac{\widehat{\mathbb{IF}}_{22, k} ( \widehat{\Omega}_{k}^{-1} ) - \Bias_{\theta, k} ( \hat{\psi}_{1} )}{\widehat{\se} \left[ \hat{\psi}_{1} \right]} \frac{\widehat{\se} [\hat{\psi}_{1}]}{\widehat{\se} [ \widehat{\mathbb{IF}}_{22, k} (\widehat{\Omega}_{k}^{-1} ) ]} \right) + \frac{\widehat{\se} [\hat{\psi}_{1}]}{\widehat{\se} [ \widehat{\mathbb{IF}}_{22, k} (\widehat{\Omega}_{k}^{-1} ) ]} \left( \frac{\Bias_{\theta, k} ( \hat{\psi}_{1} )}{\widehat{\se} \left[ \hat{\psi}_{1} \right]} - \delta \right) \\
= & \; \left( \frac{\widehat{\mathbb{IF}}_{22, k} ( \widehat{\Omega}_{k}^{-1} ) - \mathbb{E}_\theta [ \widehat{\mathbb{IF}}_{22, k} ( \widehat{\Omega}_{k}^{-1} ) ]}{\widehat{\se} \left[ \hat{\psi}_{1} \right]} \frac{\widehat{\se} [\hat{\psi}_{1}]}{\widehat{\se} [ \widehat{\mathbb{IF}}_{22, k} (\widehat{\Omega}_{k}^{-1} ) ]} \right) + \frac{\mathit{EB}_{\theta, k} ( \widehat{\Omega}_{k}^{-1} )}{\widehat{\se} \left[ \hat{\psi}_{1} \right]} \frac{\widehat{\se} [\hat{\psi}_{1}]}{\widehat{\se} [ \widehat{\mathbb{IF}}_{22, k} (\widehat{\Omega}_{k}^{-1} ) ]} \\
& \; + \frac{\widehat{\se} [\hat{\psi}_{1}]}{\widehat{\se} [ \widehat{\mathbb{IF}}_{22, k} (\widehat{\Omega}_{k}^{-1} ) ]} \left( \frac{\Bias_{\theta, k} ( \hat{\psi}_{1} )}{\widehat{\se} \left[ \hat{\psi}_{1} \right]} - \delta \right) \\
= & \; \left\{ \left( \frac{\widehat{\mathbb{IF}}_{22, k} ( \widehat{\Omega}_{k}^{-1} ) - \mathbb{E}_\theta [ \widehat{\mathbb{IF}}_{22, k} ( \widehat{\Omega}_{k}^{-1} ) ]}{\se_\theta \left[ \widehat{\mathbb{IF}}_{22, k} (\widehat{\Omega}_{k}^{-1} ) \right]} \right) + \left( \frac{\Bias_{\theta, k} ( \hat{\psi}_{1} ) + \mathit{EB}_{\theta, k} ( \widehat{\Omega}_{k}^{-1} )}{\se_\theta \left[ \hat{\psi}_{1} \right]} - \delta \right) \frac{\se_\theta [ \hat{\psi}_{1} ]}{\se_\theta [ \widehat{\mathbb{IF}}_{22, k} (\widehat{\Omega}_{k}^{-1} ) ]} \right\} (1 + o_{P_\theta}(1)) \\
= & \; \left\{ \left( \frac{\widehat{\mathbb{IF}}_{22, k} ( \widehat{\Omega}_{k}^{-1} ) - \mathbb{E}_\theta [ \widehat{\mathbb{IF}}_{22, k} ( \widehat{\Omega}_{k}^{-1} ) ]}{\se_\theta \left[ \widehat{\mathbb{IF}}_{22, k} (\widehat{\Omega}_{k}^{-1} ) \right]} \right) + \left( \gamma + \frac{\mathit{EB}_{\theta, k} ( \widehat{\Omega}_{k}^{-1} )}{\se_\theta \left[ \hat{\psi}_{1} \right]} - \delta \right) \frac{\se_\theta [ \hat{\psi}_{1} ]}{\se_\theta [ \widehat{\mathbb{IF}}_{22, k} (\widehat{\Omega}_{k}^{-1} ) ]} \right\} (1 + o_{P_\theta}(1)) \\
= & \; \left\{ \underbrace{\left( \frac{\widehat{\mathbb{IF}}_{22, k} ( \widehat{\Omega}_{k}^{-1} ) - \mathbb{E}_\theta [ \widehat{\mathbb{IF}}_{22, k} ( \widehat{\Omega}_{k}^{-1} ) ]}{\se_\theta \left[ \widehat{\mathbb{IF}}_{22, k} (\widehat{\Omega}_{k}^{-1} ) \right]} \right) - \left( \delta - \gamma \right) \frac{\se_\theta [ \hat{\psi}_{1} ]}{\se_\theta [ \widehat{\mathbb{IF}}_{22, k} (\widehat{\Omega}_{k}^{-1} ) ]}}_{A} + \underbrace{\frac{\mathit{EB}_{\theta, k} ( \widehat{\Omega}_{k}^{-1} )}{\se_\theta [ \widehat{\mathbb{IF}}_{22, k} (\widehat{\Omega}_{k}^{-1} ) ]}}_{B} \right\} (1 + o_{P_\theta}(1)).
\end{split}
\end{equation}

The effect of estimating $\Omega_{k}^{-1}$ on the asymptotic validity of the test $\widehat\chi_{k} (\widehat{\Omega}_{k}^{-1}, \zeta_{k}, \delta)$ of $\H_{0, k} (\delta)$ thus depends on the orders of terms A and B. A has variance 1 and mean $- \left( \delta - \gamma \right) \frac{\se_\theta [ \hat{\psi}_{1} ]}{\se_\theta [ \widehat{\mathbb{IF}}_{22, k} (\widehat{\Omega}_{k}^{-1} ) ]}$. B depends on the estimation bias due to estimating $\Omega_{k}^{-1}$ by $\widehat{\Omega}_{k}^{-1}$. Hence if we have:
\begin{enumerate}[label=(\arabic*)]
\item $\frac{\widehat{\mathbb{IF}}_{22, k} ( \widehat{\Omega}_{k}^{-1} ) - \mathbb{E}_\theta [ \widehat{\mathbb{IF}}_{22, k} ( \widehat{\Omega}_{k}^{-1} ) ]}{\se_\theta \left[ \widehat{\mathbb{IF}}_{22, k} (\widehat{\Omega}_{k}^{-1} ) \right]}$ is asymptotically $N(0, 1)$ conditional on the training sample;
\item $\frac{\se_\theta [\hat{\psi}_{1}]}{\se_\theta [ \widehat{\mathbb{IF}}_{22, k} (\widehat{\Omega}_{k}^{-1} ) ]} / \frac{\se_\theta [\hat{\psi}_{1}]}{\se_\theta [ \widehat{\mathbb{IF}}_{22, k} ]} \rightarrow 1$;
\item $B = o(1)$ under $\H_{0, k} (\delta)$ or fixed alternatives to $\H_{0, k} (\delta)$;
\end{enumerate}
\begin{enumerate}[label=(\arabic*')]
\setcounter{enumi}{2}
\item $B \ll - \left( \delta - \gamma \right) \frac{\se_\theta [ \hat{\psi}_{1} ]}{\se_\theta [ \widehat{\mathbb{IF}}_{22, k} (\widehat{\Omega}_{k}^{-1} ) ]}$ under diverging alternatives to $\H_{0, k} (\delta)$ i.e. $\delta - \gamma = c$ for some $c \rightarrow \infty$ (at any rate).
\end{enumerate}
$\widehat{\chi}_{k}^{(1)} (\widehat{\Omega}_{k}^{-1}; \zeta_{k}, \delta)$ is an asymptotically valid level $1 - \Phi(\zeta_{k})$ one-sided test for $\H_{0, k} (\delta)$ and rejects the null with probability approaching 1 under both fixed or diverging alternatives to $\H_{0, k} (\delta)$. (Similarly, $\widehat{\chi}_{k}^{(2)} (\widehat{\Omega}_{k}^{-1}; \zeta_{k}, \delta)$ is an asymptotically level $2 - 2 \Phi(\zeta_{k})$ two-sided test for $\H_{0, k} (\delta)$ and rejects the null with probability approaching 1 under diverging alternatives to $\H_{0, k} (\delta)$.)

Both (1) and (2) hold for $\widehat{\mathbb{IF}}_{22 \rightarrow 33, k} ( [\widehat{\Omega}_{k}^\mathsf{tr}]^{-1} )$ (see \Cref{rem:anormal}), implied by \Cref{rem:simpler} and \Cref{prop:var_if2233} respectively.

In terms of (3) and (3'), \citet[Theorem 4]{mukherjee2017semiparametric} implies that, under the conditions of \Cref{prop:test} and $\H_{0, k}(\delta)$, $\mathit{EB}_{\theta, 2, k} ( [\widehat{\Omega}_{k}^\mathsf{tr}]^{-1} ) = O (\BL_{2, b, k} \BL_{2, p, k} \sqrt{\frac{k\mathsf{log}(k)}{n}} ) \lesssim \frac{\sqrt{k \log(k)}}{n}$. Therefore $\frac{\mathit{EB}_{\theta, 2, k} ( [\widehat{\Omega}_{k}^\mathsf{tr}]^{-1} )}{\se_\theta [\widehat{\IIFF}_{22, k} ([\widehat{\Omega}_{k}^{\tr}]^{-1})]}$ could be of order $\sqrt{\log(k)}$ when $\se_\theta [\widehat{\IIFF}_{22, k} ([\widehat{\Omega}_{k}^{\tr}]^{-1})]$ is of order $\sqrt{k} / n$ (see \Cref{thm:soif}). However, \citet[Theorem 4]{mukherjee2017semiparametric} implies that $\mathit{EB}_{\theta, 3, k} ( [\widehat{\Omega}_{k}^\mathsf{tr}]^{-1} ) \lesssim \frac{k \log(k)}{n^{3/2}}$ which is of smaller order than $\sqrt{k} / n$. Thus if instead using $\widehat{\mathbb{IF}}_{22 \rightarrow 33, k} ( [\widehat{\Omega}_{k}^\mathsf{tr}]^{-1} )$: 
\begin{itemize}
\item Under $\H_{0, k} (\delta)$ or fixed alternatives to $\H_{0, k} (\delta)$, $B = o(1)$. Hence (3) is satisfied.
\item Under diverging alternatives to $\H_{0, k} (\delta)$ i.e. $\gamma - \delta = c \rightarrow \infty$, $(\gamma - \delta) \se_{\theta} (\hat{\psi}_{1}) \asymp \BL_{2, b, k} \BL_{2, p, k}$, 
\begin{align*}
B & \lesssim \frac{\BL_{2, b, k} \BL_{2, p, k}}{\se_\theta [ \widehat{\mathbb{IF}}_{22 \rightarrow 33, k} ([\widehat{\Omega}_{k}^{\tr}]^{-1} ) ]} \frac{k \log(k)}{n} \ll \frac{\BL_{2, b, k} \BL_{2, p, k}}{\se_\theta [ \widehat{\mathbb{IF}}_{22 \rightarrow 33, k} ([\widehat{\Omega}_{k}^{\tr}]^{-1} ) ]} \\
& \asymp - \left( \delta - \gamma \right) \frac{\se_\theta [ \hat{\psi}_{1} ]}{\se_\theta [ \widehat{\mathbb{IF}}_{22 \rightarrow 33, k} ([\widehat{\Omega}_{k}^{\tr}]^{-1} ) ]}.
\end{align*}
Hence (3') is satisfied.
\end{itemize}

In summary, $\widehat{\chi}_{33, k}^{(1)} ([\widehat{\Omega}_{k}^{\tr}]^{-1}; \zeta_{k}, \delta)$ is an asymptotically valid level $1 - \Phi(\zeta_{k})$ one-sided test for $\H_{0, k} (\delta)$ when $\psi (\theta) = \BE_{\theta}[\var_{\theta}(A | X)]$ and rejects the null with probability approaching 1 under both fixed and diverging alternatives to $\H_{0, k} (\delta)$. Similarly, $\widehat{\chi}_{33, k}^{(2)} ([\widehat{\Omega}_{k}^{\tr}]^{-1}; \zeta_{k}, \delta)$ is an asymptotically valid level $2 - 2 \Phi(\zeta_{k})$ two-sided test for $\H_{0, k} (\delta)$ when $\psi (\theta) = \BE_{\theta}[\cov_{\theta}(A, Y | X)]$ and rejects the null with probability approaching 1 under diverging alternatives to $\H_{0, k} (\delta)$.

\section{On the variances of higher order influence function estimators}\label{sec:est.var}
In this section, to avoid overloading notations, we abbreviate $\Omega_{k}^{- 1 / 2} \Zbar_{k} \hat{\varepsilon}_{b}$ as $\B$ and $\Omega_{k}^{- 1 / 2} \Zbar_{k} \hat{\varepsilon}_{p}$ as $\P$.

\subsection{On $\var_{\theta} \left[ \widehat{\IIFF}_{22, k} \right]$ and its estimator}\label{sec:est.var.if22.oracle}
We first write down the explicit formula of $\var_{\theta} \left[ \widehat{\IIFF}_{22, k} \right]$:
\begin{equation}\label{eq:exact.var}
\begin{split}
& \ \var_{\theta} \left[ \widehat{\mathbb{IF}}_{22, k} \right] \\
= & \ \underbrace{\frac{1}{n (n - 1)} \left\{ \BE_{\theta} \left[ \B^{\top} \BE_{\theta} \left[ \P \P^{\top} \right] \B \right] + \BE_{\theta} \left[ \P^{\top} \BE_{\theta} \left[ \B \P^{\top} \right] \B \right] \right\}}_{\mathsf{(I)}} \\
& + \underbrace{\frac{n - 2}{n (n - 1)} \left\{ \BE_{\theta} \left[ \B \right]^{\top} \BE_{\theta} \left[ \P \P^{\top} \right] \BE_{\theta} \left[ \B \right] + \BE_{\theta} \left[ \P \right]^{\top} \BE_{\theta} \left[ \B \B^{\top} \right] \BE_{\theta} \left[ \P \right] \right\}}_{\mathsf{(II.1)}} \\
& + \underbrace{\frac{2n - 4}{n (n - 1)} \BE_{\theta} \left[ \B \right]^{\top} \BE_{\theta} \left[ \P \B^{\top} \right] \BE_{\theta} \left[ \P \right]}_{\mathsf{(II.2)}} - \underbrace{\frac{4n - 6}{n (n - 1)} \BE_{\theta} \left[ \B \right]^{\top} \BE_{\theta} \left[ \P \right] \BE_{\theta} \left[ \B \right]^{\top} \BE_{\theta} \left[ \P \right]}_{\mathsf{(III)}}.
\end{split}
\end{equation}
Thus we have, under \Cref{cond:w},
\begin{align*}
\var_{\theta} \left[ \widehat{\mathbb{IF}}_{22, k} \right] = O \left( \underbrace{\frac{k}{n^{2}}}_{\mathsf{(I)}} + \frac{1}{n} \left\{ \underbrace{\BL_{2, b, k}^{2} + \BL_{2, p, k}^{2}}_{\mathsf{(II.1)}} + \underbrace{\BL_{2, p, k} \BL_{2, b, k}}_{\mathsf{(II.2)}} \right\} \right)
\end{align*}

\cref{eq:exact.var} also facilitates the construction of the following estimator $\widehat{\var} \left[ \widehat{\mathbb{IF}}_{22, k} \right]$:
$\widehat{\var} \left[ \widehat{\mathbb{IF}}_{22, k} \right]$ is constructed by unbiasedly estimating each piece in \cref{eq:exact.var} using U-statistics separately. So
\begin{equation}\label{eq:est.var}
\begin{split}
\widehat{\var} \left[ \widehat{\mathbb{IF}}_{22, k} \right] = \widehat{\mathsf{(I)}} + \widehat{\mathsf{(II.1)}} + \widehat{\mathsf{(II.2)}} - \widehat{\mathsf{(III)}}
\end{split}
\end{equation}
where
\begin{align*}
\widehat{\mathsf{(I)}} & = \frac{1}{n^{2} (n - 1)^{2}} \sum_{1 \leq i_{1} \neq i_{2} \leq n} \B_{i_{1}}^{\top} \P_{i_{2}} \P_{i_{2}}^{\top} \B_{i_{1}} + \B_{i_{1}}^{\top} \P_{i_{2}} \B_{i_{2}}^{\top} \P_{i_{1}}, \\
\widehat{\mathsf{(II.1)}} & = \frac{1}{n^{2} (n - 1)^{2}} \sum_{1 \leq i_{1} \neq i_{2} \neq i_{3} \leq n} \B_{i_{1}}^{\top} \P_{i_{2}} \P_{i_{2}}^{\top} \B_{i_{3}} + \P_{i_{1}}^{\top} \B_{i_{2}} \B_{i_{2}}^{\top} \P_{i_{3}}, \\
\widehat{\mathsf{(II.2)}} & = \frac{2}{n^{2} (n - 1)^{2}} \sum_{1 \leq i_{1} \neq i_{2} \neq i_{3} \leq n} \B_{i_{1}}^{\top} \P_{i_{2}} \B_{i_{2}}^{\top} \P_{i_{3}}, \\
\widehat{\mathsf{(III)}} & = \frac{4 n - 6}{n^{2} (n - 1)^{2} (n - 2) (n - 3)} \sum_{1 \leq i_{1} \neq i_{2} \neq i_{3} \neq i_{4} \leq n} \B_{i_{1}}^{\top} \P_{i_{2}} \B_{i_{3}}^{\top} \P_{i_{4}}.
\end{align*}

In particular, $\widehat{\var} \left[ \widehat{\mathbb{IF}}_{22, k} \right]$ satisfies
$
\frac{\widehat{\var} \left[ \widehat{\mathbb{IF}}_{22, k} \right]}{\var_{\theta} \left[ \widehat{\mathbb{IF}}_{22, k} \right]} = 1 + o_{P_{\theta}} (1)
$
because: (1) $\BE_{\theta} \left[ \widehat{\var} \left[ \widehat{\mathbb{IF}}_{22, k} \right] \right] = \var_{\theta} \left[ \widehat{\mathbb{IF}}_{22, k} \right]$, i.e. $\widehat{\var} \left[ \widehat{\mathbb{IF}}_{22, k} \right]$ is an unbiased estimator of $\var_{\theta} \left[ \widehat{\mathbb{IF}}_{22, k} \right]$; and (2) $\var_{\theta} \left[ \frac{\widehat{\var} \left[ \widehat{\mathbb{IF}}_{22, k} \right]}{\var_{\theta} \left[ \widehat{\mathbb{IF}}_{22, k} \right]} \right] \rightarrow 0.$ As for (2), it is easy (though tedious) to show that $\var_{\theta}[\widehat{\var} [\widehat{\mathbb{IF}}_{22, k}]] \asymp \frac{1}{n} \left( \frac{k^{2}}{n^{4}} + \frac{1}{n^{2}} \{ \BL_{2, b, k}^{4} + \BL_{2, p, k}^{4} \} \right) \ll \{ \var_{\theta}[\widehat{\mathbb{IF}}_{22, k}] \}^{2} \asymp \frac{k^{2}}{n^{4}} + \frac{1}{n^{2}} \{ \BL_{2, b, k}^{4} + \BL_{2, p, k}^{4} \}$.

The variance of $\widehat{\IIFF}_{22, k} ([\widehat{\Omega}_{k}^{\tr}]^{-1})$ and its estimator can be obtained similarly with $\Omega_{k}^{-1}$ replaced by $[\widehat{\Omega}_{k}^{\tr}]^{-1}$ in the definition of $\B$ and $\P$. In \Cref{tab:var}, we show that in finite sample the estimated variances of $\widehat{\IIFF}_{22, k}$ for different $k$ are close to their Monte Carlo variances (MCvars). Moreover, for $\widehat{\IIFF}_{22, k} ([\widehat{\Omega}_{k}^{\tr}]^{-1})$, the estimated variances are quite close to their MCvars even when it is not well conditioned (blowing up) at $k = 2048, 4096$.

\begin{table}[tbp]
\caption{Simulation setup I: $\psi(\theta) = \mathbb{E}_\theta [ \mathsf{var}_\theta (A | X) ]$, regression functions estimated by nonparametric kernel regression with cross validation}
\label{tab:var}\centering
\begin{tabular}{l|cccc}
\hline
$k$ & MCvar $[ \widehat{\mathbb{IF}}_{22, k} ]$ & MCav $\widehat{\var} [ \widehat{\mathbb{IF}}_{22, k} ]$ & MCvar $[ \widehat{\mathbb{IF}}_{22, k} ([\widehat{\Omega}_{k}^{\tr}]^{-1}) ]$ & MCav $\widehat{\var} [ \widehat{\mathbb{IF}}_{22, k} ([\widehat{\Omega}_{k}^{\tr}]^{-1}) ]$ \\ 
\hline
$64$ & 0.611 & 0.558 & 0.642 & 0.600 \\
$128$ & 0.690 & 0.668 & 0.732 & 0.750 \\ 
$256$ & 2.084 & 1.677 & 2.405 & 2.104 \\ 
$512$ & 2.161 & 2.035 & 3.049 & 2.996 \\ 
$1024$ & 2.974 & 2.718 & 6.978 & 6.437 \\ 
$2048$ & 5.686 & 4.444 & 1882.454 & 1838.906 \\ 
$4096$ & 10.392 & 9.353 & $1.033 \times 10^{21}$ & $1.183 \times 10^{21}$ \\ 
\hline
\end{tabular}
\newline
All the numbers in the table should be multiplied by $10^{-4}$. For more details on the data generating mechanism, see \Cref{sec:simulations}. Note that MCvar stands for ``Monte Carlo variance''.
\end{table}

\begin{rem}\label{sec:var_quasi}
We construct ``working'' variance estimator of $\widehat{\mathbb{IF}}_{22,k}^{\mathsf{quasi}}([\widehat{\Omega}_{k}^{\mathsf{est}}]^{-1})$ following the same idea in the construction of the variance estimator of $\widehat{\IIFF}_{22, k}$, by estimating the variances of and the covariance between the two components in $\widehat{\IIFF}_{22, k}^{\quasi} ([\widehat{\Omega}_{k}^{\est}]^{-1})$ (see \cref{equasi}) separately by U-statistics, pretending that $\widehat{\Omega}_{k}^{\est}$ is independent of the estimation sample.

The finite sample performance is shown in \Cref{tab:var_quasi}, demonstrating that $\widehat{\var} [\widehat{\mathbb{IF}}_{22,k}^{\mathsf{quasi}}([\widehat{\Omega}_{k}^{\mathsf{est}}]^{-1})]$ at different $k$ are indeed quite close to $\var_{\theta} [\widehat{\mathbb{IF}}_{22,k}^{\mathsf{quasi}}([\widehat{\Omega}_{k}^{\mathsf{est}}]^{-1})]$. 

However, due to the dependence in the ``U-statistic'' kernel on all the estimation sample, it is difficult to show that $\widehat{\var} [\widehat{\mathbb{IF}}_{22,k}^{\mathsf{quasi}}([\widehat{\Omega}_{k}^{\mathsf{est}}]^{-1})]$ is close to $\var_{\theta} [\widehat{\mathbb{IF}}_{22,k}^{\mathsf{quasi}}([\widehat{\Omega}_{k}^{\mathsf{est}}]^{-1})]$.
\end{rem}

\begin{table}[tbp]
\caption{Simulation setup I: $\psi(\theta) = \mathbb{E}_\theta [ \mathsf{var}_\theta (A | X) ]$, regression functions estimated by nonparametric kernel regression with cross validation}
\label{tab:var_quasi}\centering
\begin{tabular}{l|cc}
\hline
$k$ & MCvar $[ \widehat{\mathbb{IF}}_{22, k}^{\quasi} ([\widehat{\Omega}_{k}^{\est}]^{-1}) ]$ & MCav $\widehat{\var} [ \widehat{\mathbb{IF}}_{22, k}^{\quasi} ([\widehat{\Omega}_{k}^{\est}]^{-1}) ]$ \\ 
\hline 
$64$ & 0.616 & 0.539 \\ 
$132$ & 0.690 & 0.625 \\
$256$ & 2.012 & 1.820 \\
$512$ & 2.261 & 2.202 \\ 
$1024$ & 2.982 & 2.912 \\ 
$2048$ & 4.911 & 5.028 \\ 
$4096$ & 3.938 & 3.878 \\ 
\hline
\end{tabular}
\newline
All the numbers in the table should be multiplied by $10^{-4}$. For more details on the data generating mechanism, see \Cref{sec:simulations}. Note that MCvar stands for ``Monte Carlo variance''.
\end{table}

\begin{rem}\label{rem:var_shrink}
We also construct the variance estimator of $\widehat{\mathbb{IF}}_{22, k} ([\widehat{\Omega}_{k}^{\mathsf{shrink}}]^{-1})$ in the same way as the variance estimator of $\widehat{\IIFF}_{22, k}$ proposed in \cref{eq:est.var}, just replacing $\Omega_{k}^{-1}$ by $[\widehat{\Omega}_{k}^{\mathsf{shrink}}]^{-1}$. In \Cref{tab:var_shrink}, we demonstrate that in finite sample, the variance estimators of $\widehat{\mathbb{IF}}_{22, k} ([\widehat{\Omega}_{k}^{\mathsf{shrink}}]^{-1})$ for different $k$ are close to their MCvars, when $\widehat{\mathbb{IF}}_{22, k} ([\widehat{\Omega}_{k}^{\mathsf{shrink}}]^{-1})$ does not blow up. For example, at $k = 1024$, $\widehat{\mathbb{IF}}_{22, k} ([\widehat{\Omega}_{k}^{\mathsf{shrink}}]^{-1})$ blows up and its variance estimator blows up and is negative. We use this empirical observation to determine at a given $k$ if $\widehat{\mathbb{IF}}_{22, k} ([\widehat{\Omega}_{k}^{\mathsf{shrink}}]^{-1})$ can be used as an estimator of $\Bias_{\theta, k} (\hat{\psi}_{1})$ in our simulations.
\end{rem}

\begin{table}[tbp]
\caption{Simulation setup I: $\psi(\theta) = \mathbb{E}_\theta [ \mathsf{var}_\theta (A | X) ]$, regression functions estimated by nonparametric kernel regression with cross validation}
\label{tab:var_shrink}\centering
\begin{tabular}{l|cc}
\hline
$k$ & MCvar $[ \widehat{\mathbb{IF}}_{22, k} ([\widehat{\Omega}_{k}^{\shrink}]^{-1}) ]$ & MCav $\widehat{\var} [ \widehat{\mathbb{IF}}_{22, k} ([\widehat{\Omega}_{k}^{\shrink}]^{-1}) ]$ \\ 
\hline 
$512$ & 2.467 & 2.308 \\ 
$1024$ & $3.803 \times 10^{13}$ & $-4.977 \times 10^{-14}$ \\ 
$2048$ & 6.871 & 5.638 \\ 
$4096$ & 13.980 & 10.024 \\ 
\hline
\end{tabular}
\newline
All the numbers in the table should be multiplied by $10^{-4}$. For more details on the data generating mechanism, see \Cref{sec:simulations}. Note that MCvar stands for ``Monte Carlo variance''.
\end{table}

\subsection{On $\var_{\theta} \left[ \widehat{\IIFF}_{33, k} \left( [\widehat{\Omega}_{k}^{\tr}]^{-1} \right) \right]$ and its estimator}\label{sec:est.var.if33}
As before, we first write down the explicit formula for $\var_{\theta} \left[ \widehat{\IIFF}_{33, k} \left( [\widehat{\Omega}_{k}^{\tr}]^{-1} \right) \right]$. In this section, unlike the previous two sections, we denote $[\widehat{\Omega}_{k}^{\tr}]^{- 1 / 2} \Zbar_{k} \hat{\varepsilon}_{b}$ as $\B$ and $[\widehat{\Omega}_{k}^{\tr}]^{- 1 / 2} \Zbar_{k} \hat{\varepsilon}_{p}$ as $\P$. We also denote $M = [\widehat{\Omega}_{k}^{\tr}]^{- 1 / 2} \left( \Zbar_{k} \Zbar_{k}^{\top} - \widehat{\Omega}_{k}^{\tr} \right) [\widehat{\Omega}_{k}^{\tr}]^{- 1 / 2}$. 

As in \cite{mukherjee2017semiparametric}, in this section we assume the following: (1) under \Cref{cond:w} and within in the event that $\widehat{\Omega}_{k}^{\tr}$ is invertible, (2) $k \rightarrow \infty$ as $n \rightarrow \infty$ and (3) $k = o(n / \log^{2}(n))$.

\begin{equation}\label{eq:exact.var.if33}
\begin{split}
& \ \var_{\theta} \left[ \widehat{\IIFF}_{33, k} ([\widehat{\Omega}_{k}^{\tr}]^{-1}) \right] \\
= & \ \underbrace{\frac{1}{n (n - 1) (n - 2)} \left\{ \begin{array}{c}
\BE_{\theta} \left[ \B_{1}^{\top} M_{2} \P_{3} \B_{1}^{\top} M_{2} \P_{3} \right] + \BE_{\theta} \left[ \B_{1}^{\top} M_{2} \P_{3} \B_{1}^{\top} M_{3} \P_{2} \right] + \BE_{\theta} \left[ \B_{1}^{\top} M_{2} \P_{3} \B_{2}^{\top} M_{1} \P_{3} \right] \\
+ \ \BE_{\theta} \left[ \B_{1}^{\top} M_{2} \P_{3} \B_{2}^{\top} M_{3} \P_{1} \right] + \BE_{\theta} \left[ \B_{1}^{\top} M_{2} \P_{3} \B_{3}^{\top} M_{1} \P_{2} \right] + \BE_{\theta} \left[ \B_{1}^{\top} M_{2} \P_{3} \B_{3}^{\top} M_{2} \P_{1} \right]
\end{array} \right\}}_{\mathsf{(I)}} \\
& + \underbrace{\frac{n - 3}{n (n - 1) (n - 2)} \left\{ \begin{array}{c}
\BE_{\theta} \left[ \B_{1}^{\top} M_{2} \P_{3} \B_{1}^{\top} M_{2} \P_{4} \right] + \BE_{\theta} \left[ \B_{1}^{\top} M_{2} \P_{3} \B_{2}^{\top} M_{1} \P_{4} \right] + \BE_{\theta} \left[ \B_{1}^{\top} M_{2} \P_{3} \B_{2}^{\top} M_{3} \P_{4} \right] \\
+ \ \BE_{\theta} \left[ \B_{1}^{\top} M_{2} \P_{3} \B_{3}^{\top} M_{2} \P_{4} \right] + \BE_{\theta} \left[ \B_{1}^{\top} M_{2} \P_{3} \B_{1}^{\top} M_{3} \P_{4} \right] + \BE_{\theta} \left[ \B_{1}^{\top} M_{2} \P_{3} \B_{3}^{\top} M_{1} \P_{4} \right] \\
+ \ \BE_{\theta} \left[ \B_{1}^{\top} M_{2} \P_{3} \B_{1}^{\top} M_{4} \P_{2} \right] + \BE_{\theta} \left[ \B_{1}^{\top} M_{2} \P_{3} \B_{2}^{\top} M_{4} \P_{1} \right] + \BE_{\theta} \left[ \B_{1}^{\top} M_{2} \P_{3} \B_{2}^{\top} M_{4} \P_{3} \right] \\
+ \ \BE_{\theta} \left[ \B_{1}^{\top} M_{2} \P_{3} \B_{3}^{\top} M_{4} \P_{2} \right] + \BE_{\theta} \left[ \B_{1}^{\top} M_{2} \P_{3} \B_{1}^{\top} M_{4} \P_{3} \right] + \BE_{\theta} \left[ \B_{1}^{\top} M_{2} \P_{3} \B_{3}^{\top} M_{4} \P_{1} \right] \\
+ \ \BE_{\theta} \left[ \B_{1}^{\top} M_{2} \P_{3} \B_{4}^{\top} M_{1} \P_{2} \right] + \BE_{\theta} \left[ \B_{1}^{\top} M_{2} \P_{3} \B_{4}^{\top} M_{2} \P_{1} \right] + \BE_{\theta} \left[ \B_{1}^{\top} M_{2} \P_{3} \B_{4}^{\top} M_{2} \P_{3} \right] \\
+ \ \BE_{\theta} \left[ \B_{1}^{\top} M_{2} \P_{3} \B_{4}^{\top} M_{3} \P_{2} \right] + \BE_{\theta} \left[ \B_{1}^{\top} M_{2} \P_{3} \B_{4}^{\top} M_{1} \P_{3} \right] + \BE_{\theta} \left[ \B_{1}^{\top} M_{2} \P_{3} \B_{4}^{\top} M_{3} \P_{1} \right]
\end{array} \right\}}_{\mathsf{(II)}} \\
& + \underbrace{\frac{(n - 3) (n - 4)}{n (n - 1) (n - 2)} \left\{ \begin{array}{c}
\BE_{\theta} \left[ \B_{1}^{\top} M_{2} \P_{3} \B_{1}^{\top} M_{4} \P_{5} \right] + \BE_{\theta} \left[ \B_{1}^{\top} M_{2} \P_{3} \B_{2}^{\top} M_{4} \P_{5} \right] + \BE_{\theta} \left[ \B_{1}^{\top} M_{2} \P_{3} \B_{3}^{\top} M_{4} \P_{5} \right] \\
+ \ \BE_{\theta} \left[ \B_{1}^{\top} M_{2} \P_{3} \B_{4}^{\top} M_{2} \P_{5} \right] + \BE_{\theta} \left[ \B_{1}^{\top} M_{2} \P_{3} \B_{4}^{\top} M_{1} \P_{5} \right] + \BE_{\theta} \left[ \B_{1}^{\top} M_{2} \P_{3} \B_{4}^{\top} M_{3} \P_{5} \right] \\
+ \ \BE_{\theta} \left[ \B_{1}^{\top} M_{2} \P_{3} \B_{4}^{\top} M_{5} \P_{3} \right] + \BE_{\theta} \left[ \B_{1}^{\top} M_{2} \P_{3} \B_{4}^{\top} M_{5} \P_{2} \right] + \BE_{\theta} \left[ \B_{1}^{\top} M_{2} \P_{3} \B_{4}^{\top} M_{5} \P_{1} \right]
\end{array} \right\}}_{\mathsf{(III)}} \\
& - \underbrace{\frac{3 (3n^{2} - 15n + 20)}{n (n - 1) (n - 2)} \left\{ \BE_{\theta} [ \B_{1}]^{\top} \BE_{\theta} [M_{2}] \BE_{\theta} [\P_{3}] \right\}^{2}}_{\mathsf{(IV)}}.
\end{split}
\end{equation}

First we have $\mathsf{(IV)} \lesssim \dfrac{1}{n} \BL_{2, b, k}^{2} \BL_{2, p, k}^{2} \dfrac{k \log(k)}{n}$ following \citet[Theorem 4]{mukherjee2017semiparametric}.

Under \Cref{cond:w}, $\left\Vert \Pi \left[ b - \hat{b} \vert \Zbar_{k} \right] \right\Vert_{\infty}$ and $\left\Vert \Pi \left[ p - \hat{p} \vert \Zbar_{k} \right] \right\Vert_{\infty}$ are bounded. Thus we have
\begin{equation}\label{eq:var_if33_order}
\begin{split}
\var_{\theta} \left[ \widehat{\IIFF}_{33, k} ([\widehat{\Omega}_{k}^{\tr}]^{-1}) \right] = O \left( \begin{array}{c}
\underbrace{\frac{k^{2}}{n^{3}}}_{\mathsf{(I)}} + \underbrace{\frac{k}{n^{2}} \left\{ \BL_{2, b, k}^{2} + \BL_{2, p, k}^{2} \right\}}_{\mathsf{(II)}} + \underbrace{\dfrac{1}{n} \BL_{2, b, k}^{2} \BL_{2, p, k}^{2} \dfrac{k \log(k)}{n}}_{\mathsf{(IV)}} \\
+ \ \underbrace{\frac{1}{n} \left\{ \begin{array}{c}
\underbrace{\BE_{\theta} \left[ \Pi \left[ b - \hat{b} \vert \zbar_{k} (X) \right] (b(X) - \hat{b}(X)) \Pi \left[ p - \hat{p} \vert \zbar_{k} (X) \right]^{2} \right]}_{\mathsf{(III.2)} + \mathsf{(III.5)}} \\
+ \ \underbrace{\BE_{\theta} \left[ \Pi \left[ p - \hat{p} \vert \zbar_{k} (X) \right] (p(X) - \hat{p}(X)) \Pi \left[ b - \hat{b} \vert \zbar_{k} (X) \right]^{2} \right]}_{\mathsf{(III.6)} + \mathsf{(III.8)}} \\
+ \ \underbrace{\BE_{\theta} \left[ \Pi \left[ b - \hat{b} \vert \zbar_{k} (X) \right]^{2} \Pi \left[ p - \hat{p} \vert \zbar_{k} (X) \right]^{2} \right]}_{\mathsf{(III.4)}}
\end{array} \right\}}_{\mathsf{(III)}}
\end{array} \right).
\end{split}
\end{equation}
Under \Cref{cond:w} in which we assume $\hat{p} - p$ and $\hat{b} - b$ to be bounded, by \citet[Examples 3.8-3.10]{belloni2015some}, $\Pi \left[ b - \hat{b} \vert \zbar_{k} (X) \right]$ and $\Pi \left[ p - \hat{p} \vert \zbar_{k} (X) \right]$ are also bounded when $\zbar_{k}(\cdot)$ are Cohen-Daubechies-Vial wavelet series, local polynomial partition series or spline series. 

For $\var_{\theta} \left[ \widehat{\IIFF}_{22 \rightarrow 33, k} ([\widehat{\Omega}_{k}^{\tr}]^{-1}) \right]$, we have:
\begin{align}
\var_{\theta} \left[ \widehat{\IIFF}_{22 \rightarrow 33, k} ([\widehat{\Omega}_{k}^{\tr}]^{-1}) \right] \leq 2 \var_{\theta} \left[ \widehat{\IIFF}_{22, k} ([\widehat{\Omega}_{k}^{\tr}]^{-1}) \right] + 2 \var_{\theta} \left[ \widehat{\IIFF}_{33, k} ([\widehat{\Omega}_{k}^{\tr}]^{-1}) \right]. \label{order:var_if2233}
\end{align}

One may wonder if we could obtain $\var_{\theta} \left[ \widehat{\IIFF}_{22 \rightarrow 33, k} ([\widehat{\Omega}_{k}^{\tr}]^{-1}) \right] \asymp \var_{\theta} \left[ \widehat{\IIFF}_{22, k} \right]$. This is indeed the case. We consider $\psi(\theta) = \BE_{\theta}[\var_{\theta}(A | X)]$ and $\psi(\theta) = \BE_{\theta}[\cov_{\theta}(A, Y | X)]$ separately: 
\begin{itemize}
\item For $\psi(\theta) = \BE_{\theta}[\var_{\theta}(A | X)]$, terms in $\mathsf{(III)}$ of \cref{eq:var_if33_order} can be further bounded by:
\begin{align*}
& \ \frac{1}{n} \left\vert \BE_{\theta} \left[ \Pi \left[ p - \hat{p} \vert \zbar_{k} (X) \right] (p(X) - \hat{p}(X)) \Pi \left[ p - \hat{p} \vert \zbar_{k} (X) \right]^{2} \right] \right\vert, \\
\leq & \ \frac{1}{n} \Vert \Pi \left[ p - \hat{p} \vert \zbar_{k} (X) \right] (p(X) - \hat{p}(X)) \Vert_{\infty} \BE \left[ \Pi \left[ p - \hat{p} \vert \zbar_{k} (X) \right]^{2} \right] \lesssim \frac{1}{n} \Bias_{\theta, k} (\hat{\psi}_{1}), \\
& \ \frac{1}{n} \BE_{\theta} \left[ \Pi \left[ p - \hat{p} \vert \zbar_{k} (X) \right]^{2} \Pi \left[ p - \hat{p} \vert \zbar_{k} (X) \right]^{2} \right], \\
\leq & \frac{1}{n} \Vert \Pi \left[ p - \hat{p} \vert \zbar_{k} (X) \right]^{2} \Vert_{\infty} \BE \left[ \Pi \left[ p - \hat{p} \vert \zbar_{k} (X) \right]^{2} \right] \lesssim \frac{1}{n} \Bias_{\theta, k} (\hat{\psi}_{1}).
\end{align*}
We thus have
$$
\var_{\theta} \left[ \widehat{\IIFF}_{33, k} ([\widehat{\Omega}_{k}^{\tr}]^{-1}) \right] \lesssim \var_{\theta} \left[ \widehat{\IIFF}_{22, k} ([\widehat{\Omega}_{k}^{\tr}]^{-1}) \right]
$$
and hence
$$
\var_{\theta} \left[ \widehat{\IIFF}_{22 \rightarrow 33, k} ([\widehat{\Omega}_{k}^{\tr}]^{-1}) \right] \asymp \var_{\theta} \left[ \widehat{\IIFF}_{22, k} \right].
$$
\item For $\psi(\theta) = \BE_{\theta}[\cov_{\theta}(A, Y | X)]$, terms in $\mathsf{(III)}$ of \cref{eq:var_if33_order} can be further bounded by:
\begin{align*}
& \ \frac{1}{n} \left\vert \BE_{\theta} \left[ \Pi [ b - \hat{b} \vert \zbar_{k} (X) ] (b(X) - \hat{b}(X)) \Pi [ p - \hat{p} \vert \zbar_{k} (X) \right]^{2} ] \right\vert, \\
\leq & \ \frac{1}{n} \Vert \Pi [ b - \hat{b} \vert \zbar_{k} (X) ] (b(X) - \hat{b}(X)) \Vert_{\infty} \BE \left[ \Pi \left[ p - \hat{p} \vert \zbar_{k} (X) \right]^{2} \right] \lesssim \frac{1}{n} \BL_{2, p, k}^{2}, \\
& \ \frac{1}{n} \left\vert \BE_{\theta} \left[ \Pi [ p - \hat{p} \vert \zbar_{k} (X) ] (p(X) - \hat{p}(X)) \Pi [ b - \hat{b} \vert \zbar_{k} (X) \right]^{2} ] \right\vert, \\
\leq & \ \frac{1}{n} \Vert \Pi [ p - \hat{p} \vert \zbar_{k} (X) ] (p(X) - \hat{p}(X)) \Vert_{\infty} \BE \left[ \Pi \left[ b - \hat{b} \vert \zbar_{k} (X) \right]^{2} \right] \lesssim \frac{1}{n} \BL_{2, b, k}^{2}, \\
& \ \frac{1}{n} \BE_{\theta} \left[ \Pi \left[ b - \hat{b} \vert \zbar_{k} (X) \right]^{2} \Pi \left[ p - \hat{p} \vert \zbar_{k} (X) \right]^{2} \right], \\
\leq & \frac{1}{n} \Vert \Pi [ b - \hat{b} \vert \zbar_{k} (X) ]^{2} \Vert_{\infty} \BE \left[ \Pi \left[ p - \hat{p} \vert \zbar_{k} (X) \right]^{2} \right] \lesssim \frac{1}{n} \BL_{2, p, k}^{2}.
\end{align*}
We want to remark that the above upper bound is not optimal but it is good enough for our purpose to show that
$$
\var_{\theta} \left[ \widehat{\IIFF}_{33, k} ([\widehat{\Omega}_{k}^{\tr}]^{-1}) \right] \lesssim \var_{\theta} \left[ \widehat{\IIFF}_{22, k} ([\widehat{\Omega}_{k}^{\tr}]^{-1}) \right]
$$
and hence
$$
\var_{\theta} \left[ \widehat{\IIFF}_{22 \rightarrow 33, k} ([\widehat{\Omega}_{k}^{\tr}]^{-1}) \right] \asymp \var_{\theta} \left[ \widehat{\IIFF}_{22, k} \right].
$$
\end{itemize}

We summarize the above calculation in the following proposition:
\begin{proposition}\label{prop:var_if2233}
Under \Cref{cond:w}, and in the event that $\widehat{\Omega}_{k}^{\tr}$ is invertible, if $k \rightarrow \infty$ as $n \rightarrow \infty$ and $k = o(n / \log^{2}(n))$, $\var_{\theta} \left[ \widehat{\IIFF}_{22 \rightarrow 33, k} ([\widehat{\Omega}_{k}^{\tr}]^{-1}) \right] \asymp \var_{\theta} \left[ \widehat{\IIFF}_{22, k} \right]$ for both $\psi(\theta) = \BE_{\theta}[\var_{\theta}(A | X)]$ and $\psi(\theta) = \BE_{\theta}[\cov_{\theta}(A, Y | X)]$.
\end{proposition}

Similarly, we can construct $\widehat{\var} \left[ \widehat{\IIFF}_{22 \rightarrow 33, k} ([\widehat{\Omega}_{k}^{\tr}]^{-1}) \right]$, the variance estimator of $\widehat{\IIFF}_{22 \rightarrow 33, k} ([\widehat{\Omega}_{k}^{\tr}]^{-1})$ by
\begin{align}\label{eq:est.var_if2233}
\widehat{\var} \left[ \widehat{\IIFF}_{22 \rightarrow 33, k} ([\widehat{\Omega}_{k}^{\tr}]^{-1}) \right] = \widehat{\var} \left[ \widehat{\IIFF}_{22, k} ([\widehat{\Omega}_{k}^{\tr}]^{-1}) \right] + \widehat{\var} \left[ \widehat{\IIFF}_{33, k} ([\widehat{\Omega}_{k}^{\tr}]^{-1}) \right] + 2 \widehat{\cov} \left[ \widehat{\IIFF}_{22, k} ([\widehat{\Omega}_{k}^{\tr}]^{-1}), \widehat{\IIFF}_{33, k} ([\widehat{\Omega}_{k}^{\tr}]^{-1}) \right].
\end{align}
$\widehat{\cov} \left[ \widehat{\IIFF}_{22, k} ([\widehat{\Omega}_{k}^{\tr}]^{-1}), \widehat{\IIFF}_{33, k} ([\widehat{\Omega}_{k}^{\tr}]^{-1}) \right]$ is again an unbiased estimator for $\cov_{\theta} \left[ \widehat{\IIFF}_{22, k} ([\widehat{\Omega}_{k}^{\tr}]^{-1}), \widehat{\IIFF}_{33, k} ([\widehat{\Omega}_{k}^{\tr}]^{-1}) \right]$ based on U-statistics using the estimation strategy in \Cref{sec:est.var.if22.oracle}.

We did not implement $\widehat{\var} \left[ \widehat{\IIFF}_{33, k} ([\widehat{\Omega}_{k}^{\tr}]^{-1}) \right]$ in our simulations because it involves a 6th-order U-statistics. We did not use $\widehat{\IIFF}_{22 \rightarrow 33, k} ([\widehat{\Omega}_{k}^{\tr}]^{-1})$ to construct the data-adaptive estimator $\widehat{\IIFF}_{22, k} (\widehat{\Omega}_{k}^{-1})$ reported in \Cref{tab:intro} and in the simulations in later \Cref{sec:simulations} because in \Cref{sec:cov} we have shown that $\widehat{\IIFF}_{22, k}^{\quasi} ([\widehat{\Omega}_{k}^{\est}]^{-1})$ has much more stable numerical performance than $\widehat{\IIFF}_{22 \rightarrow 33, k} ([\widehat{\Omega}_{k}^{\tr}]^{-1})$ in simulations.

\section{On the data-adaptive estimator $\widehat{\mathbb{IF}}_{22, k} (\widehat{\Omega}_{k}^{-1})$}\label{sec:adaptive} 
We denote the data-adaptive estimator as $\widehat{\mathbb{IF}}_{22, k} (\widehat{\Omega}_{k}^{-1})$. We consider $\psi(\theta) = \mathbb{E}_\theta [ \mathsf{var}_\theta (A | X) ]$ as we know $\Bias_{\theta, k} ( \hat{\psi}_{1} )$ increases with $k$. Simulation results show that:
\begin{enumerate}
\item The estimation bias of $\widehat{\mathbb{IF}}_{22, k}^\mathsf{quasi} ([\widehat{\Omega}_{k}^\mathsf{est}]^{-1} )$ increases with $k$, reflected by that the MCav of $\widehat{\mathbb{IF}}_{22, k}^\mathsf{quasi} ( [\widehat{\Omega}_{k}^\mathsf{est}]^{-1} )$ can decrease when $k$ is near $n$; see \Cref{sec:quasi}

\item $\widehat{\mathbb{IF}}_{22, k} ([\widehat{\Omega}_{k}^\mathsf{shrink}]^{-1})$ blows up when its variance estimator blows up; see \Cref{rem:var_shrink}.
\end{enumerate}
Though in \Cref{sec:cov}, we describe the finite sample performance of $\widehat{\mathbb{IF}}_{22, k}^\mathsf{quasi} ([\widehat{\Omega}_{k}^\mathsf{est}]^{-1} )$ and $\widehat{\mathbb{IF}}_{22, k} ([\widehat{\Omega}_{k}^\mathsf{shrink}]^{-1})$ with one simulation, we observe the above two phenomenon in all our simulations.

Thus we design a data-adaptive algorithm to decide at each $k$, whether $\widehat{\mathbb{IF}}_{22, k}^\mathsf{quasi} ( [\widehat{\Omega}_{k}^\mathsf{est}]^{-1} )$ can be used as $\widehat{\mathbb{IF}}_{22, k} (\widehat{\Omega}_{k}^{-1})$ and if not whether $\widehat{\mathbb{IF}}_{22, k} ([\widehat{\Omega}_{k}^\mathsf{shrink}]^{-1})$ can be used. 
Ideally, a part of the estimation sample or a totally independent sample should be reserved for the implementation of the data-adaptive algorithm.

Since we lack theory on the estimation bias of $\widehat{\mathbb{IF}}_{22, k}^\mathsf{quasi} ( [\widehat{\Omega}_{k}^\mathsf{est}]^{-1} )$ and $\widehat{\mathbb{IF}}_{22, k} ( [\widehat{\Omega}_{k}^\mathsf{shrink}]^{-1} )$, the algorithm is developed based on the empirical observations from our simulation studies. This strategy may be modified after the statistical properties of $\widehat{\mathbb{IF}}_{22, k}^\mathsf{quasi} ( [\widehat{\Omega}_{k}^\mathsf{est}]^{-1} )$ and $\widehat{\mathbb{IF}}_{22, k} ( [\widehat{\Omega}_{k}^\mathsf{shrink}]^{-1} )$ are established.

In contrast to $\Bias_{\theta, k} ( \hat{\psi}_{1} )$ monotonically increasing with $k$, the MCav of $\widehat{\mathbb{IF}}_{22, k}^\mathsf{quasi} ([\widehat{\Omega}_{k}^\mathsf{est}]^{-1})$ starts to decrease when $k$ is close to $n$ as the estimation bias of $\widehat{\mathbb{IF}}_{22, k}^\mathsf{quasi}([\widehat{\Omega}_{k}^\mathsf{est}]^{-1})$ increases with $k$. Therefore, the first step of the data adaptive algorithm is to identify the point $k^\text{quasi}$ at which $\widehat{\mathbb{IF}}_{22, k}^\mathsf{quasi} ([\widehat{\Omega}_{k}^\mathsf{est}]^{-1})$ stops increasing. For $k \le k^\mathsf{quasi}$, we choose $\widehat{\mathbb{IF}}_{22, k}(\widehat{\Omega}_{k}^{-1}) = \widehat{\mathbb{IF}}_{22, k}^\mathsf{quasi} ([\widehat{\Omega}_{k}^\mathsf{est}]^{-1})$; for $k > k^\mathsf{quasi}$, we decide whether $\widehat{\mathbb{IF}}_{22, k} ([\widehat{\Omega}_{k}^\mathsf{shrink}]^{-1})$ can be used as $\widehat{\mathbb{IF}}_{22, k}(\widehat{\Omega}_{k}^{-1})$.

We now describe the algorithm step by step. Suppose that we are given following ordered set $\{ k_1 < k_2 < \cdots < k_J \}$ of all candidate $k$'s. For $\tilde{j} = 1, \cdots, J$, when $\tilde{j} = j$, for some user-specified parameter $c_{\tilde{j}} > 0$\footnote{Here one could choose $c_j$ as 1 as a preliminary default setting.}:

\begin{itemize}
\item If $\widehat{\mathbb{IF}}_{22, k_{\tilde{j} + 1}}^\mathsf{quasi} ([\widehat{\Omega}_{k_{\tilde{j} + 1}}^\mathsf{est}]^{-1} ) < \widehat{\mathbb{IF}}_{22, k_{\tilde{j}}}^\mathsf{quasi} ( [\widehat{\Omega}_{k_{\tilde{j}}}^\mathsf{est}]^{-1} ) - c_{\tilde{j}} \widehat{\mathsf{var}} [ \widehat{\mathbb{IF}}_{22, k_{\tilde{j}}}^\mathsf{quasi} ( [\widehat{\Omega}_{k_{\tilde{j}}}^\mathsf{est}]^{-1} ) ]^{1 / 2}$, the iteration terminates and outputs $j^\mathsf{quasi} = \tilde{j}$ (and $k^\mathsf{quasi} \equiv k_{j^\mathsf{quasi}} = k_{\tilde{j}}$). For $k \le k^\mathsf{quasi}$, the algorithm outputs $\widehat{\mathbb{IF}}_{22, k}(\widehat{\Omega}_{k}^{-1}) = \widehat{\mathbb{IF}}_{22, k}^\mathsf{quasi} ([\widehat{\Omega}_{k}^\mathsf{est}]^{-1})$.

\item Otherwise, $\tilde{j} = j +1$ and repeat the above procedure.

\begin{itemize}
\item If $\tilde{j} = J + 1$, the entire data-adaptive algorithm terminates. For all $k \le K_J$, the algorithm outputs $\widehat{\mathbb{IF}}_{22, k}(\widehat{\Omega}_{k}^{-1}) = \widehat{\mathbb{IF}}_{22, k}^\mathsf{quasi} ( [\widehat{\Omega}_{k}^\mathsf{est}]^{-1} )$. 

\item Otherwise, we need to decide the lowest $k$ such that $\widehat{\IIFF}_{22, k} ( [\widehat{\Omega}_{k}^{\shrink}]^{-1} )$ can be used as $\widehat{\IIFF}_{22, k}(\widehat{\Omega}_{k}^{-1})$ for $\tilde{j'} > j^{\quasi}$. For $\tilde{j'} = j^{\quasi} + 1, j^{\quasi} + 2, \dots, J$, when $\tilde{j'} = j'$, for some user-specified parameter $v_{\tilde{j'}} > 0$ (see \Cref{rem:cutoff}):

\begin{itemize}
\item If $\frac{\widehat{\var} [ \widehat{\IIFF}_{22, k_{\tilde{j'}}} ( [\widehat{\Omega}_{k_{\tilde{j'}}}^{\shrink}]^{-1} ) ]}{\widehat{\var} [ \widehat{\IIFF}_{22, k^{\quasi}}^{\quasi} ( [\widehat{\Omega}_{k^{\quasi}}^{\est}]^{-1} ) ]} \le v_{j'}$, the algorithm outputs $j_{\shrink} = \tilde{j'}$ (and $k_{\shrink} \equiv k_{j_{\shrink}} = k_{\tilde{j'}}$). Then we need to decide the largest $k$ such that $\widehat{\IIFF}_{22, k} ( [\widehat{\Omega}_{k}^{\shrink}]^{-1} )$ can be used as $\widehat{\IIFF}_{22, k}(\widehat{\Omega}_{k}^{-1})$ for $\tilde{j^{\prime \prime}} \ge j_{\shrink}$. For $\tilde{j^{\prime \prime}} = j_{\shrink}, j_{\shrink} + 1, \cdots, J$, when $\tilde{j^{\prime \prime}} = j^{\prime \prime}$, for some user-specified parameter $w_{\tilde{j^{\prime \prime }}} > 0$ (see \Cref{rem:cutoff}):

\begin{itemize}
\item If $\frac{\widehat{\var} [ \widehat{\IIFF}_{22, k_{\tilde{j^{\prime \prime}} + 1}} ( [\widehat{\Omega}_{k_{\tilde{j^{\prime \prime}} + 1}}^{\shrink}]^{-1} ) ]}{\widehat{\var} [ \widehat{\IIFF}_{22, k_{\tilde{j^{\prime \prime}}}} ( [\widehat{\Omega}_{k_{\tilde{j^{\prime \prime}}}}^{\shrink}]^{-1} ) ]} > w_{\tilde{j^{\prime \prime}}}$, the entire data-adaptive algorithm terminates and outputs $j^\ast = j^{\shrink} = \tilde{j^{\prime \prime}}$ (and $k^\ast \equiv k_{j^\ast} = k^{\shrink} \equiv k_{j^{\shrink}} = k_{\tilde{j^{\prime \prime}}}$). For $k_{\shrink} \le k \le k^{\shrink}$, the algorithm outputs $\widehat{\IIFF}_{22, k}(\widehat{\Omega}_{k}^{-1}) = \widehat{\IIFF}_{22, k} ( [\widehat{\Omega}_{k}^{\shrink}]^{-1} )$. 

\item Otherwise, $\tilde{j^{\prime \prime}} = j^{\prime \prime} + 1$ and repeat the above procedure. The entire algorithm terminates when $\tilde{j^{\prime \prime}} = J + 1$.
\end{itemize}

\item Otherwise, $\tilde{j'} = j'+ 1$ and repeat the above procedure.

\begin{itemize}
\item If $\tilde{j'} = J + 1$, the entire data-adaptive algorithm terminates. For any $k > k^{\quasi}$, the algorithm outputs $\widehat{\IIFF}_{22, k}(\widehat{\Omega}_{k}^{-1}) = \text{NA}$. 
\end{itemize}
\end{itemize}
\end{itemize}
\end{itemize}

In the end, the algorithm outputs $k^{\quasi}$, the largest $k$ such that $\widehat{\IIFF}_{22, k} (\widehat{\Omega}_{k}^{-1}) = \widehat{\IIFF}_{22, k}^{\quasi} ( [\widehat{\Omega}_{k}^{\est}]^{-1} )$, $k_{\shrink}$ and $k^{\shrink}$, the smallest and the largest $k$ such that $\widehat{\IIFF}_{22, k} (\widehat{\Omega}_{k}^{-1}) = \widehat{\IIFF}_{22, k} ( [\widehat{\Omega}_{k}^{\shrink}]^{-1} )$. For $k \le k^{\quasi}$, the algorithm assigns $\widehat{\IIFF}_{22, k} (\widehat{\Omega}_{k}^{-1}) = \widehat{\IIFF}_{22, k}^{\quasi} ( [\widehat{\Omega}_{k}^{\est}]^{-1} )$; for $k_{\shrink} \le k \le k^{\shrink}$, the algorithm assigns $\widehat{\IIFF}_{22, k}(\widehat{\Omega}_{k}^{-1}) = \widehat{\IIFF}_{22, k} ([\widehat{\Omega}_{k}^{\shrink}]^{-1})$; for $k^{\quasi} < k < k_{\shrink}$ or $k > k_{\shrink}$, the algorithm assigns $\widehat{\IIFF}_{22, k} (\widehat{\Omega}_{k}^{-1}) = \text{NA}$. 

\begin{rem}\label{rem:cutoff} 
To decide $k_\mathsf{shrink}$ and $k^\mathsf{shrink}$, we need to specify the cutoff $v_j > 0$ and $w_j > 0$. Since the variance of $\widehat{\mathbb{IF}}_{22, k} (\widehat{\Omega}_{k}^{-1} )$ is of order $k / n^2$, one would expect the variance of $\widehat{\mathbb{IF}}_{22, k} (\widehat{\Omega}_{k}^{-1} )$ to grow linearly with $k$. When choosing $v_j$ to decide $k_\mathsf{shrink}$, we compare if $\frac{\widehat{\mathsf{var}} [ \widehat{\mathbb{IF}}_{22, k_j} ( [\widehat{\Omega}_{k_j}^\mathsf{shrink}]^{-1}) ]}{\widehat{\mathsf{var}} [ \widehat{\mathbb{IF}}_{22, k^\mathsf{quasi}}^\mathsf{quasi} ( [\widehat{\Omega}_{k^\mathsf{quasi}}^\mathsf{est}]^{-1} ) ]} \le v_j$. Thus a reasonable choice is to set $v_j$ proportional to $C (k_j / k^\mathsf{quasi})$ for some $C > 0$. When choosing $w_j$ to decide $k^\mathsf{shrink}$, we compare if $\frac{\widehat{\mathsf{var}} [ \widehat{\mathbb{IF}}_{22, k_{j + 1}} ( [\widehat{\Omega}_{k_{j + 1}}^\mathsf{shrink}]^{-1} ) ]}{\widehat{\mathsf{var}} [ \widehat{\mathbb{IF}}_{22, k_j} ( [\widehat{\Omega}_{k_j}^\mathsf{shrink}]^{-1} ) ]} \le w_j$, again a reasonable choice is to choose $w_j$ to be $C (k_{j + 1} / k_j)$ for some $C > 0$. In terms of the constant $C > 0$, as a heuristic, one can plot the ratios $\frac{\widehat{\mathsf{var}} [ \widehat{\mathbb{IF}}_{22, k_{j + 1}}^\mathsf{quasi} ( [\widehat{\Omega}_{k_{j + 1}}^\mathsf{est}]^{-1} ) ] \cdot k_j}{\widehat{\mathsf{var}} [ \widehat{\mathbb{IF}}_{22, k_j}^\mathsf{quasi} ( [\widehat{\Omega}_{k_j}^\mathsf{est}]^{-1} ) ] \cdot k_{j + 1}}$ for all $k_j < k^\mathsf{quasi}$ against $k_j$ as information on reasonable range of the constant $C$.
\end{rem}

For $\BE_\theta [ \cov_\theta (A, Y | X) ]$, however, $\Bias_{\theta, k} (\hat{\psi}_{1})$ is not guaranteed to increase with $k$ even under \Cref{cond:b}. To circumvent such non-monotonicity, we first find $k^{\quasi; b}$ for $\BE_\theta [ \var_\theta (A | X) ]$ and $k^{\quasi; p}$ for $\BE_\theta [ \var_\theta (A | X) ]$ respectively using the strategy described above for $\psi(\theta) = \mathbb{E}_\theta [ \mathsf{var}_\theta (A | X) ]$. Then we choose $k^{\quasi} = \mathsf{min} \{ k^{\quasi; b}, k^{\quasi; p} \}$. For any $k \le k^\text{quasi}$, we choose $\widehat{\IIFF}_{22, k} (\widehat{\Omega}_{k}^{-1}) = \widehat{\IIFF}_{22, k}^{\quasi} ( [\widehat{\Omega}_{k}^{\est}]^{-1} )$.

Then for $k > k^{\quasi}$, we use the same variance comparison strategy to determine $k_{\shrink}$ and $k^{\shrink}$.

Finally for $\psi(\theta) = \BE_{\theta}[\var_{\theta}(A | X)]$ we define the following data-adaptive one-sided test and upper confidence bound for $\mathsf{H}_{0, k} (\delta)$ when $\Omega_{k}^{-1}$ is unknown: 
\begin{align}
\widehat\chi_{k}^{(1)} (\widehat{\Omega}_{k}^{-1}; \zeta_{k}, \delta) = \mathbbm{1} \left\{ \frac{\widehat{\IIFF}_{22, k} (\widehat{\Omega}_{k}^{-1})}{\widehat{\mathsf{s.e.}} (\hat{\psi}_{1})} - \zeta_{k} \frac{\widehat{\mathsf{s.e.}} (\widehat{\IIFF}_{22, k} (\widehat{\Omega}_{k}^{-1}))}{\widehat{\mathsf{s.e.}} (\hat{\psi}_{1})} > \delta \right\}, \label{oneside-adapt} \\
\mathsf{UCB}^{(1)} (\widehat{\Omega}_{k}^{-1}; \alpha, \alpha^{\dag}) \coloneqq \TC_\alpha \left( \left[ \frac{\widehat{\IIFF}_{22, k} (\widehat{\Omega}_{k}^{-1}) - z_{\alpha^{\dag}} \widehat{\mathsf{s.e.}} [ \widehat{\IIFF}_{22, k} (\widehat{\Omega}_{k}^{-1}) ]}{\widehat{\mathsf{s.e.}} [\hat{\psi}_{1}]} \right] \right). \label{ucbone-adapt}.
\end{align}
For $\psi(\theta) = \BE_{\theta}[\cov_{\theta}(A, Y | X)]$ we define the following data-adaptive two-sided test $\mathsf{H}_{0, k} (\delta)$ when $\Omega_{k}^{-1}$ is unknown:
\begin{align}
\widehat\chi_{k}^{(2)} (\widehat{\Omega}_{k}^{-1}; \zeta_{k}, \delta) = \mathbbm{1} \left\{ \frac{\vert \widehat{\IIFF}_{22, k} (\widehat{\Omega}_{k}^{-1}) \vert}{\widehat{\mathsf{s.e.}} (\hat{\psi}_{1})} - \zeta_{k} \frac{\widehat{\mathsf{s.e.}} (\widehat{\IIFF}_{22, k} (\widehat{\Omega}_{k}^{-1}))}{\widehat{\mathsf{s.e.}} (\hat{\psi}_{1})} > \delta \right\}. \label{twoside-adapt}
\end{align}

\section{An alternative oracle test targeting $\CSBias_{\theta, k} (\hat{\psi}_{1})$ for $\psi (\theta) = \BE_{\theta}[\cov_{\theta} (A, Y | X)]$}\label{sec:csbias}
In \Cref{sec:test_cov}, we have seen that it is possible to empirically falsify the CS null hypothesis $\H_{0, CS} (\delta)$ by testing $\H_{0, k} (\delta)$ using the two-sided test $\widehat\chi_{k}^{(2)} (\zeta_{k}, \delta)$. However, it is entirely possible that $\H_{0, k} (\delta)$ is true whereas $\H_{0, CS} (\delta)$ is false because we only have $\Bias_{\theta, k} (\hat{\psi}_{1}) \leq \CSBias_{\theta} (\hat{\psi}_{1})$. Then can we still reject the CS null hypothesis $\H_{0, CS} (\delta)$ by finding a more direct empirical test? We now show that the answer is yes by constructing a higher-order U-statistic test for $\CSBias_\theta^{\langle 2 \rangle} ( \hat{\psi}_{1} ) \coloneqq \{ \CSBias_\theta ( \hat{\psi}_{1} ) \}^2 \equiv \BL_{2, b}^{2} \BL_{2, p}^{2}$, where for notational convenience, $\BL_{2, b} \coloneqq \{ \BE_{\theta} [(b(X) - \hat{b}(X))^{2}] \}^{1 / 2}$ and $\BL_{2, p} \coloneqq \{ \BE_{\theta} [(p(X) - \hat{p}(X))^{2}] \}^{1 / 2}$. Specifically we consider the operationalized pair 
\begin{eqnarray*}
\NH_{0, CS}^{\langle 2 \rangle} &:& \CSBias_\theta^{\langle 2 \rangle} (\hat{\psi}_{1}) \equiv \left\{ \CSBias_{\theta} (\hat{\psi}_{1}) \right\}^{1/2} = o (n^{-1}), \\
\H_{0, CS}^{\langle 2 \rangle} (\delta) &:& \frac{\CSBias_\theta^{\langle 2 \rangle} (\hat{\psi}_{1})}{\mathsf{var}_\theta [ \hat{\psi}_{1} ]} < \delta^2.
\end{eqnarray*}

Next define the surrogate operationalized pair
\begin{equation*}
\CSBias_{\theta, k}^{\langle 2 \rangle} (\hat{\psi}_{1}) \coloneqq \CSBias_{\theta, k} (\hat{\psi}_{1})
\end{equation*}
and the operationalized pair 
\begin{eqnarray*}
\NH_{0, CS, k}^{\langle 2 \rangle} &:& \CSBias_{\theta, k}^{\langle 2 \rangle} (\hat{\psi}_{1}) = o (n^{-1}) \\
\H_{0, CS, k}^{\langle 2 \rangle} (\delta) &:& \frac{\CSBias_{\theta, k}^{\langle 2 \rangle} (\hat{\psi}_{1})}{\mathsf{var}_\theta [ \hat{\psi}_{1} ]} < \delta^2.
\end{eqnarray*}
Unlike in \Cref{sec:cov}, we denote $k \coloneqq (k_{b}, k_{p})$ as a tuple rather than a scalar integer. Hence $\CSBias_{\theta, k} (\hat{\psi}_{1}) \coloneqq \BL_{2, b, k_{b}} \BL_{2, p, k_{p}}$ and $\CSBias_{\theta, k}^{\langle 2 \rangle} (\hat{\psi}_{1}) \coloneqq \BL_{2, b, k_{b}}^{2} \BL_{2, p, k_{p}}^{2}$.

Then following \cref{eq:cs_bound}, we have the following corollary of \Cref{lem:cs_logic}:
\begin{cor}\label{cor:cs_logic}
Under \Cref{cond:b}, $\NH_{0, CS}^{\langle 2 \rangle} \Rightarrow \NH_{0, CS, k}^{\langle 2 \rangle}$ for every $k$ and similarly $\H_{0, CS}^{\langle 2 \rangle} (\delta) \Rightarrow \H_{0, CS, k}^{\langle 2 \rangle} (\delta)$ for every $k$.
\end{cor}

In particular, it is straightforward to check that the following 4th order U-statistic is an unbiased estimator of $\CSBias_{\theta, k}^{\langle 2 \rangle} (\hat{\psi}_{1})$:
\begin{align*}
\widehat{\mathbb{IF}}_{44, k}^{\langle 2 \rangle} & \equiv \widehat{\mathbb{IF}}_{44, k}^{\langle 2 \rangle} ( (\Omega_{k}^{-1})) = \frac{(n - 4)!}{n!} \sum_{1 \le i_1 \neq i_2 \neq i_3 \neq i_4 \le n} \widehat{\IF}_{44, k, (i_1, i_2, i_3, i_4)}^{\langle 2 \rangle} (\Omega_{k}^{-1}) \text{ where} \\
& \widehat{\IF}_{44, k, (i_1, i_2, i_3, i_4)}^{\langle 2 \rangle} (\Omega_{k}^{-1}) = \hat{\varepsilon}_{b, i_1} \zbar_{k}(X_{i_1})^{\top} \Omega_{k}^{-1} \zbar_{k}(X_{i_2}) \hat{\varepsilon}_{b, i_2} \hat{\varepsilon}_{p, i_3} \zbar_{k} (X_{i_3})^{\top} \Omega_{k}^{-1} \zbar_{k} (X_{i_4}) \hat{\varepsilon}_{p, i_4}
\end{align*}

The following theorem summarizes the statistical properties of $\widehat{\IIFF}_{44, k}^{\langle 2 \rangle}$:
\begin{proposition}\label{thm:foif} 
Under \Cref{cond:w}, with $k_{b}, k_{p} \rightarrow \infty$, conditional on the training sample, $\widehat{\mathbb{IF}}_{44, k}^{\langle 2 \rangle}$ is unbiased for $\CSBias_{\theta, k}^{\langle 2 \rangle} (\hat{\psi}_{1})$ with variance of order $$\frac{1}{n^{2}} \max \left\{ \frac{k_{b} k_{p}}{n^{2}}, k_{p} \BL_{2, b, k_{b}}^{4}, k_{b} \BL_{2, p, k_{p}}^{4} \right\}.$$
In particular, $\se_{\theta} [\widehat{\mathbb{IF}}_{44, k}^{\langle 2 \rangle}] \coloneqq \{ \var_{\theta} [\widehat{\mathbb{IF}}_{44, k}^{\langle 2 \rangle}] \}^{1/2}$ can be estimated by $\widehat{\se} [\widehat{\mathbb{IF}}_{44, k}^{\langle 2 \rangle}] \coloneqq \{ \widehat{\mathsf{var}} [\widehat{\mathbb{IF}}_{44, k}^{\langle 2 \rangle}] \}^{1/2}$ defined in \Cref{sec:est.var.cs_if44.oracle} satisfying $\frac{\widehat{\se} [\widehat{\mathbb{IF}}_{44, k}^{\langle 2 \rangle}]}{\se_{\theta} [\widehat{\mathbb{IF}}_{44, k}^{\langle 2 \rangle}]} = 1 + o_{P_{\theta}}(1)$.

If further $k_{b} k_{p} = o(n^{2})$, $\var_{\theta} (\widehat{\mathbb{IF}}_{44, k}^{\langle 2 \rangle}) = o(1)$.
\end{proposition}

\begin{rem}\label{rem:order_var_if44}
We now characterize the conditions for the order of $\se_{\theta} \left[ \widehat{\mathbb{IF}}_{44, k}^{\langle 2 \rangle} \right]$ to be of the same order as $\var_{\theta} (\hat{\psi}_{1}) \asymp n^{-1}$. Without loss of generality, we assume that $\sqrt{k_{b}} \BL_{2, p, k_{p}} \gtrsim \sqrt{k_{p}} \BL_{2, b, k_{b}}$. Then
\begin{align*}
\se_{\theta} \left[ \widehat{\mathbb{IF}}_{44, k}^{\langle 2 \rangle} \right] = O \left( \frac{1}{n} \left\{ \frac{\sqrt{k_{b} k_{p}}}{n} + \sqrt{k_{b}} \BL_{2, p, k_{p}}^{2} \right\} \right),
\end{align*}
of which only the last term can be made order $1 / n$. Hence:
\begin{itemize}
\item when $\BL_{2, p, k_{p}}^{2} \lesssim \frac{\sqrt{k_{p}}}{n}$, $\se_{\theta} \left[ \widehat{\mathbb{IF}}_{44, k}^{\langle 2 \rangle} \right] \asymp \frac{\sqrt{k_{b} k_{p}}}{n^{2}}$;
\item when $\BL_{2, p, k_{p}}^{2} \gg \frac{\sqrt{k_{p}}}{n}$, $\se_{\theta} \left[ \widehat{\mathbb{IF}}_{44, k}^{\langle 2 \rangle} \right] \asymp \frac{\sqrt{k_{b}}}{n} \BL_{2, p, k_{p}}^{2}$.
\end{itemize}
\end{rem}

Based on the statistical properties of $\widehat{\mathbb{IF}}_{44, k}^{\langle 2 \rangle}$ summarized above, we now consider the properties of the following one-sided test $\widehat\chi_{CS, k}^{\langle 2 \rangle} (\zeta_{k}, \delta)$ of the null hypothesis $\H_{0, CS, k}^{\langle 2 \rangle} (\delta): \frac{\CSBias_{\theta, k}^{\langle 2 \rangle}}{\mathsf{var}_\theta [
\hat{\psi}_{1} ]} < \delta^2$, where
\begin{align}
\widehat\chi_{CS, k}^{\langle 2 \rangle} (\zeta_{k}, \delta) \coloneqq \mathbbm{1} \left\{ \frac{\widehat{\IIFF}_{44, k}^{\langle 2 \rangle}}{\widehat{\var}(\hat{\psi}_{1})} - \zeta_{k} \frac{\widehat{\se}[\widehat{\IIFF}_{44, k}^{\langle 2 \rangle}]}{\widehat{\var}(\hat{\psi}_{1})} > \delta^{2} \right\} \label{oneside_cs}
\end{align}
for user-specified $\zeta_{k}, \delta > 0$. We use a one-sided test because the sign of $\CSBias_{\theta, k}(\hat{\psi}_{1}) \geq 0$ is known {\it a priori}.

\Cref{thm:foif} characterizes the asymptotic properties of $\widehat\chi_{CS, k}^{\langle 2 \rangle} ((\alpha^{\dag})^{-1/2}, \delta)$ as a test for $\H_{0, CS, k}^{\langle 2 \rangle} (\delta)$:

\begin{proposition}\label{thm:test_cov_cs}
For $\psi(\theta) = \BE_{\theta}[\cov_\theta(A, Y | X)]$, under \Cref{cond:w}, when $k \rightarrow \infty$ but $k = o(n)$, for any given $\delta, \zeta_{k} > 0$, suppose that $\frac{\CSBias_{\theta, k}^{\langle 2 \rangle} (\hat{\psi}_{1})}{\var_\theta [\hat{\psi}_{1}]} \equiv \frac{\BL_{2, b, k}^{2} \BL_{2, p, k}^{2}}{\var_\theta [\hat{\psi}_{1}]} = \gamma^{2}$ for some (sequence) $\gamma^{2} = \gamma(n)^{2}$ (where $\gamma(n)$ can diverge with $n$), then the asymptotic rejection probability of $\widehat\chi_{CS, k}^{\langle 2 \rangle} ((\alpha^{\dag})^{-1/2}, \delta)$ is
\begin{align}\label{rejection_cs}
\lim_{n \rightarrow \infty} P_{\theta} \left\{ \frac{\widehat{\IIFF}_{44, k}^{\langle 2 \rangle} - \BE_{\theta} [\widehat{\IIFF}_{44, k}^{\langle 2 \rangle}]}{\se_{\theta} (\widehat{\IIFF}_{44, k}^{\langle 2 \rangle})} \geq (\alpha^{\dag})^{-1/2} - \lim_{n \rightarrow \infty} \left( \gamma^{2} - \delta^{2} \right) \frac{\var_\theta [\hat{\psi}_{1}]}{\se_\theta [\widehat{\IIFF}_{44, k}^{\langle 2 \rangle}]} \right\}
\end{align}
In particular,
\begin{enumerate}[label=(\arabic*)]
\item under $\H_{0, CS, k}^{\langle 2 \rangle} (\delta): \gamma^{2} \leq \delta^{2}$, $\widehat\chi_{CS, k}^{\langle 2 \rangle} ((\alpha^{\dag})^{-1/2}, \delta)$ rejects the null with probability less than or equal to $\alpha^{\dag}$, as $n \rightarrow \infty$;

\item under the following alternative to $\H_{0, CS, k}^{\langle 2 \rangle} (\delta)$: $\gamma^{2} = \delta^{2} + c^{2}$, for some sequence $c^{2} = c^{2}(n)$ such that $c^{2}(n) \gg \max\{ \sqrt{k_{b}} \BL_{2, p, k_{p}}^{2}, \sqrt{k_{p}} \BL_{2, b, k_{b}}^{2} \}$, $\widehat\chi_{CS, k}^{\langle 2 \rangle} ((\alpha^{\dag})^{-1/2}, \delta)$ rejects the null with probability converging to 1, as $n \rightarrow \infty$;
\end{enumerate}
\end{proposition}

\begin{rem}\label{rem:normal44}
We have not yet obtained the limiting distribution of $\frac{\widehat{\IIFF}_{44, k}^{\langle 2 \rangle} - \BE_{\theta}[\widehat{\IIFF}_{44, k}^{\langle 2 \rangle}]}{\se_{\theta} (\widehat{\IIFF}_{44, k}^{\langle 2 \rangle})}$. If the limiting distribution of $\frac{\widehat{\IIFF}_{44, k}^{\langle 2 \rangle} - \BE_{\theta}[\widehat{\IIFF}_{44, k}^{\langle 2 \rangle}]}{\se_{\theta} (\widehat{\IIFF}_{44, k}^{\langle 2 \rangle})}$ under $\H_{0, CS, k} (\delta)$ were known, we could have selected the cutoff based on the quantiles of the limiting distribution rather than using Chebyshev's inequality.
\end{rem}

\begin{rem}\leavevmode\label{rem:test_cov_cs}
The rejection probability of $\widehat\chi_{CS, k}^{\langle 2 \rangle} ((\alpha^{\dag})^{-1/2}, \delta)$ follows from a similar calculation as in \Cref{app:oracletest}, except that we do not use normality. Note that in \cref{rejection_cs}, $\frac{\widehat{\IIFF}_{44, k}^{\langle 2 \rangle} - \BE_{\theta} [\widehat{\IIFF}_{44, k}^{\langle 2 \rangle}]}{\se_{\theta} (\widehat{\IIFF}_{44, k}^{\langle 2 \rangle})}$ is $O_{P_{\theta}}(1)$. We next prove that \cref{rejection_cs} implies \Cref{thm:test_cov_cs}(1)-(2).
\begin{itemize}
\item Regarding (1), under $\H_{0, CS, k}^{\langle 2 \rangle} (\delta): \gamma^{2} \leq \delta^{2}$, we have
$$
- (\gamma^{2} - \delta^{2}) \frac{\var_\theta [\hat{\psi}_{1}]}{\se_\theta [\widehat{\IIFF}_{44, k}^{\langle 2 \rangle}]} \geq 0
$$
which implies that the rejection probability is less than $\alpha^{\dag}$ by Chebyshev's inequality.

\item Regarding (2), under any alternative to $\H_{0, CS, k}^{\langle 2 \rangle} (\delta): \gamma^{2} - \delta^{2} = c > 0$, without loss of generality, assume that $\sqrt{k_{b}} \BL_{2, p, k_{p}}^{2} \gtrsim \sqrt{k_{p}} \BL_{2, b, k_{b}}^{2}$. If $\BL_{2, p, k_{p}}^{2} \lesssim \frac{\sqrt{k_{p}}}{n}$, we must have $\BL_{2, b, k_{b}}^{2} \lesssim \frac{\sqrt{k_{b}}}{n}$ and hence $\BL_{2, p, k_{p}}^{2} \BL_{2, b, k_{b}}^{2} \lesssim \frac{\sqrt{k_{p} k_{b}}}{n} \ll \frac{1}{n}$. But if $\BL_{2, p, k_{p}}^{2} \gg \frac{\sqrt{k_{p}}}{n}$, it follows from \Cref{rem:order_var_if44} and \cref{rejection_cs} that the asymptotic rejection probability equals to
\begin{align*}
\lim_{n \rightarrow \infty} P_{\theta} \left( O_{P_{\theta}}(1) \geq (\alpha^{\dag})^{-1/2} - \lim_{n \rightarrow \infty} \Theta(b, \hat{b}, p, \hat{p}, f_{X}, \Zbar_{k_{b}}, \Zbar_{k_{p}}) c^{2} \frac{1}{\sqrt{k_{b}} \BL_{2, p, k_{p}}^{2}} \right)
\end{align*}
where $\Theta(b, \hat{b}, p, \hat{p}, f_{X}, \Zbar_{k_{b}}, \Zbar_{k_{p}})$ is some positive constant depending on the true regression functions $b$ and $p$, the estimated functions $\hat{b}, \hat{p}$ from the training sample, the density $f_{X}$ of $X$ and the chosen basis functions $\Zbar_{k_{b}}$ for $b$ and $\Zbar_{k_{p}}$ for $p$. The rejection probability, under the alternative to $\H_{0, CS, k}^{\langle 2 \rangle} (\delta)$, approaches 1 if $c^{2} \gg \sqrt{k_{b}} \BL_{2, p, k_{p}}^{2}$. For example:
\begin{itemize}
\item If $\BL_{2, p, k_{p}}^{2} = O (1)$, $c^{2} \gg \sqrt{k_{b}}$; 
\item If $\BL_{2, p, k_{p}}^{2} = o (1)$, $c^{2} \gtrsim \sqrt{k_{b}}$; 
\item Suppose $k_{b} \gg k_{p}$, and $\BL_{2, p, k_{p}}^{2} = O (\sqrt{k_{b}} / n)$, $c^{2} \gg \sqrt{k_{b} k_{p}} / n$.
\end{itemize}
\end{itemize}
\end{rem}

The following corollary, implied by \Cref{thm:test_cov_cs} and \Cref{lem:cs_logic}, summarizes the implication of the result of the test $\widehat\chi_{CS, k}^{\langle 2 \rangle} ((\alpha^{\dag})^{-1/2}, \delta)$ on the actual null hypothesis of interest $\H_{0, CS} (\delta)$.
\begin{cor}\leavevmode\label{cor:test_cov_cs}
Under the conditions in \Cref{thm:test_cov_cs}, $\widehat\chi_{CS, k}^{\langle 2 \rangle} ((\alpha^{\dag})^{-1/2}, \delta)$ is an asymptotically level $\alpha^{\dag}$ one-sided test for $\H_{0, CS} (\delta)$.
\end{cor}

\Cref{thm:test_cov_cs} is not quite satisfying because under alternatives to $\H_{0, CS, k}^{\langle 2 \rangle} (\delta)$ ($\gamma^{2} - \delta^{2} = c^{2}$), we may not be able to guarantee the power of $\widehat\chi_{CS, k}^{\langle 2 \rangle} (\zeta_{k}, \delta)$ converging to 1 over a large parameter space given a fixed tuple $k$. For example, in \Cref{rem:test_cov_cs}, we showed that if $\hat{p}$ is a $L_{2}(P_{\theta})$-consistent estimator of $p$, we still need $c^{2} \gtrsim \sqrt{k_{b}}$ to ensure that the power of the test approaches 1. To increase the chance of rejecting the null hypothesis of interest $\H_{0, CS}(\delta)$ when it is in fact false, we consider the following test, given $\mathcal{K} = \{ k_{1} = (k_{b, 1}, k_{p, 1}), \ldots, (k_{b, m}, k_{p, m}) \}$ with $m$ candidate tuples, we define
\begin{align}\label{joint_cs}
\widehat\chi_{CS}^{\langle 2 \rangle} ((\alpha^{\dag})^{-1/2}, \delta) = \max\{ \widehat\chi_{CS, k}^{\langle 2 \rangle} ((\alpha^{\dag} / m)^{-1/2}, \delta), k \in \mathcal{K} \}
\end{align}
where we use $\alpha^{\dag} / m$ in each $\widehat\chi_{CS, k}^{\langle 2 \rangle} ((\alpha^{\dag} / m)^{-1/2}, \delta)$ to adjust for multiple testing ($m$ tests in total).

The following corollary, which is a consequence of \Cref{thm:test_cov_cs} and \Cref{cor:test_cov_cs}, summarizes the asymptotic level and power of the test given in \cref{joint_cs}.
\begin{cor}\label{cor_cs}
Under the conditions in \Cref{thm:test_cov_cs},
\begin{enumerate}[label=(\arabic*)]
\item under $\H_{0, CS}^{\langle 2 \rangle} (\delta)$, $\widehat\chi_{CS}^{\langle 2 \rangle} ((\alpha^{\dag})^{-1/2}, \delta)$ in \cref{joint_cs} is an asymptotic level $\alpha^{\dag}$ test of the null;
\item under the alternative to $\H_{0, CS}^{\langle 2 \rangle} (\delta)$, if there exists at least a tuple $k \in \mathcal{K}$ such that $\H_{0, CS, k}^{\langle 2 \rangle} (\delta)$ is false, and $\gamma^{2} - \delta^{2} = c^{2} \gg \max\{ \sqrt{k_{b}} \BL_{2, p, k_{p}}^{2}, \sqrt{k_{p}} \BL_{2, b, k_{b}}^{2} \}$, $\widehat\chi_{CS}^{\langle 2 \rangle} ((\alpha^{\dag})^{-1/2}, \delta)$ rejects $\H_{0, CS}^{\langle 2 \rangle} (\delta)$ with probability approaching 1.
\end{enumerate}
\end{cor}

\begin{rem}
We choose $m$ to be bounded for technical reasons. If one had exponential inequalities on 4-th order U-statistics with explicit constants that can be estimated from data, we can generalize bounded $m$ to $m \rightarrow \infty$ as $n \rightarrow \infty$. In future work, we plan to study the following problem: under classical smoothness/sparsity assumptions on $b$ and $p$ but with unknown smoothness/sparsity levels, whether $\widehat\chi_{CS}^{\langle 2 \rangle} ((\alpha^{\dag})^{-1/2}, \delta)$ is the optimal adaptive test for $\H_{0, CS}^{\langle 2 \rangle} (\delta)$.
\end{rem}

\subsection{On $\var_{\theta} \left[ \widehat{\IIFF}_{44, k}^{\langle 2 \rangle} \right]$ and its estimator}\label{sec:est.var.cs_if44.oracle}
Similar to \Cref{sec:est.var}, we denote $\Omega_{k_{b}}^{-1/2} \Zbar_{k_{b}} \hat{\varepsilon}_{b}$ as $\B$ and $\Omega_{k_{p}}^{-1/2} \Zbar_{k_{p}} \hat{\varepsilon}_{p}$ as $\P$. We first write down the explicit formula of $\var_{\theta} \left[ \widehat{\IIFF}_{44, k}^{\langle 2 \rangle} \right]$.
\begin{equation}\label{eq:exact.var.cs_if44.oracle}
\begin{split}
& \ \var_{\theta} \left[ \widehat{\IIFF}_{44, k}^{\langle 2 \rangle} \right] \\
= & \ \underbrace{\frac{1}{n (n - 1) (n - 2) (n - 3)} \left\{ \begin{array}{c}
4 \BE_{\theta} \left[ \B_{1}^{\top} \B_{2} \P_{3}^{\top} \P_{4} \B_{1}^{\top} \B_{2} \P_{3}^{\top} \P_{4} \right] + 4 \BE_{\theta} \left[ \B_{1}^{\top} \B_{2} \P_{3}^{\top} \P_{4} \B_{3}^{\top} \B_{4} \P_{1}^{\top} \P_{2} \right] \\
+ \ 16 \BE_{\theta} \left[ \B_{1}^{\top} \B_{2} \P_{3}^{\top} \P_{4} \B_{2}^{\top} \B_{4} \P_{1}^{\top} \P_{3} \right]
\end{array} \right\}}_{\mathsf{(I)}} \\
& + \underbrace{\frac{n - 4}{n (n - 1) (n - 2) (n - 3)} \left\{ \begin{array}{c}
8 \BE_{\theta} \left[ \B_{1}^{\top} \B_{2} \P_{3}^{\top} \P_{4} \B_{1}^{\top} \B_{2} \P_{3}^{\top} \P_{5} \right] + 8 \BE_{\theta} \left[ \B_{1}^{\top} \B_{2} \P_{3}^{\top} \P_{4} \B_{1}^{\top} \B_{5} \P_{3}^{\top} \P_{4} \right] \\
+ \ 8 \BE_{\theta} \left[ \B_{1}^{\top} \B_{2} \P_{3}^{\top} \P_{4} \B_{3}^{\top} \B_{4} \P_{1}^{\top} \P_{5} \right] + 8 \BE_{\theta} \left[ \B_{1}^{\top} \B_{2} \P_{3}^{\top} \P_{4} \B_{3}^{\top} \B_{5} \P_{1}^{\top} \P_{2} \right] \\
+ \ 16 \BE_{\theta} \left[ \B_{1}^{\top} \B_{2} \P_{3}^{\top} \P_{4} \B_{1}^{\top} \B_{3} \P_{2}^{\top} \P_{5} \right] + 16 \BE_{\theta} \left[ \B_{1}^{\top} \B_{2} \P_{3}^{\top} \P_{4} \B_{3}^{\top} \B_{5} \P_{1}^{\top} \P_{4} \right] \\
+ \ 16 \BE_{\theta} \left[ \B_{1}^{\top} \B_{2} \P_{3}^{\top} \P_{4} \B_{1}^{\top} \B_{3} \P_{4}^{\top} \P_{5} \right] + 16 \BE_{\theta} \left[ \B_{1}^{\top} \B_{2} \P_{3}^{\top} \P_{4} \B_{1}^{\top} \B_{5} \P_{2}^{\top} \P_{4} \right]
\end{array} \right\}}_{\mathsf{(II)}} \\
& + \underbrace{\frac{(n - 4) (n - 5)}{n (n - 1) (n - 2) (n - 3)} \left\{ \begin{array}{c}
4 \BE_{\theta} \left[ \B_{1}^{\top} \B_{2} \P_{3}^{\top} \P_{4} \B_{1}^{\top} \B_{2} \P_{5}^{\top} \P_{6} \right] + 4 \BE_{\theta} \left[ \B_{1}^{\top} \B_{2} \P_{3}^{\top} \P_{4} \B_{5}^{\top} \B_{6} \P_{3}^{\top} \P_{4} \right] \\
+ \ 4 \BE_{\theta} \left[ \B_{1}^{\top} \B_{2} \P_{3}^{\top} \P_{4} \B_{3}^{\top} \B_{4} \P_{5}^{\top} \P_{6} \right] + 4 \BE_{\theta} \left[ \B_{1}^{\top} \B_{2} \P_{3}^{\top} \P_{4} \B_{5}^{\top} \B_{6} \P_{1}^{\top} \P_{2} \right] \\
+ \ 16 \BE_{\theta} \left[ \B_{1}^{\top} \B_{2} \P_{3}^{\top} \P_{4} \B_{1}^{\top} \B_{3} \P_{5}^{\top} \P_{6} \right] + 16 \BE_{\theta} \left[ \B_{1}^{\top} \B_{2} \P_{3}^{\top} \P_{4} \B_{5}^{\top} \B_{6} \P_{1}^{\top} \P_{3} \right] \\
+ \ 16 \BE_{\theta} \left[ \B_{1}^{\top} \B_{2} \P_{3}^{\top} \P_{4} \B_{2}^{\top} \B_{5} \P_{3}^{\top} \P_{6} \right] + 8 \BE_{\theta} \left[ \B_{1}^{\top} \B_{2} \P_{3}^{\top} \P_{4} \B_{3}^{\top} \B_{5} \P_{2}^{\top} \P_{6} \right]
\end{array} \right\}}_{\mathsf{(III)}} \\
& + \underbrace{\frac{(n - 4) (n - 5) (n - 6)}{n (n - 1) (n - 2) (n - 3)} \left\{ \begin{array}{c}
4 \BE_{\theta} \left[ \B_{1}^{\top} \B_{2} \P_{3}^{\top} \P_{4} \B_{1}^{\top} \B_{5} \P_{6}^{\top} \P_{7} \right] + 4 \BE_{\theta} \left[ \B_{1}^{\top} \B_{2} \P_{3}^{\top} \P_{4} \B_{5}^{\top} \B_{6} \P_{3}^{\top} \P_{7} \right] \\
+ \ 8 \BE_{\theta} \left[ \B_{1}^{\top} \B_{2} \P_{3}^{\top} \P_{4} \B_{5}^{\top} \B_{6} \P_{1}^{\top} \P_{7} \right]
\end{array} \right\}}_{\mathsf{(IV)}} \\
& - \underbrace{\frac{8 (2n^{3} - 21n^{2} + 79n - 105)}{n (n - 1) (n - 2) (n - 3)} \left( \BE_{\theta} \left[ \B^{\top} \right] \BE_{\theta} \left[ \B \right] \BE \left[ \P^{\top} \right] \BE_{\theta} \left[ \P \right] \right)^{2}}_{\mathsf{(V)}}
\end{split}
\end{equation}
Thus we have, under \Cref{cond:w},
\begin{align*}
\var_{\theta} \left[ \widehat{\mathbb{IF}}_{44, k}^{\langle 2 \rangle} \right] = O \left( \begin{array}{c}
\underbrace{\frac{k_{b} k_{p}}{n^{4}}}_{\mathsf{(I)}} + \underbrace{\frac{k_{p}}{n^{3}} \BL_{2, b, k_{b}}^{2} + \frac{k_{b}}{n^{3}} \BL_{2, p, k_{p}}^{2}}_{\mathsf{(II)}} \\
+ \ \underbrace{\frac{1}{n^{2}} \left\{ k_{p} \BL_{2, b, k_{b}}^{4} + k_{b} \BL_{2, p, k_{p}}^{4} + \sqrt{k_{b} k_{p}} \BL_{2, b, k_{b}}^{2} \BL_{2, p, k_{p}}^{2} + \BL_{2, b, k_{b}}^{3} \BL_{2, p, k_{p}} + \BL_{2, b, k_{b}} \BL_{2, p, k_{p}}^{3} \right\}}_{\mathsf{(III)}} \\
+ \ \underbrace{\frac{\BL_{2, p, k_{p}}^{2} \BL_{2, b, k_{b}}^{2}}{n} \left\{ \BL_{2, b, k_{b}}^{2} + \BL_{2, p, k_{p}}^{2} + \BL_{2, p, k_{p}} \BL_{2, b, k_{b}} \right\}}_{\mathsf{(IV)}} 
\end{array} \right)
\end{align*}

$\widehat{\var} \left[ \widehat{\IIFF}_{44, k}^{\langle 2 \rangle} \right]$, the variance estimator of $\widehat{\IIFF}_{44, k}^{\langle 2 \rangle}$ that also satisfies 
$
\frac{\widehat{\var} \left[ \widehat{\mathbb{IF}}_{44, k}^{\langle 2 \rangle} \right]}{\var_{\theta} \left[ \widehat{\mathbb{IF}}_{44, k}^{\langle 2 \rangle} \right]} = 1 + o_{P_{\theta}} (1),
$
can be constructed in a similar fashion to $\widehat{\var} \left[ \widehat{\IIFF}_{22, k} \right]$ by estimating each term appeared in \cref{eq:exact.var.cs_if44.oracle} by a corresponding unbiased U-statistic. Due to its complicated form, we do not report it here. For example, term $\mathsf{(V)}$ of \cref{eq:exact.var.cs_if44.oracle} can be estimated by the following 8-th order U-statistic:
\begin{align*}
\widehat{\mathsf{(V)}} = \frac{8 (2n^{3} - 21n^{2} + 79n - 105)}{n (n - 1) (n - 2) (n - 3)} \frac{(n - 8)!}{n!} \sum_{1 \leq i_{1} \neq \cdots \neq i_{8} \leq n} \B_{i_{1}}^{\top} \B_{i_{2}} \P_{i_{3}}^{\top} \P_{i_{4}} \B_{i_{5}}^{\top} \B_{i_{6}} \P_{i_{7}}^{\top} \P_{i_{8}}.
\end{align*}

\section{Details in the sequential test}\label{app:hierarchy}
\subsection{Testing $\H_{0, 2, k} (\delta)$ for any given $k = o(n^2)$}\label{q1} 
We start by developing a test for a single null hypothesis $\H_{0, 2, k} (\delta)$ for any given $k \in \mathcal{K}_{J}$. $\H_{0, 2, k_0} (\delta)$ with $k_0 = o(n)$ is then a special case. As $\mathsf{Bias}_\theta (\hat{\psi}_{2, k}) \equiv \mathsf{TB}_{\theta, k} (\hat{\psi}_{1})$ is not consistently estimable without further assumptions on $p$ and $\hat{p}$, we instead consider testing the following surrogate hypothesis, for some $k^{\prime }\in \mathcal{K}_{J}$ such that $k = o(k^{\prime })$, 
\begin{align}  \label{h0kk}
\mathsf{H}_{0, 2, k \rightarrow k^{\prime }}(\delta): \frac{\mathsf{Bias}_{\theta, k^{\prime }} (\hat{\psi}_{2, k})}{ \mathsf{s.e.}_\theta [\hat{\psi}_{2, k}] } < \delta
\end{align}
where $\mathsf{Bias}_{\theta, k^{\prime }} (\hat{\psi}_{2, k}) \coloneqq \mathsf{Bias}_\theta (\hat{\psi}_{2, k}) - \mathsf{Bias}_\theta (\hat{\psi}_{2, k^{\prime }}) \equiv \mathbb{E}_\theta [ \widehat{\mathbb{IF}}_{22, k^{\prime }} - \widehat{\mathbb{IF}}_{22, k}]$ is an estimable part of $\mathsf{Bias}_\theta (\hat{\psi}_{2, k})$. Similar to \Cref{lem:var}, we have the following:

\begin{lem}\label{lem:var_{k}} 
For $\psi(\theta) = \mathbb{E}_\theta [ \mathsf{var}_\theta (A | X) ]$, and $k, k^{\prime }, k_{1}, k_{2} \in \mathcal{J}$:

\begin{enumerate}
\item Given $k < k_1 < k_2$, $\mathsf{Bias}_{\theta, k_1} (\hat{\psi}_{2, k}) \le \mathsf{Bias}_{\theta, k_2} (\hat{\psi}_{2, k}) \le \mathsf{Bias}_\theta (\hat{\psi}_{2, k})$;

\item Given $k_1 < k_2 < k^{\prime }$, $\mathsf{Bias}_{\theta, k^{\prime }} (\hat{\psi}_{2, k_1}) \ge \mathsf{Bias}_{\theta, k^{\prime }} (\hat{\psi}_{2, k_2})$;

\item For any $k' > k$, $\mathsf{H}_{0, 2, k} (\delta) \Rightarrow \mathsf{H}_{0, 2, k \rightarrow k^{\prime }}(\delta)$.
\end{enumerate}
\end{lem}

\begin{proof}
(1) - (3) directly follows from $\Bias_{\theta, k'} (\hat{\psi}_{2, k}) = \BE_\theta [ \widehat{\IIFF}_{22, k'} - \widehat{\IIFF}_{22, k} ]$ and the larger $k'$ (or the smaller $k$), the larger the difference $\BE_\theta [ \widehat{\IIFF}_{22, k'} - \widehat{\IIFF}_{22, k} ]$ for $\psi(\theta) = \BE_\theta [ \var_\theta (A | X) ]$.
\end{proof}

Similar to $\widehat\chi_{k}^{(1)} (\zeta_{k}, \delta)$ \eqref{oneside}, we define the following test statistic\footnote{As the convention in this paper, $\widehat\chi_{2, k \rightarrow k^{\prime}} (\zeta_{k \rightarrow k^{\prime }}, \delta) \equiv \widehat\chi_{2, k \rightarrow k^{\prime }} (\Omega_{k}^{-1}, \Omega_{k^{\prime }}^{-1}; \zeta_{k \rightarrow k^{\prime }}, \delta)$. \newline $\widehat{\mathsf{s.e.}} [ \widehat{\mathbb{IF}}_{22, k^{\prime }} ]$ were defined in \Cref{thm:soif}. We use $\widehat{\mathsf{s.e.}} [ \widehat{\mathbb{IF}}_{22, k^{\prime }} ]$ instead of $\widehat{\mathsf{s.e.}} [ \widehat{\mathbb{IF}}_{22, k^{\prime}} - \widehat{\mathbb{IF}}_{22, k} ]$ because when $k = o(k^{\prime })$, $\frac{\mathsf{s.e.}_{\theta}[\widehat{\mathbb{IF}}_{22, k^{\prime }}]}{\mathsf{s.e.}_{\theta}[\widehat{\mathbb{IF}}_{22, k^{\prime }} - \widehat{\mathbb{IF}}_{22, k}]} \rightarrow 1$. We choose $\widehat{\mathsf{s.e.}} [ \hat{\psi}_{2, k_{0}} ] = \widehat{\mathsf{s.e.}} (\hat{\psi}_{1})$ as we assumed that $\mathsf{s.e.}_{\theta}(\widehat{\mathbb{IF}}_{22, k_{0}}) \asymp \sqrt{k_{0}} / n \ll \mathsf{s.e.}_{\theta} (\hat{\psi}_{1})$ in \Cref{sec:hierarchy}.}, given $k, k^{\prime }\in \mathcal{K}_{J}$ and $k = o(k^{\prime })$, 
\begin{equation}  \label{test_k}
\widehat\chi_{2, k \rightarrow k^{\prime }} (\zeta_{k \rightarrow k^{\prime }}, \delta) \coloneqq \mathbbm{1} \left\{ \frac{\widehat{\mathbb{IF}}_{22, k^{\prime }} - \widehat{\mathbb{IF}}_{22, k}}{\widehat{\mathsf{s.e.}} [\hat{\psi}_{2, k}]} - \zeta_{k \rightarrow k^{\prime }} \frac{\widehat{\mathsf{s.e.}} [ \widehat{\mathbb{IF}}_{22, k^{\prime }} ]}{\widehat{\mathsf{s.e.}} [ \hat{\psi}_{2, k} ]} > \delta \right\}.
\end{equation}
When $k > n$, we choose $\widehat{\mathsf{s.e.}} (\hat{\psi}_{2, k}) = \widehat{\mathsf{s.e.}} [ \widehat{\mathbb{IF}}_{22, k} ]$ as $\se_{\theta} [ \widehat{\mathbb{IF}}_{22, k} ] \gg \se_{\theta} (\hat{\psi}_{1})$, whereas for $k_{0}$, we choose $\widehat{\mathsf{s.e.}} (\hat{\psi}_{2, k_{0}}) = \widehat{\se} (\hat{\psi}_{1})$ as we have assumed that $\se_{\theta} (\hat{\psi}_{2, k_{0}}) \asymp \sqrt{k_{0}} / n$. Note that $\widehat{\mathbb{IF}}_{22, k^{\prime }} - \widehat{\mathbb{IF}}_{22, k} = \hat{\psi}_{2, k^{\prime }} - \hat{\psi}_{2, k}$. Thus the test statistic $\widehat\chi_{2, k \rightarrow k^{\prime }} (\zeta_{k \rightarrow k^{\prime}}, \delta)$ corresponds exactly to the comparisons that were described in \Cref{rem:heu}. The following proposition characterizes the asymptotic level and power of the test $\widehat\chi_{k \rightarrow k^{\prime }} (\zeta_{k \rightarrow k^{\prime }}, \delta)$ of $\H_{0, 2, k \rightarrow k^{\prime }}(\delta)$:

\begin{proposition}
\label{prop:test_{k}}\leavevmode
Under \Cref{cond:w}, given $k^{\prime }, k \in \mathcal{K}_{J}$ and $k = o(k^{\prime })$, for any given $\delta, \zeta_{k \rightarrow k^{\prime }} > 0$, suppose that $\frac{\mathsf{Bias}_{\theta, k^{\prime }} (\hat{\psi}_{2, k})}{\mathsf{s.e.}_\theta (\hat{\psi}_{2, k})} = \gamma$ for some (sequence) $\gamma = \gamma(n)$, then the rejection probability of $\widehat\chi_{2, k \rightarrow k^{\prime }} (\zeta_{k \rightarrow k^{\prime }}, \delta)$ converges to $1 - \Phi \left( \zeta_{k \rightarrow k'} - \lim_{n \rightarrow \infty} (\gamma - \delta) \frac{\mathsf{s.e.}_\theta (\hat{\psi}_{2, k})}{\mathsf{s.e.}_\theta [\widehat{\mathbb{IF}}_{22, k'}]} \right)$ where $\frac{\se_{\theta} (\hat{\psi}_{2, k})}{\se_{\theta} [\widehat{\mathbb{IF}}_{22, k^{\prime }}]} \rightarrow \sqrt{\frac{\max\{k, n\}}{k'}} C$ for some constant $C > 0$, as $n \rightarrow \infty$. In particular,

\begin{enumerate}[label=(\arabic*)]

\item under $\H_{0, 2, k \rightarrow k'} (\delta)$, $\widehat\chi_{2, k \rightarrow k'} (\zeta_{k \rightarrow k'}, \delta)$ rejects the null hypothesis with probability less than or equal to $1 - \Phi (\zeta_{k \rightarrow k'})$, as $n \rightarrow \infty$;

\item under the following alternative to $\H_{0, 2, k \rightarrow k'} (\delta)$ \eqref{h0kk} with $\gamma = \delta + c$ for some $c > 0$,
\begin{enumerate}[label=(\roman*)]
\item if $c = o \left(\sqrt{\frac{k'}{\max\{k, n\}}}\right)$\footnote{The relevance of the local alternative regime $\gamma - \delta \asymp \sqrt{\frac{k'}{\max\{ k, n \}}}$ is a consequence of the variance of the statistic $\frac{\widehat{\mathbb{IF}}_{22, k'}}{\widehat{\se} (\hat{\psi}_{2, k})}$ being of order $k' / \max\{k, n\}$.}, $\widehat\chi_{2, k \rightarrow k^{\prime }} (\zeta_{k \rightarrow k'}, \delta)$ rejects the null with probability approaching $1 - \Phi(\zeta_{k \rightarrow k'})$, as $n \rightarrow \infty$;

\item if $c \rightarrow C' \sqrt{\frac{k'}{\max\{k, n\}}}$ as $n \rightarrow \infty$ for some constant $C' > 0$, $\widehat\chi_{2, k \rightarrow k^{\prime }} (\zeta_{k \rightarrow k'}, \delta)$ rejects the null with probability approaching $1 - \Phi(\zeta_{k \rightarrow k'} - C C' )$, as $n \rightarrow \infty$;

\item if $c \gg \sqrt{\frac{k'}{\max\{k, n\}}}$, then $\widehat\chi_{2, k \rightarrow k'} (\zeta_{k \rightarrow k'}, \delta)$ rejects the null with probability approaching 1, as $n \rightarrow \infty$.
\end{enumerate}
\end{enumerate}
\end{proposition}

\begin{rem}\label{rem:multiple}
Consider the problem of testing $\mathsf{H}_{0, 2, k} (\delta)$, for which
the surrogate hypotheses are $\mathsf{H}_{0, 2, k \rightarrow k^{\prime}}
(\delta): \frac{\mathsf{Bias}_{\theta, k^{\prime }} (\hat{\psi}_{2, k})}{%
\mathsf{s.e.}_{\theta} [\hat{\psi}_{2, k}]} \leq \delta$ for all $%
k^{\prime}\in \mathcal{K}_{J}$ with $k^{\prime }> k$. Under the alternative
to $\mathsf{H}_{0, 2, k \rightarrow k^{\prime }} (\delta)$, %
\Cref{prop:test_{k}}(2) states that the test $\widehat{\chi}_{2, k
\rightarrow k^{\prime }} (\zeta_{k \rightarrow k^{\prime }}, \delta)$
rejects the surrogate null $\mathsf{H}_{0, 2, k \rightarrow k^{\prime }}
(\delta)$ with probability converging to 1, only under a diverging
alternative to $\mathsf{H}_{0, 2, k \rightarrow k^{\prime }} (\delta)$ when $%
\frac{\mathsf{Bias}_{\theta, k^{\prime }} (\hat{\psi}_{2, k})}{\mathsf{s.e.}%
_{\theta} [\hat{\psi}_{2, k}]} - \delta \gg \sqrt{\frac{k^{\prime }}{k}}$,
with $\sqrt{\frac{k^{\prime }}{k}}$ growing with sample size when $k =
o(k^{\prime})$. With increasing $k^{\prime }$, even though $\frac{\mathsf{%
Bias}_{\theta, k^{\prime }} (\hat{\psi}_{2, k})}{\mathsf{s.e.}_{\theta} [%
\hat{\psi}_{2, k}]}$ increases by \Cref{lem:var_{k}}(1) (or equivalently the
surrogate null $\mathsf{H}_{0, 2, k \rightarrow k^{\prime }} (\delta)$ is
more likely to be false), the parameter space under the alternative for
which $\widehat{\chi}_{2, k \rightarrow k^{\prime }} (\zeta_{k \rightarrow k^{\prime }}, \delta)$ does not have power approaching 1 to reject also expands. Hence we choose to test multiple surrogate hypotheses to improve the chance of rejecting $\mathsf{H}_{0, 2, k} (\delta)$ when it is in fact false. See \cref{test_joint} in \Cref{sec:hierarchy}.
\end{rem}

\section{Simulation experiments}\label{sec:simulations}
In this section, we first describe the setups of our simulation studies. We focus on two parameters of interest: $\psi (\theta) = \BE_{\theta} [\var_{\theta} (A | X)]$ and $\psi (\theta) = \BE_{\theta} [\cov_{\theta} (Y, A | X)]$. We consider two simulation setups. In simulation setup I, we draw 101 replicates of datasets, each with sample size $n = 5000$ in the following way: 
\begin{itemize}
\item Draw $X_{j}$ for $j = 1, 2$ (so $d = 2$) with the density function $f_{X}$ supported on $[0, 1]$ such that the marginal densities $f_{X_{j}}$ for $j = 1, 2$ in each direction satisfies $f_{X_{j}} \in \text{H\"{o}lder}(s_{f} = 0.1)$, based on the algorithm described in \Cref{app:multiX}. 
\item Then draw $A$ and $Y$ according to the following data generating mechanism:
\begin{equation*}
A \sim p(X) + N(0, 1) \text{ where } p(X) = \sum_{j = 1}^{2} \tau_{p, j} h_{p} (X_{j}; 0.25)
\end{equation*}
and
\begin{equation*}
Y \sim \mathsf{Bernoulli}\left( b(X) \equiv \mathsf{expit} \left\{ \sum_{j = 1}^{2} \tau_{b, j} h_{b} (X_{j}; 0.25) \right\} \right).
\end{equation*}
\end{itemize}
We fix one of the 101 replicates as the training sample and the rest 100 replicates as the estimation sample. In simulation setup II, we only change $f_{X_{j}} \in \text{H\"{o}lder}(s_{f} = 0.1)$ to $f_{X_{j}} \in \text{H\"{o}lder}(s_{f} = 0.4)$ and fix everything else in the data generating mechanism. Because of the additive structure across the $d = 2$ dimensions in the definitions of $p(X)$ and $b(X)$, the minimax rates of convergence in $L_2(P_{\theta})$ norm for estimating $b$ and $p$ are $n^{- \frac{s_b}{1 + 2 s_b}} \lesssim n^{-1/6}$ and $n^{-\frac{s_p}{1 + 2 s_p}} \lesssim n^{-1/6}$, which are dimension free \citep{stone1985additive}.

In both simulation setups, while constructing the second order influence function estimators and tests, we 
\begin{itemize}
\item For each $k$, the basis $\zbar_{k}(X)$ is the concatenation of the DB6 father wavelets at level $\mathsf{log}_2(k / 2)$ applied to each of the $d = 2$ dimensions of $X$. The chosen basis functions satisfy \Cref{cond:b}. We plan to compare the performance of our proposed assumption free test statistics using different basis functions or even using data-driven algorithms to select basis functions from a given dictionary of functions in future works.

\item To compute $\widehat{\mathbb{IF}}_{22, k}$, we need the true $\Omega_{k}$. Since the analytical form of $\Omega_{k}$ is difficult to derive, we ``estimate'' $\Omega_{k}$ with sample covariance matrix from an independent sample of extremely large size of $5 \times 10^7$ (or $X$-semisupervised dataset) as the ``oracle'' $\Omega_{k}$.
\end{itemize}

\subsection{Simulation setup I: nonsmooth $f_X$}\label{sec:main_design}
We display the following results in a similar format to \Cref{tab:intro}. For regression functions estimators other than kernel regression with cross validation, we do not report results for $k = 4096$.

\begin{table}[tbp]
\caption{}
\label{tab:smooth_nn_var}
\resizebox{\columnwidth}{!}{
\begin{tabular}{l | lclc}
\hline
$k$ & $\oracleSOIF$ & \shortstack{MC Coverage \\ ($\hat{\psi}_{2, k} $ 90\% Wald CI)} & $\mathsf{Bias} ( \oraclepsi )$ & $\widehat\chi_{k}^{(1)} ( \Omega_{k}^{-1}; z_{0.10}, \delta = 0.75 (1.5))$ \\
\hline
$0$ & 0 (0) & 0\% & 0.392 (0.0229) & 0\% (0\%) \\
$64$ & 0.245 (0.0189) & 0\% & 0.147 (0.0137) & 100\% (100\%) \\
$128$ & 0.251 (0.0209) & 0\% & 0.143 (0.0144) & 100\% (100\%) \\
$256$ & 0.331 (0.0295) & 0\% & 0.0652 (0.0114) & 100\% (100\%) \\
$512$ & 0.332 (0.0275) & 0\% & 0.0653 (0.0139) & 100\% (100\%) \\
$1024$ & 0.332 (0.0269) & 2\% & 0.0634 (0.0163) & 100\% (100\%) \\
$2048$ & 0.366 (0.0385) & 74\% & 0.0300 (0.0244) & 100\% (100\%) \\
\hline
$k$ & $\genericSOIF$ & \shortstack{MC Coverage \\ ($\hat{\psi}_{2, k} (\widehat{\Omega}_{k}^{-1} )$ 90\% Wald CI)} & $\mathsf{Bias} \left( \genericpsi \right)$ & $\widehat\chi_{k}^{(1)} ( \widehat{\Omega}_{k}^{-1}; z_{0.10}, \delta = 0.75 (1.5))$ \\
\hline
$0$ & 0 (0) & 0\% & 0.392 (0.0229) & 0\% (0\%) \\
$64$ & 0.249 (0.0188) & 0\% & 0.144 (0.0126) & 100\% (100\%) \\
$128$ & 0.257 (0.0207) & 0\% & 0.137 (0.0131) & 100\% (100\%) \\
$256$ & 0.345 (0.0297) & 0\% & 0.0505 (0.00993) & 100\% (100\%) \\
$512$ & 0.357 (0.0265) & 6\% & 0.0396 (0.0121) & 100\% (100\%) \\
$1024$ & 0.367 (0.0245) & 32\% & 0.0277 (0.0126) & 100\% (100\%) \\
$2048$ & 0.373 (0.0326) & 64\% & 0.0230 (0.0178) & 100\% (100\%) \\
\hline
\end{tabular}
} \newline
Simulation setup I ($\psi(\theta) = \mathbb{E}_\theta [ \mathsf{var}_\theta (A | X) ]$): We reported the MCavs of point estimates and MCsds (first column in each panel) of $\widehat{\mathbb{IF}}_{22, k}$ and $\widehat{\mathbb{IF}}_{22, k} (\widehat{\Omega}_{k}^{-1})$, together with the coverage probability of 90\% confidence intervals (second column in each panel) of $\hat{\psi}_{2, k}$ and $\hat{\psi}_{2, k} (\widehat{\Omega}_{k}^{-1})$, the MCavs of the bias and MCsds (third column in each panel) of $\hat{\psi}_{2, k}$ and $\hat{\psi}_{2, k} (\widehat{\Omega}_{k}^{-1})$ and the empirical rejection rate based on the test statistic $\widehat\chi_{k}^{(1)} (\zeta_{k}, \delta = 0.75 \text{ or } 1.5)$ and $\widehat\chi_{k}^{(1)} (\widehat{\Omega}_{k}^{-1}; \zeta_{k}, \delta = 0.75 \text{ or } 1.5)$ (see \Cref{sec:var}) with $\zeta_{k} = z_{\alpha^{\dag} = 0.10} = 1.28$ (fourth column in each panel). Nuisance functions are estimated by neural networks.
\end{table}

\subsubsection{$\psi(\theta) = \BE_{\theta} [\var_{\theta}(A | X)] \equiv 1$}
In this section, we report other complementary results to those shown in \Cref{tab:intro}, with the nuisance function $p$ estimated by other methods like neural networks instead of nonparametric kernel regression with cross validation. We are still estimating $\psi(\theta) = \BE_{\theta} [\var_{\theta}(A | X)] \equiv 1$ and trying to falsify if the Wald CI centered at $\hat{\psi}_{1}$ undercovers using the one-sided tests $\widehat{\chi}_{k}^{(1)} (\zeta_{k}, \delta)$ or $\widehat{\chi}_{k}^{(1)} (\widehat{\Omega}_{k}^{-1}; \zeta_{k}, \delta)$ with $\delta = 0.75$ or 1.5 and $\zeta_{k} = z_{0.10} = 1.28$. Hence the tests are nominal 0.10 level two-sided tests.


\Cref{tab:smooth_nn_var} reports the simulation results when the nuisance function $p$ is estimated by fully connected neural networks with 5 layers, each layer with 20 neurons (width = 20 for all layers). The default rectified linear unit (ReLU) activation function was used in every intermediate layer and we set the learning rate parameter to be $2 \times 10^{-4}$. Recently, there are some very interesting theoretical analyses on the convergence rates of neural network estimators of functions in \Holder{}-type of function classes \citep{schmidt2017nonparametric}. However, it is still quite difficult to choose the ``right'' architecture based on these theoretical results because the corresponding width and depth parameters are only optimal up to constants. In the simulation, we did not try to optimize the network architectures in order to obtain ``optimal'' prediction of $b$ and $p$. With the current architecture setup, the DRML estimator based on neural network nuisance estimators indeed has a slightly larger bias (e.g. MC bias is 0.392 for $\psi (\theta) = \BE_{\theta} [\var_{\theta} (A | X)]$ in simulation setup I) than that based on nonparametric kernel regression nuisance estimators (e.g. MC bias is 0.229). All implementation was done using the R interface to TensorFlow. As suggested by \Cref{tab:smooth_nn_var}, both the test statistics $\widehat\chi_{k}^{(1)} (\zeta_{k}, \delta)$ and $\widehat\chi_{k}^{(1)} (\widehat{\Omega}_{k}^{-1}; \zeta_{k}, \delta)$ are able to reject $\H_{0, k} (\delta)$ and hence reject $\H_{0} (\delta)$ for various $k$ ranging from 64 to 2048 in 100 out of 100 simulations.

\subsubsection{$\psi(\theta) = \BE_{\theta} [\cov_{\theta}(A, Y | X)] \equiv 0$}\label{sec:cov_sim}
In \Cref{tab:smooth_kern_cov}, unlike in the previous setup, we are estimating $\psi (\theta) = \BE_{\theta} [\cov_{\theta}(A, Y | X)] \equiv 0$ and trying to falsify if the Wald CI centered at $\hat{\psi}_{1}$ undercovers using the two-sided tests $\widehat{\chi}_{k}^{(2)} (\zeta_{k}, \delta)$ or $\widehat{\chi}_{k}^{(2)} (\widehat{\Omega}_{k}^{-1}; \zeta_{k}, \delta)$ with $\delta = 0.10$ and $\zeta_{k} = z_{0.05} = 1.64$. Hence the tests are nominal 0.10 level two-sided tests. Choosing $\delta = 0.10$ corresponds to tolerating the coverage rate of a 90\% Wald CI centered at $\hat{\psi}_{1}$ under the null hypothesis of $\H_{0, k}(\delta)$ is no smaller than 89.3\%, which is a quite stringent requirement. We also display the results when the regression functions are estimated by nonparametric kernel regression with cross validation and neural networks as described in the previous section. 

Reading from the first row of \Cref{tab:smooth_kern_cov}, the MC bias of $\hat{\psi}_{1}$ is -0.0124, almost 2 times the MCsd of $\hat{\psi}_{1}$ which is 0.00725. As expected, the MC coverage rate of $\hat{\psi}_{1} \pm 1.64 \widehat{\se}(\hat{\psi}_{1})$ is 100\%. Hence the null hypothesis $\H_{0, k} (\delta)$ is false for $\delta$ small, e.g. $\delta = 0.01$. Indeed, when $k$ is relatively large compared to $n$, our test rejects the null hypothesis with high probability: e.g. when $k = 256$, our test rejects $\H_{0, k} (\delta)$ in 94 out of 100 simulations. Interestingly, when $k$ increases, the rejection probability of the two-sided tests $\widehat\chi_{k}^{(2)} (z_{0.05}, \delta = 0.10)$ and $\widehat\chi_{k}^{(2)} (\widehat{\Omega}_{k}^{-1}; z_{0.05}, \delta = 0.10)$ could also decrease, because of the greater variance of $\widehat{\IIFF}_{22, k}$ (the first column of \Cref{tab:smooth_kern_cov}). When $\Omega_{k}^{-1}$ is estimated from data (lower panel of \Cref{tab:smooth_kern_cov}), we observe similar results to the case of known $\Omega_{k}^{-1}$ displayed in the upper panel of \Cref{tab:smooth_kern_cov}.

\Cref{tab:smooth_nn_cov} reports the simulation results when the regression functions $b$ and $p$ are estimated by neural networks, whose architecture is described in the previous section. As in the previous section, $\hat{\psi}_{1}$ estimated by neural networks has larger bias (MC bias is -0.0216) compared to that estimated by nonparametric kernel regression (MC bias is -0.0124). Thus as expected, the rejection probabilities of the 0.10 level two-sided tests $\widehat\chi_{k}^{(2)} (z_{0.05}, \delta = 0.10)$ and $\widehat\chi_{k}^{(2)} (\widehat{\Omega}_{k}^{-1}; z_{0.05}, \delta = 0.10)$ are in general closer to 100\%, compared to the case with $b$ and $p$ estimated by nonparametric kernel regression with cross validation (\Cref{tab:smooth_kern_cov}).

\begin{table}[tbp]
\caption{}
\label{tab:smooth_kern_cov}
\resizebox{\columnwidth}{!}{
\begin{tabular}{l | lclc}
\hline
$k$ & $\oracleSOIF$ & \shortstack{MC Coverage \\ ($\hat{\psi}_{2, k} $ 90\% Wald CI)} & $\mathsf{Bias} ( \oraclepsi )$ & $\widehat\chi_{k}^{(2)} ( \Omega_{k}^{-1}; z_{0.05}, \delta = 0.1)$ \\
\hline
$0$ & 0 (0) & 48\% & -0.0124 (0.00734) & 0\% \\
$64$ & -0.00472 (0.00186) & 77\% & -0.00769 (0.00725) & 66\% \\
$128$ & -0.00506 (0.00210) & 78\% & -0.00735 (0.00730) & 61\% \\
$256$ & -0.00910 (0.00288) & 86\% & -0.00331 (0.00714) & 89\% \\
$512$ & -0.00920 (0.00347) & 87\% & -0.00321 (0.00760) & 81\% \\
$1024$ & -0.00919 (0.00366) & 88\% & -0.00322 (0.00766) & 62\% \\
$2048$ & -0.0109 (0.00525) & 89\% & -0.00151 (0.00855) & 51\% \\
$4096$ & -0.0117 (0.00716) & 89\% & -0.000710 (0.00988) & 44\% \\
\hline
$k$ & $\genericSOIF$ & \shortstack{MC Coverage \\ ($\hat{\psi}_{2, k} (\widehat{\Omega}_{k}^{-1} )$ 90\% Wald CI)} & $\mathsf{Bias} \left( \genericpsi \right)$ & $\widehat\chi_{k}^{(2)} ( \widehat{\Omega}_{k}^{-1}; z_{0.05}, \delta = 0.1)$ \\
\hline
$0$ & 0 (0) & 48\% & -0.0124 (0.00734) & 0\% \\
$64$ & -0.00479 (0.00187) & 78\% & -0.00762 (0.00719) & 68\% \\
$128$ & -0.00522 (0.00218) & 78\% & -0.00719 (0.00727) & 68\% \\
$256$ & -0.00951 (0.00308) & 88\% & -0.00290 (0.00706) & 90\% \\
$512$ & -0.00975 (0.00372) & 88\% & -0.00266 (0.00755) & 88\% \\
$1024$ & -0.00976 (0.00447) & 90\% & -0.00265 (0.00806) & 76\% \\
$2048$ & -0.0113 (0.00618) & 86\% & -0.00114 (0.00940) & 73\% \\
$4096$ & -0.0145 (0.00724) & 88\% & 0.00206 (0.00995) & 49\% \\
\hline
\end{tabular}
} \newline
Simulation setup I ($\psi(\theta) = \mathbb{E}_\theta [ \mathsf{cov}_\theta (Y, A | X) ]$): We reported the MCavs of point estimates and MCsds (first column in each panel) of $\widehat{\mathbb{IF}}_{22, k}$ and $\widehat{\mathbb{IF}}_{22, k} (\widehat{\Omega}_{k}^{-1})$, together with the coverage probability of 90\% confidence intervals (second column in each panel) of $\hat{\psi}_{2, k}$ and $\hat{\psi}_{2, k} (\widehat{\Omega}_{k}^{-1})$, the MCavs of the bias and MCsds (third column in each panel) of $\hat{\psi}_{2, k}$ and $\hat{\psi}_{2, k} (\widehat{\Omega}_{k}^{-1})$ and the empirical rejection rate based on the test statistic $\widehat\chi_{k}^{(2)} (\zeta_{k}, \delta = 0.1)$ and $\widehat\chi_{k}^{(2)} (\widehat{\Omega}_{k}^{-1}; \zeta_{k}, \delta = 0.1)$ (see \Cref{sec:cov}) with $\zeta_{k} = z_{\alpha^{\dag} = 0.05} = 1.64$ (fourth column in each panel). Nuisance functions are estimated by nonparametric kernel regression with cross validation.
\end{table}

\begin{table}[tbp]
\caption{}
\label{tab:smooth_nn_cov}
\resizebox{\columnwidth}{!}{
\begin{tabular}{l | lclc}
\hline
$k$ & $\oracleSOIF$ & \shortstack{MC Coverage \\ ($\hat{\psi}_{2, k} $ 90\% Wald CI)} & $\mathsf{Bias} ( \oraclepsi )$ & $\widehat\chi_{k}^{(2)} ( \Omega_{k}^{-1}; z_{0.05}, \delta = 0.1)$ \\
\hline
$0$ & 0 (0) & 13\% & -0.0216 (0.00775) & 0\% \\
$64$ & -0.0139 (0.00366) & 75\% & -0.00769 (0.00704) & 97\% \\
$128$ & -0.0142 (0.00377) & 77\% & -0.00728 (0.00710) & 98\% \\
$256$ & -0.0186 (0.00416) & 86\% & -0.00335 (0.00698) & 100\% \\
$512$ & -0.0186 (0.00474) & 86\% & -0.00324 (0.00750) & 98\% \\
$1024$ & -0.0189 (0.00530) & 88\% & -0.00301 (0.00754) & 95\% \\
$2048$ & -0.0209 (0.00646) & 87\% & -0.00134 (0.00844) & 93\% \\
\hline
$k$ & $\genericSOIF$ & \shortstack{MC Coverage \\ ($\hat{\psi}_{2, k} (\widehat{\Omega}_{k}^{-1} )$ 90\% Wald CI)} & $\mathsf{Bias} \left( \genericpsi \right)$ & $\widehat\chi_{k}^{(2)} ( \widehat{\Omega}_{k}^{-1}; z_{0.05}, \delta = 0.1)$ \\
\hline
$0$ & 0 (0) & 48\% & -0.0124 (0.00734) & 0\% \\
$64$ & -0.0141 (0.00370) & 75\% & -0.000750 (0.00695) & 97\% \\
$128$ & -0.0146 (0.00385) & 78\% & -0.00689 (0.00703) & 99\% \\
$256$ & -0.0195 (0.00437) & 88\% & -0.00249 (0.00684) & 100\% \\
$512$ & -0.0199 (0.00510) & 87\% & -0.00194 (0.00733) & 99\% \\
$1024$ & -0.0203 (0.00607) & 88\% & -0.00159 (0.00776) & 98\% \\
$2048$ & -0.0213 (0.00746) & 89\% & -0.000937 (0.00933) & 98\% \\
\hline
\end{tabular}
} \newline
Simulation setup I ($\psi(\theta) = \mathbb{E}_\theta [ \mathsf{cov}_\theta (Y, A | X) ]$): We reported the MCavs of point estimates and MCsds (first column in each panel) of $\widehat{\mathbb{IF}}_{22, k}$ and $\widehat{\mathbb{IF}}_{22, k} (\widehat{\Omega}_{k}^{-1})$, together with the coverage probability of 90\% confidence intervals (second column in each panel) of $\hat{\psi}_{2, k}$ and $\hat{\psi}_{2, k} (\widehat{\Omega}_{k}^{-1})$, the MCavs of the bias and MCsds (third column in each panel) of $\hat{\psi}_{2, k}$ and $\hat{\psi}_{2, k} (\widehat{\Omega}_{k}^{-1})$ and the empirical rejection rate based on the test statistic $\widehat\chi_{k}^{(2)} (\zeta_{k}, \delta = 0.1)$ and $\widehat\chi_{k}^{(2)} (\widehat{\Omega}_{k}^{-1}; \zeta_{k}, \delta = 0.1)$ (see \Cref{sec:cov}) with $\zeta_{k} = z_{\alpha^{\dag} = 0.05} = 1.64$ (fourth column in each panel). Nuisance functions are estimated by neural networks.
\end{table}

\subsection{Simulation setup II: smooth $f_X$}\label{sec:nonsmooth}
Due to the similarity of the results to the previous section, we only display the results when the regression functions are estimated by nonparametric kernel regression with cross validation in \Cref{tab:nonsmooth_kern_var} and \Cref{tab:nonsmooth_kern_cov}. Since the overall messages of \Cref{tab:nonsmooth_kern_var} and \Cref{tab:nonsmooth_kern_cov} are very similar to that of \Cref{tab:intro} and \Cref{tab:smooth_kern_cov} for the simulation setup I with nonsmooth $f_X$, we do not further describe the results in detail.

\begin{table}[tbp]
\caption{}
\label{tab:nonsmooth_kern_var}
\resizebox{\columnwidth}{!}{
\begin{tabular}{l | lclc}
\hline
$k$ & $\oracleSOIF$ & \shortstack{MC Coverage \\ ($\hat{\psi}_{2, k} $ 90\% Wald CI)} & $\mathsf{Bias} ( \oraclepsi )$ & $\widehat\chi_{k}^{(1)} ( \Omega_{k}^{-1}; z_{0.10}, \delta = 0.75 (1.5))$ \\
\hline
$0$ & 0 (0) & 0\% & 0.217 (0.0145) & 0\% (0\%) \\
$64$ & 0.0366 (0.00645) & 0\% & 0.180 (0.0138) & 95\% (5\%) \\
$128$ & 0.0392 (0.00692) & 0\% & 0.177 (0.0139) & 96\% (10\%) \\
$256$ & 0.114 (0.0120) & 0\% & 0.103 (0.0125) & 100\% (100\%) \\
$512$ & 0.116 (0.0136) & 0\% & 0.101 (0.0144) & 100\% (100\%) \\
$1024$ & 0.116 (0.0140) & 0\% & 0.101 (0.0144) & 100\% (100\%) \\
$2048$ & 0.142 (0.0192) & 0\% & 0.0751 (0.0164) & 100\% (100\%) \\
$4096$ & 0.164 (0.0277) & 31\% & 0.0531 (0.0228) & 100\% (100\%) \\
\hline
$k$ & $\genericSOIF$ & \shortstack{MC Coverage \\ ($\hat{\psi}_{2, k} (\widehat{\Omega}_{k}^{-1} )$ 90\% Wald CI)} & $\mathsf{Bias} \left( \genericpsi \right)$ & $\widehat\chi_{k}^{(1)} ( \widehat{\Omega}_{k}^{-1}; z_{0.10}, \delta = 0.75 (1.5))$ \\
\hline
$0$ & 0 (0) & 0\% & 0.217 (0.0145) & 0\% (0\%) \\
$64$ & 0.0372 (0.00646) & 0\% & 0.180 (0.0139) & 95\% (7\%) \\
$128$ & 0.0403 (0.00706) & 0\% & 0.176 (0.0139) & 98\% (15\%) \\
$256$ & 0.119 (0.0120) & 0\% & 0.0973 (0.0126) & 100\% (100\%) \\
$512$ & 0.124 (0.0135) & 0\% & 0.0924 (0.0144) & 100\% (100\%) \\
$1024$ & 0.129 (0.0161) & 0\% & 0.0877 (0.0165) & 100\% (100\%) \\
$2048$ & 0.147 (0.0232) & 8\% & 0.0692 (0.0214) & 100\% (100\%) \\
$4096$ & 0.177 (0.0280) & 50\% & 0.0393 (0.0236) & 100\% (100\%) \\
\hline
\end{tabular}
} \newline
Simulation setup II ($\psi(\theta) = \mathbb{E}_\theta [ \mathsf{var}_\theta (A | X) ]$): We reported the MCavs of point estimates and MCsds in the parenthesis (first column in each panel) of $\widehat{\mathbb{IF}}_{22, k}$ and $\widehat{\mathbb{IF}}_{22, k} (\widehat{\Omega}_{k}^{-1})$, together with the coverage probability of 90\% CIs (second column in each panel) of $\hat{\psi}_{2, k}$ and $\hat{\psi}_{2, k} (\widehat{\Omega}_{k}^{-1})$, the MCavs of the bias and MCsds in the parenthesis (third column in each panel) of $\hat{\psi}_{2, k}$ and $\hat{\psi}_{2, k} (\widehat{\Omega}_{k}^{-1})$ and the empirical rejection rate based on the test statistic $\widehat\chi_{k}^{(1)} (\zeta_{k}, \delta = 0.75 \text{ or } 1.5)$ and $\widehat\chi_{k}^{(1)} (\widehat{\Omega}_{k}^{-1}; \zeta_{k}, \delta = 0.75 \text{ or } 1.5)$ (see \Cref{sec:var}) with $\zeta_{k} = z_{0.10} = 1.28$ (fourth column in each panel). Nuisance functions are estimated by nonparametric kernel regression with cross validation.
\end{table}

\begin{table}[tbp]
\caption{}
\label{tab:nonsmooth_kern_cov}
\resizebox{\columnwidth}{!}{
\begin{tabular}{l | lclc}
\hline
$k$ & $\oracleSOIF$ & \shortstack{MC Coverage \\ ($\hat{\psi}_{2, k} $ 90\% Wald CI)} & $\mathsf{Bias} ( \oraclepsi )$ & $\widehat\chi_{k}^{(2)} ( \Omega_{k}^{-1}; z_{0.05}, \delta = 0.1)$ \\
\hline
$0$ & 0 (0) & 60\% & -0.0111 (0.00748) & 0\% \\
$64$ & -0.00376 (0.00165) & 77\% & -0.00735 (0.00754) & 55\% \\
$128$ & -0.00385 (0.00187) & 76\% & -0.00726 (0.00755) & 49\% \\
$256$ & -0.00799 (0.00247) & 91\% & -0.00313 (0.00708) & 84\% \\
$512$ & -0.00807 (0.00274) & 88\% & -0.00305 (0.00719) & 70\% \\
$1024$ & -0.00813 (0.00379) & 86\% & -0.00299 (0.00774) & 51\% \\
$2048$ & -0.00899 (0.00537) & 89\% & -0.00213 (0.00858) & 40\% \\
$4096$ & -0.00974 (0.00722) & 89\% & -0.00138 (0.00925) & 30\% \\
\hline
$k$ & $\genericSOIF$ & \shortstack{MC Coverage \\ ($\hat{\psi}_{2, k} (\widehat{\Omega}_{k}^{-1} )$ 90\% Wald CI)} & $\mathsf{Bias} \left( \genericpsi \right)$ & $\widehat\chi_{k}^{(2)} ( \widehat{\Omega}_{k}^{-1}; z_{0.05}, \delta = 0.1)$ \\
\hline
$0$ & 0 (0) & 48\% & -0.0124 (0.00734) & 0\% \\
$64$ & -0.00383 (0.00168) & 77\% & -0.00729 (0.00755) & 58\% \\
$128$ & -0.00396 (0.00194) & 79\% & -0.00716 (0.00756) & 51\% \\
$256$ & -0.00832 (0.00253) & 93\% & -0.00279 (0.00709) & 90\% \\
$512$ & -0.00865 (0.00281) & 88\% & -0.00247 (0.00715) & 82\% \\
$1024$ & -0.00905 (0.00456) & 89\% & -0.00207 (0.00835) & 73\% \\
$2048$ & -0.00892 (0.00587) & 89\% & -0.00220 (0.00913) & 68\% \\
$4096$ & -0.0111 (0.00735) & 90\% & -$5.682 \times 10^{-6}$ (0.00905) & 32\% \\
\hline
\end{tabular}
} \newline
Simulation setup II ($\psi(\theta) = \mathbb{E}_\theta [ \mathsf{var}_\theta (A | X) ]$): We reported the MCavs of point estimates and MCsds in the parenthesis (first column in each panel) of $\widehat{\mathbb{IF}}_{22, k}$ and $\widehat{\mathbb{IF}}_{22, k} (\widehat{\Omega}_{k}^{-1})$, together with the coverage probability of 90\% CIs (second column in each panel) of $\hat{\psi}_{2, k}$ and $\hat{\psi}_{2, k} (\widehat{\Omega}_{k}^{-1})$, the MCavs of the bias and MCsds in the parenthesis (third column in each panel) of $\hat{\psi}_{2, k}$ and $\hat{\psi}_{2, k} (\widehat{\Omega}_{k}^{-1})$ and the empirical rejection rate based on the test statistic $\widehat\chi_{k}^{(2)} (\zeta_{k}, \delta = 0.1)$ and $\widehat\chi_{k}^{(2)} (\widehat{\Omega}_{k}^{-1}; \zeta_{k}, \delta = 0.1)$ (see \Cref{sec:cov}) with $\zeta_{k} = z_{0.05} = 1.64$ (fourth column in each panel). For more details on the simulation setup, see \Cref{sec:simulations}. Nuisance functions are estimated by nonparametric kernel regression with cross validation.
\end{table}

\subsection{Generating functions from \Holder{} spaces in simulation studies}\label{app:holder} 
The functions $h_{f}$, $h_{b}$ and $h_{p}$ appeared in \Cref{sec:simulations} are of the following forms: 
\begin{align}
h_{f} (x; s_{f}) & \propto 1 + \text{exp} \left\{ \frac{1}{2} \sum_{i \in \mathcal{I}, \ell \in \mathbb{Z}} 2^{- i (s_{f} + 0.25)} \alpha^{\dag}_{i, \ell} (x) \right\},  \label{f} \\
h_{b} (x; s_{b}) & = \sum_{i \in \mathcal{I}, \ell \in \mathbb{Z}} 2^{- i (s_{b} + 0.25)} \alpha^{\dag}_{i, \ell} (x),  \label{b} \\
h_{p} (x; s_{p}) & = \text{expit} \left\{ - 2 \sum_{i \in \mathcal{I}, \ell \in \mathbb{Z}} 2^{- i (s_{p} + 0.25)} \alpha^{\dag}_{i, \ell} (x) \right\}
\label{p}
\end{align}
where $\mathcal{I} = \{ 0, 3, 6, 9, 10, 16 \}$ and $\alpha^{\dag}_{i, \ell} (\cdot)$ is the D12 (or equivalently db6) father wavelets function dilated at resolution $i$, shifted by $\ell$ \citep{daubechies1992ten, hardle1998wavelets, mallat1999wavelet}. \cite{hardle1998wavelets}[Theorem 9.6] indeed implies that $h_{f} (\cdot; s_{f}) \in \text{H\"{o}lder}(s_{f})$, $h_{b} (\cdot; s_{b}) \in \text{H\"{o}lder}(s_{b})$ and $h_{p} (\cdot; s_{p}) \in \text{H\"{o}lder}(s_{p})$. We fix $s_{b} = s_{p} = 0.25$. In simulation setup I we choose $s_{f} = 0.1$, so $f_X$ has smoothness lower than those of $b$ and $p$; whereas in simulation setup II, we choose $s_{f} = 0.4$, so $f_{X}$ has smoothness higher than those of $b$ and $p$.

\begin{table}[!htb]
\begin{tabular}{c|c|c}
\hline
$j$ & $\tau_{b, j}$ & $\tau_{p, j}$ \\ 
\hline
1 & 0.5 & 0.5 \\ 
2 & -0.5 & -0.5 \\
\hline
\end{tabular}
\caption{Coefficients used in constructing $b$ and $p$ in \Cref{sec:simulations}.}
\label{tab:s1}
\end{table}

In Table \ref{tab:s1}, we provide the numerical values for $\left( \tau_{b, j}, \tau_{p, j} \right)_{j = 1}^{2}$ used in generating the simulation experiments in \Cref{sec:simulations}.

\subsection{Generating correlated multidimensional covariates $X$ with fixed non-smooth marginal densities}\label{app:multiX} 
In the simulation study conducted in \Cref{sec:simulations}, one key step of generating the simulated datasets is to draw correlated multidimensional covariates $X \in [0, 1]^{d}$ ($d = 2$) with fixed non-smooth marginal densities. First, we fix the marginal densities of $X$ in each dimension proportional to $h_{f} (\cdot)$ (\cref{f}). Then we draw independently $\tilde{X}_{i, j}$, $i = 1, \ldots, 2K$ with sample size $2K$, from $h_{f}$ for every $j = 1, \ldots, d$ so $\tilde{X} = (\tilde{X}_{1, \cdot}, \ldots, \tilde{X}_{2K, \cdot})^{\top} \in [0, 1]^{2K \times d}$. Next, to create correlations between different dimensions, we follow the strategy proposed in \citet{baker2008order}. First we group every two consecutive draws: $(\tilde{X}_{1, \cdot}, \tilde{X}_{2, \cdot})^{\top}, (\tilde{X}_{3, \cdot}, \tilde{X}_{4, \cdot})^{\top}, \ldots, (\tilde{X}_{2K - 1, \cdot}, \tilde{X}_{2K, \cdot})^{\top}$. Then for each pair $(\tilde{X}_{2 i - 1, \cdot}, \tilde{X}_{2 i, \cdot})^{\top}$ for $i = 1, \ldots, K$, we form the following $d$-dimensional random vectors 
\begin{align*}
U_{i} \coloneqq (\mathsf{max}(\tilde{X}_{2 i - 1, 1}, \tilde{X}_{2 i, 1}), \ldots, \mathsf{max}(\tilde{X}_{2 i - 1, d}, \tilde{X}_{2 i, d}))^{\top}, \\
V_{i} \coloneqq (\mathsf{min}(\tilde{X}_{2 i - 1, 1}, \tilde{X}_{2 i, 1}), \ldots, \mathsf{min}(\tilde{X}_{2 i - 1, d}, \tilde{X}_{2 i, d}))^{\top}.
\end{align*}
Lastly, we construct $K$ independent $d$-dimensional vectors $X$ by the following rule: for each $i = 1, \ldots, K$, we draw a Bernoulli random variable $B_{i}$ with probability 1 / 2, and if $B_{i} = 0$, $X_{i, \cdot} = U_{i}$, otherwise $X_{i, \cdot} = V_{i}$. Following the above strategy, we conserve the marginal density of $X_{\cdot, j}$ as that of $\tilde{X}_{\cdot, j}$ but create dependence between different dimensions.

\section{Supplementary Figures}
\subsection{Histograms of the upper confidence bound in simulation setup I}\label{app:ucb} 
\begin{figure}[!htb]
\caption{Histograms of $\mathsf{UCB}^{(1)}(\Omega_{k}^{-1}; \alpha = 0.10, \alpha^{\dag} = 0.10)$ and $\mathsf{UCB}^{(1)}(\widehat\Omega_{k}^{-1}; \alpha = 0.10, \alpha^{\dag} = 0.10)$ at $k = 2048$ in simulation experiment described in \Cref{sec:simulations}, for $\psi (\theta) = \BE_{\theta}[\var_{\theta}(A | X)]$}
\label{fig:ucb}\centering
\includegraphics[width=0.8\textwidth]{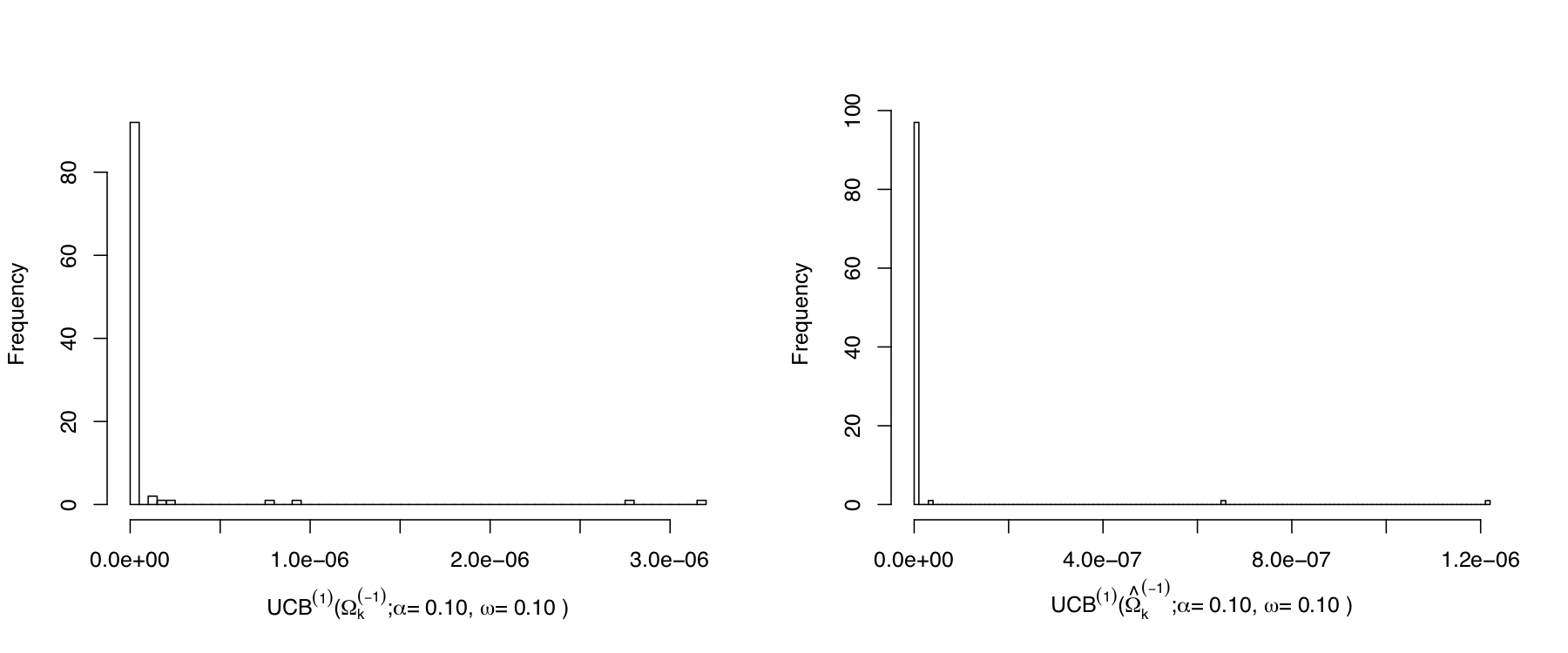}
\end{figure}

In \Cref{fig:ucb}, we display the histograms of $\mathsf{UCB}^{(1)}(\Omega_{k}^{-1}; \alpha = 0.10, \alpha^{\dag} = 0.10)$ and $\mathsf{UCB}^{(1)}(\widehat{\Omega}_{k}^{-1}; \alpha = 0.10, \alpha^{\dag} = 0.10)$ at $k = 2048$ in simulation experiment described in \Cref{sec:simulations}. 

\subsection{qqplots of $\widehat{\mathbb{IF}}_{22, k}$, $\widehat{\mathbb{IF}}_{22, k}^\mathsf{quasi} ( [\widehat\Omega_{k}^\mathsf{est}]^{-1} )$, and $\widehat{\mathbb{IF}}_{22, k} ( [\widehat{\Omega}_{k}^\mathsf{shrink}]^{-1})$ in simulation setup I} \label{app:norm} 
\begin{figure}[!htb]
\caption{qqplots of $\widehat{\mathbb{IF}}_{22, k}$, $\widehat{\mathbb{IF}}_{22, k}^{\mathsf{quasi}} ( [\widehat\Omega_{k}^\mathsf{est}]^{-1} )$ and $\widehat{\mathbb{IF}}_{22, k} ( [\widehat{\Omega}_{k}^\mathsf{shrink}]^{-1} )$ in simulation setup I, for $\psi (\theta) = \BE_{\theta}[\var_{\theta}(A | X)]$}
\label{fig:normal}\centering
\includegraphics[width=0.6\textwidth]{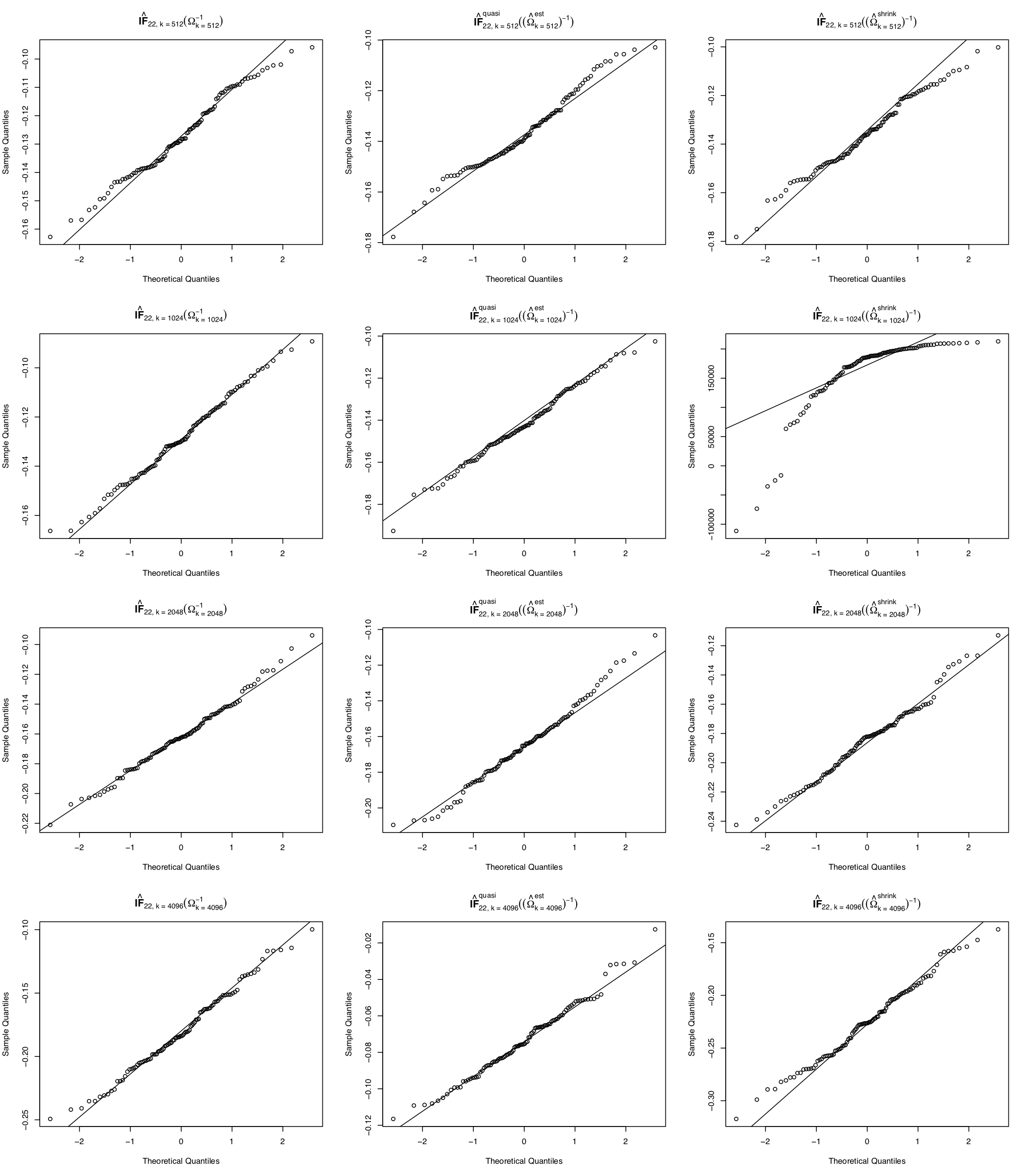} \newline
Left panel: qqplots for $\widehat{\mathbb{IF}}_{22, k}$ with $k = 512$ (the 1st row), $k = 1024$ (the 2nd row), $k = 2048$ (the 3rd row), and $k = 4096$ (the 4th row); Middle panel: qqnorm plots for $\widehat{\mathbb{IF}}_{22, k}^\mathsf{quasi} ( [\widehat{\Omega}_{k}^\mathsf{est}]^{-1} )$ with $k = 512$ (the 1st row), $k = 1024$ (the 2nd row), $k = 2048$ (the 3rd row), and $k = 4096$ (the 4th row); Right panel: qqnorm plots for $\widehat{\mathbb{IF}}_{22, k} ( [\widehat{\Omega}_{k}^\mathsf{shrink}]^{-1} )$ with $k = 512$ (the 1st row), $k = 1024$ (the 2nd row), $k = 2048$ (the 3rd row), and $k = 4096$ (the 4th row).
\end{figure}

In \Cref{fig:normal}, we display the qqplots of $\widehat{\mathbb{IF}}_{22, k}$, $\widehat{\mathbb{IF}}_{22, k}^\mathsf{quasi} ( [\widehat{\Omega}_{k}^\mathsf{est}]^{-1} )$, and $\widehat{\mathbb{IF}}_{22, k} ( [\widehat{\Omega}_{k}^\mathsf{shrink}]^{-1} )$ in simulation experiment described in \Cref{sec:simulations} over $k = 512, 1024, 2048, 4096$ from top to bottom, for $\psi (\theta) = \BE_{\theta}[\var_{\theta}(A | X)]$. Most of the statistics displayed are close to a normal distribution. The one that deviates from normal the most is the 2nd row of the right panel, displaying $\widehat{\mathbb{IF}}_{22, k} ( [\widehat{\Omega}_{k}^\mathsf{shrink}]^{-1} )$ with $k = 1024$. This is because $\widehat{\mathbb{IF}}_{22, k} ( [\widehat{\Omega}_{k}^\mathsf{shrink}]^{-1} )$ numerically blows up at $k = 1024$, which has been shown in \Cref{tab:smooth_shrink} and \Cref{tab:var_shrink}.

\end{document}